\documentclass{article}
\usepackage[eandd, final]{neurips_2026}      %

\usepackage[utf8]{inputenc}
\usepackage{hyperref}
\usepackage{url}

\usepackage[T1]{fontenc}
\usepackage{iftex}
\ifPDFTeX\else
  \usepackage{newunicodechar}
  \newunicodechar{–}{--}
  \newunicodechar{—}{---}
  \newunicodechar{×}{\ensuremath{\times}}
  \newunicodechar{“}{``}
  \newunicodechar{”}{''}
  \newunicodechar{’}{'}
\fi

\usepackage{booktabs}
\usepackage{amsfonts}
\usepackage{nicefrac}
\usepackage{microtype}
\usepackage{xcolor}
\usepackage{graphicx}
\usepackage{tikz}
\usetikzlibrary{positioning, arrows.meta, shapes.misc, calc, fit}
\usepackage{subcaption}
\usepackage{tabularx}
\usepackage{longtable}
\usepackage{enumitem}
\setlist[itemize]{noitemsep, topsep=0pt, parsep=0pt, partopsep=0pt}
\setlist[enumerate]{noitemsep, topsep=0pt, parsep=0pt, partopsep=0pt}
\usepackage{xurl}
\usepackage{amsmath}
\usepackage{amssymb}
\usepackage{mathtools}
\usepackage{amsthm}
\usepackage[capitalize,noabbrev]{cleveref}
\usepackage[disable,textsize=tiny]{todonotes}

\theoremstyle{plain}

\theoremstyle{definition}

\theoremstyle{remark}

\title{Rethinking SAE Evaluation: An Atomic Interpretable Unit Framework Reveals Hidden Polysemanticity and Redundancy}

\author{%
  Riteng Zhang \\
  Purdue University \\
  \texttt{zhan5982@purdue.edu} \\
  \And
  Xiaoqian Wang \\
  Purdue University \\
  \texttt{joywang@purdue.edu} \\
}

\begin{document}
\maketitle

\begin{abstract}
Sparse autoencoders (SAEs) were originally motivated by discovering monosemantic features, a property that characterizes individual features but does not specify what the learned feature set should look like as a whole. As SAEs have increasingly been adopted for attribution, circuit discovery, and intervention, researchers have begun evaluating metrics that describe global properties of the feature set, such as sparsity, dead feature rates, reconstruction fidelity, and disentanglement, a diverse set that extends far beyond monosemanticity. However, this implicit shift beyond monosemanticity has not been formalized as a clear, well-defined objective that comes before these evaluation metrics. Without such an objective, existing metrics remain ad hoc, with their necessity unjustified. In this work, we formalize this ultimate objective by introducing a new paradigm for SAE evaluation: requiring that SAE features, and more generally dictionary-based decompositions of model representations, form an atomic interpretable unit basis, a coordinate system for semantically meaningful explanation. Under this paradigm, we organize existing and new measurements such as DAME (Dominance-Aware Monosemanticity Evaluation) as proxy metrics and introduce a comprehensive benchmarking framework. Evaluating hundreds of SAEs across multiple architectures and model families, we uncover failure modes in which some SAEs appear monosemantic under existing evaluation metrics while remaining functionally polysemantic: the most dominant features, those contributing most to model computation, are systematically less interpretable and more redundant. These failure modes warrant attention from SAE researchers as targets for improved training and architectures, and motivate dominance-aware evaluation as a standard check for practitioners before SAE features are deployed for attribution, circuit tracing, or intervention.

\end{abstract}

\section{Introduction}

Sparse autoencoders (SAEs) entered mechanistic interpretability with a clear and well-motivated objective: addressing superposition by recovering monosemantic features \citep{elhage2022toy}. Superposition arises when neural networks---particularly transformer-based language models---represent more meaningful concepts than they have activation dimensions, forcing multiple concepts to share the same representational space and rendering individual neurons polysemantic \citep{elhage2022toy, scherlis2022polysemanticity}.

SAEs address this challenge by learning an expanded, sparse dictionary of feature directions. Concretely, they decompose a layer's $d_{\text{in}}$-dimensional activations into $m \gg d_{\text{in}}$ learned features, with the goal that each feature corresponds to a single interpretable concept. Early work demonstrated that this sparsity-based dictionary learning approach can successfully disentangle superposed representations \citep{sharkey2022interim, cunningham2024sparse, bricken2023towards}, isolating coherent conceptual directions from an otherwise entangled activation space and offering a compelling solution to interpretability in large language models.

As SAEs began to be used for LLM attribution, circuit discovery, and intervention \citep{templeton2024scaling, marks2024circuit, oneill2024sparse}, researchers introduced diverse evaluation metrics: reconstruction fidelity, sparsity levels, dead feature rates, disentanglement, and intervention performance \citep{gao2024scaling, rajamanoharan2024jumprelu, chanin2024absorption}. Different works emphasized different subsets \citep{bricken2023towards, cunningham2024sparse, gao2024scaling}, while benchmarking efforts like SAEBench \citep{karvonen2025saebench} organized them into systematic taxonomies containing eight diverse metrics spanning four capability categories (reconstruction, concept detection, interpretability, and feature disentanglement). Yet a foundational question remains unaddressed: \emph{what objective do these metrics collectively serve?} When one of these proposed metrics degrades, the SAE is deemed deficient---but the underlying notion of quality remains undefined. Connections between metrics are unclear, and their necessity is justified only post hoc.

This gap reflects a deeper issue: monosemanticity, the original objective proposed for SAEs, is insufficient. A feature is monosemantic if it activates on a single coherent concept rather than responding to multiple unrelated patterns. As a \emph{local} property of individual features, monosemanticity has nothing to do with metrics like reconstruction fidelity or dead feature counts that concern the feature set \emph{globally}. The proliferation of metrics reveals an implicit shift toward a global objective---but one that remains theoretically unclear and unspecified.

Some works have begun to explicitly propose foundational criteria for what properties the feature set itself should possess, such as atomicity, uniqueness, stability, and completeness \citep{leask2025canonical, bussmann2024showing, ayonrinde2024interpretability}. However, these properties and metrics are typically invoked without justification for why this particular set is necessary and sufficient, and crucially, without first specifying \emph{what function SAE features are meant to serve}---the underlying objective that would determine which properties matter and why, which in turn determines what metrics should be used to evaluate them. Appendix~\ref{appendix:extended-related-work} provides a more detailed discussion of the history of SAEs and the evolution of SAE evaluation.

\begin{figure}[t]
  \centering
  \begin{subfigure}[t]{0.49\linewidth}
    \centering
    \includegraphics[width=\linewidth]{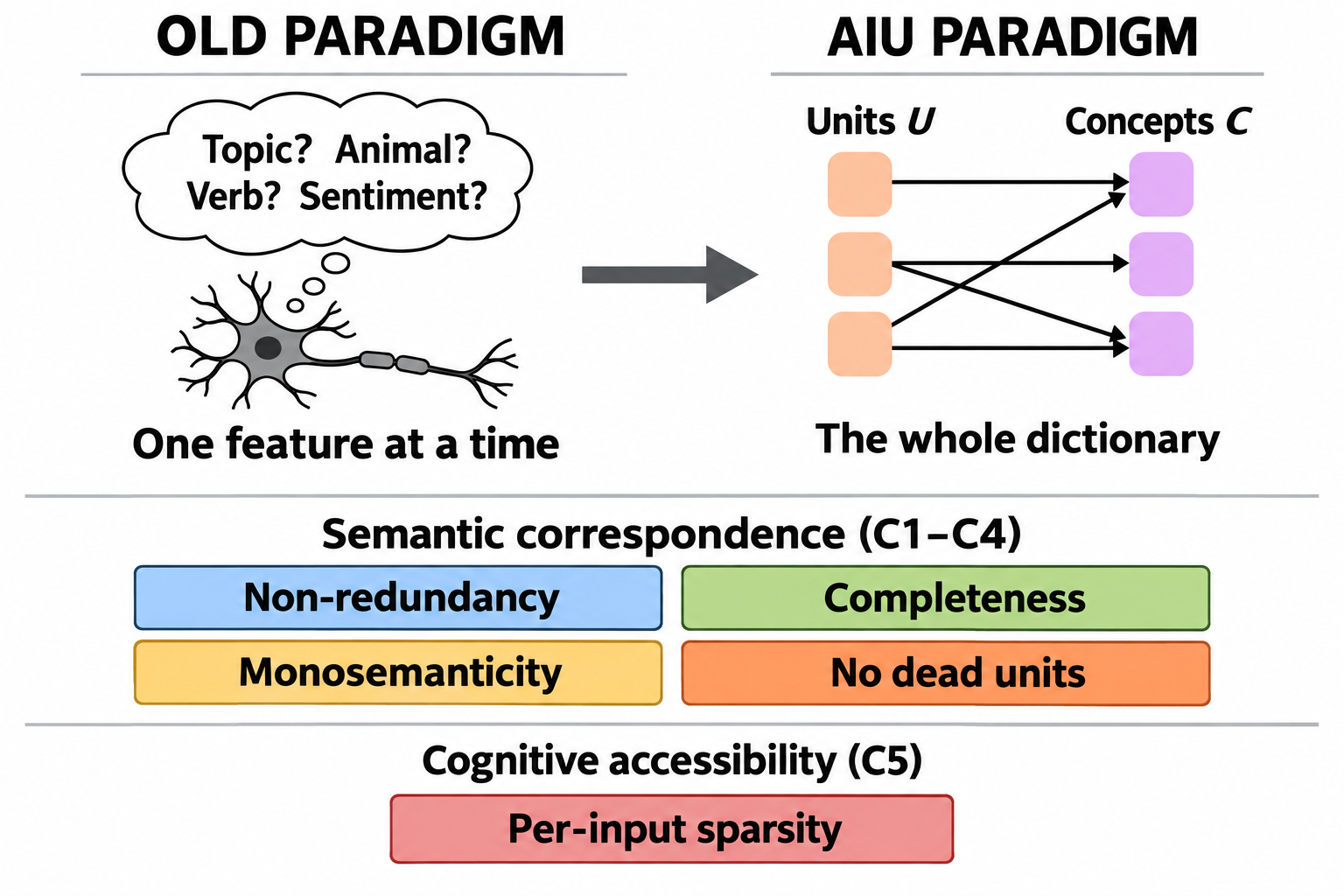}
    \caption{}
    \label{fig:interpretability-paradigm}
  \end{subfigure}
  \hfill
  \begin{subfigure}[t]{0.49\linewidth}
    \centering
    \includegraphics[width=\linewidth]{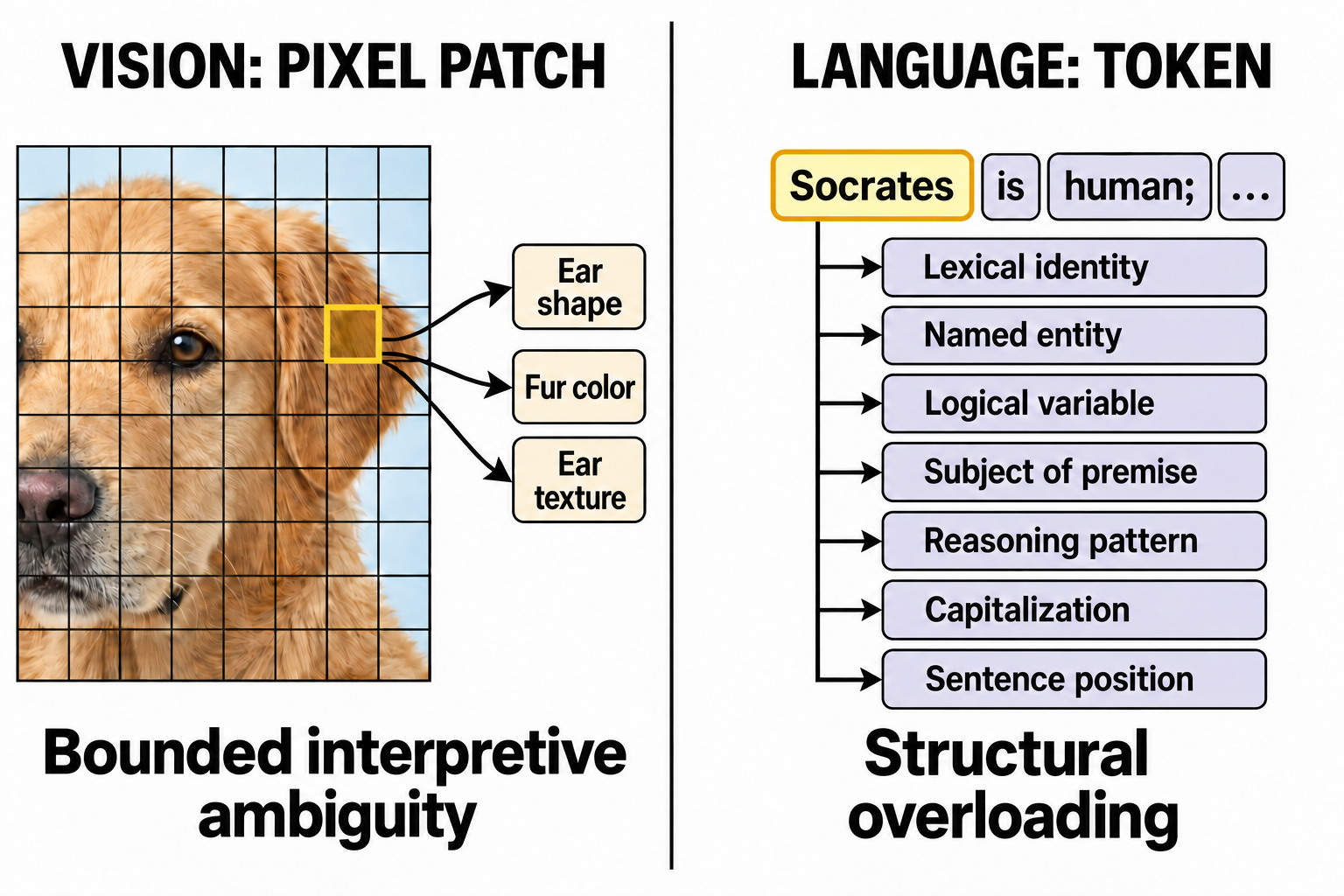}
    \caption{}
    \label{fig:atomic-units-vision-vs-language}
  \end{subfigure}
  \caption{\textbf{(a)} Paradigm shift in SAE evaluation: from expecting isolated monosemantic features to requiring that features jointly form an atomic interpretable unit basis. Under the old paradigm, newly proposed metrics function as \emph{tweak-and-patch} additions, analogous to epicycles. \textbf{(b)} Natural atomic units in vision vs.\ language. Pixel patches exhibit bounded interpretive ambiguity, while individual tokens are structurally overloaded with unrelated roles.}
  \label{fig:overview}
\end{figure}

In this work, we address this foundational gap by formalizing a new paradigm for SAE evaluation (Figure~\ref{fig:interpretability-paradigm}).
We specify (1) the ultimate objective that SAE features should serve: forming an \emph{atomic interpretable unit} (AIU) \emph{basis} (Section~\ref{sec:atomic-interpretable-units})---a coordinate system in which attribution and intervention are semantically meaningful.
Under this paradigm, we derive (2) five necessary criteria for interpretable feature dictionaries---non-redundancy, completeness, monosemanticity, no dead units, and per-input sparsity---as direct consequences of the atomic interpretable unit objective (Section~\ref{sec:atomic-interpretable-units}).

Building on this theoretical foundation, we develop a comprehensive SAE-specific evaluation framework (Section~\ref{sec:criteria} and Appendix~\ref{appendix:other_metrics}) that reorganizes existing measurements and introduces (3) novel metrics---particularly DAME (Dominance-Aware Monosemanticity Evaluation) for assessing monosemanticity, and decoder-based redundancy metrics with theoretical and geometric justification for why non-redundancy should be enforced on decoder directions (Appendix~\ref{appendix:decoder-sim})---to operationalize these criteria at scale. Our newly proposed metrics expose (4) critical failure modes---SAEs that appear monosemantic under existing metrics while remaining functionally polysemantic due to dominance (a feature's per-input fraction of total activation; Section~\ref{sec:local-to-global}) and redundancy (distinct features with near-identical decoder directions; Section~\ref{sec:redundancy}) effects previously undetectable by prior evaluation schemes (Section~\ref{sec:results}).

Beyond these core contributions, we also: (5) provide a comprehensive open-source benchmark (Appendix~\ref{appendix:benchmark_other_results}) across existing SAE families including SAEBench, JBloom, and CanRager, as well as (6) a large suite of newly trained SAE variants spanning multiple architectures and three different LLMs to systematically explore the hyperparameter space. {\def\UrlBreaks{\do\/\do-}\def\UrlBigBreaks{\do\:}Code is available at \url{https://github.com/ritengzhang/SAE-AIU} and the checkpoints of our newly trained SAEs at \url{https://huggingface.co/ritengzhang/SAE-AIU}.}

\section{The Atomic Interpretable Unit Framework}
\label{sec:atomic-interpretable-units}

Understanding how language models make decisions requires explanations that are both \textbf{faithful} (accurately reflecting internal computation) and \textbf{human-understandable} (expressed in meaningful terms). Modern attribution methods identify which units contribute to predictions, but they do not determine \emph{which units are appropriate to explain over}. Faithful attribution over the wrong units yields explanations that are computationally faithful yet not meaningfully interpretable to humans. To formalize what makes attribution meaningful, we introduce two key objects:

\paragraph{Concepts ($\mathcal{C}$).}
The set of concepts we want to explain---objects, syntactic roles, semantic relationships, reasoning patterns, and other meaningful computational units that matter for understanding model behavior.

\paragraph{Interpretable units ($U$).}
The units we actually attribute to during explanation. Each unit $u \in U$ must be interpretable to humans and can relate to one or more concepts in $\mathcal{C}$. Unlike raw computational elements (individual pixels, tokens, or neuron activations), interpretable units are chosen specifically to support meaningful attribution. Here, an atomic explanatory unit is a basic coordinate to which attribution is assigned directly, without further decomposition within the chosen explanatory interface.

\subsection{Why Natural Coordinates Fail: Intuition}

For natural model coordinates, $U$ is formed from or composed of basic architectural elements. We can examine whether these natural choices of $U$ satisfy even the minimum injectivity requirement (Figure~\ref{fig:atomic-units-vision-vs-language}).

\paragraph{Vision: partial success.}
For traditional image models that don't involve complex reasoning steps, we can take $U$ to be pixel patches. This approximately achieves $\mathcal{C} \to U$ injectivity: visual concepts (objects, parts, textures) occupy compact, limited-overlapping spatial regions, so different concepts typically map to different patches. A pixel patch participates in only a few human-meaningful concepts; moreover,
spatially coherent patches have \emph{bounded interpretive ambiguity}. Given a patch
that participates in a concept, a human observer can usually identify a small, tightly
related set of plausible interpretations—e.g., a dog ear’s shape, color, texture, or
the broader category ``ear.'' These alternatives are few and semantically coherent,
so input patches can serve as useful interpretable units for visual attribution, even
if they do not correspond perfectly to a single concept.

\paragraph{LLMs fail at input level.}
For language models, the natural choice is $U$ = token spans. However, tokens
simultaneously participate in many unrelated conceptual roles spanning multiple
levels of abstraction. Consider the token \emph{``Socrates''} appearing in the
syllogism: ``Socrates is human; all humans are mortal; therefore Socrates is
mortal.'' Attribution to this token cannot distinguish whether the model is
responding to its lexical identity, its role as a named entity, its function as
a logical variable, its position as the subject of a premise, participation in
a syllogistic reasoning pattern, or surface cues such as capitalization or
sentence-initial position. These interpretations are not minor variants of the
same concept but belong to fundamentally different conceptual categories.
As a result, the mapping $\mathcal{C} \to U$ is highly non-injective: many
unrelated concepts collapse onto the same token, making attribution inherently
ambiguous.

\paragraph{LLMs fail at activation level.}
At the activation level, we might take $U$ to be individual neurons or groups of neurons. However, language models exhibit \emph{superposition}~\citep{elhage2022toy}: they represent far more features than activation dimensions. Each neuron or residual stream coordinate already mixes many unrelated internal features. Because of superposition, $\mathcal{C} \to U$ fails injectivity: multiple distinct concepts are superposed onto the same neuron, so neuron-level attribution cannot meaningfully distinguish which concept is active.

\paragraph{The minimum requirement revealed.}
These failures across different choices of $U$ reveal a fundamental requirement: for attribution to be semantically meaningful, distinct concepts must map to distinct units—formally, the mapping $\mathcal{C} \to U$ must be injective. Vision achieves this through spatial locality; LLMs fail because tokens participate in many overlapping concepts and neurons exhibit superposition. Without injectivity, when a unit activates, we cannot tell which concept is responsible, making attribution ambiguous.

\subsection{Atomic Interpretable Units via SAEs}
Sparse autoencoders offer a fundamentally different approach. Given raw activations $\phi(x) \in \mathbb{R}^{d_{\text{in}}}$ at some layer for an input token position $x$, an SAE learns an encoder and decoder that transform activations as:
\[
\begin{aligned}
f_i(x) &= \text{ReLU}\!\left(W (\phi(x) - b_d) + b_e\right)_i, \\
\hat{\phi}(x) &= b_d + \sum_{i=1}^{m} f_i(x)\, d_i .
\end{aligned}
\]
where $W \in \mathbb{R}^{m \times d_{\text{in}}}$ is the encoder weight matrix, $b_e \in \mathbb{R}^m$ is the encoder bias, $b_d \in \mathbb{R}^{d_{\text{in}}}$ is the decoder bias, and $d_i \in \mathbb{R}^{d_{\text{in}}}$ are the decoder directions. The dictionary size $m \gg d_{\text{in}}$ allows the SAE to represent more features than the original activation dimension.

\paragraph{The key distinction.}
For natural coordinates (tokens, neurons, pixel patches), $U$ is fixed by the model architecture, composed from its basic elements; we can only hope for partial coverage of $\mathcal{C}$ and must work with whatever $U$ the architecture provides. For SAE features, $U$ is learned from scratch to \emph{be} a concept set, not constrained by architectural choices, and we control $|U| = m$ (the number of features).

Each SAE feature is a candidate atomic unit: $f_i(x)$ measures \emph{how much} the unit is used on input $x$, while the decoder direction $d_i$ specifies \emph{what} it represents (the vector it writes into the residual stream). Thus SAEs provide an interpretability interface by mapping raw activations to feature coordinates $f(x)$ on which attribution can be performed directly.

\subsection{What Makes Good Atomic Units?}
\begin{figure}[t]
\centering
\captionsetup{justification=raggedright,singlelinecheck=false}
\begin{minipage}[t]{0.53\linewidth}
  \vspace{0pt}\centering
  \setlength{\tabcolsep}{3pt}
  \renewcommand{\arraystretch}{1.45}
  \captionof{table}{Interpretability criteria: semantic-correspondence requirements (C1--C4) derived from mapping properties, and the input-conditioned accessibility requirement (C5).}
  \label{tab:criteria_mapping}
  \vspace{2pt}
  {\fontsize{7}{9}\selectfont
  \begin{tabularx}{\linewidth}{@{}*{3}{>{\raggedright\arraybackslash}X}@{}}
  \toprule
  \textbf{Requirement} & \textbf{Implication} & \textbf{Criterion}\\
  \midrule
  \multicolumn{3}{@{}l}{\emph{Semantic-correspondence requirements}}\\
  Injective $U\!\to\!\mathcal{C}$      & One unit per concept            & Non-redundancy (C1)\\
  Surjective $U\!\to\!\mathcal{C}$   & All concepts covered            & Completeness (C2)\\
  Injective $\mathcal{C}\!\to\! U$     & One concept per unit             & Monosemanticity (C3)\\
  Surjective $\mathcal{C}\!\to\! U$     & No unused units                 & No dead units (C4)\\
  \midrule
  \multicolumn{3}{@{}l}{\emph{Input-conditioned accessibility requirement}}\\
  Cognitive accessibility               & Few active units per input      & Per-input sparsity (C5)\\
  \bottomrule
  \end{tabularx}
  }
\end{minipage}\hfill%
\begin{minipage}[t]{0.43\linewidth}
  \vspace{0pt}\centering
  \includegraphics[width=\linewidth]{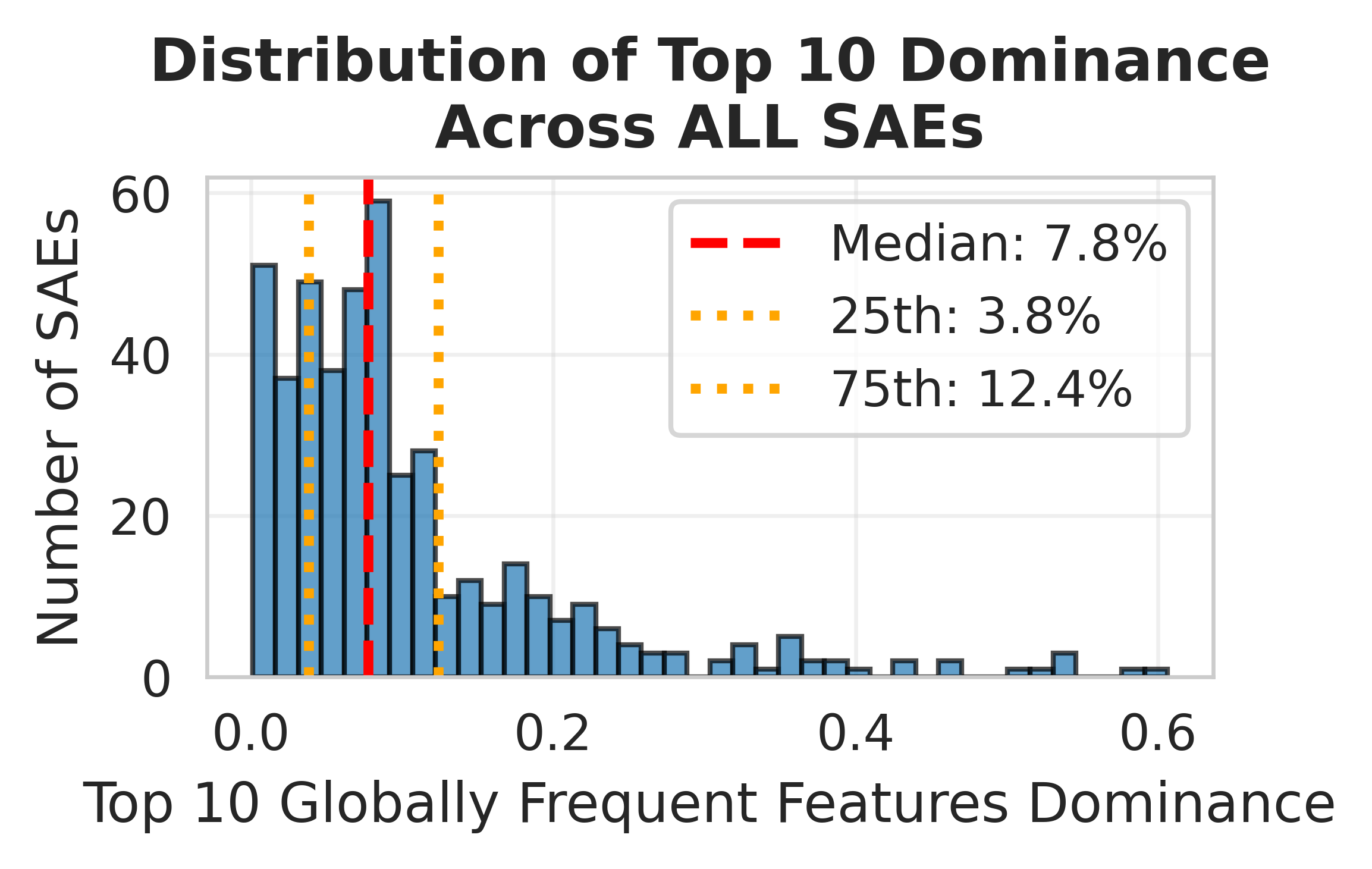}
  \captionof{figure}{Distribution of top-10 feature dominance (the 10 most frequently firing features) across all evaluated SAEs. Median 7.8\% of per-input activation, with a long right tail: in some SAEs the top 10 features alone account for over 30\% of per-input activation.}
  \label{fig:dominance_top10}
\end{minipage}
\end{figure}

Given this key distinction—that SAEs explicitly construct $U$ to approximate $\mathcal{C}$—the goal for SAE features is not merely to function as an atomic interpretable unit basis, but to directly learn features that approximate a meaningful concept set. Table~\ref{tab:criteria_mapping} summarizes the criteria.

\paragraph{The one-to-one ideal.}
Ideally, features and concepts should be in one-to-one correspondence. This requires two conditions: each feature represents one concept $\Rightarrow$ \textbf{monosemanticity} (C3): features shouldn't mix unrelated concepts; and each concept has one feature $\Rightarrow$ \textbf{non-redundancy} (C1): features shouldn't duplicate representations.

\paragraph{The granularity problem.}
Even with one-to-one correspondence, the SAE must learn features at the right level of detail. Let $m$ denote the number of learned features and let
$n_{\mathrm{ideal}}$ denote the appropriate number of concepts for this layer. If $m < n_{\mathrm{ideal}}$, the SAE recovers too few concepts: without enforcing surjectivity $U \to \mathcal{C}$, this causes \textbf{incompleteness} (C2)---some meaningful concepts remain unrecovered---while forcing surjectivity requires units to cover multiple concepts, breaking \textbf{monosemanticity} (C3). If $m > n_{\mathrm{ideal}}$, the SAE learns too many features: without forcing surjectivity $\mathcal{C} \to U$, this creates \textbf{dead features} (violating C4)---units that never activate meaningfully---while forcing surjectivity requires multiple features to map to the same concept, breaking \textbf{non-redundancy} (C1).

\paragraph{Sparsity: cognitive accessibility.}
Even with correct feature-concept correspondence at appropriate granularity, explanations are only human-usable when few features activate per input. C1--C4 constrain semantic correspondence between concepts and units, whereas C5 independently constrains the number of units participating in a particular explanation: a dictionary could have a clean one-to-one concept--unit correspondence yet activate thousands of individually meaningful features on one input, and a thousand-term attribution would not be cognitively usable. C5 therefore follows from the human-usable part of the AIU objective rather than from the correspondence mapping. \textbf{Per-input sparsity} (C5) ensures that $\|f(x)\|_0 \le k$ for small $k$, making explanations cognitively accessible.

\paragraph{Checkable evidence for C3.}
C3 is the injectivity condition $\mathcal{C}\to U$ for a target concept set fixed at a stated explanatory granularity. With ground-truth concept--unit relations, as in a synthetic or supervised setting, it can be checked directly; for open-ended LLM representations neither $\mathcal{C}$ nor the concept--unit relation is exhaustively observed, so auto-interpretability methods instead provide checkable behavioral evidence relevant to C3: whether a proposed explanation predicts held-out feature behavior beyond an uninformative baseline (Appendix~\ref{appendix:c3-lift}). This evidence depends on the chosen concept granularity, evaluation distribution, explainer, and evaluator.

\section{Evaluating the Five Criteria for SAE}
\label{sec:criteria}

The five criteria from Section~\ref{sec:atomic-interpretable-units} establish necessary conditions for SAE features to function as an atomic interpretable unit basis. These conditions are necessary rather than sufficient for causal use, which additionally depends on whether and how features are recruited by downstream layers' computation under a particular context, intervention, and outcome. However, these criteria were derived at the \emph{conceptual} level---describing what properties the feature set must satisfy---without specifying how to operationalize them as concrete measurements in the SAE context.\footnote{Many existing SAE evaluation metrics act as joint proxies for multiple criteria out of the five; a detailed correlation study and per-evaluation mapping (covering Sparse Probing, SCR, TPP, Feature Absorption, RAVEL, and Unlearning from SAEBench~\citep{karvonen2025saebench}) is provided in Appendix~\ref{appendix:saebench-correlations}.}

For \textbf{C2 (Completeness)}, \textbf{C4 (No dead units)}, and \textbf{C5 (Per-input sparsity)}, certain standard proxy metrics from prior work map naturally to our framework: for example, reconstruction fidelity metrics for completeness; firing frequency statistics for dead units; and mean $L_0$ (the number of features active on an input) for sparsity. We adopt these established measurements in our benchmarking with one addition: an effective sparsity metric that accounts for activation magnitude distribution (detailed in Appendix~\ref{appendix:other_metrics}).

\subsection{The Local-to-Global Aggregation Problem for C1/C3}
\label{sec:local-to-global}
The remaining two criteria---\textbf{C1 (Non-redundancy)} and \textbf{C3 (Monosemanticity)}---present a qualitatively different measurement challenge. Unlike \textbf{C2/C4/C5}, which admit \emph{direct dictionary-level proxy measurements}, \textbf{C1/C3} are defined as \emph{local properties}: monosemanticity is a property of individual features, while non-redundancy is a property of feature pairs.

Consequently, determining whether an SAE satisfies \textbf{C1/C3} as \emph{global requirements for an atomic interpretable unit basis}
demands an explicit \emph{local-to-global aggregation principle}. Yet most existing SAE evaluations effectively assume one by default:
they aggregate local measurements across features (or feature pairs) using uniform weighting to produce a single dictionary-level score. This assumption is fragile because feature usage is highly imbalanced: a small subset of units accounts for a disproportionate share of
activation (and therefore attribution), and thus disproportionately influences SAE-based interpretability analyses (e.g., circuit tracing);
their quality should therefore be weighted more heavily.

We stratify features by \emph{mean dominance}, $\bar{\rho}_i = \mathbb{E}_x[\rho_i(x)]$, where $\rho_i(x) = |f_i(x)| / \sum_{j=1}^m |f_j(x)|$ is the fraction of total activation contributed by feature $i$ on input $x$. When a non-atomic feature accounts for a large share of a representation, misinterpreting it distorts a large portion of what the SAE represents at that input. Complementary firing-frequency and activation-magnitude analyses, including their definitions and distinction from mean dominance, are in Appendix~\ref{sec:sae_feature_firing_behavior}.

Figure~\ref{fig:dominance_top10} shows that even in SAEs with thousands of features (4,000--128,000), the 10 most frequently firing features contribute a median of 7.8\% of per-input activation; the top-30 distribution (median 13.0\%) is in Appendix~\ref{sec:sae_feature_firing_behavior} (Figure~\ref{fig:dominance_top30}). Remarkably, higher sparsity (lower $L_0$) correlates with \emph{higher} dominance across all evaluated SAE families---e.g., in GPT2-Small SAEs (where all SAE architectures show the trend), 94\% of width-architecture combinations (17 of 18) show positive correlations (Appendix~\ref{sec:sae_feature_firing_behavior}). This raises a fundamental open question: whether increasing sparsity constraints truly decompose features into finer-grained atomic units, or instead primarily reduce sparsity loss by suppressing non-dominant features while concentrating computation into a smaller set of increasingly dominant ones.

\subsection{C3: Monosemanticity---DAME}
\label{mono_dom}
Monosemanticity concerns whether inputs with sufficiently large activation
(e.g., $\{x : |f_i(x)| \ge \epsilon\}$) share a coherent semantic property. In practice, this is assessed using the
standard auto-interpretability pipeline \citep{bills2023language}, which proceeds in
two stages. First, for each feature $i$, an LLM (the \emph{explainer}) is shown a small set of high activation
examples and produces a natural-language explanation describing the feature's firing
behavior. Second, the proposed explanation is evaluated on held-out validation inputs $S_i$ using an LLM-based judge: the judge predicts the feature's activation behavior, with labels derived from the precomputed activations $f_i(x)$ for $x \in S_i$.

\paragraph{Validation framing.}
Prior approaches frame validation in ways that fail to test whether explanations can
distinguish high from low activation, particularly under severe feature firing
frequency imbalance.

\textit{Binary classification} asks whether a feature fired or not \citep{paulo2024automatically}. However, if a rare feature never receives a positive example during validation, the explanation cannot be tested at all. For instance, suppose a polysemantic feature $i$ truly activates on two unrelated but rare triggers---mentions of \emph{Socrates} and mentions of \emph{Los Angeles}---so that $f_i(x)>0$ on only $\sim$0.01\% of tokens. If $S_i$ is sampled uniformly at random from the corpus, it is likely that no input in $S_i$ contains either trigger, yielding an all-negative validation set. In that case, any explanation (even an incorrect one that only describes ``Socrates'' and ignores ``Los Angeles'') will appear to achieve perfect accuracy simply by always predicting ``does not fire.'' The same issue applies symmetrically to high-frequency features: when a feature fires nearly 100\% of the time, negative examples become impossible to sample, so false positives cannot be detected.

\textit{Discretized magnitude prediction} bins activations into fixed intervals (e.g., 10 bins) normalized by each feature's maximum activation \citep{bills2023language}. This remains problematic: when features exhibit outliers, the maximum activation skews the bin boundaries, compressing inputs with meaningfully different activation magnitudes into the same label. Important distinctions go untested.

\paragraph{DAME.} We propose Dominance-Aware Monosemanticity Evaluation (DAME), which addresses both challenges:
\begin{enumerate}
    \item \textbf{Dominance-stratified feature sampling.} To address the dominance weighting problem, we rank features by mean dominance $\bar{\rho}_i$ and sample from each dominance quantile separately, ensuring evaluation coverage across the full dominance spectrum. Dominance is used for stratified sampling and aggregation; it is not an input to any feature's score.

    \item \textbf{Activation-balanced contrastive validation.} For each feature, we bin all inputs into three groups (low, medium, high activation) based on that feature's own activation distribution using quantile splits. Each validation question presents one example from each bin, without showing activation values, and asks the LLM to rank the three examples by activation strength given only the explanation, guaranteeing adaptive coverage across the full activation spectrum within every validation question. We score each feature by \emph{perfect ranking accuracy}: the fraction of validation questions for which the LLM predicts the exact activation order (Appendix~\ref{appendix:dame_details}).
\end{enumerate}

In a controlled test of high-versus-middle activation reversals that preserve binary firing support, DAME ranks the correct explanation above the reversed one for all 80 synthetic features, compared with 67 for discretized magnitude and at most 22 for binary/fuzz-style scoring (Appendix~\ref{appendix:scorer-comparison}).

\subsection{C1: Non-Redundancy---Decoder Cosine Similarity}
\label{sec:redundancy}
Non-redundancy requires that distinct features encode distinct concepts. Though not
often evaluated, when researchers do measure redundancy \citep{karvonen2025saebench},
they use decoder cosine similarity,
$\text{sim}(i, j) = \frac{d_i \cdot d_j}{\|d_i\| \|d_j\|}$.
However, this decoder-side measure is adopted
without explaining why decoder redundancy (rather than encoder redundancy) is the
appropriate notion for C1. We provide a justification below and report per-quantile
decoder cosine similarity in Section~\ref{sec:results}.

\noindent\textbf{Encoder-decoder asymmetry.}
Each encoder weight vector $w_i$ (the $i$-th row of $W$) functions as a sparse classifier that carves out a small firing region in activation space. Because of the per-input sparsity constraint, most features must fire on only a tiny fraction of inputs. However, transformer activations cluster near a dominant mean direction (Appendix Figure~\ref{fig:surface_coverage_95pct}), causing encoder weights to be almost orthogonal to this mean (typically near $90^\circ$; Appendix Figure~\ref{fig:mean_direction_divergence}). If $w_i$ were close to the mean input direction (say, within $40^\circ$), the feature would activate on nearly all inputs, violating sparsity. Thus encoder directions cannot reliably align with semantic features.

This encoder--decoder angular mismatch is natural and benign
(Appendix Figure~\ref{fig:angular_distances_aggregation}): the encoder $w_i$ specifies a decision
boundary that controls \emph{when} a feature fires, while the decoder $d_i$ specifies
the concept---the semantic direction it writes.

\noindent\textbf{Sparsity eliminates encoder redundancy.}
Moreover, sparsity constraints already regulate encoder-side redundancy. If two features have $w_i \approx w_j$, they fire on the same inputs with the same magnitude, so the SAE can merge them into a single feature with decoder $d_i + d_j$, achieving identical reconstruction while reducing the sparsity penalty. Decoder redundancy, however, is not directly penalized by the standard SAE objective
(reconstruction plus sparsity), and can therefore persist.

\noindent\textbf{Empirical validation of the proxy.}
We additionally verify that decoder cosine similarity tracks \emph{semantic} alignment: computing per-feature explanation embeddings, mean embedding similarity rises monotonically with decoder cosine similarity (pooled Spearman $r_s{=}0.42$, $p{<}10^{-10}$); details in Appendix~\ref{appendix:decoder-to-embedding}.

\section{Results}
\label{sec:results}
We evaluated 600+ SAEs across 7 SAE families (Appendix~\ref{appendix:experiment_details}; 28,000+ features sampled). Our findings reveal that \textbf{dominant features---those contributing most to model computation---are systematically less monosemantic and more redundant} (detailed benchmark results for all five AIU criteria in Appendix~\ref{appendix:benchmark_other_results}). In this section, for both monosemanticity and redundancy metrics, we report results across four dominance quantiles (Q1--Q4), where Q1 contains the most dominant features (highest mean dominance $\bar{\rho}_i$) and Q4 the least.

\subsection{Dominant Features Show Lower Monosemanticity}
\label{subsec:dame}

Across all SAEs evaluated with our DAME framework, top-quartile (Q1) features have
systematically lower monosemanticity than the rest of the dictionary (Q2--Q4). The
effect is statistically significant at the per-SAE level (Wilcoxon signed-rank test:
$p = 6.76 \times 10^{-8}$, Cohen's $d = -0.278$). We count an SAE as having a \emph{Q1 deficit} when its mean perfect ranking accuracy is lower for Q1 than for Q2--Q4; 65.2\% of SAEs individually
show a Q1 deficit. On average across all
evaluated SAEs (per-SAE mean), Q1 features achieve 61.8\% perfect ranking
accuracy compared to 64.1\% for Q2--Q4. This
inverse relationship between dominance and monosemanticity---where the most
computationally important features are the least monosemantic---directly contradicts
the expectation that dominant features should be monosemantic.

\begin{table}[t]
\centering
\caption{DAME results across SAE architectures. \textbf{Left:} per-architecture Q1--Q4 perfect ranking accuracy. \textbf{Right:} per-SAE Wilcoxon signed-rank test of Q1 vs.\ Q2--Q4; negative Cohen's $d$ indicates Q1 underperforms.}
\label{tab:dame_combined}
\begin{minipage}[t]{0.49\linewidth}
  \centering\footnotesize
  \begin{tabular}{lccccc}
\toprule
Architecture & Q1 & Q2 & Q3 & Q4 & Diff \\
\midrule
BatchTopK   & \textbf{61.5} & 65.2 & 67.1 & 69.9 & \textbf{-5.9} \\
Gated    & 60.8          & 57.8 & 58.9 & 60.6 & \phantom{-}1.7 \\
JumpReLU    & 58.7          & 58.0 & 56.9 & 58.0 & \phantom{-}1.1 \\
Matryoshka  & \textbf{61.6} & 64.5 & 65.8 & 66.5 & \textbf{-4.0} \\
PAnneal     & \textbf{67.1} & 67.4 & 71.0 & 72.3 & \textbf{-3.2} \\
Standard ReLU        & \textbf{60.5} & 60.8 & 61.9 & 62.1 & \textbf{-1.1} \\
TopK        & \textbf{59.2} & 64.4 & 64.0 & 64.6 & \textbf{-5.1} \\
\bottomrule
\end{tabular}

\end{minipage}\hfill
\begin{minipage}[t]{0.49\linewidth}
  \centering\footnotesize
  \begin{tabular}{lrr}
    \toprule
    Architecture & Wilcoxon $p$ & Cohen's $d$ \\
    \midrule
    BatchTopK  & $3.2{\times}10^{-7}$ & $-1.11$ \\
    TopK       & $3.5{\times}10^{-8}$ & $-0.78$ \\
    Matryoshka & $1.2{\times}10^{-3}$ & $-0.59$ \\
    PAnneal    & $0.05$               & $-0.44$ \\
    Standard ReLU & $9.2{\times}10^{-3}$ & $-0.22$ \\
    Gated      & n.s.                 & --- \\
    JumpReLU   & n.s.                 & --- \\
    \midrule
    \textbf{Overall} & $\mathbf{6.76{\times}10^{-8}}$ & $\mathbf{-0.278}$ \\
    \bottomrule
  \end{tabular}
\end{minipage}
\end{table}

\begin{figure}[t]
\centering
\includegraphics[width=\linewidth]{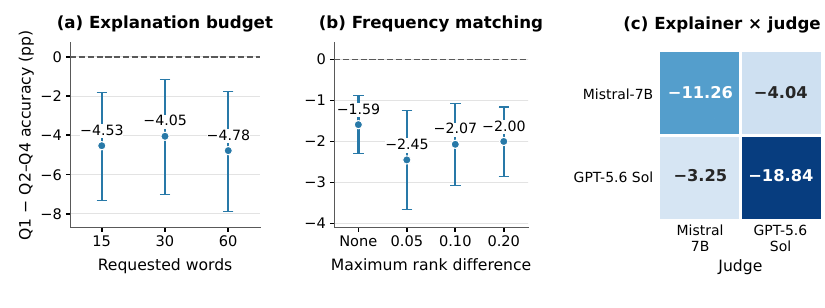}
\caption{Robustness of the DAME result. All values are mean Q1-minus-Q2--Q4 accuracy differences in percentage points; negative values indicate lower Q1 accuracy. \textbf{(a)} Longer explanation instructions on 28 SAEs, with 95\% hierarchical-bootstrap intervals. \textbf{(b)} Firing-frequency matching at three maximum rank differences (``None'': unmatched), with 95\% SAE-bootstrap intervals. \textbf{(c)} Six-SAE explainer-by-judge check; rows are explainers and columns are judges. Panels use different SAE sets; protocols and full results are in Appendices~\ref{appendix:length-budget}, \ref{appendix:frequency-matched}, and \ref{appendix:cross-judge}.}
\label{fig:dame_robustness}
\end{figure}

The Q1 deficit holds across most architectures
(Table~\ref{tab:dame_combined}, left) and is significant in 5 of 7
(Table~\ref{tab:dame_combined}, right); the n.s.\ result for Gated and JumpReLU
might only reflect high $L_0$ produced by the default sparsity hyperparameters
we adopted for these architectures (median $L_0$ in
Appendix~\ref{appendix:per_arch_l0}). Though the Q1 deficit varies across SAE families and
training configurations (Appendix~\ref{appendix:c3_dame}), the key point is
that it is \emph{broadly observed across the SAEs we evaluate} and warrants
attention from dominance-aware evaluation.

To assess the robustness of the observed Q1 deficit, we vary explanation budgets,
match features by firing frequency, and cross two explainers with two judges on
six SAEs. The mean Q1 deficit persists across these checks (Figure~\ref{fig:dame_robustness};
Appendices~\ref{appendix:length-budget}, \ref{appendix:frequency-matched}, and
\ref{appendix:cross-judge}). Separate evaluations extend the analysis to public
Qwen3 (8B--32B) and Gemma-2-9B SAEs (Appendix~\ref{appendix:larger-models}).

We further test the finding against known concepts without relying on an LLM
judge. Across 210 synthetic SAEs at widths 128--1024, Q1 features show lower
prevalence-corrected held-out support containment than Q2--Q4 overall (one-sided
SAE-level sign test, $p = 1.6 \times 10^{-12}$;
Appendix~\ref{appendix:support-containment}). This measures whether a feature
fires within a single concept without penalizing incomplete concept coverage.
A complementary correlation-gap analysis finds a significant Q1 deficit in
71 of 105 SAEs (Appendix~\ref{appendix:synthetic-validation}).

\subsection{Dominant Features Show More Redundancy}
\label{subsec:redundancy}

\begin{figure}[t]
\centering
\includegraphics[width=\linewidth]{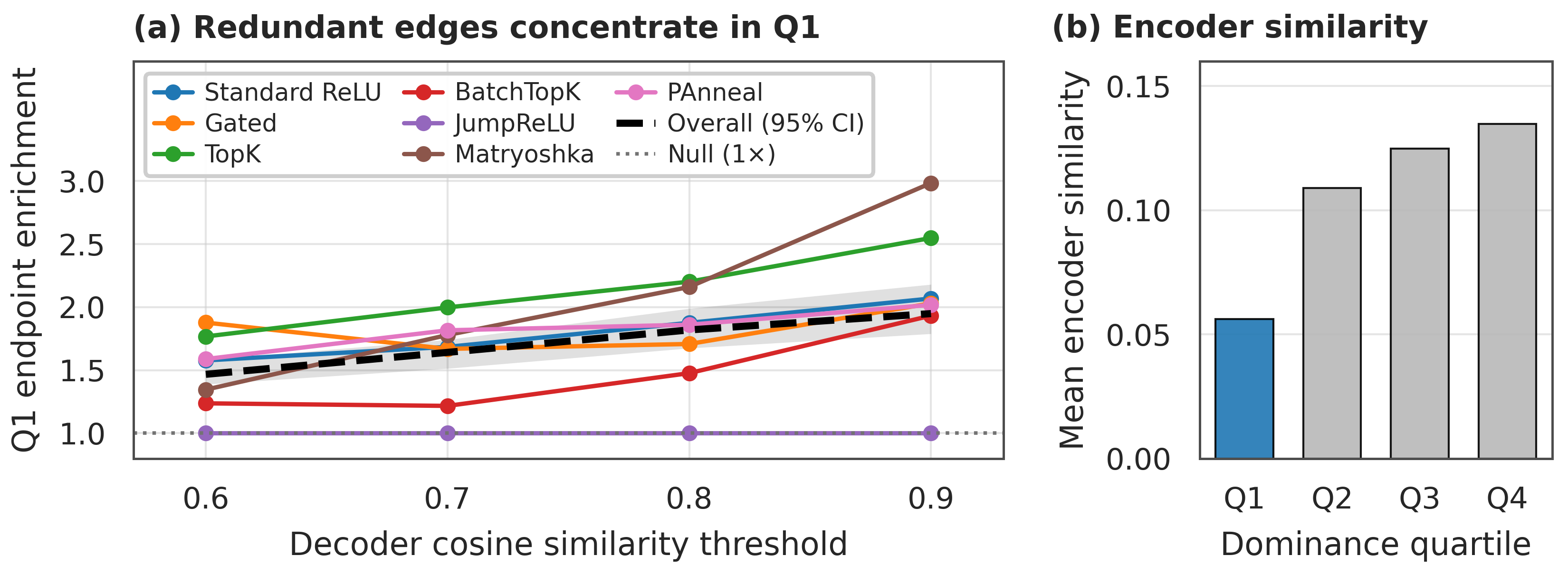}
\caption{Redundancy among dominant features. \textbf{(a)} Q1 endpoint enrichment among redundant edges as a function of the decoder cosine similarity threshold (full-dictionary dominance quartiles, symmetric active-only filter): median over SAEs for each architecture; the dashed line is the median over all SAEs with its 95\% SAE-bootstrap confidence interval and the dotted line is the $1\times$ null. \textbf{(b)} Mean encoder similarity over all feature pairs by dominance quartile.}
\label{fig:redundancy_sweep}
\end{figure}

\paragraph{Dominant features concentrate redundancy.}
We report redundancy at a primary decoder cosine threshold of 0.7 and test robustness across thresholds of 0.6--0.9 (Figure~\ref{fig:redundancy_sweep}a). We use full-dictionary dominance quartiles and remove pairs involving features that never fire in the evaluation sample, applying the same rule to all four groups. To account for unequal group sizes after filtering, we normalize each quartile's share of redundant-edge endpoints by its share of active features. This \emph{endpoint enrichment} equivalently compares the group's mean number of redundant connections per feature with the mean across all active features; Q1 enrichment above $1\times$ therefore implies a higher mean than Q2--Q4 combined. On the complete graph, including cross-quartile pairs, the median Q1 enrichment is $1.64\times$ (95\% SAE-bootstrap CI 1.51--1.74), with 71.5\% of SAEs above the $1\times$ null. Within-quartile-only results and per-architecture breakdowns are in Appendix~\ref{appendix:c1_redundancy}.
The most dominant features are especially affected: for Standard ReLU SAEs,
26.1\% of top-10 features participate in Q1 redundant pairs, with 4.7 pairs
forming within the top-10 alone (Appendix Table~\ref{tab:top10_redundancy}); 20.5\% of
top-100 features are involved in Q1 redundant pairs, with 296.3 pairs forming
within the top-100 (Appendix Table~\ref{tab:top100_redundancy}).

\paragraph{Redundant features differ in encoder direction.}
Q1 features exhibit the lowest encoder similarity (mean: 0.056) compared to
Q2--Q4 (0.109--0.135)
(Figure~\ref{fig:redundancy_sweep}b; Appendix Table~\ref{tab:encoder_similarity_by_quantile}). Critically, this pattern
persists even among redundant pairs: Q1 redundant pairs have encoder similarity
of 0.428 versus 0.661 for Q2--Q4 redundant pairs. Decoder-aligned pairs can therefore have dissimilar encoders; whether their firing supports are complementary must be assessed from activation data.

\paragraph{The feature sharding failure mode.}
We use \textbf{feature sharding} to describe a pairwise representational pattern in which features share (nearly) the same decoder direction $d_i \approx d_j$, have dissimilar encoders $w_i \not\approx w_j$, and fire on complementary regions of activation space while writing the same semantic direction. Feature sharding can preserve reconstruction fidelity and per-input sparsity while reducing dead features, and may therefore be favored by objectives with explicit dead-feature penalties. An SAE can thus appear healthy under common diagnostics even as its effective semantic dictionary collapses to far fewer directions.

Concretely, consider a semantic direction encoding the concept of \emph{color}. Rather than learning a single feature that activates whenever color is relevant, an SAE may shard this direction into multiple features whose encoders gate activation on additional, unrelated contextual factors (e.g., gendered language, syntactic role, or discourse position), effectively learning features of the form \emph{color} $\wedge$ \emph{context}. During attribution, whenever color contributes to the model's computation, exactly one such shard fires---maintaining sparsity---so attribution is tied to the shard's gating context rather than to the underlying concept of color itself.

\paragraph{A functional sharding example.}
In a TopK-65k SAE trained on Pythia-160M, feature 41902 activates on definite-article contexts (\emph{the}/\emph{The}), whereas feature 46571 activates on possessive/genitive-determiner contexts (\emph{its}, \emph{his}, \emph{their}, \emph{whose}, and possessive \emph{'s}). Their encoder and decoder cosine similarities are 0.2066 and 0.7405, respectively. Their natural supports are largely separated under the SAE's exact TopK gate: 316 versus 57 firings, only 7 co-firings, and Jaccard 0.019. On an independent fixed target-prompt family, we injected equal-$L_2$ decoder directions only at positions where both native TopK gates were naturally inactive (18/20 prompts); protocol in Appendix~\ref{appendix:functional-sharding-protocol}. Both writes increased the fixed target token \emph{aim} in the actual downstream forward pass: mean logit changes were $+0.556$ and $+0.611$, with 15/18 and 16/18 positive prompt-level effects. The induced effect vectors had cosine 0.943 (relative to support-size- and decoder-norm-matched low-decoder-similarity controls, the mean candidate-pair advantage was $+0.472$ logit; bootstrap 95\% CI $[+0.347, +0.597]$). Together, the separated natural supports and aligned intervention effects provide a functional sharding case: the two features specialize to different determiner contexts while producing similar downstream decoder writes.

Sharding differs from \emph{feature absorption}~\citep{chanin2024absorption}, in which a general-concept feature fails to activate on more specific cases while other features take over its role; the two can coexist. In terms of firing positions, sharding corresponds to decoder-aligned features that fire at largely separate positions, whereas absorption is compatible with one feature's firing positions being largely contained within the other's. In a broader analysis of decoder-aligned pairs, 21.5\% fire at largely separate positions and in 15.3\% one feature's firing positions are largely contained within the other's (equal-weight means of within-SAE percentages; Appendix~\ref{appendix:activation-support-audit}).

\section{Discussion}
\label{sec:discussion}

Our findings reveal three recurring correlated phenomena: (1) higher sparsity often correlates
with higher dominance concentration (Section~\ref{sec:local-to-global}),
(2) dominant features show lower monosemanticity (Section~\ref{subsec:dame}),
and (3) dominant features exhibit more redundancy
(Section~\ref{subsec:redundancy}). The case study also illustrates \emph{feature sharding},
in which features fire in distinct contexts but produce similar decoder writes.
Together, these suggest sparsity constraints
in SAE training may be satisfied in ways that do not improve interpretability:
dominant features either remain polysemantic or are duplicated across features with similar decoder directions—both allow
SAEs to meet sparsity and dead features penalties while leaving the most computationally important
structure entangled, raising concerns for SAE-based attribution and circuit tracing.

\paragraph{Recommendations for practitioners and researchers.}
For these reasons, before SAE features are deployed for attribution, circuit
tracing, or intervention, we recommend dominance-stratified evaluation as a
standard check: (1) monosemanticity across dominance quantiles, and (2) decoder
similarity among high-dominance features. These failure modes also warrant
attention from SAE researchers as targets for improved training objectives
and architectures that reduce polysemanticity and redundancy in high-dominance
features.

Limitations of the present work are discussed in Appendix~\ref{appendix:limitations}.

\clearpage
\begin{ack}
This work was partially supported by NSF IIS \#2146091, IIS \#2345235, and NIH NIA \#R01AG081951.

\subsection*{Competing Interests}
The authors declare no competing interests.
\end{ack}

\newpage
\bibliographystyle{plainnat}
\bibliography{refs}

@article{elhage2022toy,
  title={Toy Models of Superposition},
  author={Elhage, Nelson and Hume, Tristan and Olsson, Catherine and Schiefer, Nicholas and Henighan, Tom and Kravec, Shauna and Hatfield-Dodds, Zac and Lasenby, Robert and Drain, Dawn and Chen, Carol and Grosse, Roger and McCandlish, Sam and Kaplan, Jared and Amodei, Dario and Wattenberg, Martin and Olah, Christopher},
  journal={Transformer Circuits Thread},
  year={2022},
  eprint={2209.10652},
  archivePrefix={arXiv},
  url={https://transformer-circuits.pub/2022/toy_model/index.html}
}

@article{scherlis2022polysemanticity,
  title={Polysemanticity and Capacity in Neural Networks},
  author={Scherlis, Adam and Sachan, Kshitij and Jermyn, Adam S. and Benton, Joe and Shlegeris, Buck},
  journal={arXiv preprint arXiv:2210.01892},
  year={2022}
}

@misc{sharkey2022interim,
  title={[Interim research report] Taking features out of superposition with sparse autoencoders},
  author={Sharkey, Lee and Braun, Dan and Millidge, Beren},
  year={2022},
  howpublished={AI Alignment Forum},
  url={https://www.alignmentforum.org/posts/z6QQJbtpkEAX3Aojj/interim-research-report-taking-features-out-of-superposition}
}

@article{bricken2023towards,
  title={Towards Monosemanticity: Decomposing Language Models With Dictionary Learning},
  author={Bricken, Trenton and Templeton, Adly and Batson, Joshua and Chen, Brian and Jermyn, Adam and Conerly, Tom and Turner, Nicholas L and Anil, Cem and Denison, Carson and Askell, Amanda and Lasenby, Robert and Wu, Yifan and Kravec, Shauna and Schiefer, Nicholas and Maxwell, Tim and Joseph, Nicholas and Hatfield-Dodds, Zac and Tamkin, Alex and Nguyen, Karina and McLean, Brayden and Burke, Josiah E and Hume, Tristan and Carter, Shan and Henighan, Tom and Olah, Chris},
  journal={Transformer Circuits Thread},
  year={2023},
  url={https://transformer-circuits.pub/2023/monosemantic-features/index.html}
}

@inproceedings{cunningham2024sparse,
  title={Sparse Autoencoders Find Highly Interpretable Features in Language Models},
  author={Cunningham, Hoagy and Ewart, Aidan and Riggs, Logan and Huben, Robert and Sharkey, Lee},
  booktitle={Proceedings of the Twelfth International Conference on Learning Representations (ICLR)},
  year={2024},
  eprint={2309.08600},
  archivePrefix={arXiv},
  url={https://arxiv.org/abs/2309.08600}
}

@article{templeton2024scaling,
  title={Scaling Monosemanticity: Extracting Interpretable Features from Claude 3 Sonnet},
  author={Templeton, Adly and Conerly, Tom and Marcus, Jonathan and Lindsey, Jack and Bricken, Trenton and Chen, Brian and Pearce, Adam and Citro, Craig and Ameisen, Emmanuel and Jones, Andy and Cunningham, Hoagy and Turner, Nicholas L and McDougall, Callum and MacDiarmid, Monte and Tamkin, Alex and Durmus, Esin and Hume, Tristan and Mosconi, Francesco and Freeman, C. Daniel and Sumers, Theodore R. and Rees, Edward and Batson, Joshua and Jermyn, Adam and Carter, Shan and Olah, Chris and Henighan, Tom},
  journal={Transformer Circuits Thread},
  year={2024},
  url={https://transformer-circuits.pub/2024/scaling-monosemanticity/index.html}
}

@inproceedings{gao2024scaling,
  title={Scaling and Evaluating Sparse Autoencoders},
  author={Gao, Leo and Dupr{\'e} la Tour, Tom and Tillman, Henk and Goh, Gabriel and Troll, Rajan and Radford, Alec and Sutskever, Ilya and Leike, Jan and Wu, Jeffrey},
  booktitle={Proceedings of the Thirteenth International Conference on Learning Representations (ICLR)},
  year={2025},
  eprint={2406.04093},
  archivePrefix={arXiv},
  url={https://openreview.net/forum?id=tcsZt9ZNKD},
  note={Oral presentation}
}

@article{rajamanoharan2024jumprelu,
  title={Jumping Ahead: Improving Reconstruction Fidelity with JumpReLU Sparse Autoencoders},
  author={Rajamanoharan, Senthooran and Lieberum, Tom and Sonnerat, Nicolas and Conmy, Arthur and Varma, Vikrant and Kram{\'a}r, J{\'a}nos and Nanda, Neel},
  journal={arXiv preprint arXiv:2407.14435},
  year={2024},
  url={https://arxiv.org/abs/2407.14435}
}

@inproceedings{bussmann2024batchtopk,
  title={{BatchTopK} Sparse Autoencoders},
  author={Bussmann, Bart and Leask, Patrick and Nanda, Neel},
  booktitle={NeurIPS 2024 Workshop on Scientific Methods for Understanding Deep Learning (SciForDL)},
  year={2024},
  eprint={2412.06410},
  archivePrefix={arXiv},
  url={https://openreview.net/forum?id=d4dpOCqybL}
}

@article{bussmann2025matryoshka,
  title={Learning Multi-Level Features with Matryoshka Sparse Autoencoders},
  author={Bussmann, Bart and Nabeshima, Noa and Karvonen, Adam and Nanda, Neel},
  journal={arXiv preprint arXiv:2503.17547},
  year={2025},
  url={https://arxiv.org/abs/2503.17547}
}

@inproceedings{rajamanoharan2024gated,
  title={Improving Sparse Decomposition of Language Model Activations with Gated Sparse Autoencoders},
  author={Rajamanoharan, Senthooran and Conmy, Arthur and Smith, Lewis and Lieberum, Tom and Varma, Vikrant and Kram{\'a}r, J{\'a}nos and Shah, Rohin and Nanda, Neel},
  booktitle={Advances in Neural Information Processing Systems},
  volume={37},
  year={2024},
  eprint={2404.16014},
  archivePrefix={arXiv},
  url={https://neurips.cc/virtual/2024/poster/92961}
}

@misc{wright2024addressing,
  title={Addressing Feature Suppression in SAEs},
  author={Wright, Benjamin and Sharkey, Lee},
  year={2024},
  howpublished={AI Alignment Forum},
  url={https://www.alignmentforum.org/posts/3JuSjTZyMzaSeTxKk/addressing-feature-suppression-in-saes}
}

@article{ayonrinde2024adaptive,
  title={Adaptive Sparse Allocation with Mutual Choice \& Feature Choice Sparse Autoencoders},
  author={Ayonrinde, Kola},
  journal={arXiv preprint arXiv:2411.02124},
  year={2024},
  url={https://arxiv.org/abs/2411.02124}
}

@inproceedings{karvonen2025saebench,
  title={{SAEBench}: A Comprehensive Benchmark for Sparse Autoencoders in Language Model Interpretability},
  author={Karvonen, Adam and Rager, Can and Lin, Johnny and Tigges, Curt and Bloom, Joseph and Chanin, David and Lau, Yeu-Tong and Farrell, Eoin and McDougall, Callum and Ayonrinde, Kola and Till, Demian and Wearden, Matthew and Conmy, Arthur and Marks, Samuel and Nanda, Neel},
  booktitle={Proceedings of the 42nd International Conference on Machine Learning},
  series={Proceedings of Machine Learning Research},
  volume={267},
  pages={29223--29264},
  year={2025},
  publisher={PMLR},
  url={https://arxiv.org/abs/2503.09532}
}

@article{wu2025axbench,
  title={AxBench: Steering LLMs? Even Simple Baselines Outperform Sparse Autoencoders},
  author={Wu, Zhengxuan and Arora, Aryaman and Geiger, Atticus and Wang, Zheng and Huang, Jing and Jurafsky, Dan and Manning, Christopher D. and Potts, Christopher},
  journal={arXiv preprint arXiv:2501.17148},
  year={2025},
  url={https://arxiv.org/abs/2501.17148}
}

@inproceedings{chanin2024absorption,
  title={A is for Absorption: Studying Feature Splitting and Absorption in Sparse Autoencoders},
  author={Chanin, David and Wilken-Smith, James and Dulka, Tom{\'a}{\v{s}} and Bhatnagar, Hardik and Golechha, Satvik and Bloom, Joseph},
  booktitle={Advances in Neural Information Processing Systems},
  volume={38},
  year={2025},
  eprint={2409.14507},
  archivePrefix={arXiv},
  url={https://arxiv.org/abs/2409.14507},
  note={Oral presentation}
}

@misc{bills2023language,
  title={Language Models Can Explain Neurons in Language Models},
  author={Bills, Steven and Cammarata, Nick and Mossing, Dan and Tillman, Henk and Gao, Leo and Goh, Gabriel and Sutskever, Ilya and Leike, Jan and Wu, Jeff and Saunders, William},
  year={2023},
  howpublished={OpenAI Blog},
  url={https://openaipublic.blob.core.windows.net/neuron-explainer/paper/index.html}
}

@inproceedings{kantamneni2025useful,
  title={Are Sparse Autoencoders Useful? A Case Study in Sparse Probing},
  author={Kantamneni, Subhash and Engels, Joshua and Rajamanoharan, Senthooran and Tegmark, Max and Nanda, Neel},
  booktitle={Proceedings of the 42nd International Conference on Machine Learning},
  series={Proceedings of Machine Learning Research},
  volume={267},
  pages={29018--29049},
  year={2025},
  publisher={PMLR},
  url={https://arxiv.org/abs/2502.16681}
}

@article{makelov2024towards,
  title={Towards Principled Evaluations of Sparse Autoencoders for Interpretability and Control},
  author={Makelov, Aleksandar and Lange, Georg and Nanda, Neel},
  journal={arXiv preprint arXiv:2405.08366},
  year={2024},
  url={https://arxiv.org/abs/2405.08366}
}

@article{karvonen2024evaluating,
  title={Evaluating Sparse Autoencoders on Targeted Concept Erasure Tasks},
  author={Karvonen, Adam and Rager, Can and Marks, Samuel and Nanda, Neel},
  journal={arXiv preprint arXiv:2411.18895},
  year={2024},
  url={https://arxiv.org/abs/2411.18895}
}

@article{gurnee2023finding,
  title={Finding Neurons in a Haystack: Case Studies with Sparse Probing},
  author={Gurnee, Wes and Nanda, Neel and Pauly, Matthew and Harvey, Katherine and Troitskii, Dmitrii and Bertsimas, Dimitris},
  journal={Transactions on Machine Learning Research},
  year={2023},
  eprint={2305.01610},
  archivePrefix={arXiv},
  url={https://openreview.net/forum?id=JYs1R9IMJr}
}

@inproceedings{huang2024ravel,
  title={{RAVEL}: Evaluating Interpretability Methods on Disentangling Language Model Representations},
  author={Huang, Jing and Wu, Zhengxuan and Potts, Christopher and Geva, Mor and Geiger, Atticus},
  booktitle={Proceedings of the 62nd Annual Meeting of the Association for Computational Linguistics (Volume 1: Long Papers)},
  pages={8669--8687},
  year={2024},
  publisher={Association for Computational Linguistics},
  eprint={2402.17700},
  archivePrefix={arXiv},
  url={https://aclanthology.org/2024.acl-long.470/}
}

@article{farrell2024evaluating,
  title={Applying Sparse Autoencoders to Unlearn Knowledge in Language Models},
  author={Farrell, Eoin and Lau, Yeu-Tong and Conmy, Arthur},
  journal={arXiv preprint arXiv:2410.19278},
  year={2024},
  url={https://arxiv.org/abs/2410.19278}
}

@article{paulo2024automatically,
  title={Automatically Interpreting Millions of Features in Large Language Models},
  author={Paulo, Gon{\c{c}}alo and Mallen, Alex and Juang, Caden and Belrose, Nora},
  journal={arXiv preprint arXiv:2410.13928},
  year={2024},
  url={https://arxiv.org/abs/2410.13928}
}

@inproceedings{leask2025canonical,
  title={Sparse Autoencoders Do Not Find Canonical Units of Analysis},
  author={Leask, Patrick and Bussmann, Bart and Pearce, Michael and Bloom, Joseph and Tigges, Curt and Al Moubayed, Noura and Sharkey, Lee and Nanda, Neel},
  booktitle={Proceedings of the Thirteenth International Conference on Learning Representations (ICLR)},
  year={2025},
  eprint={2502.04878},
  archivePrefix={arXiv},
  url={https://arxiv.org/abs/2502.04878}
}

@article{paulo2025sparse,
  title={Sparse Autoencoders Trained on the Same Data Learn Different Features},
  author={Paulo, Gon{\c{c}}alo and Belrose, Nora},
  journal={arXiv preprint arXiv:2501.16615},
  year={2025},
  url={https://arxiv.org/abs/2501.16615}
}

@misc{bussmann2024showing,
  title={Showing {SAE} Latents Are Not Atomic Using Meta-{SAE}s},
  author={Bussmann, Bart and Pearce, Michael and Leask, Patrick and Bloom, Joseph and Sharkey, Lee and Nanda, Neel},
  year={2024},
  howpublished={LessWrong},
  url={https://www.lesswrong.com/posts/TMAmHh4DdMr4nCSr5/showing-sae-latents-are-not-atomic-using-meta-saes}
}

@inproceedings{peng2025use,
  title={Position: Use Sparse Autoencoders to Discover Unknowns},
  author={Peng, Kenny and Movva, Rajiv and Kleinberg, Jon and Pierson, Emma and Garg, Nikhil},
  booktitle={Proceedings of the 43rd International Conference on Machine Learning},
  year={2026},
  eprint={2506.23845},
  archivePrefix={arXiv},
  url={https://arxiv.org/abs/2506.23845}
}

@article{ayonrinde2024interpretability,
  title={Interpretability as Compression: Reconsidering SAE Explanations of Neural Activations with MDL-SAEs},
  author={Ayonrinde, Kola and Pearce, Michael T. and Sharkey, Lee},
  journal={arXiv preprint arXiv:2410.11179},
  year={2024},
  url={https://arxiv.org/abs/2410.11179}
}

@inproceedings{movva2025hypothesis,
  title={Sparse Autoencoders for Hypothesis Generation},
  author={Movva, Rajiv and Peng, Kenny and Garg, Nikhil and Kleinberg, Jon and Pierson, Emma},
  booktitle={Proceedings of the 42nd International Conference on Machine Learning},
  series={Proceedings of Machine Learning Research},
  volume={267},
  pages={44997--45023},
  year={2025},
  publisher={PMLR},
  url={https://proceedings.mlr.press/v267/movva25a.html}
}

@article{oneill2024sparse,
  title={Sparse Autoencoders Enable Scalable and Reliable Circuit Identification in Language Models},
  author={O'Neill, Charles and Bui, Thang},
  journal={arXiv preprint arXiv:2405.12522},
  year={2024},
  url={https://arxiv.org/abs/2405.12522}
}

@inproceedings{marks2024circuit,
  title={Sparse Feature Circuits: Discovering and Editing Interpretable Causal Graphs in Language Models},
  author={Marks, Samuel and Rager, Can and Michaud, Eric J. and Belinkov, Yonatan and Bau, David and Mueller, Aaron},
  booktitle={Proceedings of the Thirteenth International Conference on Learning Representations (ICLR)},
  year={2025},
  eprint={2403.19647},
  archivePrefix={arXiv},
  url={https://openreview.net/forum?id=I4e82CIDxv},
  note={Oral presentation}
}

@article{anthropic2025circuittracing,
  title={Circuit Tracing: Revealing Computational Graphs in Language Models},
  author={Ameisen, Emmanuel and Lindsey, Jack and Pearce, Adam and Gurnee, Wes and Turner, Nicholas L. and Chen, Brian and Citro, Craig and Abrahams, David and Carter, Shan and Hosmer, Basil and Marcus, Jonathan and Sklar, Michael and Templeton, Adly and Bricken, Trenton and McDougall, Callum and Cunningham, Hoagy and Henighan, Thomas and Jermyn, Adam and Jones, Andy and Persic, Andrew and Qi, Zhenyi and Thompson, T. Ben and Zimmerman, Sam and Rivoire, Kelley and Conerly, Thomas and Olah, Chris and Batson, Joshua},
  journal={Transformer Circuits Thread},
  year={2025},
  url={https://transformer-circuits.pub/2025/attribution-graphs/methods.html}
}

@inproceedings{dunefsky2024transcoders,
  title={Transcoders Find Interpretable {LLM} Feature Circuits},
  author={Dunefsky, Jacob and Chlenski, Philippe and Nanda, Neel},
  booktitle={Advances in Neural Information Processing Systems},
  year={2024},
  eprint={2406.11944},
  archivePrefix={arXiv},
  url={https://arxiv.org/abs/2406.11944}
}

@inproceedings{braun2024e2e,
  title={Identifying Functionally Important Features with End-to-End Sparse Dictionary Learning},
  author={Braun, Dan and Taylor, Jordan and Goldowsky-Dill, Nicholas and Sharkey, Lee},
  booktitle={Advances in Neural Information Processing Systems},
  year={2024},
  eprint={2405.12241},
  archivePrefix={arXiv},
  url={https://arxiv.org/abs/2405.12241}
}

@article{paulo2025transcoders,
  title={Transcoders Beat Sparse Autoencoders for Interpretability},
  author={Paulo, Gon{\c{c}}alo and Shabalin, Stepan and Belrose, Nora},
  journal={arXiv preprint arXiv:2501.18823},
  year={2025},
  url={https://arxiv.org/abs/2501.18823}
}

@article{lindsey2024crosscoders,
  title={Sparse Crosscoders for Cross-Layer Features and Model Diffing},
  author={Lindsey, Jack and Templeton, Adly and Marcus, Jonathan and Conerly, Thomas and Batson, Joshua and Olah, Christopher},
  journal={Transformer Circuits Thread},
  year={2024},
  url={https://transformer-circuits.pub/2024/crosscoders/index.html}
}

@misc{bloom2024saelens,
  title={{SAELens}},
  author={Bloom, Joseph and Tigges, Curt and Duong, Anthony and Chanin, David},
  year={2024},
  howpublished={\url{https://github.com/jbloomAus/SAELens}}
}

@misc{smith2024negative,
  title={Negative Results for {SAE}s On Downstream Tasks and Deprioritising {SAE} Research ({GDM} Mech Interp Team Progress Update \#2)},
  author={Smith, Lewis and Rajamanoharan, Senthooran and Conmy, Arthur and McDougall, Callum and Lieberum, Tom and Kram{\'a}r, J{\'a}nos and Shah, Rohin and Nanda, Neel},
  year={2025},
  howpublished={AI Alignment Forum},
  url={https://www.alignmentforum.org/posts/4uXCAJNuPKtKBsi28}
}

@article{sharkey2025open,
  title={Open Problems in Mechanistic Interpretability},
  author={Sharkey, Lee and Chughtai, Bilal and Batson, Joshua and Lindsey, Jack and Wu, Jeff and Bushnaq, Lucius and Goldowsky-Dill, Nicholas and Heimersheim, Stefan and Ortega, Alejandro and Bloom, Joseph and Biderman, Stella and Garriga-Alonso, Adria and Conmy, Arthur and Nanda, Neel and Rumbelow, Jessica and Wattenberg, Martin and Schoots, Nandi and Miller, Joseph and Michaud, Eric J. and Casper, Stephen and Tegmark, Max and Saunders, William and Bau, David and Todd, Eric and Geiger, Atticus and Geva, Mor and Hoogland, Jesse and Murfet, Daniel and McGrath, Tom},
  journal={arXiv preprint arXiv:2501.16496},
  year={2025},
  url={https://arxiv.org/abs/2501.16496}
}

@article{lindsey2025attribution,
  title={On the Biology of a Large Language Model},
  author={Lindsey, Jack and Gurnee, Wes and Ameisen, Emmanuel and Chen, Brian and Pearce, Adam and Turner, Nicholas L. and Citro, Craig and Abrahams, David and Carter, Shan and Hosmer, Basil and Marcus, Jonathan and Sklar, Michael and Templeton, Adly and Bricken, Trenton and McDougall, Callum and Cunningham, Hoagy and Henighan, Thomas and Jermyn, Adam and Jones, Andy and Persic, Andrew and Qi, Zhenyi and Thompson, T. Ben and Zimmerman, Sam and Rivoire, Kelley and Conerly, Thomas and Olah, Chris and Batson, Joshua},
  journal={Transformer Circuits Thread},
  year={2025},
  url={https://transformer-circuits.pub/2025/attribution-graphs/biology.html}
}

@inproceedings{karvonen2024board,
  title={Measuring Progress in Dictionary Learning for Language Model Interpretability with Board Game Models},
  author={Karvonen, Adam and Wright, Benjamin and Rager, Can and Angell, Rico and Brinkmann, Jannik and Smith, Logan and Verdun, Claudio Mayrink and Bau, David and Marks, Samuel},
  booktitle={Advances in Neural Information Processing Systems},
  volume={37},
  year={2024},
  eprint={2408.00113},
  archivePrefix={arXiv},
  url={https://arxiv.org/abs/2408.00113}
}

@article{venhoff2024sage,
  title={{SAGE}: Scalable Ground Truth Evaluations for Large Sparse Autoencoders},
  author={Venhoff, Constantin and Calinescu, Anisoara and Torr, Philip and Schroeder de Witt, Christian},
  journal={arXiv preprint arXiv:2410.07456},
  year={2024},
  url={https://arxiv.org/abs/2410.07456}
}

@article{radford2019language,
  title={Language Models are Unsupervised Multitask Learners},
  author={Radford, Alec and Wu, Jeffrey and Child, Rewon and Luan, David and Amodei, Dario and Sutskever, Ilya},
  journal={OpenAI Blog},
  volume={1},
  number={8},
  pages={9},
  year={2019}
}

@misc{eldan2023tinystories,
  title={TinyStories: How Small Can Language Models Be and Still Speak Coherent English?},
  author={Ronen Eldan and Yuanzhi Li}, 
  year={2023},
  eprint={2305.07759},
  archivePrefix={arXiv},
  primaryClass={cs.CL},
  doi={10.48550/arXiv.2305.07759}
}

@misc{zhang2022opt,
  title={OPT: Open Pre-trained Transformer Language Models},
  author={Susan Zhang and Stephen Roller and Naman Goyal and Mikel Artetxe and Moya Chen and Shuohui Chen and Christopher Dewan and Mona Diab and Xian Li and Xi Victoria Lin and Todor Mihaylov and Myle Ott and Sam Shleifer and Kurt Shuster and Daniel Simig and Punit Singh Koura and Anjali Sridhar and Tianlu Wang and Luke Zettlemoyer},
  year={2022},
  eprint={2205.01068},
  archivePrefix={arXiv},
  primaryClass={cs.CL}
}

@inproceedings{biderman2023pythia,
  title={Pythia: A Suite for Analyzing Large Language Models Across Training and Scaling},
  author={Biderman, Stella and Schoelkopf, Hailey and Anthony, Quentin and Bradley, Herbie and O'Brien, Kyle and Hallahan, Eric and Khan, Mohammad Aflah and Purohit, Shivanshu and Prashanth, USVSN Sai and Raff, Edward and Skowron, Aviya and Sutawika, Lintang and van der Wal, Oskar},
  booktitle={Proceedings of the 40th International Conference on Machine Learning},
  pages={2397--2430},
  year={2023},
  volume={202},
  series={Proceedings of Machine Learning Research},
  publisher={PMLR},
  eprint={2304.01373},
  archivePrefix={arXiv},
  url={https://proceedings.mlr.press/v202/biderman23a.html}
}

@article{gemma2team2024gemma2,
  title={Gemma 2: Improving Open Language Models at a Practical Size},
  author={{Gemma Team}},
  journal={arXiv preprint arXiv:2408.00118},
  year={2024},
  url={https://arxiv.org/abs/2408.00118}
}

\newpage
\appendix
\section*{Appendix}
This appendix provides comprehensive supplementary material supporting the main paper. The appendices are organized as follows:

\begin{itemize}
    \item \textbf{Appendix~\ref{appendix:extended-related-work}: Extended Related Work} -- Full discussion of the three threads framing this work: SAE architectures, evaluation/benchmarking, and feature-set desiderata.

    \item \textbf{Appendix~\ref{appendix:limitations}: Limitations} -- Discussion of limitations of the present work: model scale, proxy nature of the monosemanticity judge, conditional ground-truth concept sets, and hyperparameter transfer.

    \item \textbf{Appendix~\ref{appendix:experiment_details}: Experiment Details} -- Comprehensive details on our benchmarking experiments, including SAE families evaluated, configurations, and experimental setup.

    \item \textbf{Appendix~\ref{appendix:other_metrics}: Proxy Metrics for the Five AIU Criteria} -- Detailed implementation and methodology for all proxy metrics used to evaluate the five necessary criteria for atomic interpretable unit bases.

    \item \textbf{Appendix~\ref{appendix:previous-mono-measurement}: Automated Interpretability Scoring: Prior Work} -- Overview of prior approaches to measuring monosemanticity, including \cite{bills2023language} and \cite{paulo2024automatically}.

    \item \textbf{Appendix~\ref{appendix:benchmark_other_results}: Benchmarking Results} -- Comprehensive benchmarking results across all evaluated SAEs.

    \item \textbf{Appendix~\ref{sec:sae_feature_firing_behavior}: SAE Feature Activation Behavior and Dominance Imbalance} -- Analysis of firing frequency imbalance, magnitude distribution imbalance, per-input dominance, and their relationships.

    \item \textbf{Appendix~\ref{appendix:dominant_redundant_explanations}: Dominant Features: Explanation Analysis} -- Detailed analysis of the explanations of dominant features identified in our evaluation.

    \item \textbf{Appendix~\ref{appendix:decoder-sim}: SAE Geometry: Mean-Direction Clustering and Encoder--Decoder Mismatch} -- Theoretical and geometric justification for enforcing non-redundancy on decoder directions rather than encoder weights.

    \item \textbf{Appendix~\ref{appendix:additional-validation}: Additional Validation and Robustness Analyses} -- Controlled comparison of per-feature scoring rules; DAME and binary AutoInterp on the same features; explainer and judge sensitivity; explanation length budget; AND/OR alignment and support containment on the synthetic benchmark; frequency-matched control; sharding-compatible pairs at low $L_0$; the functional sharding case protocol; larger-model DAME results; and activation-support geometry with a blinded context audit.

    \item \textbf{Appendix~\ref{appendix:notation}: Notation and Terminology} -- Symbols and terms used throughout the paper.
\end{itemize}

\section{Extended Related Work}
\label{appendix:extended-related-work}

This appendix discusses related work along the three threads that frame our
contribution: (i) sparse autoencoders for mechanistic interpretability,
(ii) SAE evaluation metrics and benchmarking, and (iii) feature-set
desiderata.

\subsection{Sparse Autoencoders for Mechanistic Interpretability}
Sparse autoencoders address superposition \citep{elhage2022toy} by learning overcomplete sparse dictionaries that isolate monosemantic features \citep{sharkey2022interim, cunningham2024sparse, bricken2023towards, templeton2024scaling}. Numerous architectural variants have improved reconstruction fidelity through different sparsity mechanisms \citep{gao2024scaling, rajamanoharan2024jumprelu, rajamanoharan2024gated, bussmann2024batchtopk, karvonen2024board, ayonrinde2024adaptive, bussmann2025matryoshka}, while alternative approaches explore different training objectives such as transcoders \citep{paulo2025transcoders, dunefsky2024transcoders}, crosscoders \citep{lindsey2024crosscoders}, and end-to-end sparse dictionary learning \citep{braun2024e2e}.

\subsection{SAE Evaluation Metrics and Benchmarking}

SAE evaluation initially focused on the foundational goal of monosemanticity. \citet{bills2023language} introduced automated interpretability scoring, using language models to assess whether individual features respond to coherent, singular concepts based on their activation patterns across input sequences.

Basic sanity checks measured necessary conditions for SAE quality: \textbf{reconstruction fidelity (MSE, loss recovered)} \citep{bricken2023towards, cunningham2024sparse}---if an SAE cannot faithfully reconstruct activations, it cannot be a good decomposition---and \textbf{sparsity (L0 norm)} \citep{bricken2023towards, cunningham2024sparse}---high per-input activation counts can make an attribution difficult to inspect even when individual features are monosemantic. Additionally, \textbf{dead feature rates} \citep{bricken2023towards, templeton2024scaling} track features that rarely activate, indicating wasted capacity. Multiple early works \citep{bricken2023towards, cunningham2024sparse, bills2023language} extended evaluation to measure whether SAEs \textbf{preserve model performance} when their reconstructions replace original activations, testing reconstruction fidelity via downstream loss (cross-entropy, KL divergence). \citet{karvonen2025saebench} systematized evaluation with SAEBench, a comprehensive benchmark spanning reconstruction, interpretability (automated interpretability scoring), disentanglement, and concept detection metrics \citep{chanin2024absorption}.

\subsection{Feature Set Desiderata}
The lack of ground-truth features for comparison \citep{makelov2024towards} led researchers to develop supervised evaluation methods using tasks with known feature structure \citep{gurnee2023finding, gao2024scaling, huang2024ravel, marks2024circuit, farrell2024evaluating, karvonen2024evaluating, karvonen2024board, venhoff2024sage, karvonen2025saebench}. However, numerous works found that SAE features underperform compared to supervised methods on these supervised tasks \citep{kantamneni2025useful, wu2025axbench, smith2024negative}. \citet{peng2025use} reframed these negative results by distinguishing SAE objectives: even if SAEs underperform on pre-specified supervised tasks, they excel at \emph{discovering unknown concepts} \citep{movva2025hypothesis} and enabling circuit discovery \citep{marks2024circuit, oneill2024sparse, anthropic2025circuittracing, lindsey2025attribution, lindsey2024crosscoders}.

To evaluate such learned unknown feature sets, some researchers addressed pathological behaviors including feature suppression \citep{wright2024addressing}, feature absorption \citep{chanin2024absorption}, feature splitting \citep{bricken2023towards}, and redundancy \citep{karvonen2025saebench}, while others raised stability concerns---\citet{paulo2025sparse} found only 30\% feature sharing across random seeds. \citet{leask2025canonical} examined whether SAEs find ``canonical units of analysis,'' defining such a set as simultaneously \emph{unique} (no variations across runs), \emph{complete} (encompassing all necessary features), and \emph{atomic} (indivisible), finding evidence against all three through SAE stitching and meta-SAEs.

While these properties and failure modes have been proposed and discussed, they lack derivation from a principled objective specifying what function SAE features should serve \citep{sharkey2025open}---a gap our framework fills.

\section{Limitations}
\label{appendix:limitations}

We highlight several limitations of the present work.

\paragraph{Model-scale coverage.}
The primary benchmark uses models in the 70M--2.7B parameter range
(TinyStories, Pythia-70M / 160M, GPT-2-Small, OPT-125M, and Gemma-2-2B).
Additional public-SAE evaluations on Qwen3-8B / 14B / 32B and Gemma-2-9B
extend the observed scope of the Q1 token-DAME deficit
(Appendix~\ref{appendix:larger-models}), but cover limited model, layer, and
SAE-family combinations. These checks do not establish scale invariance,
a monotonic scaling trend, or larger-model replication of every failure
mode, including redundancy. The contribution is the framework itself;
broader scaling behavior remains open.

\paragraph{Proxy nature of monosemanticity scoring.}
Any monosemanticity metric that uses an LLM as a judge---including the
auto-interpretability scores of \citet{bills2023language} and
\citet{paulo2024automatically}, the SAEBench AutoInterp evaluation, and our
DAME---inherits a shared limitation: the judge can itself produce noisy or
polysemantic explanations, and its scores are bounded by the judging LLM's
own representational fidelity for the concepts it is asked to label.
DAME does not escape this limitation; its contribution is dominance-aware
\emph{stratification} of the same proxy, not a replacement of LLM-based
judging. Improvements to the judging LLM, the prompting strategy, or the
validation question construction would change all such metrics' absolute
numbers in correlated ways. We expect the relative ordering across SAEs
and across dominance quantiles to be more robust than the absolute values,
but this is itself a proxy property and the limitation should be
acknowledged.

\paragraph{Ground-truth concept sets are conditional, not unobtainable.}
A common framing of SAE evaluation is that ``ground-truth concepts are
unavailable for real LLMs.'' The situation is more nuanced: a ground-truth
concept set can in principle be inferred for a real LLM, \emph{but only
conditional on} (i) the chosen SAE width $m$ being close to the true
cardinality of the underlying concept set, and (ii) the granularity of the
discovered features lying within a range that humans can verify. Both
conditions are difficult to certify for any individual SAE, which is why we
rely on proxy metrics that provide empirical evidence relevant to the AIU criteria. Our synthetic
validation (Appendix~\ref{appendix:synthetic-validation}) sidesteps both
conditions by constructing the concept set itself; the limitation is
therefore not that ground truth is intrinsically unavailable, but that
recovering it for a real LLM requires assumptions we have not yet verified.

\paragraph{Hyperparameter transfer and sparsity grid coverage.}
For the SAEs we trained ourselves we adopted hyperparameters from
SAEBench's Pythia and Gemma configurations as a starting point
(Appendix~\ref{appendix:per_arch_l0}). This choice yields a wide divergence
in effective sparsity across architectures: top-$k$ variants enforce a
target $L_0$ by construction, while Gated and JumpReLU rely on a soft
penalty that does not transfer cleanly to LLMs of different scales and
activation distributions. A more comprehensive per-LLM hyperparameter
sweep, targeting comparable effective $L_0$ across architectures, would
likely sharpen and possibly reverse some of the architecture-specific
patterns we report (the synthetic validation in
Appendix~\ref{appendix:synthetic-validation}, which controls sparsity
directly during training, illustrates this: JumpReLU shows the strongest
Q1 deficit on synthetic data even though it is non-significant on the
real-LLM SAEs we train with our default hyperparameters).

\section{Experiment Details}
\label{appendix:experiment_details}

This appendix provides comprehensive details on our benchmarking experiments, covering the SAE families evaluated, their configurations, and the experimental setup. Our evaluation follows the Atomic Interpretable Unit (AIU) framework introduced in Section~\ref{sec:atomic-interpretable-units}, which establishes five necessary criteria for interpretable feature dictionaries: \textbf{C1 (Non-redundancy)}, \textbf{C2 (Completeness)}, \textbf{C3 (Monosemanticity)}, \textbf{C4 (No dead units)}, and \textbf{C5 (Per-input sparsity)}. We benchmark SAEs using both existing proxy metrics from prior work and newly proposed metrics that operationalize these criteria, as introduced in Section~\ref{sec:criteria}. Comprehensive implementation details for all metrics and evaluation methodology are provided in Appendix~\ref{appendix:other_metrics}. Detailed benchmarking results across all evaluated SAEs are presented in Appendix~\ref{appendix:benchmark_other_results}.

\subsection{Overview of Evaluated SAE Families}

Our comprehensive benchmark evaluates hundreds of SAEs across multiple families, architectures, and model configurations. We organize our evaluation into two main categories:

\begin{enumerate}
    \item \textbf{Existing SAE families} from prior work: SAEBench \citep{karvonen2025saebench}, JBloom OAI-v5 SAEs \citep{bloom2024saelens}, and CanRager SAEs (\url{https://huggingface.co/canrager}).
    \item \textbf{Newly trained SAE variants}: 162 SAEs we trained across three different LLMs using SAELens \citep{bloom2024saelens}, systematically exploring the hyperparameter space across six architectures, three widths, and multiple sparsity levels.
\end{enumerate}

\subsection{Existing SAE Families}

\subsubsection{SAEBench}

SAEBench \citep{karvonen2025saebench} provides a comprehensive suite of baseline SAEs designed for systematic evaluation. We evaluate all SAEBench models, which include:

\textbf{Architectures:} Seven SAE varieties:
\begin{itemize}
    \item \textbf{Standard (ReLU)}: Anthropic April Update architecture \citep{templeton2024scaling}
    \item \textbf{TopK}: Top-$k$ activation architecture \citep{gao2024scaling}
    \item \textbf{BatchTopK}: Batched top-$k$ architecture \citep{bussmann2024batchtopk}
    \item \textbf{JumpReLU}: Jump ReLU activation \citep{rajamanoharan2024jumprelu}
    \item \textbf{Gated}: Gated activation architecture \citep{rajamanoharan2024gated}
    \item \textbf{P-anneal}: P-annealing architecture \citep{karvonen2024board}
    \item \textbf{Matryoshka BatchTopK}: Matryoshka batched top-$k$ architecture \citep{bussmann2025matryoshka}
\end{itemize}

\textbf{Base Models:}
\begin{itemize}
    \item \textbf{Pythia-160M}: All architectures evaluated on layer 8 (\texttt{resid\_post\_layer\_8}) \citep{biderman2023pythia}
    \item \textbf{Gemma-2-2B}: All architectures evaluated on layer 12 (\texttt{resid\_post\_layer\_12}) \citep{gemma2team2024gemma2}
\end{itemize}

\textbf{Widths:} Three feature dictionary sizes:
\begin{itemize}
    \item \textbf{4k}: 4,096 features
    \item \textbf{16k}: 16,384 features
    \item \textbf{65k}: 65,536 features
\end{itemize}

\textbf{Sparsity Levels:} Six trainers per configuration (trainer\_0 through trainer\_5), representing different sparsity parameter settings. Each trainer corresponds to a different sparsity level, with trainer\_0 being the most sparse and trainer\_5 being the least sparse.

\textbf{Total SAEBench Models:} 252 SAEs (7 architectures $\times$ 3 widths $\times$ 2 base models $\times$ 6 trainers). All SAEBench models are loaded from HuggingFace repositories:
\begin{itemize}
    \item Pythia-160M: \nolinkurl{adamkarvonen/saebench_pythia-160m-deduped_width-2pow{12,14,16}_date-0108}
    \item Gemma-2-2B 4k: \nolinkurl{adamkarvonen/saebench_gemma-2-2b_width-2pow12_date-0108}
    \item Gemma-2-2B 16k/65k: \nolinkurl{canrager/saebench_gemma-2-2b_width-2pow{14,16}_date-0107}
\end{itemize}

\subsubsection{JBloom OAI-v5 SAEs}

We evaluate the JBloom OAI-v5 SAE family \citep{bloom2024saelens}, which consists of Standard ReLU SAEs trained on GPT-2 Small using the OpenAI v5 training methodology. These SAEs are provided in SAELens Standard Format and cover:

\textbf{Architecture:} Standard ReLU SAE (OpenAI v5 training method)

\textbf{Base Model:} GPT-2 Small (~124M parameters, $d_{\text{in}}=768$) \citep{radford2019language}

\textbf{Dataset:} OpenWebText (Skylion007/openwebtext)

\textbf{Widths:} Two feature dictionary sizes:
\begin{itemize}
    \item \textbf{32k}: 32,768 features
    \item \textbf{128k}: 131,072 features
\end{itemize}

\textbf{Hook Points:} Four activation types:
\begin{itemize}
    \item \texttt{resid\_post}: Residual stream post-attention
    \item \texttt{resid\_mid}: Residual stream mid-block
    \item \texttt{mlp\_out}: MLP output
    \item \texttt{attn\_out}: Attention output
\end{itemize}

\textbf{Layers:} All 12 layers (layers 0--11) for each hook point and width combination

\textbf{Total JBloom Models:} 96 SAEs (2 widths $\times$ 4 hook points $\times$ 12 layers). Models are loaded from HuggingFace repositories:
\begin{itemize}
    \item \nolinkurl{jbloom/GPT2-Small-OAI-v5-{32k,128k}-{resid-post,resid-mid,mlp-out,attn-out}-SAEs}
\end{itemize}

\subsubsection{CanRager SAEs}

We evaluate the CanRager SAE family, which consists of TopK and Vanilla (Standard ReLU) SAEs trained on Gemma-2-2B. These SAEs are hosted in separate HuggingFace repositories and provide:

\textbf{Architectures:} Two SAE varieties:
\begin{itemize}
    \item \textbf{TopK}: Top-$k$ activation architecture \citep{gao2024scaling}
    \item \textbf{Vanilla (Standard ReLU)}: Anthropic April Update architecture \citep{templeton2024scaling}
\end{itemize}

\textbf{Base Model:} Gemma-2-2B \citep{gemma2team2024gemma2}

\textbf{Widths:} Three feature dictionary sizes:
\begin{itemize}
    \item \textbf{4k}: 4,096 features
    \item \textbf{16k}: 16,384 features
    \item \textbf{65k}: 65,536 features
\end{itemize}

\textbf{Layers:} Three layers (layers 5, 12, and 19)

\textbf{Trainers:} Six training runs per configuration (trainer\_0 through trainer\_5)

\textbf{Total CanRager Models:} 108 SAEs (2 architectures $\times$ 3 widths $\times$ 3 layers $\times$ 6 trainers). Models are loaded from HuggingFace repositories:
\begin{itemize}
    \item \nolinkurl{canrager/saebench_gemma-2-2b_width-2pow{12,14,16}_date-1109} (\url{https://huggingface.co/canrager})
\end{itemize}

\subsection{Newly Trained SAE Variants}

To systematically explore the hyperparameter space and ensure comprehensive coverage across architectures, we trained hundreds of new SAE variants using SAELens \citep{bloom2024saelens}. Our training covers three different LLMs from distinct model families, enabling evaluation across diverse architectural characteristics.

\subsubsection{Base Models}

We trained SAEs on three language models:

\textbf{GPT-2 Small:} \citep{radford2019language}
\begin{itemize}
    \item Model: \texttt{gpt2} (HookedTransformer)
    \item Activation dimension: $d_{\text{in}} = 768$ (MLP output)
    \item Target layer: Layer 6 (middle layer of 12 total layers)
    \item Hook: \texttt{blocks.6.hook\_mlp\_out}
    \item Dataset: \texttt{apollo-research/Skylion007-openwebtext-tokenizer-gpt2}
\end{itemize}

\textbf{TinyStories-1L-21M:} \citep{eldan2023tinystories}
\begin{itemize}
    \item Model: \texttt{tiny-stories-1L-21M} (HookedTransformer)
    \item Activation dimension: $d_{\text{in}} = 1024$ (MLP output)
    \item Target layer: Layer 0 (single-layer model)
    \item Hook: \texttt{blocks.0.hook\_mlp\_out}
    \item Dataset: \texttt{apollo-research/roneneldan-TinyStories-tokenizer-gpt2}
\end{itemize}

\textbf{OPT-125M:} \citep{zhang2022opt}
\begin{itemize}
    \item Model: \texttt{facebook/opt-125m} (HookedTransformer)
    \item Activation dimension: $d_{\text{in}} = 768$ (MLP output)
    \item Target layer: Layer 6 (middle layer of 12 total layers)
    \item Hook: \texttt{blocks.6.hook\_mlp\_out}
    \item Dataset: \texttt{apollo-research/Skylion007-openwebtext-tokenizer-gpt2}
\end{itemize}

\subsubsection{SAE Architectures and Configurations}

For each base model, we trained six SAE architectures across three widths and multiple sparsity levels:

\textbf{Architecture 1: Standard (ReLU)}
\begin{itemize}
    \item Configuration class: \texttt{StandardTrainingSAEConfig}
    \item Sparsity parameter: \texttt{l1\_coefficient}
    \item Sparsity values: [1.0, 2.0, 5.0]
    \item Key hyperparameters: L1 regularization coefficient controls sparsity
\end{itemize}

\textbf{Architecture 2: Gated}
\begin{itemize}
    \item Configuration class: \texttt{GatedTrainingSAEConfig}
    \item Sparsity parameter: \texttt{l1\_coefficient}
    \item Sparsity values: [1.0, 2.0, 5.0]
    \item Key hyperparameters: L1 coefficient; gated activation mechanism \citep{rajamanoharan2024gated}
\end{itemize}

\textbf{Architecture 3: JumpReLU}
\begin{itemize}
    \item Configuration class: \texttt{JumpReLUTrainingSAEConfig}
    \item Sparsity parameter: \texttt{l0\_coefficient}
    \item Sparsity values: [1.0, 3.0, 5.0]
    \item Key hyperparameters: L0 regularization coefficient \citep{rajamanoharan2024jumprelu}
\end{itemize}

\textbf{Architecture 4: TopK}
\begin{itemize}
    \item Configuration class: \texttt{TopKTrainingSAEConfig}
    \item Sparsity parameter: \texttt{k} (number of active features per input)
    \item Sparsity values: [50, 100, 200]
    \item Key hyperparameters: Top-$k$ value determines structural sparsity \citep{gao2024scaling}
\end{itemize}

\textbf{Architecture 5: BatchTopK}
\begin{itemize}
    \item Configuration class: \texttt{BatchTopKTrainingSAEConfig}
    \item Sparsity parameter: \texttt{k} (number of active features per batch)
    \item Sparsity values: [50, 100, 200]
    \item Key hyperparameters: Batch-level top-$k$ sparsity \citep{bussmann2024batchtopk}
\end{itemize}

\textbf{Architecture 6: Matryoshka BatchTopK}
\begin{itemize}
    \item Configuration class: \texttt{MatryoshkaBatchTopKTrainingSAEConfig}
    \item Sparsity parameter: \texttt{k}
    \item Sparsity values: [50, 100, 200]
    \item Key hyperparameters: Top-$k$ value; nested sparsity structure with \texttt{matryoshka\_widths} set based on $m$ \citep{bussmann2025matryoshka}
\end{itemize}

\textbf{Widths:} For all architectures, we train three dictionary sizes:
\begin{itemize}
    \item \textbf{4k}: $m = 4,096$ features
    \item \textbf{16k}: $m = 16,384$ features
    \item \textbf{65k}: $m = 65,536$ features
\end{itemize}

\subsubsection{Training Configuration}

All newly trained SAEs share the following training hyperparameters (implemented via SAELens):

\textbf{Training Schedule:}
\begin{itemize}
    \item Total training steps: 18,000
    \item Training tokens per run: ~73.7M tokens
    \item Batch size: 4,096 tokens
    \item Context size: 512 tokens
\end{itemize}

\textbf{Optimization:}
\begin{itemize}
    \item Learning rate: $5 \times 10^{-5}$
    \item Learning rate scheduler: Constant (no warm-up or decay)
    \item Optimizer: Adam ($\beta_1 = 0.9$, $\beta_2 = 0.999$)
\end{itemize}

\textbf{Activation Storage:}
\begin{itemize}
    \item Buffer batches: 32 batches in activation buffer
    \item Store batch size: 8 prompts per batch
    \item Activation normalization: Expected average only (input)
    \item Activation mixing fraction: 0.5
\end{itemize}

\textbf{Feature Management:}
\begin{itemize}
    \item Feature sampling window: 1,000 steps
    \item Dead feature window: 1,000 steps
    \item Dead feature threshold: $10^{-4}$
\end{itemize}

\textbf{Evaluation:}
\begin{itemize}
    \item Evaluation batches: 10 batches
    \item Evaluation batch size: 2 prompts
\end{itemize}

\textbf{Performance Settings:}
\begin{itemize}
    \item Data type: \texttt{float32}
    \item Autocast: Enabled (mixed precision)
    \item LLM compilation: Enabled (\texttt{torch.compile})
    \item SAE compilation: Enabled (\texttt{torch.compile})
\end{itemize}

\subsubsection{Total Training Combinations}

For each of the three LLMs, we train:
\begin{itemize}
    \item 6 architectures $\times$ 3 widths $\times$ 3 sparsity levels = 54 combinations per LLM
    \item Total across all LLMs: 162 SAE training configurations
\end{itemize}

\subsubsection{Compute Resources}
\label{appendix:compute_resources}

All SAE training and evaluation experiments were run on a single workstation
equipped with 4$\times$ NVIDIA Quadro RTX 6000 GPUs (24~GB VRAM each, CUDA
12.8). The newly trained SAEs were distributed across the four GPUs in
parallel; each individual SAE training run processes 18{,}000 steps at
4{,}096 tokens per batch ($\approx$73.7M tokens) at the hyperparameters
listed above. Per-SAE wall-clock training time varied with width and
architecture but typically fell in the 30~min--2~hour range. Evaluation
(DAME, redundancy, completeness, sparsity / dead-unit metrics) on the full
600+ SAE benchmark required an additional $\sim$150--200 GPU-hours,
dominated by DAME's auto-interpretability stage. External SAE families
(SAEBench Pythia / Gemma, jbloom OAI v5, CanRager) were used as released;
we did not retrain them.

\subsubsection{Licenses for Existing Assets}
\label{appendix:licenses}

We rely on several externally released assets. Each is used in accordance
with its license; we credit the original authors and link the canonical
release.

\begin{itemize}
  \item \textbf{SAEBench} \citep{karvonen2025saebench} -- benchmark suite,
        SAE checkpoints (Pythia-160M, Gemma-2-2B), and evaluation code.
        No explicit LICENSE file is provided in the public repository
        as of submission; we use the released code and checkpoints
        under standard research-use conventions and credit the original
        authors throughout the paper.
  \item \textbf{jbloom OAI v5 SAEs} -- 32 GPT-2-Small residual-stream SAEs
        released by Joseph Bloom on HuggingFace. MIT License (declared on
        the model card); we use them only as inputs to our evaluation
        pipeline and do not redistribute the weights.
  \item \textbf{CanRager SAEs} -- TopK and Vanilla (Standard ReLU) SAEs
        on Gemma-2-2B, released on HuggingFace. No explicit license is
        declared on the model card; same usage conventions as above.
  \item \textbf{Public larger-model SAEs}
        (Appendix~\ref{appendix:larger-models}) -- BatchTopK SAEs for
        Qwen3-8B / 14B / 32B (\texttt{adamkarvonen/qwen3-8b-saes},
        \texttt{qwen3-14b-saes}, and \texttt{qwen3-32b-saes}; MIT License)
        and Gemma Scope residual-stream SAEs for Gemma-2-9B
        (\texttt{google/gemma-scope-9b-pt-res}; CC-BY-4.0), released on
        HuggingFace; used only as inputs to our evaluation pipeline.
  \item \textbf{SAELens} \citep{bloom2024saelens} -- training and loading
        framework. MIT License.
  \item \textbf{TransformerLens} -- model loading utilities. MIT License.
  \item \textbf{Models}: Pythia (Apache 2.0), GPT-2 (MIT), OPT-125M
        (OPT-MODEL-LICENSE-AGREEMENT; non-commercial research use only),
        Gemma-2-2B (Gemma Terms of Use; research-permitted), TinyStories
        model (no explicit license declared on the HuggingFace model card);
        for the larger-model evaluation
        (Appendix~\ref{appendix:larger-models}), Qwen3-8B / 14B / 32B
        (Apache 2.0) and Gemma-2-9B (Gemma Terms of Use).
  \item \textbf{Tokenizer / activation datasets}: Apollo Research's
        OpenWebText subset for GPT-2 (\texttt{apollo-research/Skylion007-%
openwebtext-tokenizer-gpt2}; no license declared on the dataset card);
        the TinyStories dataset (\texttt{roneneldan/TinyStories},
        CDLA-Sharing-1.0); analogous tokenized streams for Pythia and
        OPT-125M, no license declared on their dataset cards;
        WikiText-103 (\texttt{Salesforce/wikitext}, configuration
        \texttt{wikitext-103-raw-v1}; CC BY-SA 3.0 and GFDL) for the
        natural-support and context analyses in
        Appendices~\ref{appendix:functional-sharding-protocol}
        and~\ref{appendix:activation-support-audit}.
  \item \textbf{DAME's LLM judge}:
        \texttt{mistralai/Mistral-7B-Instruct-v0.2}, run locally
        (offline, no API). Apache 2.0 License.
  \item \textbf{Additional explainer and judge}: GPT-5.6 Sol (OpenAI),
        used for the explainer-by-judge comparison
        (Appendix~\ref{appendix:cross-judge}) and the context judgments
        (Appendix~\ref{appendix:activation-support-audit}); a hosted
        model accessed remotely under the provider's terms of use.
\end{itemize}

All asset usages comply with the corresponding license terms or, where no
explicit license is provided, with standard research-use conventions. We do
not redistribute any third-party model weights, datasets, or SAE
checkpoints beyond pointers to their original public locations.

\section{Proxy Metrics for the Five AIU Criteria}
\label{appendix:other_metrics}

This appendix details the proxy metrics we employ to operationalize each of the five AIU criteria introduced in Section~\ref{sec:atomic-interpretable-units}. Since we do not assume access to a predefined concept set as ground truth, these metrics serve as practical proxies for evaluating whether SAE features form an atomic interpretable unit basis. We organize our discussion by criterion, describing what each metric measures, which ones we include, and relevant implementation details.

\subsection{Proxy Metrics for Non-Redundancy (C1)}
\label{metrics_redundancy}

For \textbf{C1 (Non-redundancy)}, we employ dominance-aware redundancy metrics that evaluate whether distinct features encode distinct concepts. The standard approach aggregates redundancy by uniformly counting redundant pairs: $\text{RedundancyRate}_{\theta} = \frac{1}{\binom{m}{2}} \sum_{i<j} \mathbf{1}[\cos(d_i, d_j) \ge \theta]$. However, uniform aggregation ignores that redundancy among dominant features is far more problematic than redundancy among rare features.

Our dominance-aware approach includes the \textbf{top-$n$ dominance redundancy rate}, which restricts evaluation to pairs among the $n$ most dominant features (selected by mean dominance $\bar{\rho}_i$). Additional variants, including quantile-based redundancy metrics, are detailed in Section~\ref{sec:criteria} and used in our evaluation. We evaluate redundancy using decoder cosine similarity, formally justified in Section~\ref{sec:redundancy} on the grounds that encoders act as gating mechanisms while decoders define semantic content.

We compute decoder cosine similarity as $\text{sim}(i, j) = \frac{d_i \cdot d_j}{\|d_i\| \|d_j\|}$ for all feature pairs, where $d_i$ and $d_j$ are the decoder weight vectors. Features are ranked by mean dominance $\bar{\rho}_i$ and split into four quantiles (Q1--Q4), where Q1 contains the most dominant features. We count redundant pairs as those with decoder cosine similarity $\geq 0.7$ (following \citet{karvonen2025saebench}) and report redundancy rates per quantile. We also compute encoder similarity among redundant features for comparison (Section~\ref{subsec:redundancy}).

\subsubsection{Empirical Validation: Decoder Cosine Similarity vs.\ Semantic Alignment}
\label{appendix:decoder-to-embedding}

Section~\ref{sec:redundancy} provides a geometric argument for measuring
redundancy on the decoder side. Here we additionally verify that decoder cosine
similarity tracks semantic alignment between features---i.e., that the geometric
proxy reflects what the features actually encode in concept space.

\paragraph{Workflow.} For every SAE in our benchmark, we (1) collect the top-activating
contexts for each feature (window sizes $T \in \{8, 16, 32, 64\}$ tokens around the
maximum-activation position); (2) generate a natural-language explanation for each
feature using an LLM (autointerp pipeline); (3) embed each explanation with
sentence-BERT to obtain a feature-level explanation embedding $e_i \in \mathbb{R}^{384}$;
(4) for every feature pair $(i, j)$ with sufficiently distinct decoder
directions, compute the pair-level explanation similarity
$\cos(e_i, e_j)$; (5) aggregate by binning pairs by decoder cosine similarity
and computing mean explanation similarity per bin.

\paragraph{Coverage.} 136 SAEs across 5 LLM families
(Gemma-2-2b, Pythia-160M, Pythia-70M via CanRager, GPT-2-small via JBloom and our
training, and our training on OPT-125M and TinyStories) and 7 architectures
(Standard, TopK, BatchTopK, Matryoshka, JumpReLU, Gated, PAnneal). At $T{=}8$,
this yields 95{,}206 feature pairs.

\begin{figure}[t]
  \centering
  \begin{subfigure}[t]{0.49\linewidth}
    \centering
    \includegraphics[width=\linewidth]{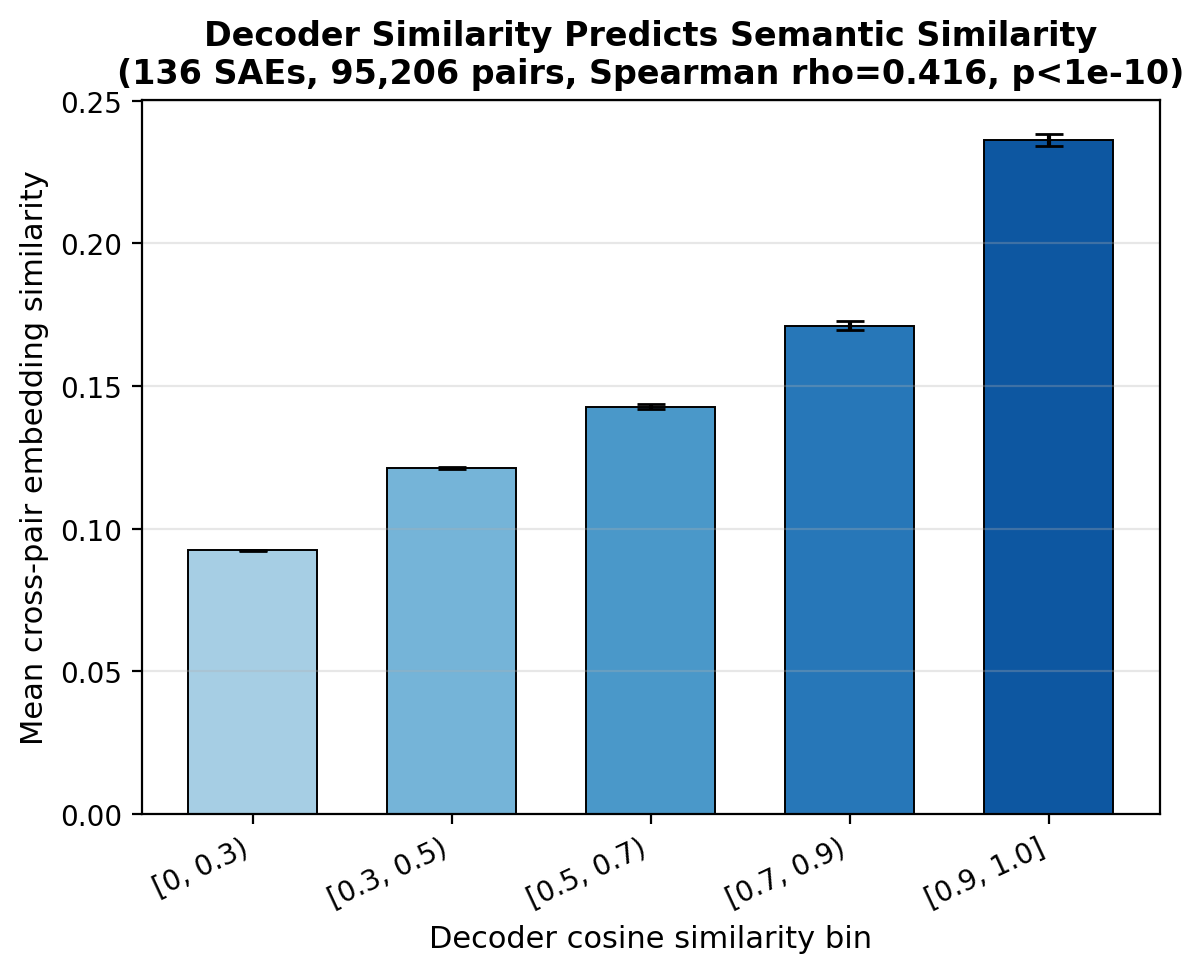}
    \caption{Mean explanation-embedding similarity binned by decoder cosine similarity (95{,}206 pairs across 136 SAEs).}
    \label{fig:decoder_to_embedding_bar}
  \end{subfigure}\hfill
  \begin{subfigure}[t]{0.49\linewidth}
    \centering
    \includegraphics[width=\linewidth]{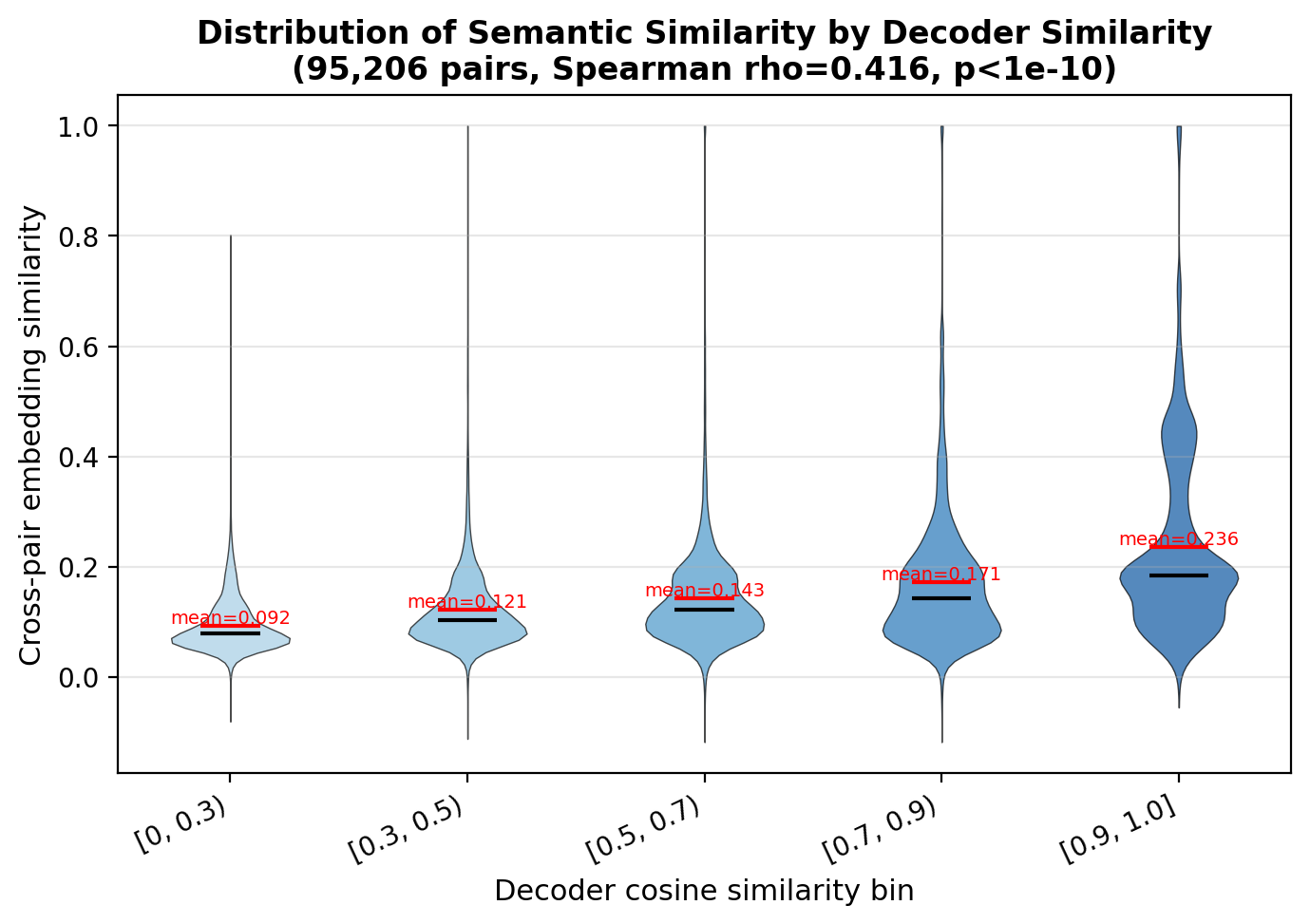}
    \caption{Per-SAE distribution of Spearman $r_s$; 130 / 136 SAEs (96\%) positive.}
    \label{fig:decoder_to_embedding_violin}
  \end{subfigure}
  \caption{Empirical validation of decoder cosine similarity as a proxy for semantic alignment. (\subref{fig:decoder_to_embedding_bar}) Pooled across all 95{,}206 feature pairs, mean explanation similarity rises monotonically with decoder cosine similarity. (\subref{fig:decoder_to_embedding_violin}) Per-SAE Spearman $r_s$ distribution shows the effect is consistent across SAEs, not driven by outliers.}
  \label{fig:decoder_to_embedding}
\end{figure}

\paragraph{Pooled result.} Mean explanation similarity rises monotonically with
decoder cosine similarity (Figure~\ref{fig:decoder_to_embedding},
panel~\ref{fig:decoder_to_embedding_bar}). Pooled Spearman $r_s = 0.42$ at
$T{=}8$, $p < 10^{-10}$. The trend is visually clear: pairs with decoder
cosine $\geq 0.7$ (our redundancy threshold) show explanation cosine
substantially higher than pairs with decoder cosine $\leq 0.3$.

\paragraph{Per-SAE result.} The effect is not driven by a few outlier SAEs.
Per-SAE Spearman correlations (Figure~\ref{fig:decoder_to_embedding},
panel~\ref{fig:decoder_to_embedding_violin}): 130 of 136 SAEs (96\%) show
\emph{positive} correlation; median per-SAE $r_s \approx 0.43$. All per-LLM
and per-architecture pools are statistically significant at
$p < 2.4 \times 10^{-5}$.

\paragraph{Robustness across context windows.} Repeating the analysis at
$T \in \{8, 16, 32, 64\}$ tokens yields qualitatively identical pooled trends,
indicating the relationship is not an artifact of the specific window choice.

\paragraph{Interpretation.} Decoder cosine similarity is therefore a
\emph{defensible} proxy for semantic alignment: it geometrically measures
direction overlap in the residual stream, and empirically that overlap
predicts independent semantic similarity (estimated from explanations
produced by an LLM). Pairs that the redundancy metric flags as redundant
($\cos \geq 0.7$) are also pairs that an external interpretability pipeline
considers semantically alike.

\subsection{Proxy Metrics for Completeness (C2)}
\label{metrics_completeness}

For \textbf{C2 (Completeness)}, we evaluate how well SAE reconstructions capture the original activation space. Since we lack a predefined concept set as ground truth, the best we can do is use reconstruction fidelity and SAE training quality to estimate completeness---if an SAE cannot faithfully reconstruct activations, it cannot form a complete basis for semantic explanation.

We employ three standard metrics that have become established in SAE evaluation:

\paragraph{Mean Squared Error (MSE).} The most direct measure of reconstruction fidelity, MSE computes the average squared error between original activations $\phi(x)$ and their reconstructions $\hat{\phi}(x)$: $\text{MSE} = \mathbb{E}_x[||\phi(x) - \hat{\phi}(x)||^2]$. Lower MSE indicates better reconstruction fidelity. This metric is widely used across SAE evaluation frameworks \citep{bricken2023towards, cunningham2024sparse, karvonen2025saebench} and provides a stable, interpretable measure of how well the SAE preserves activation information.

\paragraph{Cross-Entropy Loss Recovery.} Multiple early SAE works \citep{bricken2023towards, cunningham2024sparse} extended evaluation to measure whether SAEs preserve model performance when their reconstructions replace original activations, testing reconstruction fidelity via downstream loss. Cross-entropy loss recovery measures the cross-entropy loss when SAE reconstructions are passed through the model's downstream layers, comparing it to both the baseline loss (without SAE) and the ablation loss (when activations are set to zero). The derived $\text{ce\_loss\_score}$ metric normalizes performance preservation: $\text{ce\_loss\_score} = (\text{ce\_loss\_with\_ablation} - \text{ce\_loss\_with\_sae}) / (\text{ce\_loss\_with\_ablation} - \text{ce\_loss\_without\_sae})$. This metric captures how well the SAE maintains the model's predictive capabilities, which is critical for using SAEs in attribution and intervention tasks.

\paragraph{KL Divergence.} Kullback-Leibler divergence measures the divergence between the distribution of the model's output predictions when using original activations versus SAE reconstructions. Similar to cross-entropy loss recovery, KL divergence assesses whether the SAE preserves the model's behavior distribution, not just point estimates. The derived $\text{kl\_div\_score}$ metric normalizes by comparing to ablation: $\text{kl\_div\_score} = (\text{kl\_div\_with\_ablation} - \text{kl\_div\_with\_sae}) / \text{kl\_div\_with\_ablation}$. This metric is particularly important for ensuring that SAE reconstructions preserve the model's uncertainty and distributional behavior \citep{karvonen2025saebench}.

These three metrics are computed over validation batches, using the same activation layers as those used in training. MSE is directly computed from activation reconstructions, while cross-entropy and KL divergence require passing reconstructions through the model's decoder and output layers. Our implementation follows standard SAE evaluation practices \citep{karvonen2025saebench, bloom2024saelens}.

\begin{figure}[t]
  \centering
  \includegraphics[width=0.85\linewidth]{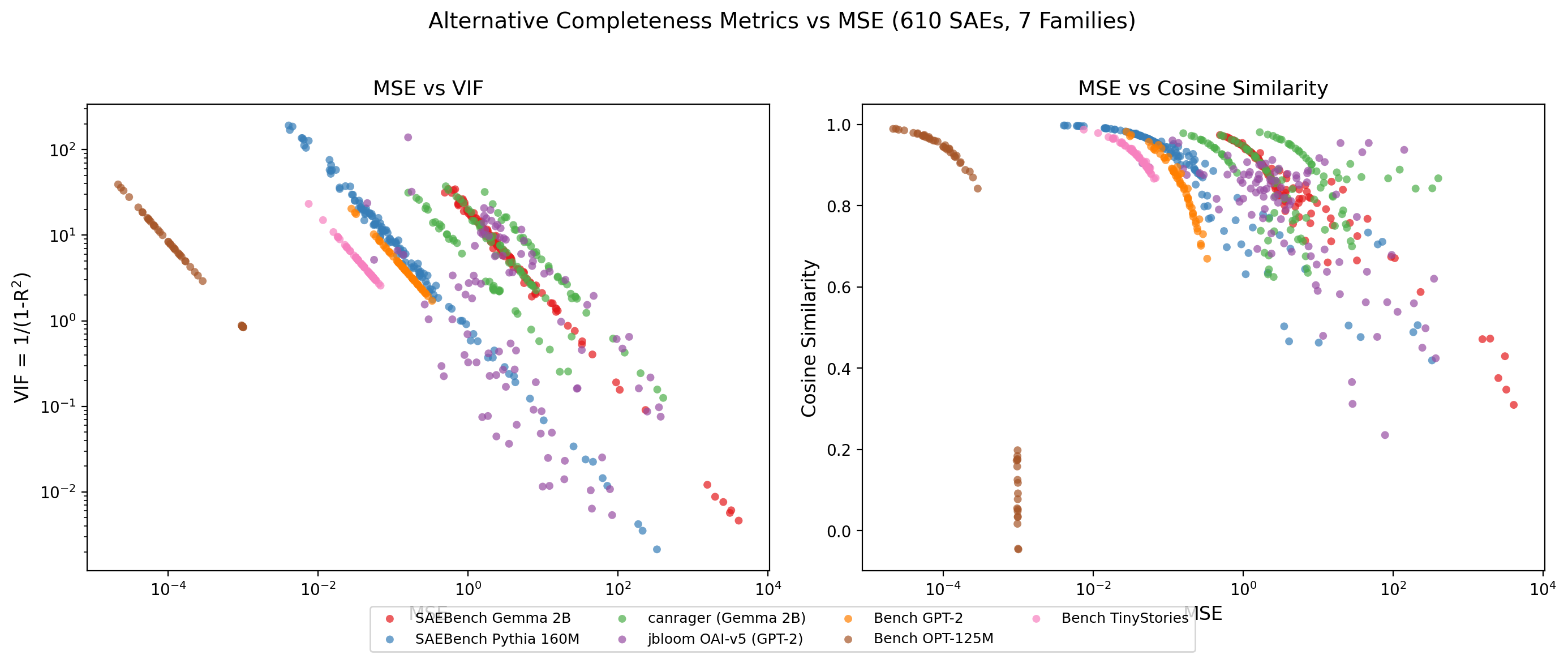}
  \caption{Pairwise scatter of MSE against $R^2$ (plotted as $\text{VIF} = 1/(1-R^2)$ for log-scale clarity) and cosine similarity, across 610 SAEs from 7 families. MSE ranks SAEs near-identically to both alternatives ($|r_s| \geq 0.97$), confirming MSE as a valid simplifying proxy for reporting.}
  \label{fig:completeness_correlations}
\end{figure}

In the main paper we report MSE alone for completeness, although cross-entropy loss recovery and KL divergence offer complementary perspectives. To check that this choice does not change relative SAE rankings, we additionally computed explained variance ($R^2$) and cosine similarity across the full benchmark and measured pairwise Spearman correlations against MSE. The agreement is near-perfect: $r_s = -0.99$ to $-1.00$ for MSE vs.\ $R^2$, $r_s = -0.97$ to $-0.99$ for MSE vs.\ cossim (all $p < 10^{-30}$). Figure~\ref{fig:completeness_correlations} shows the pairwise scatter.

\subsection{Proxy Metrics for Monosemanticity (C3)}
\label{diagnostic_mono}

For \textbf{C3 (Monosemanticity)}, we introduce the Dominance-Aware Monosemanticity Evaluation (DAME), a comprehensive evaluation framework that addresses critical limitations of existing automated interpretability scoring methods. DAME is detailed in Section~\ref{mono_dom} of the main paper; here we provide complete implementation details.

\paragraph{From the C3 criterion to checkable evidence.}\label{appendix:c3-lift}
An explainer uses construction examples to propose an activation-conditioned account $e_i$, expressed at the chosen granularity of $\mathcal{C}$, of what unit $u_i$ represents; $e_i$ is a candidate account, not an assumed optimum. An evaluator (the judge, in DAME) then uses $e_i$ to predict an activation-derived target $y_i(z)$ on a held-out item $z$, producing $p_{e_i}(z)$, and a higher-is-better scoring rule $s$ compares this prediction with the observed target. Let $p_0(z)$ denote the uninformative baseline prediction. The explanation-specific lift is
\[
P_i(e_i)=\mathbb{E}_z\bigl[s(p_{e_i}(z),y_i(z))\bigr]-\mathbb{E}_z\bigl[s(p_0(z),y_i(z))\bigr],
\]
where $\mathbb{E}_z$ averages over held-out validation items. Binary detection, magnitude prediction, and DAME instantiate different choices of $z$, $y_i$, and $s$; DAME uses held-out low-, medium-, and high-activation ordering. Positive lift means that the proposed explanation predicts held-out feature behavior beyond the uninformative baseline. It tests the predictive adequacy of that account; because it neither enumerates all concepts in $\mathcal{C}$ nor determines how many concepts map to $u_i$, it does not itself test injectivity. It is therefore checkable evidence relevant to C3, but neither a logical consequence nor a sufficient proof of C3.

\subsubsection{DAME Overview}
\label{appendix:dame_details}

DAME evaluates SAE feature interpretability using a dominance-aware approach with a 3-way ranking task. Features are ranked by mean dominance (average contribution percentage across all input tokens), not firing frequency. The evaluation uses a 3-bin strategy and asks the LLM to rank 3 examples by activation strength.

\paragraph{Feature Discovery and Dominance Calculation.}

DAME uses a two-pass activation collection strategy to enable efficient evaluation:

\textbf{Pass 0 (Fast Dominance Counting):} We compute mean dominance for each feature across $N$ batches (typically 50 for discovery). For each batch:
\begin{itemize}
    \item Get tokens: $\text{batch\_tokens} = \text{store.get\_batch\_tokens()} \to [\text{batch\_size}=32, \text{seq\_len}=16]$
    \item Get model activations: $\text{model\_acts} = \text{store.get\_activations}(\text{batch\_tokens}) \to [\text{batch}, \text{seq}, d_{\text{in}}]$
    \item Encode through SAE: $\text{sae\_acts} = \text{sae.encode}(\text{model\_acts\_flat}) \to [\text{batch} \times \text{seq}, m]$
    \item For each token position and each feature, compute dominance: $\rho_i(x) = |f_i(\phi(x))| / \sum_j |f_j(\phi(x))|$
    \item Sum dominance across all token positions
\end{itemize}

After all batches, we compute mean dominance: $\bar{\rho}_i = \mathbb{E}_x[\rho_i(x)] = \text{dominance\_sum}[i] / \text{total\_token\_positions}$. We also count firing sequences (for filtering): $\text{firing\_count}[i] = \text{count of sequences where max\_activation} > 0$.

This pass requires only $O(m)$ memory (storing dominance sums and counts, not activations) and completes in approximately 2--3 minutes for typical SAE sizes.

\paragraph{Feature Filtering and Quantile Sampling.}

We filter features to keep only those with $\geq 80$ sequence firings (ensuring enough data for the 3-bin strategy: 20 for explanation + 40 for evaluation split into bin\_A and bin\_B + 20 for bin\_C = 80 total). We then sort filtered features by \textbf{mean dominance} (descending)---not firing frequency---and split into 4 quantiles (Q1: 0--25\%, Q2: 25--50\%, Q3: 50--75\%, Q4: 75--100\%). We sample 20 features from each quantile = 80 total features per SAE.

Dominance-based ranking captures each feature's average contribution percentage, which is more theoretically sound than firing frequency and ensures fair evaluation across the dominance spectrum.

\paragraph{Collection and 3-Bin Strategy.}

We collect all sequences for sampled features by scanning the same batches as Pass 0. For each batch and each sampled feature:
\begin{itemize}
    \item Get all sequences where the feature activated
    \item Compute $\text{max\_activation}$ (for token-based) and $\text{sum\_activation}$ (for sentence-based)
    \item Store all sequences (not just top $N$)
\end{itemize}

After all batches, for each feature:
\begin{itemize}
    \item Sort all sequences by activation magnitude (descending)
    \item Reserve top 20 sequences for explanation generation (excluded from evaluation)
    \item For remaining sequences: split at 50\% mark $\to$ HIGH half (top 50\% of remaining) and LOW half (bottom 50\% of remaining)
    \item Split HIGH half in the middle $\to$ BIN A (top 20) and BIN B (next 20)
    \item LOW half $\to$ BIN C (bottom 20, may include non-firing sequences)
\end{itemize}

The 3-bin result:
\begin{itemize}
    \item \textbf{Top 20 sequences} $\to$ Reserved for explanation (excluded from questions)
    \item \textbf{BIN A (HIGH-high):} Next 20 sequences with highest activations (top half of evaluation sequences after removing explanation set)
    \item \textbf{BIN B (HIGH-low):} Next 20 sequences with lower activations (bottom half of HIGH evaluation sequences)
    \item \textbf{BIN C (LOW):} Bottom 20 sequences (lowest activations, may include non-firing)
\end{itemize}

Critically, \textbf{bin\_A has higher activations than bin\_B} (ensured by sorting before splitting). We maintain separate bins for token-based evaluation (sorted by $\text{max\_activation}$) and sentence-based evaluation (sorted by $\text{sum\_activation}$).

\paragraph{Explanation Generation.}

For each sampled feature, we generate explanations using the top 20 sequences (excluded from evaluation). We build a prompt showing each sequence with: (1) the full sentence (all tokens concatenated), and (2) the top activation token (the single token with highest activation in that sequence). We do \textbf{not} show raw activation values to the LLM, ensuring fairness across features. We call the LLM (\texttt{Mistral-7B-Instruct-v0.2}, run locally) to generate an explanation.

Constraints (V2):
\begin{itemize}
    \item \textbf{40 token limit:} $\text{max\_new\_tokens}=40$
    \item \textbf{15 word maximum:} System prompt explicitly states ``Explain in ONE complete sentence, maximum 15 words.''
    \item \textbf{No frequency mentions:} Cannot mention ``often'', ``frequently'', ``rarely'', etc.
    \item \textbf{Focus on content:} What lexical/syntactic/semantic patterns cause activation
\end{itemize}

We track which sequences were used for explanation (to exclude from evaluation), preventing data leakage.

\paragraph{3-Way Ranking Evaluation.}

We test whether explanations can correctly rank examples by activation strength using a 3-way ranking task:

\textbf{Token-Based Evaluation:} For each feature, we sample 1 sequence from each bin (A, B, C) for each of 20 questions. We pick the highest-activating token from each sequence and build a ranking question:
\begin{quote}
Given the explanation: [EXPLANATION] \\
Rank these 3 examples by activation strength (1=highest, 2=middle, 3=lowest): \\
Example A: [text with highlighted token] \\
Example B: [text with highlighted token] \\
Example C: [text with highlighted token] \\

Output only: A=?, B=?, C=?
\end{quote}

The prompt does \textbf{not} show activation values. We call the same LLM to rank the examples. The true ranking is A=1 (highest), B=2 (middle), C=3 (lowest), reflecting that bin\_A has higher activations than bin\_B.

\textbf{Sentence-Based Evaluation:} Similar to token-based, but we use $\text{bin\_A\_sum\_sequences}$, $\text{bin\_B\_sum\_sequences}$, $\text{bin\_C\_sum\_sequences}$ and ask the LLM to rank entire sentences by activation strength.

\paragraph{Scoring (4 Metrics per Question).}

For each ranking question, we compute 4 metrics:
\begin{enumerate}
    \item \textbf{Perfect ranking accuracy:} Did the LLM get the exact order correct? (A=1, B=2, C=3)
    \item \textbf{Rank correlation:} Spearman correlation between predicted and true ranking
    \item \textbf{Pairwise accuracy:} A vs B, B vs C, A vs C---fraction of pairwise comparisons correct
    \item \textbf{Rank distance:} Average absolute difference in rank positions
\end{enumerate}

We aggregate these metrics per feature (mean across questions) and then per-quantile (mean across features in each quantile).

\paragraph{Memory Management.}

The two-pass strategy enables memory-efficient evaluation:
\begin{itemize}
    \item \textbf{Pass 0:} $O(m)$ memory---only stores dominance sums and counts
    \item \textbf{Collection:} $O(80 \times \text{sequences\_per\_feature})$ memory---stores sequences only for selected features (not all features)
    \item \textbf{Binning:} In-place sorting and splitting
\end{itemize}

After each SAE evaluation, we delete all sequence storage, clear GPU cache, and garbage collect, enabling evaluation of many SAEs in sequence without memory issues.

\subsubsection{Synthetic Validation of DAME on Ground-Truth Concepts}
\label{appendix:synthetic-validation}

DAME relies on an LLM-as-judge proxy for monosemanticity. To verify that the
dominance--monosemanticity relationship reflects a genuine property of SAE
decompositions rather than a measurement artifact, we replicate the Section~\ref{subsec:dame}
analysis on a synthetic benchmark with \emph{known} ground-truth concepts.

\paragraph{Setup.}
We generate synthetic activations from 128 ground-truth concepts and train
105 SAEs spanning all 7 architectures (Standard, TopK, BatchTopK, Matryoshka,
JumpReLU, Gated, PAnneal), 3 widths, and 5 sparsity levels (15 SAEs per
architecture).

\paragraph{Ground-truth monosemanticity gap.}
For each feature $i$, we compute its activation correlation with each
ground-truth concept $c_k$ and report the gap between the best and second-best
concept correlation; a large gap indicates the feature is monosemantically
aligned with a single ground-truth concept. We then test, per SAE, whether Q1
(top-quartile dominance) features have a smaller gap than Q2--Q4 (one-sided
Mann--Whitney U test, $\alpha = 0.05$).

\paragraph{Headline.}
\textbf{71 of 105 SAEs (68\%)} show a statistically significant Q1 deficit
($p < 0.05$). The synthetic findings replicate the real-data pattern using
ground truth instead of LLM judgment, consistent with the dominance--monosemanticity
relationship.

\begin{figure}[t]
  \centering
  \includegraphics[width=0.8\linewidth]{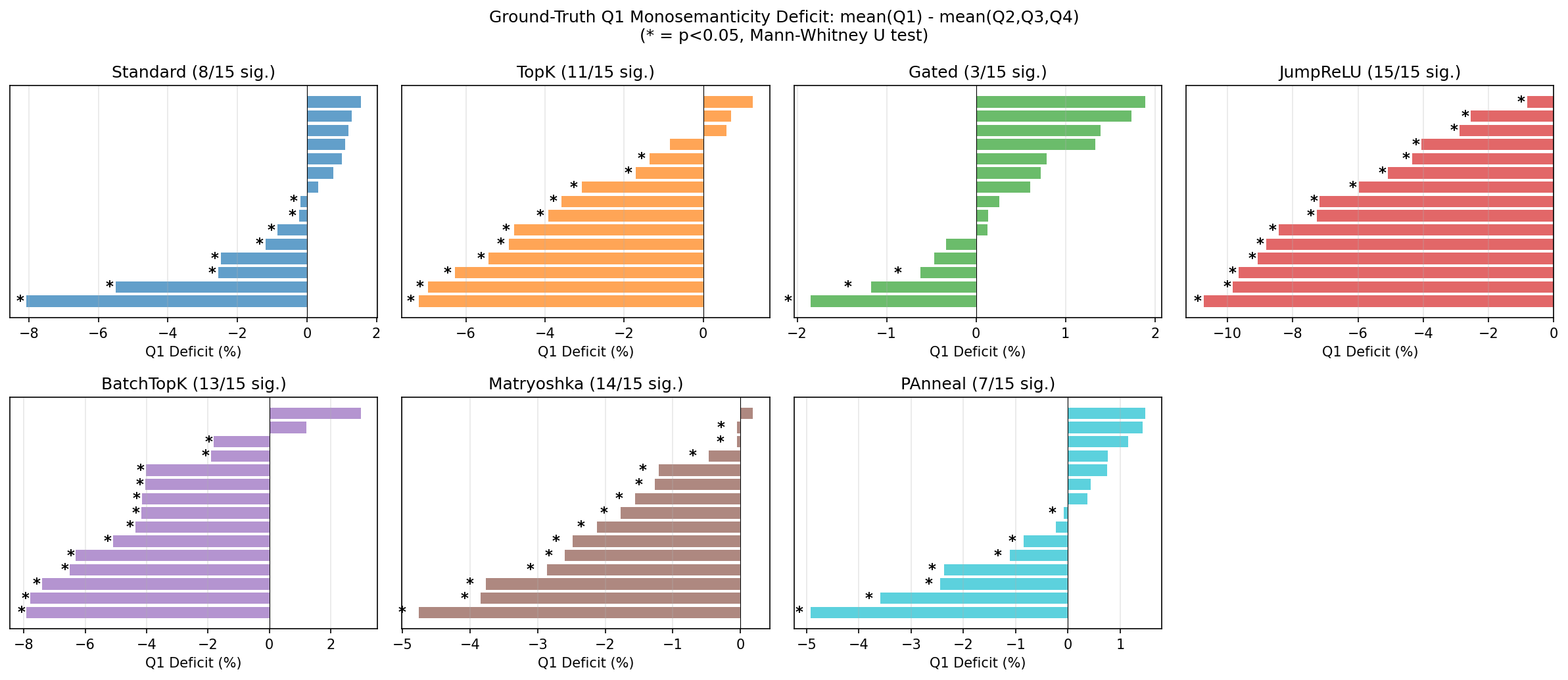}
  \caption{Synthetic validation of DAME on ground-truth concepts. For each architecture, horizontal bars show the Q1 monosemanticity deficit (mean Q1 $-$ mean Q2--Q4) of each individual SAE; stars mark statistical significance (Mann--Whitney U, $p<0.05$). Subplot titles show the per-architecture count of significant SAEs out of 15: JumpReLU 15/15 (strongest); Gated 3/15 (weakest). 71 of 105 SAEs (68\%) overall show significant Q1 deficit.}
  \label{fig:synthetic_validation}
\end{figure}

\paragraph{Per-architecture pattern.}
JumpReLU shows the strongest Q1 deficit on the synthetic benchmark
(\textbf{15 / 15} SAEs significant); Gated is weakest (\textbf{3 / 15})
(Figure~\ref{fig:synthetic_validation}). This is
the inverse of the real-data result, where JumpReLU was n.s.\ on SAEBench.
The discrepancy is explained by hyperparameter mismatch: SAEBench-default
JumpReLU has median $L_0 = 691$, far above the saturation point at which
dominance dilutes. With sparsity matched in synthetic training, JumpReLU's
architectural tendency toward dominant features re-emerges. This further
supports the interpretation in Section~\ref{subsec:dame}: the real-data n.s.\
result for JumpReLU/Gated reflects hyperparameter choice, not architectural
exemption from the Q1 deficit.

\subsection{Proxy Metrics for No Dead Units (C4)}
\label{metrics_dead_neuron}

\paragraph{Dead Features.} Dead features---those not activating meaningfully across training or evaluation data---present a persistent challenge in SAE training and evaluation. Without appropriate mitigations, a large fraction of SAE features can become dead, wasting dictionary capacity and reducing the effective number of learned features.

\citet{templeton2024scaling} reported that their 34M latent SAE had only 12M alive latents (approximately 65\% dead); without mitigations, up to 90\% of features can die during training. This problem is particularly acute in architectures with explicit sparsity constraints (e.g., TopK), where features may be initialized in regions of activation space that rarely activate.

Multiple solutions have been proposed to address dead features:
\begin{itemize}
    \item \textbf{Encoder-decoder initialization:} Initializing the encoder as the transpose of the decoder provides an initial alignment that encourages feature utilization \citep{templeton2024scaling}.
    \item \textbf{Auxiliary losses:} Providing gradient signal to dead features through auxiliary reconstruction losses ensures they receive updates even when rarely activating \citep{gao2024scaling}.
    \item \textbf{Ghost gradients:} Anthropic's ``ghost grads'' technique artificially injects gradients for dead features, simulating activation patterns to keep them alive during training \citep{templeton2024scaling}.
\end{itemize}

These techniques have proven effective: \citet{templeton2024scaling} report that OpenAI's 16M latent GPT-4 SAE achieved only 7\% dead latents through a combination of these approaches.

\paragraph{Metrics.} We evaluate dead feature rates by counting features that have zero (or near-zero, below threshold) activations across evaluation batches. We report both the \textbf{proportion of dead features} (features with firing frequency below threshold, typically $10^{-4}$ activation threshold) and \textbf{firing frequency statistics} (mean, median, quartiles of feature firing frequencies). These metrics help identify poor usage of dictionary capacity and indicate whether SAE training succeeded in utilizing the full feature set.

\subsection{Proxy Metrics for Per-Input Sparsity (C5)}
\label{metrics_sparsity}

For \textbf{C5 (Per-input sparsity)}, we evaluate how sparse SAE activations are, both for individual inputs and across datasets. High per-input activation counts can make an attribution difficult to inspect even when individual features are monosemantic.

\paragraph{Average Active Features.} The standard sparsity metric counts how many features activate per input, where a feature is considered ``active'' if its absolute activation value exceeds a threshold (typically $10^{-5}$). We compute the average number of active features across all inputs: $\text{avg\_active\_features} = \mathbb{E}_x[\sum_i \mathbf{1}[|f_i(\phi(x))| > \tau]]$. This matches the common intuition in prior work that sparsity refers to how many features fire for one input sequence.

\paragraph{Effective Sparsity.} However, simple counting can be misleading for architectures where negligible activations are not explicitly penalized (e.g., TopK-like architectures, where many features may have tiny but non-zero activations). To make sparsity metrics more accurate, we also report \textbf{75\% and 95\% effective sparsity}: for each input, we count how many features (when sorted by absolute activation) are needed to sum up to the specified percentile (75\% or 95\%) of total absolute activation.

Formally, for input $x$, let $\text{sorted\_acts} = \text{sort}(|f_i(\phi(x))|)$ (descending). We compute:
\[
\text{effective\_active}_p(x) = \min\left\{k : \frac{\sum_{i=1}^k \text{sorted\_acts}[i]}{\sum_{j} \text{sorted\_acts}[j]} \geq p/100\right\}
\]
and report $\mathbb{E}_x[\text{effective\_active}_p(x)]$ for $p \in \{75, 95\}$. This captures tail behavior: many features may be negligible though not strictly zero, and effective sparsity measures the number of features that actually contribute to the reconstruction.

\paragraph{Data-Wise vs Feature-Wise Sparsity.} We distinguish between two perspectives on sparsity that are sometimes conflated:

\textbf{Data-wise sparsity} (what most people intuitively mean): For a given input sequence, how many features fire? This is the standard per-input $L_0$ norm and matches common intuition.

\textbf{Feature-wise sparsity} (what loss functions often implicitly optimize): Across the dataset, on how many inputs does a feature fire? When designing loss functions, researchers control or minimize the overall activation amount, which also minimizes the number of inputs on which each feature fires. This matches more closely with monosemanticity---when a feature fires less often, it may in some cases imply better monosemanticity (unless there is a feature-split problem) or at least better interpretability.

In both other papers and in our work, sparsity is fundamentally and practically talking about the number of features fired for one input (data-wise sparsity). However, as sparsity highly correlates with monosemanticity both in results and in loss design, this can be confusing sometimes and cause potential problems in people's theoretical work. We report both data-wise metrics (average active features per input) and feature-wise statistics (firing frequency distributions) to provide a complete picture.

Our implementation computes these metrics across evaluation batches, processing activations in chunks to avoid memory issues for large SAEs. For effective sparsity calculations, we process inputs sequentially, computing cumulative activation sums and identifying the minimum number of features needed to reach each percentile threshold.

\subsection{Empirical Mapping to SAEBench Evaluations}
\label{appendix:saebench-correlations}

The five AIU criteria can be empirically validated against existing SAEBench
evaluations \citep{karvonen2025saebench}. Each SAEBench evaluation tests
downstream task performance under specific failure modes; we map each evaluation
to the AIU criteria it implicitly probes (treating evaluations that depend on
multiple criteria as \emph{joint proxies} rather than direct proxies for a single
criterion). We then verify each mapping empirically across 246 SAEs (SAEBench
Gemma + SAEBench Pythia, 3 widths $\in \{4\text{k}, 16\text{k}, 65\text{k}\}$;
RAVEL and Unlearning are Gemma-only because Pythia lacks an instruct variant
and is too small for reliable attribute prediction). We use SAEBench's
\texttt{ce\_loss\_score} as the C2 proxy, SAEBench's autointerp score as the C3
proxy, and our dead-feature-filtered redundancy score as the C1 proxy.

\paragraph{Joint-proxy methodology.}
For each proposed criterion pair (e.g., C2+C3), we (i) rank SAEs on each
criterion \emph{within} each (model, width) group to control for scale
confounds, (ii) take the product of the two within-group ranks as a joint
proxy that is high only when \emph{both} criteria are high, and (iii) compute
Spearman correlation between this product and the SAEBench evaluation score
across the pooled 246 SAEs. Significance markers below: $^{*}p<0.05$,
$^{**}p<0.01$, $^{***}p<0.001$, n.s.\ otherwise.

\paragraph{Directly aligned (2/8).}
\textbf{Loss Recovered} $\rightarrow$ C2 (completeness; direct). \textbf{Automated
Interpretability} \citep{bills2023language, paulo2024automatically} $\rightarrow$
C3 (monosemanticity; direct---this is what DAME refines).

\paragraph{Verified joint proxies (3/8).}
For these three evaluations, the proposed criterion pair is the strongest
predictor and the mechanism matches the eval design.

\begin{itemize}
  \item \textbf{Sparse Probing} $\rightarrow$ C2$+$C3, $r_s = +0.258^{***}$.
        Trains a $k$-sparse linear probe on the top-5 SAE features per class.
        C2 ensures the task-relevant signal is captured at all; C3 ensures each
        of those 5 features carries clean single-concept information so the
        probe can succeed with so few features.
  \item \textbf{SCR} (Spurious Correlation Removal) $\rightarrow$ C2$+$C3,
        $r_s = +0.459^{***}$. Ablates SAE features correlated with a spurious
        attribute and measures recovery toward clean behavior. C2 ensures the
        spurious signal is captured (and thus removable); C3 ensures the
        ablation is surgical---removing the spurious signal without collateral
        damage to the true task signal.
  \item \textbf{Feature Absorption} $\rightarrow$ C1$+$C3,
        $r_s = +0.248^{***}$. Tests whether broad features absorb the signal
        of more specific ones (e.g., a generic ``Paris'' feature absorbing
        first-letter knowledge). Absorption is a form of feature entanglement
        --- the opposite of C1; C3 ensures each feature captures one concept
        cleanly and so absorbs less.
\end{itemize}

\paragraph{Conceptually C1$+$C3, empirically C2-bottlenecked (2/8).}
\textbf{TPP} (Targeted Probe Perturbation) ablates the top SAE features for a
target class and measures whether the target class drops more than other
classes. \textbf{RAVEL} learns a binary mask over SAE features to isolate one
attribute of an entity, swaps masked features between entities, and decodes
through the SAE to check whether the model's output changes accordingly. Both
were proposed as C1$+$C3 evaluations: C1 because non-redundant features have
no backup copies that can compensate for ablation, and C3 because monosemantic
features make the perturbation precise. Empirically, however, both are
dominated by C2 (TPP: $r_{s,\mathrm{C2}} = +0.768^{***}$;
RAVEL: $r_{s,\mathrm{C2}} = +0.971^{***}$): the encode-modify-decode pipeline requires
the SAE to first \emph{capture} a concept faithfully before any C1/C3
intervention can be tested. If the SAE cannot reconstruct the activations that
encode the target attribute, ablation or swapping cannot have the intended
effect regardless of feature distinctness or interpretability.

The pooled C2$\leftrightarrow$C3 anti-correlation (next paragraph) makes this
visible at small widths: high-C3 SAEs tend to be low-C2 at $m=4\text{k}$,
so the C1$+$C3 product correlates \emph{negatively} with TPP/RAVEL there. As
width grows and C2 stops being the bottleneck, the C1$+$C3 signal begins to
surface (Table~\ref{tab:tpp_ravel_width_breakdown}). At 65k, TPP shows a
significant positive C1$+$C3 correlation; RAVEL approaches non-significance.
The framework thus \emph{predicts} that these two evaluations conflate
disentanglement with reconstruction.

\begin{table}[h]
\centering
\small
\begin{tabular}{lrr|rr}
\toprule
 & \multicolumn{2}{c}{TPP} & \multicolumn{2}{c}{RAVEL} \\
Width & C2 alone & C1$+$C3 & C2 alone & C1$+$C3 \\
\midrule
4k  & $+0.832^{***}$ & $-0.178$ n.s.   & $+0.983^{***}$ & $-0.453^{**}$ \\
16k & $+0.755^{***}$ & $+0.152$ n.s.   & $+0.975^{***}$ & $-0.332^{*}$  \\
65k & $+0.711^{***}$ & $+0.231^{*}$    & $+0.955^{***}$ & $-0.187$ n.s. \\
\bottomrule
\end{tabular}
\caption{Per-width Spearman correlations of C2 alone vs.\ the C1$+$C3 joint
proxy with TPP and RAVEL. The C2 bottleneck dominates at all widths; the
intended C1$+$C3 signal only begins to emerge at 65k, after the
C2$\leftrightarrow$C3 anti-correlation weakens.}
\label{tab:tpp_ravel_width_breakdown}
\end{table}

\paragraph{Special case: Unlearning (1/8).}
\textbf{Unlearning} ablates the top SAE features and reports the largest drop
in dangerous-knowledge accuracy (WMDP-bio) achievable while preserving general
capabilities (MMLU $\geq 0.99$). All three of C1, C2, C3 should matter in
principle: C1 because redundant copies of dangerous-knowledge features survive
ablation of any one copy; C2 because the dangerous concept must be captured
to be ablated; C3 because monosemantic features cleanly separate dangerous
from safe content. Empirically, however, $r_{s,\mathrm{C3}} = +0.494^{***}$ dominates,
$r_{s,\mathrm{C2}} = -0.562^{***}$ (negative), and $r_{s,\mathrm{C1}}$ is not significant. The
likely reason: unlearning targets one specific concept (bioweapons knowledge),
not an aggregate, so global C1 and C2 proxies---which average across many
concepts---may not vary enough to predict whether \emph{this particular}
concept was captured. The negative C2 correlation is most plausibly inherited
from the C2$\leftrightarrow$C3 anti-correlation rather than causal.

\paragraph{C2 $\leftrightarrow$ C3 anti-correlation as a diagnostic feature.}
At small SAE widths the SAEBench analysis reveals a strong negative pooled
correlation between completeness (C2) and monosemanticity (C3): $r_s = -0.66$
overall, $-0.88$ at width 4k. This is precisely why TPP and RAVEL---both nominally
C1$+$C3 evaluations---are bottlenecked by C2 in practice: at small widths an SAE
must trade reconstruction fidelity for per-feature interpretability. The
framework thus \emph{predicts}, rather than merely describes, this downstream
failure mode.

\section{Automated Interpretability Scoring: Prior Work}
\label{appendix:previous-mono-measurement}

The landscape of automated interpretability scoring has evolved rapidly since OpenAI's foundational work, with two systems establishing the dominant paradigms: Bills et al.'s simulation-based approach \citep{bills2023language} and EleutherAI's Delphi detection-based system \citep{paulo2024automatically}. This appendix provides detailed implementation specifications for these systems and documents how major SAE papers deploy them in practice.

\subsection{Overview of Automated Interpretability Scoring}

Automated interpretability scoring aims to quantify whether individual features in neural networks respond to coherent, singular concepts—directly operationalizing the monosemanticity objective that motivated SAE development. These systems generally follow a three-stage pipeline: (1) explanation generation, where a language model produces natural language descriptions from feature activation patterns; (2) scoring, where the quality of explanations is evaluated through prediction tasks; and (3) aggregation, where feature-level scores combine into global metrics.

Two systems dominate current practice. OpenAI's approach \citep{bills2023language} introduced simulation-based scoring using GPT-4, establishing the paradigm that subsequent work built upon. EleutherAI's Delphi system \citep{paulo2024automatically} fundamentally redesigned the scoring methodology to achieve approximately 150× cost reduction through detection scoring and open-source models, making large-scale evaluation tractable. We detail each system's methodology before examining how major SAE papers adapt these approaches.

\subsection{Bills et al. (2023): Simulation-Based Scoring with GPT-4}

OpenAI's ``Language Models Can Explain Neurons in Language Models'' \citep{bills2023language} introduced the dominant automated interpretability framework through a three-step process: explanation generation, simulation, and correlation scoring. The system analyzed all 307,200 neurons in GPT-2 XL, though only approximately 1,000 (0.33\%) achieved explanation scores $\geq 0.8$.

\paragraph{Explanation Generation.}
The system uses GPT-4 to generate natural language descriptions from 5 top-activating text excerpts, each containing 64 contiguous tokens sampled from the pre-training distribution. Raw neuron activations undergo preprocessing: negative values are clamped to zero, remaining positive values are linearly scaled to a 0--10 range, then discretized to integers. For sparse neurons (activating on $<20\%$ of tokens), the system uses modified prompts that include only non-zero token-activation pairs, helping the model focus on relevant activation patterns rather than long sequences of zeros.

\paragraph{Simulation Scoring.}
The core innovation is ``simulation'' scoring: GPT-4, given only the generated explanation, predicts how the neuron would activate on new sequences. The simulator outputs integer predictions from 0--10 for each token. The Pearson correlation coefficient between simulated and actual activations serves as the primary metric:
\[
\text{Score} = \text{corr}(\text{predicted activations}, \text{actual activations}).
\]
Two evaluation modes exist: ``random-only'' using 5 uniformly random excerpts, and ``top-and-random'' using a mix of 5 top-activating plus 5 random excerpts—the latter providing variance reduction at the cost of some bias toward high-activation patterns.

\paragraph{Validation and Limitations.}
The paper validated simulation scoring against an expensive ``ablation scoring'' gold standard, computed as:
\[
1 - \frac{\mathbb{E}_x[\text{AvgJSD}(m(x, n=s(x)) \parallel m(x))]}{\mathbb{E}_x[\text{AvgJSD}(m(x, n=\mu) \parallel m(x))]},
\]
where $m(x, n=v)$ denotes the model's output distribution when neuron $n$ is set to value $v$, demonstrating that correlation scoring serves as a reasonable proxy for causal importance. Critical limitations acknowledged include: (1) explanations are correlative rather than mechanistic, (2) scores degrade significantly with larger subject models, and (3) many neurons exhibit polysemanticity—activating on ``five or six different things with no discernible pattern.''

OpenAI released code at \url{https://github.com/openai/automated-interpretability} and public datasets on Azure blob storage, including per-neuron activations and explanations with simulation results.

\subsection{EleutherAI Delphi (2024): Detection Scoring at Scale}

EleutherAI's Delphi system \citep{paulo2024automatically} fundamentally redesigns the scoring methodology to achieve approximately \$1,300 for 1.5M features versus approximately \$200,000 for prior methods using closed-source models. The key insight: detection and fuzzing scoring require far fewer tokens and enable use of open-source models like Llama 3.1 70B Instruct quantized with AWQ to INT4.

\paragraph{Detection and Fuzzing Scoring.}
The critical innovation moves from token-level simulation to sequence-level detection. Rather than asking an LLM to predict activation values for every token, detection scoring asks whether whole sequences contain a feature—a simpler classification task. Detection uses activating examples sampled from quantiles of the activation distribution plus random non-activating examples, measuring both precision and recall rather than just recall. 

Fuzzing scoring marks tokens with \texttt{<<} and \texttt{>>} delimiters and asks the model to identify which examples are correctly marked, using 7 activating examples from each of 10 quantiles (5 correctly marked, 2 incorrectly marked) for 70 total examples. This approach directly tests whether explanations capture the specific token-level patterns that trigger activation.

\begin{table}[h]
\centering
\small
\begin{tabular}{@{}lccc@{}}
\toprule
Method & Prompt Tokens & Output Tokens & Runtime (5 ex.) \\
\midrule
Explanation & 397 + 566 & 29.90 & 3.14s \\
Detection/Fuzzing & 725 + 53 & 11.99 & 4.29s \\
Simulation & 24,074.85 & 1,598.1 & 73.9s \\
\bottomrule
\end{tabular}
\caption{Token and runtime comparison between scoring methods. Detection and fuzzing achieve comparable accuracy with approximately 5× fewer tokens and 17× less runtime than simulation.}
\label{tab:delphi-efficiency}
\end{table}

\paragraph{Quantile-Based Sampling.}
Delphi's sampling strategy marks another departure from Bills et al. The team found that sampling only top-N examples produces narrow explanations that fail to capture behavior across the full distribution, while uniform sampling produces overly broad explanations. The solution: sample evenly from all activation quantiles to match train/test distributions, achieving higher scores. The system uses \texttt{<<} and \texttt{>>} delimiter syntax to highlight max-activating tokens in context rather than listing tokens with tabs.

Detection + fuzzing balanced accuracy shows Pearson correlation of 0.61 with traditional simulation scoring while costing approximately 5× fewer tokens and 17× less runtime. The EleutherAI pipeline filters problematic features before expensive explanation: positional features (identified via mutual information with position $>0.05$) and unigram features ($\leq 20$ unique tokens in top 80\% activations) are flagged and handled separately.

\subsection{Adoption in Major SAE Papers}

Each major SAE publication adapts automated interpretability differently, revealing an evolution in methodology and growing recognition of limitations.

\paragraph{Bricken et al. (2023): Direct Application of Bills et al.}
``Towards Monosemanticity'' \citep{bricken2023towards} directly applies Bills et al. methodology using GPT-4 as both explainer and simulator. The paper evaluated all 4,096 features in their A/1 SAE, computing both activation-based and logit-weight-based auto-interpretability scores. They used top activating examples plus random samples from quantiles, finding that most learned features are interpretable. The paper acknowledged that interpretations may not capture all aspects of features' behaviors.

\paragraph{Templeton et al. (2024): Rubric-Based Scoring.}
``Scaling Monosemanticity'' \citep{templeton2024scaling} developed a custom rubric-based system using Claude 3 for both explanation generation and scoring. Rather than correlation scores, features receive relevance ratings: ``highly relevant,'' ``somewhat relevant,'' or ``not relevant.'' This simplification was necessary for scale—the paper examined SAEs with 1M, 4M, and 34M latents where evaluating all features is computationally intractable. Random sampling with top-activating examples found SAE features substantially more interpretable than neurons (25\% of neurons are highly relevant). The paper noted that later layers are harder to interpret and many features exhibit ``cross-layer superposition.''

\paragraph{Cunningham et al. (2024): Automated Interpretability with Layer Analysis.}
Cunningham et al.\ \citep{cunningham2024sparse} represents the first major SAE paper systematically applying Bills et al.\ automated interpretability, demonstrating that SAE features are more interpretable than neurons, PCA directions, or random directions. They sampled random nonzero activations (up to 16) to build explanations with another 16 as true positives for recall. Key finding: interpretability scores decrease in later layers, suggesting features may be more complex or require examining output effects rather than just input patterns.

\paragraph{Gao et al. (2024): Critique of Recall Bias.}
``Scaling and Evaluating Sparse Autoencoders'' \citep{gao2024scaling} from OpenAI introduced significant methodological innovations while explicitly critiquing recall bias in Bills et al.\ scoring. They adopted Neuron2Graph (N2G), which generates collections of n-grams with wildcards rather than natural language—cheaper but less expressive. Crucially, they separated precision and recall metrics: explanation precision measures how many tokens matching explanations actually activate features, while recall measures how many activations are explained. 

Their key critique: Bills et al.\ scoring disproportionately depends on recall, meaning broad explanations score well even when inaccurate. Example: a ``stop'' explanation scores highly when the actual feature is ``don't stop'' or ``can't stop''—a problem that becomes more severe as features get sparser and more specific.

\paragraph{Karvonen et al. (2025): SAEBench Synthesis.}
SAEBench \citep{karvonen2025saebench} synthesizes prior work into a comprehensive benchmark with 8 diverse metrics. For automated interpretability specifically, they use \texttt{gpt-4o-mini} as an LLM judge with detection scoring similar to EleutherAI's Delphi system \citep{paulo2024automatically}. The generation phase collects SAE activations on webtext, selects top-k sequences plus importance-weighted samples, formats with \texttt{<<token>>} highlighting, and prompts for explanations. The scoring phase creates test sets mixing top activation sequences, importance-weighted sequences, and random sequences. 

A crucial finding: automated interpretability struggles to provide statistically significant differences between SAE architectures, with Anthropic team members noting ``most SAE variants perform comparably'' on autointerp. This suggests that automated interpretability scores alone may be insufficient for SAE evaluation.

\subsection{Other Technical Implementation Details}

The technical details of these systems involve numerous design choices that significantly affect results.

\paragraph{Activation Thresholding.}
Activation thresholding varies by architecture. TopK SAEs directly enforce sparsity by keeping only $k$ largest latents (typical $k \in \{32, 64, 128\}$), eliminating need for $L_1$ penalties and preventing activation shrinkage. JumpReLU applies learned threshold $\theta$ with formula $J_\theta(x) = x \cdot \mathbb{1}_{(x>\theta)}$, requiring approximate gradients. Standard ReLU uses implicit threshold at zero with sparsity controlled by $L_1$ coefficient $\lambda$ (typical values: $\alpha \approx 7.7 \times 10^{-2}$). Gated SAEs separate selection of which latents are active from estimation of magnitudes.

\paragraph{Feature Sampling for Evaluation.}
Sampling strategies operate at two levels. \textbf{Within-feature sampling} determines which examples to show when generating explanations: EleutherAI samples from 10 activation deciles of each feature's distribution, ensuring explanations capture both weak and strong activations; N2G uses up to 16 random nonzero activations. \textbf{Between-feature sampling} determines which features to evaluate: SAEBench samples 1,000 non-dead latents using probe attribution; standard approaches use uniform random sampling of features, which may underrepresent dominant features that contribute most to model behavior.

\paragraph{Aggregation Methods.}
Aggregation methods typically employ mean or median. SAEBench reports ``median score and interquartile (25\%--75\%) range.'' EleutherAI uses mean with error bars. OpenAI's sparsity-of-effects metric uses $(L_1/L_2)^2$ ratio for downstream logit effects—SAE latents achieve 10--14\% versus residual stream channels at approximately 60\%. Self-similarity thresholds require separate optimization for different layer SAEs, with scores higher than approximately 0.01 generally indicating successful explanation generation.

\subsection{Comparison to DAME (Our Approach)}

DAME (Dominance-Aware Monosemanticity Evaluation) addresses fundamental limitations in prior automated interpretability scoring through innovations at three stages: feature sampling, explanation generation, and aggregation. Table~\ref{tab:dame-comparison} summarizes the key differences.

\begin{table}[h]
\centering
\small
\begin{tabular}{@{}p{2.5cm}p{4.5cm}p{4.5cm}@{}}
\toprule
\textbf{Stage} & \textbf{Prior Work} & \textbf{DAME} \\
\midrule
\textbf{Feature Sampling} & Uniform random (Bills et al.) or probe-based (SAEBench) & Dominance-stratified feature sampling by mean dominance $\bar{\rho}_i$ \\
\addlinespace
\textbf{Explanation Generation} & Show activation values; unbounded explanations & Hide magnitudes; length-constrained explanations \\
\addlinespace
\textbf{Scoring} & Binary classification (Delphi) or continuous correlation (Bills et al.) & Activation-balanced contrastive validation \\
\addlinespace
\textbf{Aggregation} & Uniform mean: $\frac{1}{N}\sum_i s_i$ & Per-quantile + dominance-weighted: $\sum_i w_i^{\text{dom}} s_i$ \\
\bottomrule
\end{tabular}
\caption{Comparison of DAME to prior automated interpretability approaches.}
\label{tab:dame-comparison}
\end{table}

\paragraph{Feature Sampling: From Uniform to Dominance-Stratified.}
Prior automated interpretability systems sample features uniformly at random \citep{bills2023language, bricken2023towards, cunningham2024sparse} or based on probe attribution—correlation with downstream task performance \citep{karvonen2025saebench}. Both approaches face fundamental issues when features exhibit vastly different computational influence:

\textbf{Uniform sampling} treats all features equally, regardless of their contribution to model behavior. In a typical SAE with power-law dominance distribution, a uniform sample of 100 features might include 1--2 features from the top dominance quartile (which collectively account for $>50\%$ of reconstruction magnitude) and 50+ features from the bottom quartile (contributing $<5\%$). When these per-feature scores are averaged with equal weight, the resulting global metric primarily reflects the monosemanticity of computationally negligible features.

\textbf{Probe-based sampling} (SAEBench) selects features correlated with specific downstream tasks, but this measures task-relevant feature quality rather than the overall feature dictionary quality. A feature dictionary could excel on pre-selected supervised tasks while failing catastrophically on the dominant features that drive most model computation.

DAME introduces \textbf{dominance-stratified feature sampling}: features are ranked by mean dominance $\bar{\rho}_i$ and divided into dominance quantiles, sampling equal numbers from each quantile. This ensures evaluation coverage across the full dominance spectrum, from dominant features contributing $>1\%$ of average reconstruction to rare features contributing $<0.001\%$.

\paragraph{Explanation Generation: Ensuring Comparable Information.}
Prior work directly shows activation magnitudes during explanation generation \citep{bills2023language, templeton2024scaling, paulo2024automatically}, creating systematic bias: features with different firing frequencies and magnitude distributions receive qualitatively different information:

\begin{itemize}
\item \textbf{High-frequency features} activate on many tokens, providing numerous examples with varied contexts during explanation generation—high-entropy information that enables nuanced descriptions.
\item \textbf{Rare features} activate sparsely, providing few examples with limited context—low-entropy information that forces vague or overly specific descriptions.
\item \textbf{Heavy-tailed features} (common among dominant features) show extreme magnitude variation: displaying ``activation = 0.0001'' versus ``activation = 10.0'' signals very different feature importance, biasing explanation quality.
\end{itemize}

Additionally, Bills et al.\ and subsequent work allow \textbf{unbounded explanation length}, enabling polysemantic features (especially dominant polysemantic features with diverse activation patterns) to list multiple unrelated triggers in a single description, artificially inflating scores.

DAME addresses both issues: (1) \textbf{activation magnitudes are hidden} during explanation generation—the LLM sees only ``this token triggered the feature'' without magnitude information, ensuring comparable information across features with different magnitude distributions; (2) \textbf{explanations are length-constrained}, encouraging descriptions to identify a single coherent concept rather than enumerating multiple unrelated patterns.

\paragraph{Scoring: From Binary Classification to Ranked Bins.}
Existing scoring methodologies suffer from base rate problems when comparing features with different firing frequencies:

\textbf{Bills et al.\ simulation scoring} computes Pearson correlation between predicted and actual activation values, but this is sensitive to the distribution of activations in the test set. For always-on features (activating on $>90\%$ of tokens), even random predictions achieve high correlation by chance. For rare features (activating on $<1\%$ of tokens), correlation is dominated by the abundant negative examples.

\textbf{Delphi detection scoring} \citep{paulo2024automatically} frames scoring as binary classification: does the feature activate on this sequence or not? This improves upon simulation by measuring precision and recall explicitly, but the F1 score remains sensitive to base rate: a feature activating on 50\% of tokens has balanced classes, while a feature activating on 1\% has severe class imbalance (99:1 negative:positive ratio), making F1 scores incomparable across features.

DAME introduces \textbf{activation-balanced contrastive validation}: for each feature, all test inputs are divided into three groups (low, medium, high activation) based on that feature's own activation distribution using quantile splits, and the LLM ranks one example from each group by predicted activation strength. This design ensures:
\begin{itemize}
\item \textbf{Balanced comparison:} All features receive exactly one example from each activation level, eliminating base rate effects.
\item \textbf{Magnitude sensitivity:} Unlike binary thresholding, ranking requires distinguishing strong from weak activations, directly testing whether explanations capture the full activation distribution rather than just presence/absence.
\item \textbf{Comparable difficulty:} The 3-way ranking task has identical structure for all features, making scores directly comparable across features with wildly different activation patterns.
\end{itemize}

Ranking quality is measured through four complementary metrics (perfect ranking accuracy, rank correlation, pairwise accuracy, rank distance) that together provide robust assessment of explanation quality.

\paragraph{Aggregation: From Uniform Averaging to Dominance Weighting.}
Prior work aggregates feature-level scores through uniform averaging \citep{bills2023language, paulo2024automatically, karvonen2025saebench}:
\[
\text{Global Score}_{\text{prior}} = \frac{1}{N} \sum_{i=1}^N s_i,
\]
treating all features as equally important regardless of their computational contribution. When combined with uniform feature sampling, this produces global metrics dominated by rare, low-influence features.

DAME provides two complementary aggregation approaches:

\textbf{Per-quantile reporting} separately reports monosemanticity for each dominance quantile:
\[
\text{MonoScore}^{Q_q} = \frac{1}{|Q_q|} \sum_{i \in Q_q} s_i
\]
for quantiles $Q_1, Q_2, \ldots, Q_Q$. This reveals whether dominant features are less monosemantic than rare features—a critical diagnostic for identifying failures where they matter most for model behavior. If $\text{MonoScore}^{Q_1} \ll \text{MonoScore}^{Q_4}$ (highest-dominance quartile much worse than lowest), the SAE exhibits functional polysemanticity in its most important features despite potentially high average scores.

\textbf{Dominance-weighted aggregation} provides a single summary metric for cross-SAE comparison:
\[
\text{MonoScore}_{\text{DAME}} = \sum_{i \in \text{sample}} w_i^{\text{dom}} \cdot s_i, \quad w_i^{\text{dom}} = \frac{\bar{\rho}_i}{\sum_{j \in \text{sample}} \bar{\rho}_j},
\]
where each feature's score is weighted by its mean dominance $\bar{\rho}_i$. This ensures that failures in dominant features (which drive model behavior) penalize the global score more than failures in negligible features.

Together, DAME provides the first automated interpretability framework explicitly designed to evaluate monosemanticity where it matters most—in the dominant features that drive model computation. The innovations described above address key limitations in prior work: scores become comparable across features with different base rates (via activation-balanced contrastive validation), unbounded explanations can no longer reward polysemanticity (via length constraints), and binary thresholds no longer ignore magnitude distributions (via ranking across activation levels).

\subsection{Monosemanticity vs.\ Automated Interpretability: A Fundamental Disconnect}
\label{appendix:mono-vs-interpretability}

The unbounded explanations length problem mentioned earlier reveals a deeper conceptual issue: automated interpretability metrics fundamentally measure something different from monosemanticity. As discussed in Section~\ref{mono_dom}, when explanations can be arbitrarily long, polysemantic features can list multiple unrelated triggers, inflating scores \citep{bills2023language, paulo2024automatically}. This particularly affects frequent polysemantic features, which receive more information during explanation generation, allowing multiple concepts to be covered as they have more activated examples.

This observation exposes a critical tension in the field: while SAE research explicitly aims to address superposition and find monosemantic features—features that encode a single coherent concept—the dominant evaluation metrics used throughout the literature do not actually measure how many concepts each feature encodes. Instead, automated interpretability asks: for all the concepts that a feature can cover, are those concepts interpretable or conveyable? As long as the concepts can be conveyed (even if there are many of them), the feature is considered interpretable.

Under this definition, automated interpretability is fundamentally different from monosemanticity. A polysemantic feature encoding five unrelated concepts can achieve high interpretability scores if all five concepts can be listed in an explanation, while a truly monosemantic feature encoding a single concept receives the same score. The metric does not distinguish between these cases—it measures whether concepts are conveyable, not whether there is a single concept.

This disconnect means that claims about achieving monosemanticity based solely on automated interpretability scores are technically unsupported. A feature dictionary could achieve high average interpretability scores while remaining highly polysemantic, as long as the multiple concepts encoded by each feature can be enumerated in explanations. The field's core objective—finding features that encode single concepts—requires metrics that explicitly penalize multi-concept encoding, which standard automated interpretability approaches do not provide.

\section{Benchmarking Results}
\label{appendix:benchmark_other_results}

This appendix provides detailed results for the AIU criteria evaluation, including completeness/sparsity/dead units (C2), monosemanticity (C3), non-redundancy (C1), and a comparison section.

\subsection{C2: Completeness, Sparsity, and No Dead Units}
\label{appendix:c2_completeness}

\subsubsection{Completeness, Sparsity, and Dead Feature Analysis}

We evaluate SAE performance across three complementary criteria: \textbf{sparsity} (measured as average active features per input), \textbf{dead feature proportion} (fraction of features whose firing frequency falls below a small threshold; we report several thresholds below), and \textbf{mean squared error (MSE)} (reconstruction fidelity). These metrics capture different aspects of SAE quality: sparsity measures activation efficiency, dead feature proportion indicates feature utilization, and MSE quantifies reconstruction accuracy.

\paragraph{Overall Performance Patterns.}
Table~\ref{tab:completeness_summary_by_family} summarizes mean and standard deviation for each metric across all six SAE families. \textbf{SAEBench Gemma} shows the highest average number of active features (2013 on average, indicating less sparse activations), while \textbf{OPT-125M} achieves the lowest (280 active features, indicating sparser activations). MSE varies substantially across families, with \textbf{OPT-125M} achieving the lowest mean MSE ($3.40 \times 10^{-4}$) and \textbf{SAEBench Gemma} showing the highest ($1.46 \times 10^{2}$). Dead feature proportions (computed at the firing-frequency threshold $<10^{-3}$) vary across families, ranging from $\sim$19\% (\textbf{TinyStories}) to 50\% (\textbf{CanRager}), indicating substantial variation in feature utilization efficiency.

\begin{table}[t]
\centering
\caption{Summary statistics by SAE family for sparsity (average active features), dead feature proportion, and MSE.}
\label{tab:completeness_summary_by_family}
\begin{tabular}{lrrr}
\toprule
Family & Avg Active & Dead Prop (mean) & MSE (mean) \\
\midrule
GPT2-Small      & 605.0  & 0.214 & 1.33e-01 \\
OPT-125M        & 279.9  & 0.460 & 3.40e-04 \\
Tinystories     & 376.4  & 0.187 & 3.71e-02 \\
Canrager        & 392.2  & 0.500 & 7.15e+00 \\
Saebench Gemma  & 2013.2 & 0.294 & 1.46e+02 \\
Saebench Pythia & 1106.2 & 0.392 & 8.28e+00 \\
\bottomrule
\end{tabular}

\end{table}

\begin{figure}[t]
\centering
\includegraphics[width=0.85\linewidth]{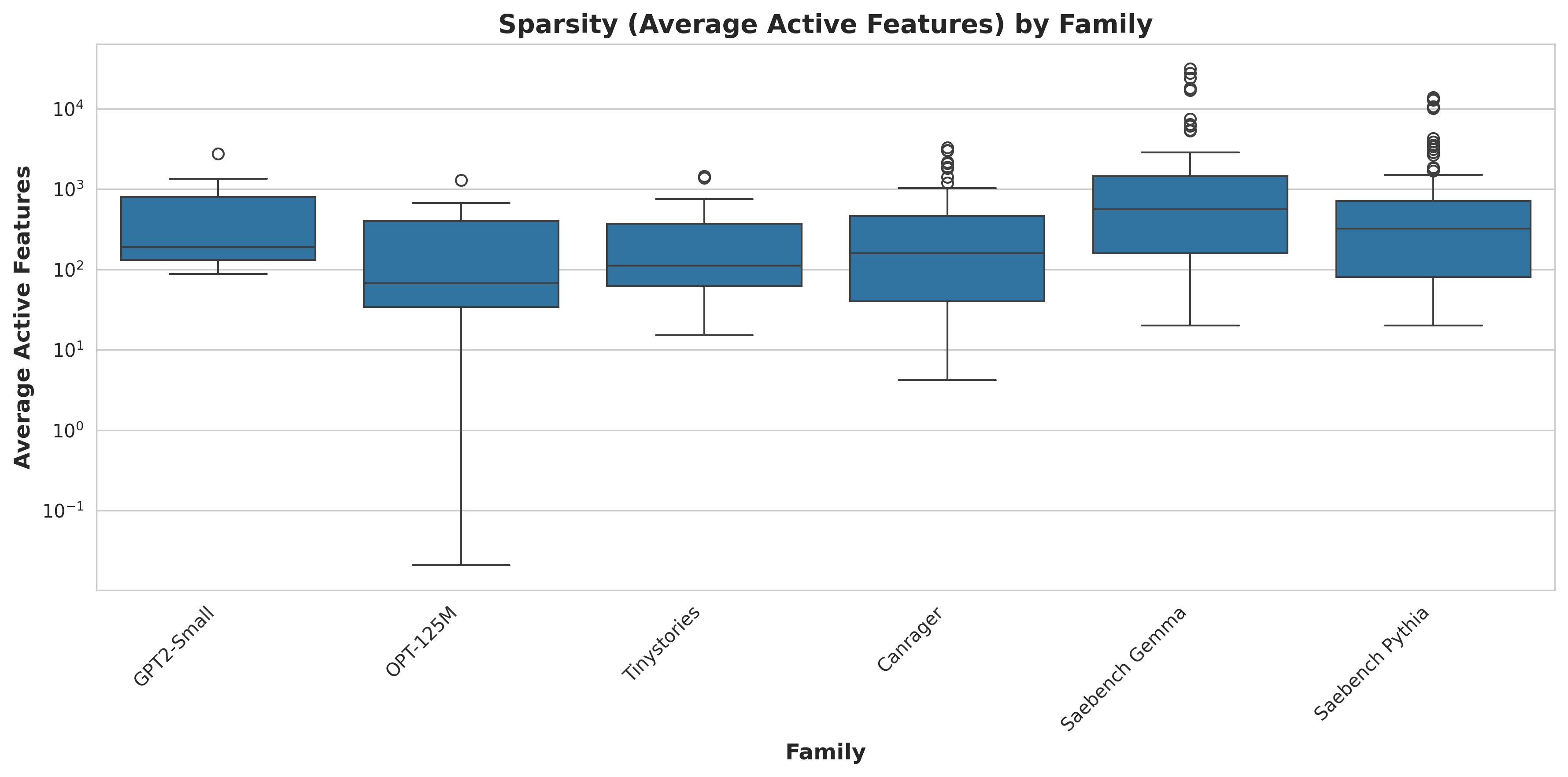}
\caption{Average active features by SAE family. Lower values indicate sparser (more efficient) activations. SAEBench Gemma shows the highest number of active features (less sparse), while OPT-125M achieves the lowest (most sparse).}
\label{fig:sparsity_by_family}
\end{figure}

\begin{figure}[t]
\centering
\includegraphics[width=0.85\linewidth]{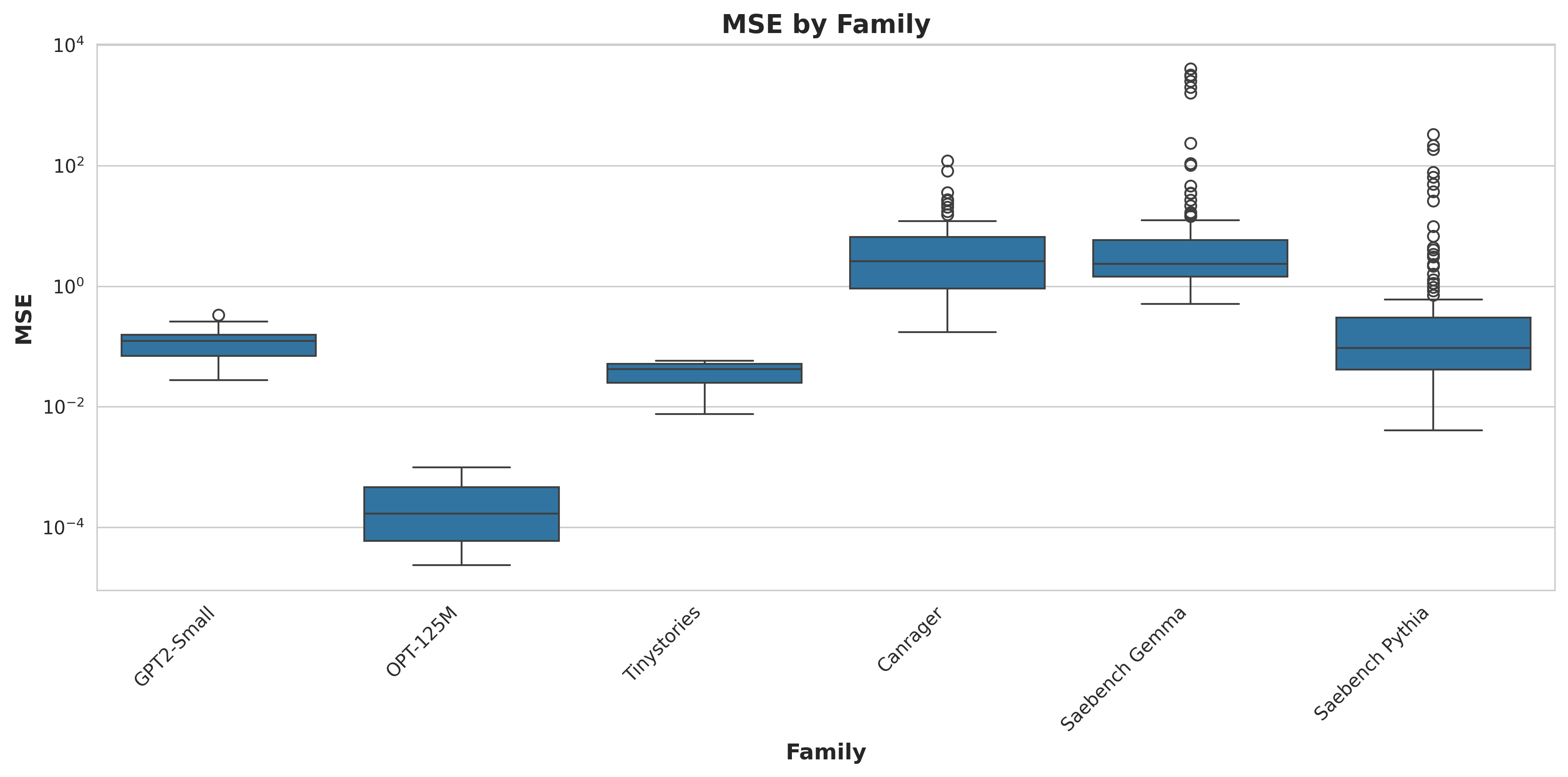}
\caption{Mean squared error (MSE) by SAE family. Lower values indicate better reconstruction fidelity. OPT-125M achieves the lowest MSE, while SAEBench Gemma shows higher reconstruction error.}
\label{fig:mse_by_family}
\end{figure}

\begin{figure}[t]
\centering
\includegraphics[width=0.85\linewidth]{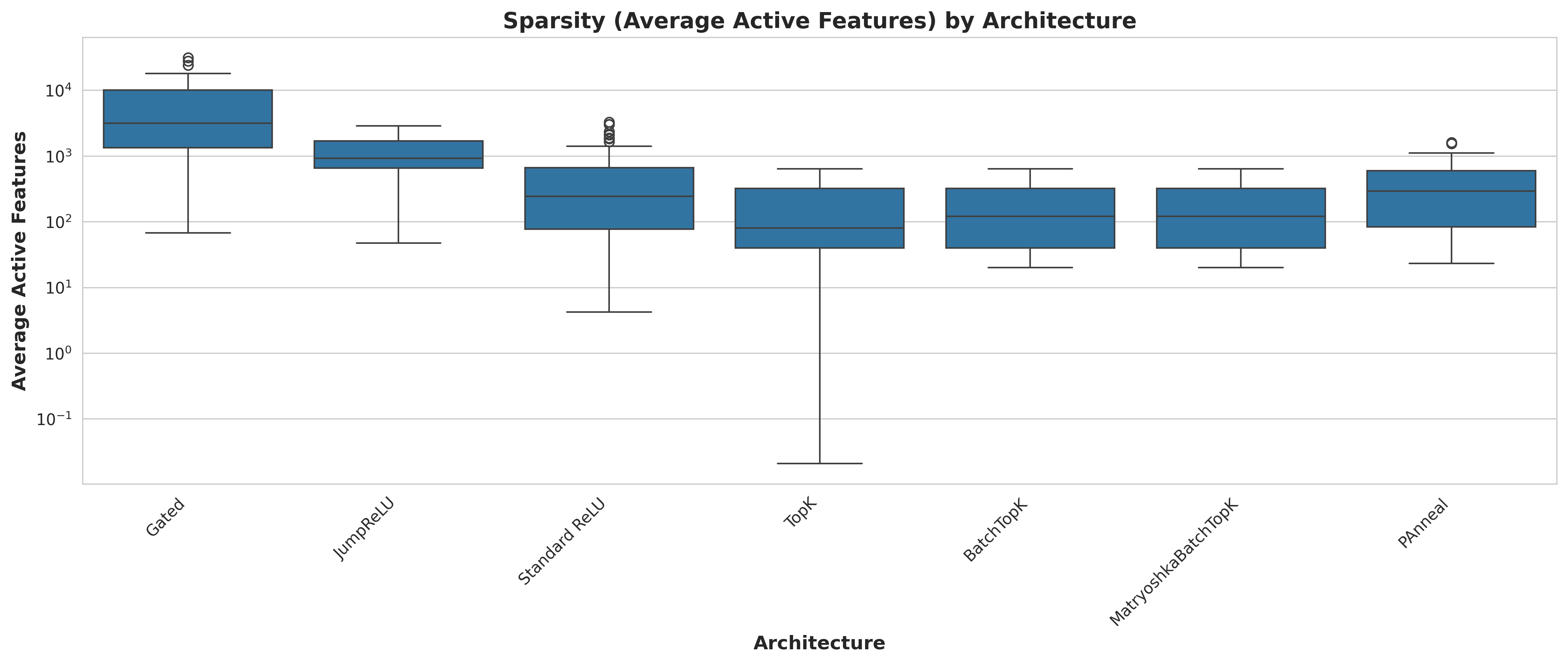}
\caption{Average active features by architecture across all families. TopK architecture achieves the lowest sparsity, while other architectures show more variable performance.}
\label{fig:sparsity_by_architecture}
\end{figure}

\begin{figure}[t]
\centering
\includegraphics[width=0.85\linewidth]{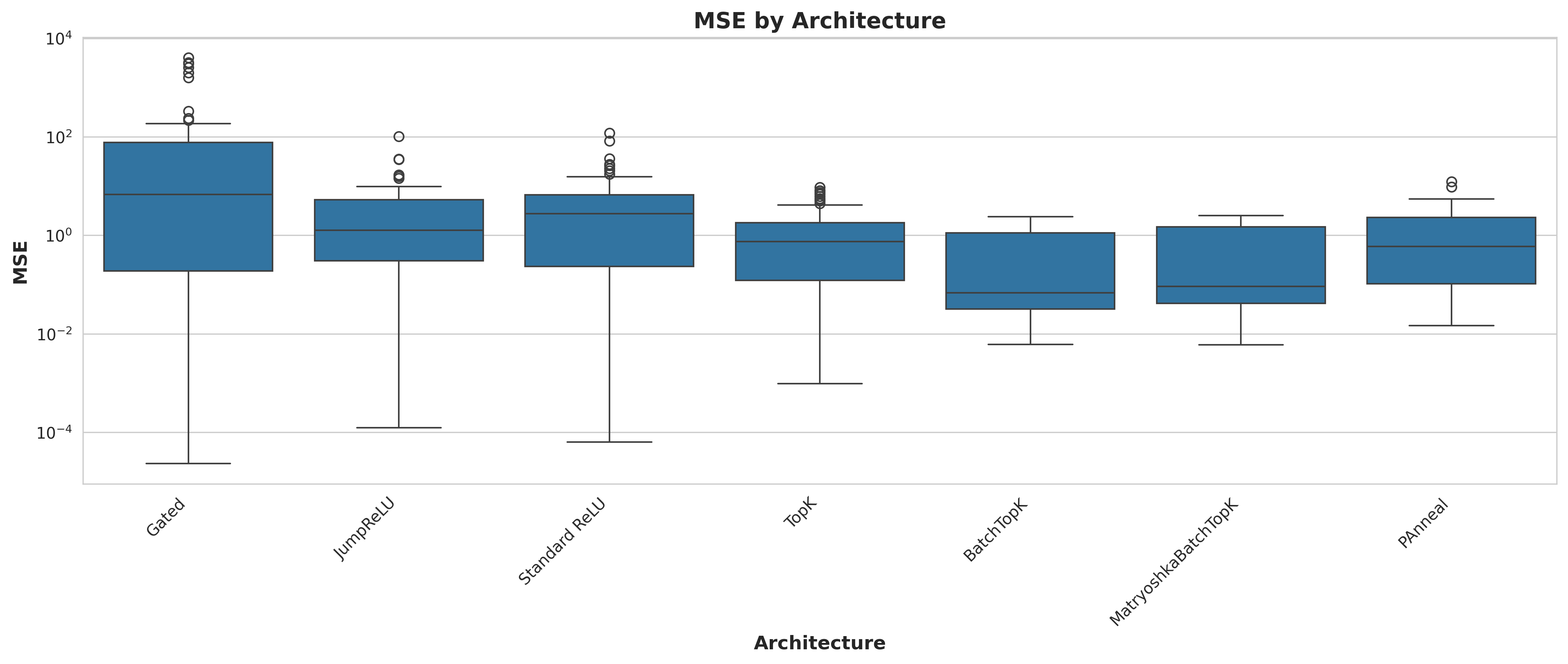}
\caption{Mean squared error by architecture across all families. Gated architecture shows the best average reconstruction fidelity, while TopK and BatchTopK show higher variance.}
\label{fig:mse_by_architecture}
\end{figure}

\paragraph{Architecture-Level Patterns.}
Figures~\ref{fig:sparsity_by_architecture} and~\ref{fig:mse_by_architecture} show performance distributions by architecture across all families. \textbf{TopK} architecture consistently achieves the lowest sparsity (fewest active features), indicating efficient activation patterns. \textbf{Gated} and \textbf{Standard ReLU} architectures show more variable sparsity, with wider distributions suggesting greater sensitivity to hyperparameter settings. For MSE, \textbf{Gated} architecture shows the best (lowest) reconstruction error on average, while \textbf{TopK} and \textbf{BatchTopK} architectures show higher variance in reconstruction fidelity.

\paragraph{Top Performers.}
Since hyperparameter settings vary substantially across families, direct cross-family comparisons may not be meaningful. We identify the top 10 performing configurations overall and within each family based on composite rankings across all three criteria (sparsity, dead feature proportion, and MSE). Tables showing these rankings are available in the supplementary materials.

\textbf{Sparsity (Average Active Features):} TopK architecture consistently achieves the best sparsity (lowest active features) across families, indicating efficient activation patterns. Within-family top performers show consistent patterns, with TopK architectures dominating top positions in most families.

\textbf{Dead Feature Proportion:} Dead feature proportions (firing frequency $<10^{-3}$) show substantial variation across families and configurations. Top performers achieve dead feature proportions near zero, indicating excellent feature utilization. However, average dead feature proportions across families range from $\sim$19\% to 50\%, suggesting that many SAE configurations have significant numbers of rarely-firing features. This variation highlights the importance of hyperparameter tuning for achieving efficient feature utilization.

\textbf{MSE (Reconstruction Fidelity):} \textbf{OPT-125M} shows the best overall MSE values across families, with top performers achieving MSE below $10^{-4}$. \textbf{CanRager} and \textbf{SAEBench} families show higher MSE values, likely due to different model scales and activation distributions. Gated architecture shows the best average reconstruction fidelity across families.

\paragraph{Key Observations.}
Several patterns emerge from this analysis:

\begin{enumerate}
\item \textbf{Architecture-specific strengths:} TopK architecture consistently achieves the best sparsity (lowest active features), while Gated architecture shows the best reconstruction fidelity (lowest MSE) on average. This suggests a potential trade-off between activation efficiency and reconstruction accuracy.

\item \textbf{Family-specific variation:} MSE values vary substantially across families (ranging from $10^{-4}$ for OPT-125M to $10^{2}$ for SAEBench Gemma), indicating that the underlying model and activation distribution significantly influence reconstruction difficulty. This variation makes cross-family comparisons less meaningful than within-family rankings.

\item \textbf{Feature utilization variation:} Across the 610 SAEs spanning all 7 architectures and 6 families, computed on $\sim$122k validation tokens, the strict ``never fires'' rate (firing count $= 0$) is near-zero for every architecture. Under a less strict threshold (firing frequency $<10^{-3}$), per-SAE median dead rates vary widely: from near-zero for Gated SAEs to roughly 0--80\% across the other architectures (with BatchTopK and Matryoshka highest). The very-rare threshold ($<10^{-4}$) gives a similarly wide spread. This variation underscores that ``dead'' is threshold-dependent and that achieving efficient feature utilization requires careful hyperparameter tuning, especially for sparse-activation architectures.

\item \textbf{Hyperparameter sensitivity:} The wide distributions within each architecture-family combination (visible in the box plots) indicate that hyperparameter settings play a crucial role in determining performance. This reinforces the importance of within-family comparisons rather than cross-family rankings.
\end{enumerate}

\paragraph{Sample-size note for the redundancy dead-feature filter.}
The dead-feature filter applied in Section~\ref{subsec:redundancy} uses a
different sample. Specifically, dead features are detected as features with
zero observed firings within the redundancy evaluation sample
($\sim$12k tokens, 3 batches $\times$ 4096 tokens). At this smaller sample
size, a feature whose true firing frequency is $\sim$$10^{-5}$ has expected count
$\sim$0.12 and is often labelled as ``dead.'' In practice this approximates a
``firing frequency $<10^{-4}$'' threshold rather than a strict ``never fires''
threshold. The dead-feature rates reported above (122k tokens) and the
per-architecture pair filter rates reported in
Appendix~\ref{appendix:c1_redundancy} therefore measure related but
non-identical quantities, and may differ on the same SAE.

\subsection{C3: Monosemanticity (DAME)}
\label{appendix:c3_dame}

\subsubsection{Per-Architecture $L_0$}
\label{appendix:per_arch_l0}

The DAME framework partitions features into dominance quantiles based on per-input
attribution share. When an SAE has very high $L_0$ (many features active per
input), each feature contributes a smaller fraction of the activation magnitude,
diluting the dominance signal. Table~\ref{tab:per_arch_l0} reports median $L_0$
per architecture in three forms: absolute count, percentage of dictionary size,
and percentage of the activation dimension $d_{\text{in}}$
of the residual stream. The $L_0 / d_{\text{in}}$ column is the most
informative diagnostic: a sparse decomposition should have $L_0 \ll d_{\text{in}}$,
otherwise the SAE is no longer compressing the activation into a sparse code.

Top-$k$-based architectures (BatchTopK, TopK, Matryoshka) sit comfortably below
$10\%$ of $d_{\text{in}}$. PAnneal, Standard ReLU, and JumpReLU range from
$\sim 20$--$50\%$. \textbf{Gated SAEs in our benchmark have median $L_0$ of
$\mathbf{126.9\%}$ of $d_{\text{in}}$}---they activate \emph{more} features
per input than the residual stream has dimensions. This regime is not a
meaningful sparse decomposition: dominance is spread thin across many active
features, and the per-feature attribution share that DAME measures is too small
to be informative. This explains why the Q1 deficit is not significant for
Gated and JumpReLU on real LLMs even though it appears strongly under controlled
synthetic sparsity (Appendix~\ref{appendix:synthetic-validation}).

\paragraph{Hyperparameter transfer caveat.}
For the architectures we trained ourselves (BatchTopK, TopK, Matryoshka,
JumpReLU, Gated, Standard ReLU, PAnneal on GPT-2, OPT-125M, and TinyStories),
we adopted hyperparameters from SAEBench's Pythia and Gemma configurations as a
starting point. The result is a wide divergence in effective sparsity across
architectures: top-$k$ variants enforce sparsity by construction (independent of
the regularizer), while Gated and JumpReLU rely on a soft penalty that does not
transfer cleanly across LLMs of different scales and activation distributions.
The high $L_0$ values reported above (especially for Gated) likely reflect this
hyperparameter mismatch rather than an intrinsic architectural property. A
proper per-LLM sparsity sweep is left to future work; the synthetic validation
(Appendix~\ref{appendix:synthetic-validation}) sidesteps this issue by
controlling sparsity directly during training.

\begin{table}[t]
\centering
\footnotesize
\caption{Per-architecture median $L_0$ in absolute count, as a percentage of dictionary size, and as a percentage of model dimensionality $d_{\text{in}}$. Computed per-SAE then aggregated across SAEs of each architecture (each SAE weighted equally). Last column is the most informative: a sparse decomposition should have $L_0 \ll d_{\text{in}}$. Gated's $\sim$127\% is in the regime where the SAE is no longer sparse; JumpReLU's $\sim$53\% is borderline.}
\label{tab:per_arch_l0}
\begin{tabular}{lrrr}
\toprule
Architecture & Median $L_0$ & $L_0$/dict (\%) & $L_0$/$d_{\text{in}}$ (\%) \\
\midrule
BatchTopK     &   49 & 0.30  & 5.21 \\
TopK          &  100 & 0.61  & 6.94 \\
Matryoshka    &   50 & 0.49  & 5.21 \\
PAnneal       &  292 & 1.31  & 20.56 \\
Standard ReLU &  354 & 2.17  & 32.81 \\
Gated         & 1383 & 17.34 & \textbf{126.89} \\
JumpReLU      &  682 & 3.05  & 52.69 \\
\bottomrule
\end{tabular}
\end{table}

\subsubsection{Perfect Ranking Accuracy Results}

\begin{figure}[t]
\centering
\includegraphics[width=0.85\linewidth]{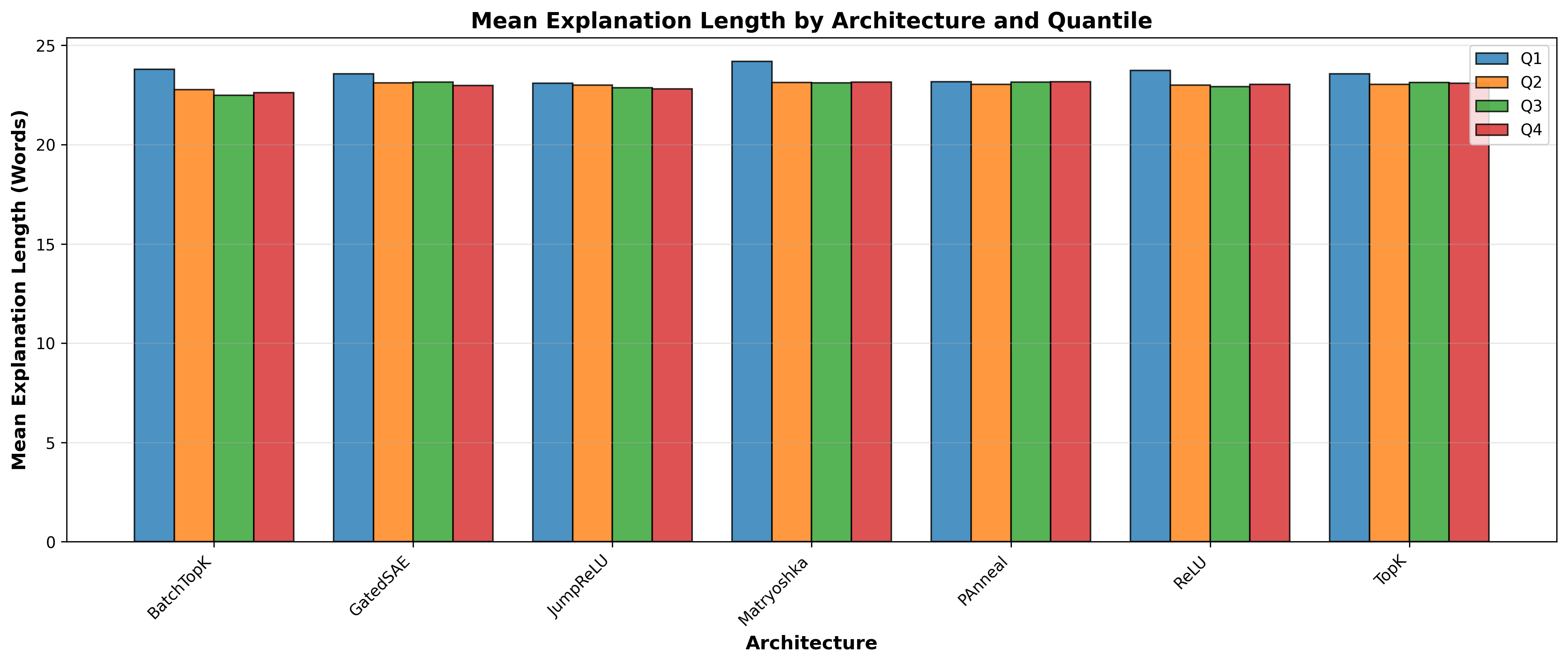}
\caption{Mean explanation length by architecture and quantile. Q1 explanations are about 0.6 words longer than Q2--Q4 on average (23.6 vs.\ 23.0 words; paired $p < 10^{-21}$).}
\label{fig:architecture_explanation_length}
\end{figure}

\begin{figure}[t]
\centering
\includegraphics[width=0.85\linewidth]{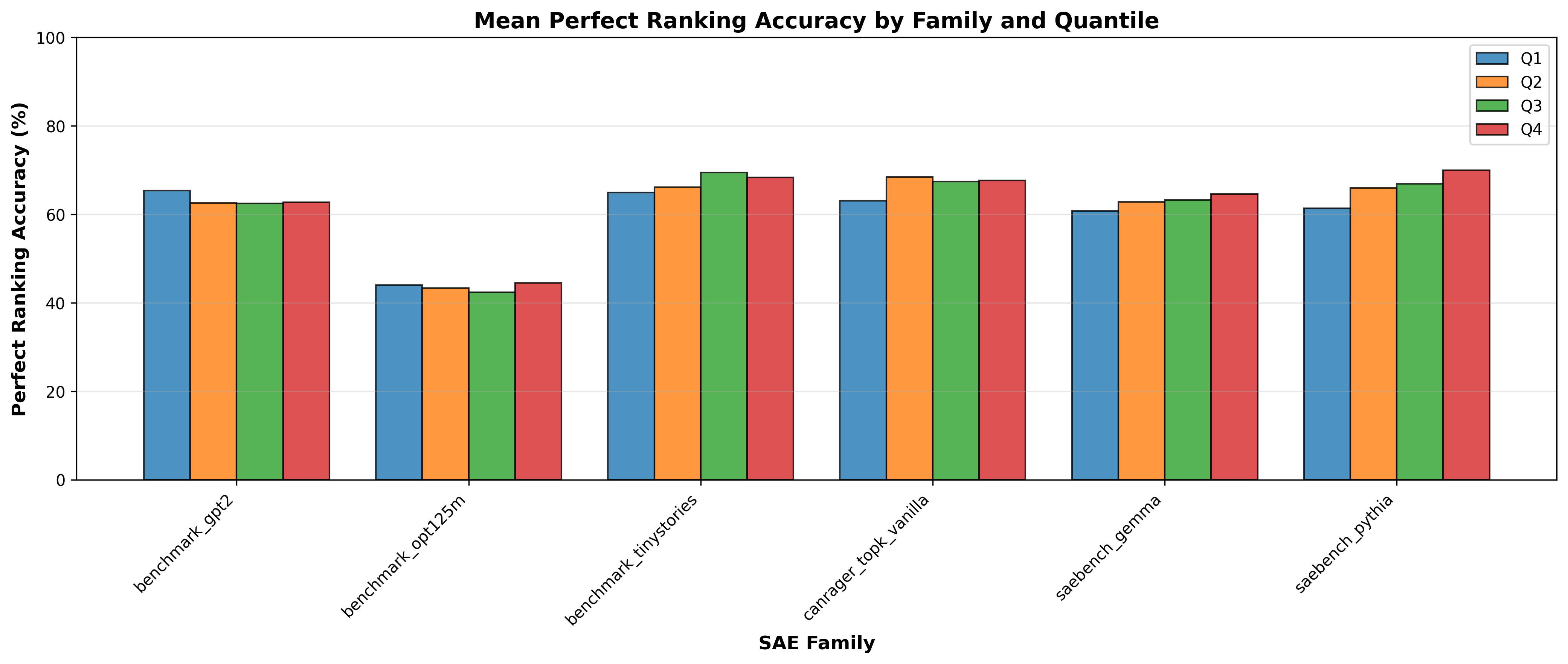}
\caption{Mean perfect ranking accuracy by SAE family and quantile (token-based). The Q1 underperformance pattern varies by family, with some families showing Q1 performing better than Q2--Q4.}
\label{fig:family_accuracy}
\end{figure}

\paragraph{Per-family quantile breakdown.}
Table~\ref{tab:family_perfect_ranking_acc_quantile} reports the per-family DAME perfect-ranking-accuracy values across quantiles. SAEBench Pythia shows the largest Q1 gap; GPT2-Small SAEs show Q1 outperforming Q2--Q4. Table~\ref{tab:saebench_pythia_architecture_quantile} drills into SAEBench Pythia by architecture, where BatchTopK shows the largest gap (-9.8 pp), followed by Matryoshka (-8.2), TopK (-8.0), and Standard ReLU (-7.2).

\begin{table}[t]
\centering
\begin{tabular}{lccccc}
\toprule
Family & Q1 & Q2 & Q3 & Q4 & Diff \\
\midrule
GPT2-Small & 65.3 & 62.5 & 62.5 & 62.7 & \phantom{-}2.7 \\
OPT125M & 44.0 & 43.3 & 42.3 & 44.5 & \phantom{-}0.6 \\
Tinystories & \textbf{64.9} & 66.1 & 69.5 & 68.3 & \textbf{-3.1} \\
Canrager & \textbf{63.1} & 68.4 & 67.4 & 67.7 & \textbf{-4.7} \\
Jbloom & \textbf{66.5} & 67.0 & 67.1 & 69.6 & \textbf{-1.4} \\
Saebench Gemma & \textbf{60.7} & 62.8 & 63.2 & 64.6 & \textbf{-2.8} \\
Saebench Pythia & \textbf{61.3} & 65.9 & 66.9 & 69.9 & \textbf{-6.3} \\
\bottomrule
\end{tabular}

\caption{Per-family average DAME perfect ranking accuracy by quantile.}
\label{tab:family_perfect_ranking_acc_quantile}
\end{table}

\begin{table}[t]
\centering
\begin{tabular}{lccccc}
\toprule
Architecture & Q1 & Q2 & Q3 & Q4 & Diff \\
\midrule
BatchTopK   & \textbf{57.3} & 68.5 & 65.6 & 67.2 & \textbf{-9.8} \\
Gated    & 62.4          & 61.1 & 63.2 & 66.3 & \textbf{-1.1} \\
JumpReLU    & \textbf{60.5} & 63.1 & 64.6 & 68.2 & \textbf{-4.8} \\
Matryoshka  & \textbf{57.4} & 64.1 & 64.5 & 68.2 & \textbf{-8.2} \\
PAnneal     & \textbf{69.2} & 71.1 & 74.1 & 77.3 & \textbf{-5.0} \\
Standard ReLU        & \textbf{63.9} & 65.3 & 72.1 & 75.8 & \textbf{-7.2} \\
TopK        & \textbf{58.4} & 68.3 & 64.2 & 66.7 & \textbf{-8.0} \\
\bottomrule
\end{tabular}

\caption{Per-architecture quantile accuracy for SAEBench Pythia family (token-based evaluation).}
\label{tab:saebench_pythia_architecture_quantile}
\end{table}

\paragraph{Per-family explanation length.}
Q1 features consistently receive longer explanations than Q2--Q4 across all
families (Table~\ref{tab:family_word_count}); yet, as reported in
Section~\ref{subsec:dame}, this additional descriptive capacity does not
translate into higher monosemanticity for Q1 features.

\begin{table}[t]
\centering
\begin{tabular}{lccccc}
\toprule
Family & Q1 & Q2 & Q3 & Q4 & Diff \\
\midrule
GPT2-Small & \textbf{25.3} & 23.0 & 22.8 & 22.6 & \phantom{-}2.5 \\
OPT125M & 23.3 & 23.8 & 23.8 & 23.7 & \textbf{-0.5} \\
Tinystories & 21.9 & 22.7 & 22.4 & 22.8 & \textbf{-0.7} \\
Canrager & \textbf{23.7} & 23.1 & 23.2 & 23.2 & \phantom{-}0.5 \\
Jbloom & \textbf{24.2} & 22.3 & 23.0 & 22.9 & \phantom{-}1.5 \\
Saebench Gemma & \textbf{24.0} & 23.0 & 23.2 & 23.2 & \phantom{-}0.9 \\
Saebench Pythia & \textbf{23.3} & 22.9 & 22.8 & 22.7 & \phantom{-}0.5 \\
\bottomrule
\end{tabular}

\caption{Per-family average explanation length (number of words) by quantile.}
\label{tab:family_word_count}
\end{table}

\paragraph{Sparsity-Monosemanticity Relationships.}
We analyzed correlations between effective sparsity (average active features) and monosemanticity across 88 SAE configurations (controlling for width and architecture). Table~\ref{tab:sparsity_correlation_summary} aggregates these results for both perfect ranking accuracy and pairwise accuracy metrics. Overall, the relationship between sparsity and interpretability is weak and inconsistent: only 3.4\% of correlations reach statistical significance (p < 0.05) for overall perfect ranking accuracy, and 2.3\% for Q1-specific accuracy. Pairwise accuracy shows slightly higher significance rates (4.5\% overall, 6.9\% for Q1), but still indicates weak effects.

Notably, for Q1 features specifically, 56.8\% of configurations show positive correlations for perfect ranking accuracy (less sparse SAEs have higher Q1 interpretability), compared to 39.8\% for overall accuracy. For pairwise accuracy, this pattern is even stronger: 59.8\% of Q1 correlations are positive versus 38.6\% overall. This suggests that sparsity may have a slightly more positive relationship with Q1 feature interpretability, though the effect remains weak overall. The lack of strong, consistent sparsity effects indicates that other factors (architecture, training procedure, model family) play more dominant roles in determining feature monosemanticity.

\begin{table}[t]
\centering
\small
\begin{tabular}{lrr}
\toprule
\textbf{Metric} & \textbf{Positive} & \textbf{Negative} \\
\midrule
Sparsity $\rightarrow$ Perfect Ranking Acc & 35 (39.8\%) & 53 (60.2\%) \\
Sparsity $\rightarrow$ Q1 Perfect Ranking Acc & 50 (56.8\%) & 38 (43.2\%) \\
Sparsity $\rightarrow$ Pairwise Acc & 34 (38.6\%) & 54 (61.4\%) \\
Sparsity $\rightarrow$ Q1 Pairwise Acc & 52 (59.8\%) & 35 (40.2\%) \\
\bottomrule
\end{tabular}
\caption{Aggregated sparsity correlation results across 88 SAE configurations (controlling for width and architecture). Positive correlations indicate that less sparse SAEs (more active features) have higher interpretability scores.}
\label{tab:sparsity_correlation_summary}
\end{table}

\begin{figure}[t]
\centering
\includegraphics[width=0.85\linewidth]{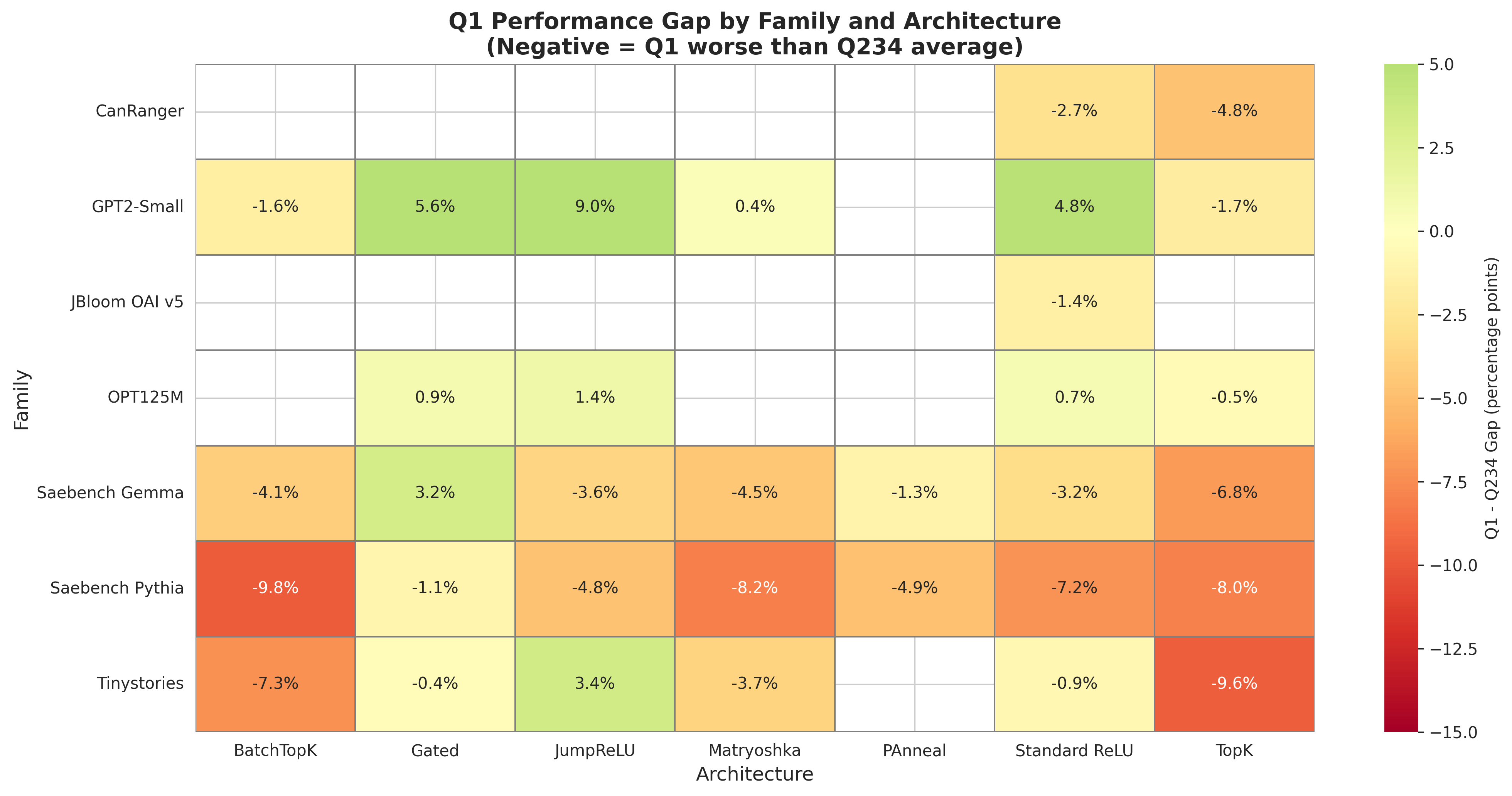}
\caption{Q1 performance gap (Q1 accuracy minus Q2--Q4 average, in percentage points) by family and architecture. Negative values indicate Q1 underperformance. The pattern varies substantially across combinations, with SAEBench Pythia showing larger gaps than GPT2-Small.}
\label{fig:family_architecture_q1_gap}
\end{figure}

\paragraph{Family-Architecture Interactions.}
The Q1 performance gap varies substantially across family-architecture combinations (Figure~\ref{fig:family_architecture_q1_gap}). \textbf{SAEBench Pythia} with Standard ReLU shows a large negative gap (-7.2 percentage points), while \textbf{GPT2-Small} with Gated shows a positive gap (+1.4 points), indicating Q1 features actually outperform Q2--Q4 in this configuration. This family-specific variation suggests that the Q1 underperformance pattern is not universal but depends on both the base model and SAE architecture.

\begin{figure}[t]
\centering
\includegraphics[width=0.85\linewidth]{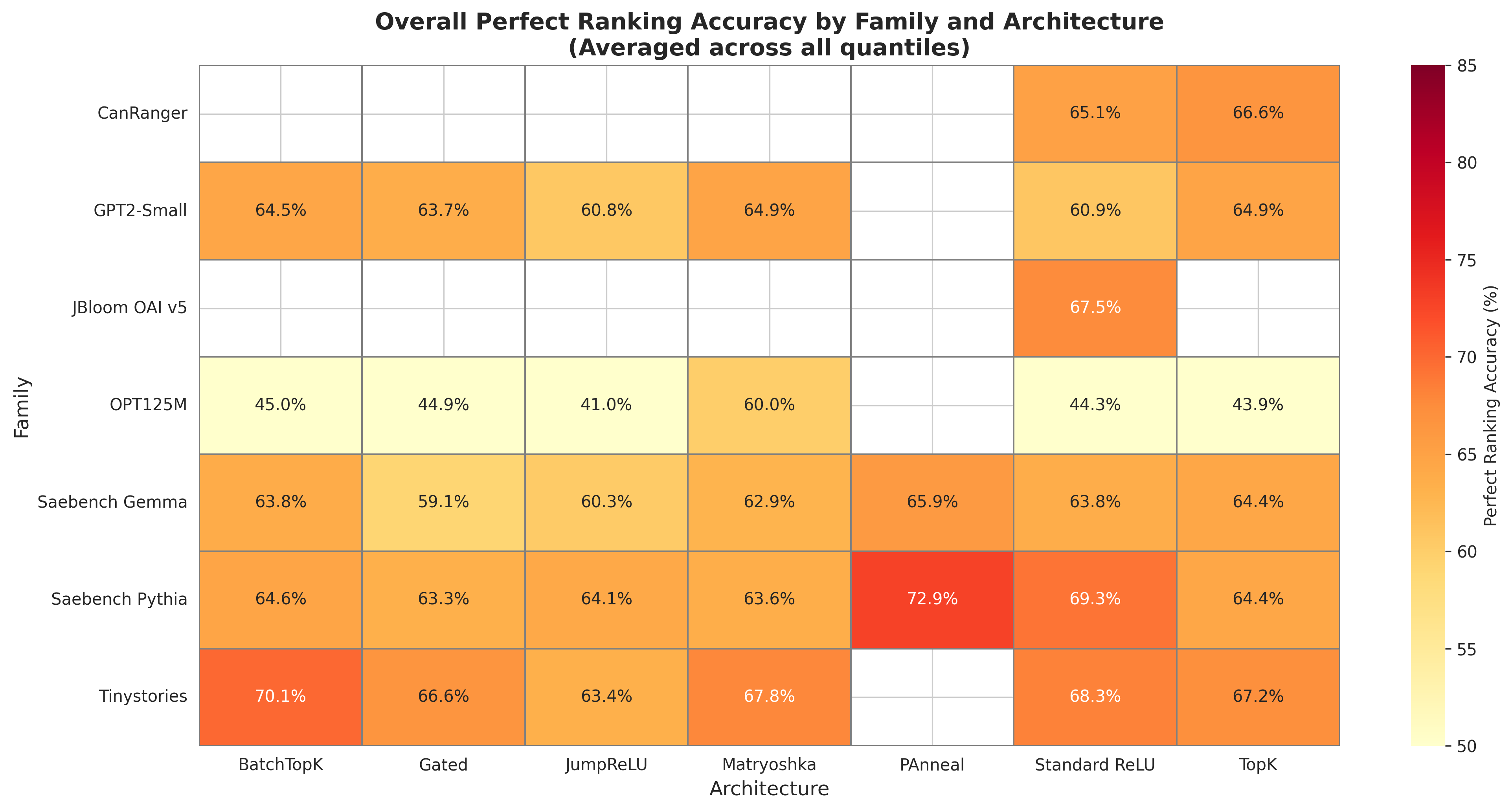}
\caption{Overall perfect ranking accuracy (averaged across all quantiles) by family and architecture. Higher values indicate better overall monosemanticity. PAnneal architecture consistently achieves high scores across families.}
\label{fig:overall_perfect_ranking_acc}
\end{figure}

\begin{figure}[t]
\centering
\includegraphics[width=0.85\linewidth]{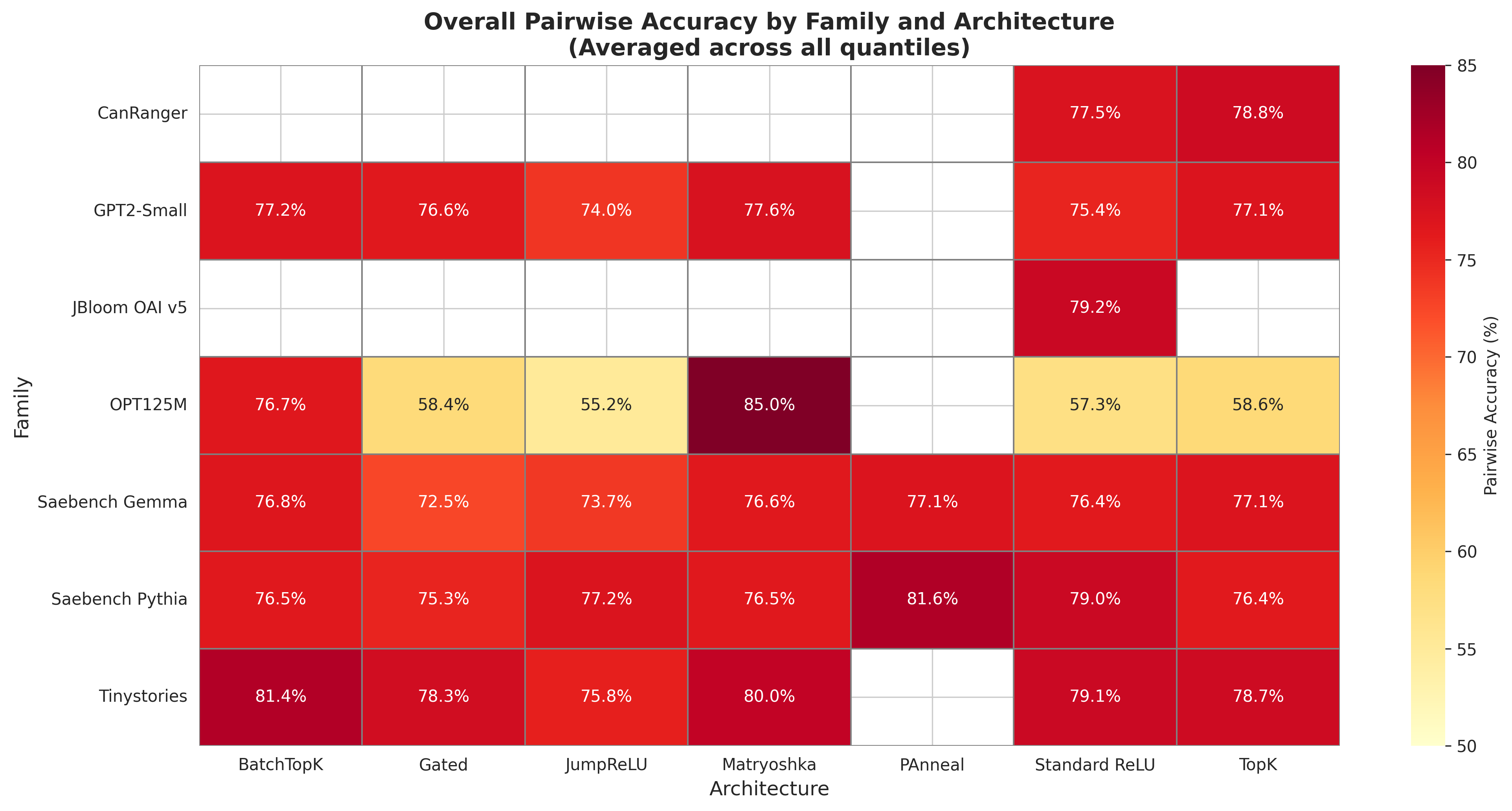}
\caption{Overall pairwise accuracy (averaged across all quantiles) by family and architecture. Higher values indicate better overall monosemanticity. Patterns are similar to perfect ranking accuracy, with PAnneal and Gated architectures performing well.}
\label{fig:overall_pairwise_acc}
\end{figure}

\paragraph{Overall Monosemanticity Patterns.}
Figures~\ref{fig:overall_perfect_ranking_acc} and~\ref{fig:overall_pairwise_acc} show overall monosemanticity (averaged across all quantiles) by family and architecture. These heatmaps reveal the general interpretability patterns across SAE configurations, complementing the Q1-specific analysis. \textbf{PAnneal} architecture consistently achieves high overall perfect ranking accuracy across families, with SAEBench Pythia PAnneal configurations reaching up to 78.8\%. \textbf{Standard ReLU} shows more variable performance, with GPT2-Small achieving higher scores than SAEBench Pythia. The pairwise accuracy heatmap shows similar patterns, with PAnneal and Gated architectures generally performing well across families.

\begin{figure}[t]
\centering
\includegraphics[width=0.85\linewidth]{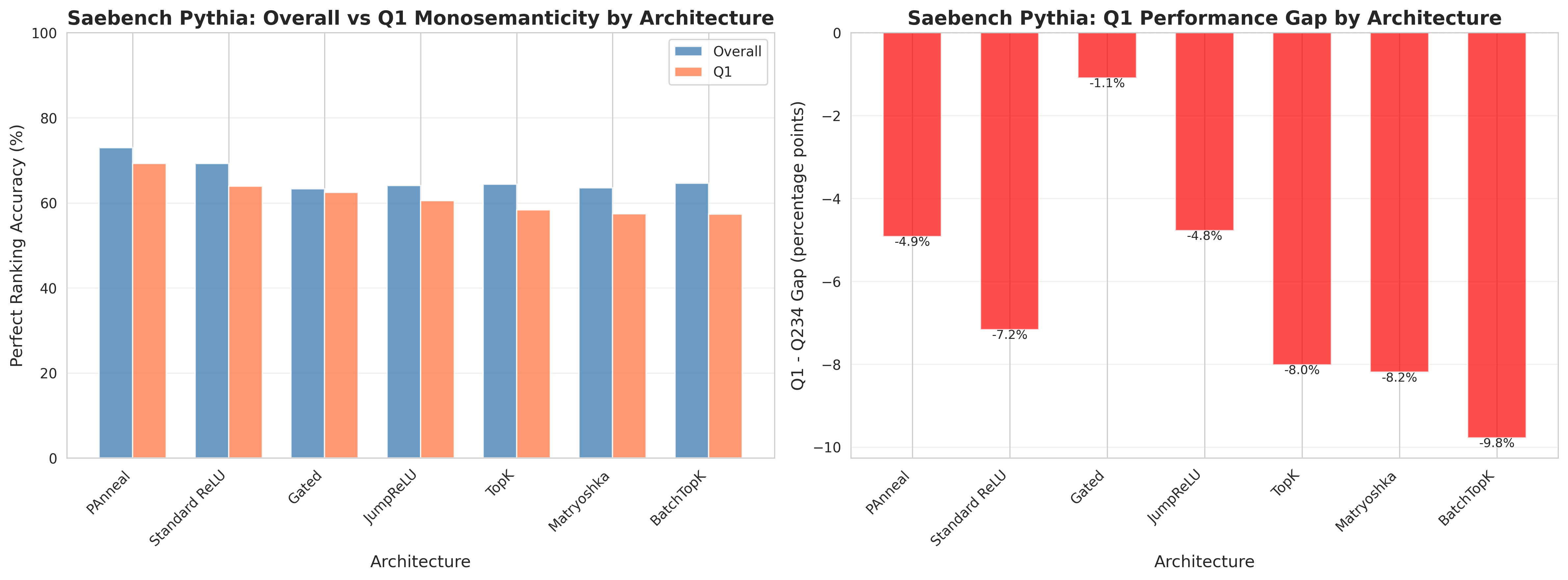}
\caption{SAEBench Pythia: Architecture comparison showing overall vs Q1 monosemanticity (left) and Q1 performance gap (right). Negative gaps indicate Q1 underperformance relative to Q2--Q4 average.}
\label{fig:saebench_pythia_architecture}
\end{figure}

\begin{figure}[t]
\centering
\includegraphics[width=0.85\linewidth]{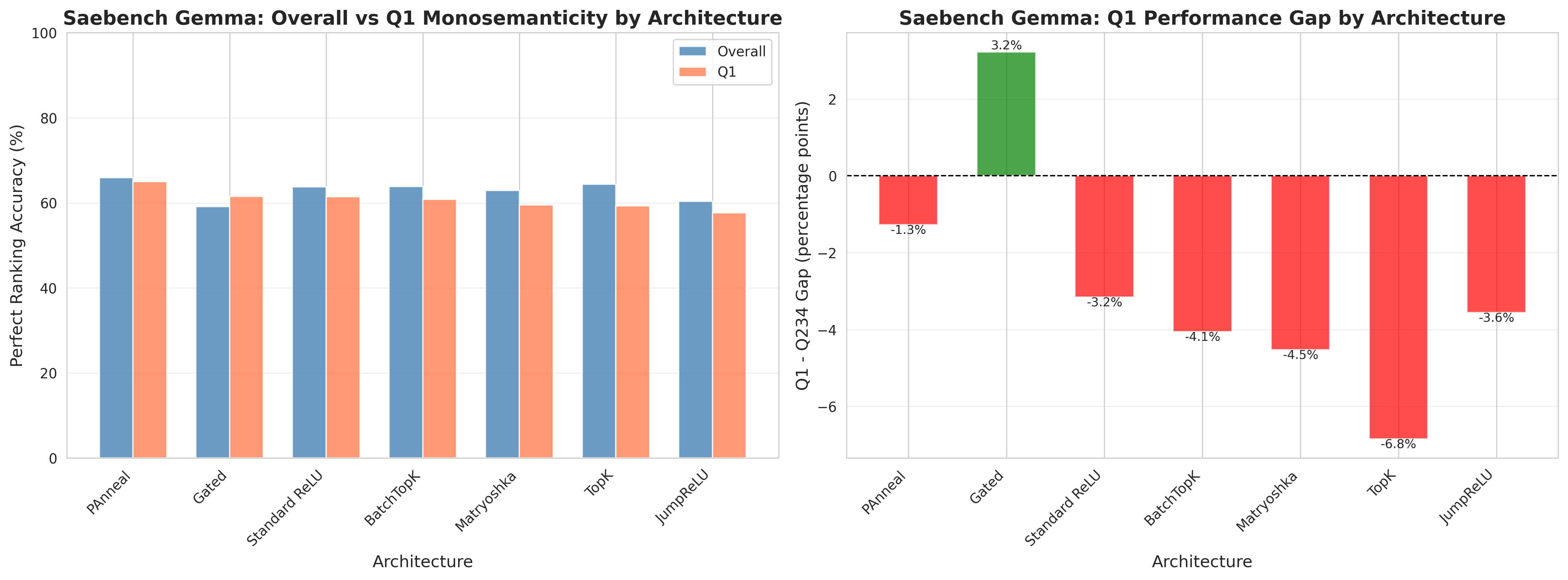}
\caption{SAEBench Gemma: Architecture comparison showing overall vs Q1 monosemanticity (left) and Q1 performance gap (right). Negative gaps indicate Q1 underperformance relative to Q2--Q4 average.}
\label{fig:saebench_gemma_architecture}
\end{figure}

\paragraph{SAEBench Pythia and Gemma: Architecture-Specific Patterns.}
Figures~\ref{fig:saebench_pythia_architecture} and~\ref{fig:saebench_gemma_architecture} provide detailed architecture comparisons for SAEBench Pythia and SAEBench Gemma, two of the largest SAE families in our evaluation. These families show distinct architecture-dependent patterns in how they handle Q1 (dominant) versus Q2--Q4 features.

For \textbf{SAEBench Pythia}, BatchTopK shows the largest Q1 gap (-9.8 pp), Standard ReLU is at -7.2 pp, and PAnneal shows the smallest gap among statistically significant architectures (-5.0 pp); Gated (-1.1) and JumpReLU (-4.8) are not statistically significant. Across architectures, Q1 features consistently underperform Q2--Q4, but the magnitude varies substantially. PAnneal achieves the highest overall accuracy (72.9\%, averaged across all quantiles) and maintains relatively good Q1 performance (69.2\%), suggesting it handles dominant features better than other architectures.

For \textbf{SAEBench Gemma}, the patterns are more mixed. While most architectures show Q1 underperformance, the gaps are generally smaller than in Pythia. Gated architecture shows particularly interesting behavior, with Q1 performance closer to overall performance compared to other architectures.

These family-specific analyses reveal that the Q1 underperformance phenomenon is not uniform across architectures, and that some architectures (notably PAnneal) are better at maintaining interpretability for dominant features.

\paragraph{Top-Performing SAE Configurations.}
Tables~\ref{tab:top10_overall_mono},~\ref{tab:top10_q1_gap}, and~\ref{tab:top10_q1_mono} identify the top 10 SAE configurations across three metrics: overall monosemanticity, Q1 vs.\ Q2--Q4 gap (largest negative gaps indicating worst Q1 performance), and Q1 monosemanticity. These tables reveal several important patterns.

\textbf{Top 10 Overall Monosemanticity:} PAnneal architecture dominates the top performers, with 5 of the top 10 configurations. SAEBench Pythia PAnneal at 16k and 4k widths both achieve 78.8\% overall accuracy---the highest in our evaluation. These top performers show relatively high dominance values (median top 10 feature dominance around 0.0017), suggesting that high overall interpretability can coexist with moderate feature dominance.

\textbf{Top 10 Q1 vs.\ Q2--Q4 Gap:} The configurations with the largest negative gaps (Q1 worst relative to Q2--Q4) are predominantly from SAEBench Pythia and TinyStories families. BatchTopK and TopK architectures feature prominently, with gaps reaching -19.6 percentage points (SAEBench Pythia BatchTopK 16k). Interestingly, these configurations show lower dominance values (median around 0.0011) compared to top overall performers, suggesting that lower dominance does not necessarily protect against Q1 underperformance.

\textbf{Top 10 Q1 Monosemanticity:} JBloom OAI v5 configurations achieve the highest Q1 scores (78.2\% at 32k width), followed by SAEBench Pythia PAnneal configurations. These top Q1 performers show very low dominance values (around 0.0003-0.0005), suggesting that extremely low dominance may be associated with better Q1 interpretability. However, they also show lower overall accuracy compared to top overall performers, indicating a trade-off.

\textbf{Overlap Analysis:} There is minimal overlap between the three top 10 lists: only 4 configurations appear in both top 10 overall and top 10 Q1, and no configurations appear in all three lists. This suggests that optimizing for overall monosemanticity, Q1 performance, or minimizing Q1 gap are largely independent objectives. The lack of overlap indicates that different SAE configurations excel at different aspects of interpretability, and there is no single ``best'' configuration that dominates across all metrics.

\begin{table}[t]
\centering
\small
\begin{tabular}{lcccrr}
\toprule
\textbf{Family} & \textbf{Arch} & \textbf{Width} & \textbf{Overall} & \textbf{Sparsity} & \textbf{Dominance (\%)} \\
\midrule
Saebench Pythia & PAnneal & 16k  & 78.8\% & 27.6   & 0.30\% \\
Saebench Pythia & PAnneal & 4k   & 78.8\% & 22.9   & 0.34\% \\
Saebench Pythia & PAnneal & 65k  & 76.5\% & 38.2   & 0.19\% \\
JBloom OAI v5   & Standard ReLU & 128k & 75.5\% & 893.9  & 0.03\% \\
Saebench Pythia & PAnneal & 16k  & 74.8\% & 74.1   & 0.11\% \\
JBloom OAI v5   & Standard ReLU & 128k & 74.8\% & 1492.3 & 0.03\% \\
Saebench Pythia & PAnneal & 65k  & 74.2\% & 99.6   & 0.07\% \\
Saebench Pythia & BatchTopK & 16k & 73.5\% & 40.0   & 0.15\% \\
Saebench Gemma  & TopK    & 16k  & 73.4\% & 40.0   & 0.14\% \\
Saebench Pythia & Standard ReLU & 16k & 73.4\% & 84.8   & 0.32\% \\
\bottomrule
\end{tabular}
\caption{Top 10 SAE configurations by overall perfect ranking accuracy. Sparsity is the mean number of active features per input (lower = more sparse). Dominance (\%) is $100 \times$ the \emph{median} dominance score among the SAE's \emph{10 most dominant features} (Top-10 median).}
\label{tab:top10_overall_mono}
\end{table}

\begin{table}[t]
\centering
\small
\begin{tabular}{lcccrrr}
\toprule
\textbf{Family} & \textbf{Arch} & \textbf{Width} & \textbf{Gap} & \textbf{Q1} & \textbf{Sparsity} & \textbf{Dominance (\%)} \\
\midrule
Saebench Pythia & BatchTopK   & 16k & -19.6\% & 50.5\% & 160.0 & 0.11\% \\
Saebench Pythia & TopK        & 65k & -19.2\% & 54.0\% & 40.0  & 0.19\% \\
Tinystories     & TopK        & 16k & -18.4\% & 55.5\% & 97.4  & 0.05\% \\
Tinystories     & Matryoshka  & 65k & -17.2\% & 56.8\% & 0.1   & 1.14\% \\
Saebench Pythia & Standard ReLU&16k & -16.4\% & 54.2\% & 455.0 & 0.06\% \\
Saebench Pythia & TopK        & 16k & -16.2\% & 46.2\% & 639.9 & 0.21\% \\
Saebench Pythia & Matryoshka  & 4k  & -16.1\% & 55.0\% & 160.0 & 0.18\% \\
Saebench Pythia & Matryoshka  & 65k & -16.1\% & 51.8\% & 160.0 & 0.13\% \\
Tinystories     & BatchTopK   & 65k & -15.7\% & 61.0\% & 0.0   & 1.05\% \\
Saebench Gemma  & TopK        & 65k & -15.2\% & 56.0\% & 160.0 & 0.04\% \\
\bottomrule
\end{tabular}
\caption{Top 10 SAE configurations with the largest Q1--Q2--4 gap (more negative = Q1 performs worse relative to Q2--4). Gap is Q1 accuracy minus the average accuracy over Q2--4, in percentage points. Dominance (\%) is $100\times$ the \emph{median} dominance score among the SAE's \emph{10 most dominant features} (Top-10 median).}
\label{tab:top10_q1_gap}
\end{table}

\begin{table}[t]
\centering
\small
\begin{tabular}{lcccrrr}
\toprule
\textbf{Family} & \textbf{Arch} & \textbf{Width} & \textbf{Q1} & \textbf{Overall} & \textbf{Sparsity} & \textbf{Dominance (\%)} \\
\midrule
JBloom OAI v5     & Standard ReLU & 32k  & 78.2\% & 72.0\% & 199.9 & 0.03\% \\
Saebench Pythia   & PAnneal       & 16k  & 76.8\% & 74.8\% & 74.1  & 0.11\% \\
Saebench Pythia   & PAnneal       & 4k   & 75.8\% & 78.8\% & 22.9  & 0.34\% \\
Saebench Pythia   & PAnneal       & 65k  & 75.7\% & 76.5\% & 38.2  & 0.19\% \\
JBloom OAI v5     & Standard ReLU & 32k  & 75.3\% & 70.7\% & 224.4 & 0.05\% \\
Tinystories       & Standard ReLU & 65k  & 75.2\% & 69.6\% & 673.5 & 0.01\% \\
JBloom OAI v5     & Standard ReLU & 128k & 74.5\% & 72.0\% & 193.3 & 0.04\% \\
Saebench Pythia   & Standard ReLU & 16k  & 74.0\% & 73.4\% & 84.8  & 0.32\% \\
JBloom OAI v5     & Standard ReLU & 128k & 73.3\% & 70.2\% & 203.0 & 0.08\% \\
Saebench Gemma    & PAnneal       & 65k  & 73.2\% & 67.9\% & 38.2  & 0.02\% \\
\bottomrule
\end{tabular}
\caption{Top 10 SAE configurations by Q1 perfect ranking accuracy. Dominance (\%) is $100\times$ the \emph{median} dominance score among the SAE's \emph{10 most dominant features} (Top-10 median).}
\label{tab:top10_q1_mono}
\end{table}

\begin{figure}[t]
\centering
\includegraphics[width=0.85\linewidth]{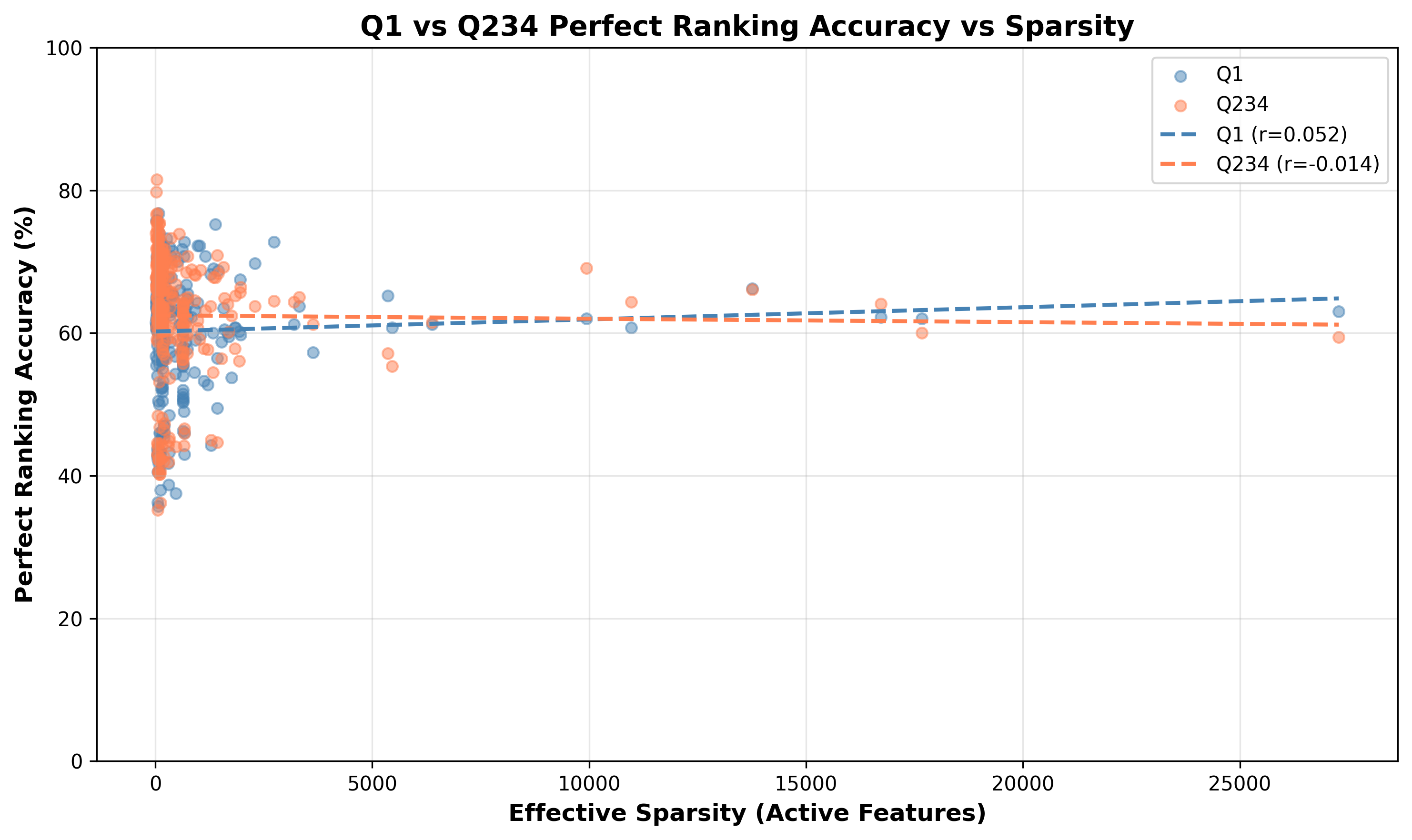}
\caption{Relationship between dominance, interpretability, and sparsity. The inverse dominance-interpretability relationship holds across different sparsity levels.}
\label{fig:dominance_interpretability_sparsity}
\end{figure}

\begin{figure}[t]
\centering
\includegraphics[width=0.85\linewidth]{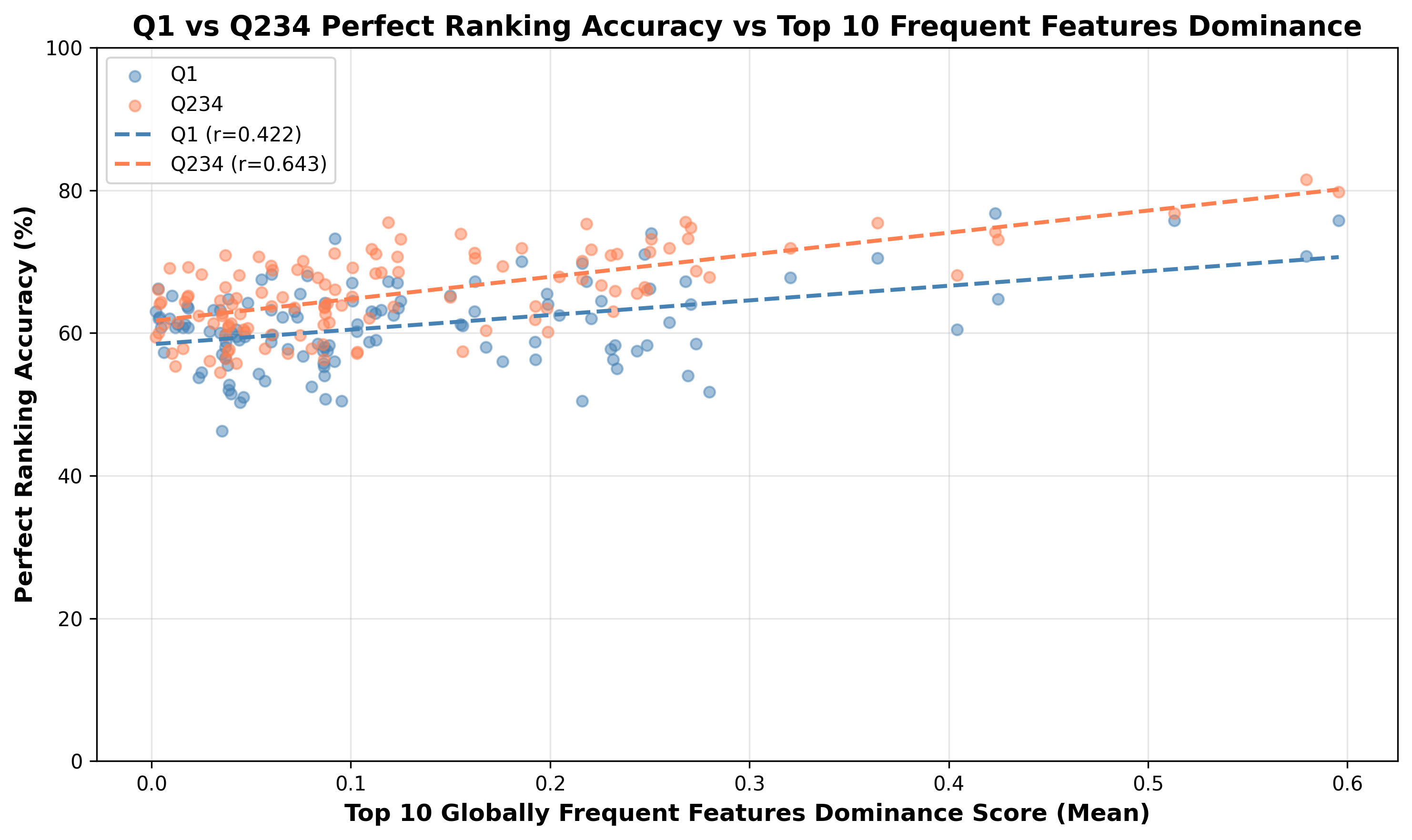}
\caption{Interpretability patterns for top 10\% most dominant features. Even among the highest-dominance features, the inverse relationship with interpretability persists.}
\label{fig:dominance_interpretability_top10}
\end{figure}

\paragraph{Additional Visualizations.}
Figure~\ref{fig:dominance_interpretability_sparsity} shows the relationship between dominance, interpretability, and sparsity across architectures, revealing that the dominance-interpretability trade-off persists regardless of sparsity level. Figure~\ref{fig:dominance_interpretability_top10} focuses on the top 10\% most dominant features, showing that even among the highest-dominance features, interpretability remains lower than less dominant features.

\subsubsection{Pairwise Accuracy Results}

The pairwise accuracy metric (measuring whether the model correctly identifies which of two examples has higher activation) shows similar patterns to perfect ranking accuracy. Across all SAE configurations, Q1 features achieve lower pairwise accuracy than Q2--Q4 features, with the gap varying by architecture and family. The sparsity-pairwise accuracy correlations (included in Table~\ref{tab:sparsity_correlation_summary}) show similar weak, inconsistent patterns as the perfect ranking accuracy correlations, with only 4.5\% of overall correlations and 6.9\% of Q1 correlations reaching statistical significance. However, pairwise accuracy shows a stronger positive correlation pattern for Q1 features (59.8\% positive) compared to perfect ranking accuracy, suggesting that sparsity may have a slightly more beneficial effect on Q1 pairwise accuracy.

\subsection{Comparison: Completeness/Sparsity/Dead Feature vs Monosemanticity}
\label{appendix:comparison}

\subsubsection{Three-Metric Composite as Proxy for Monosemanticity}

Practitioners often use a three-metric composite (dead feature proportion + sparsity + MSE) as an in-training proxy for monosemanticity, since these metrics are computationally tractable compared to full DAME evaluations. This practice is grounded in the superposition hypothesis \citep{elhage2022toy}, though prior work has questioned whether optimizing these proxies yields interpretable features \citep{karvonen2025saebench, smith2024negative, gao2024scaling}. To our knowledge, no prior work has quantitatively validated this relationship.

We compute rank correlations between the three-metric composite and DAME monosemanticity scores across all configurations (defined by SAE family, architecture, and width) that appear in both the completeness and DAME evaluations. We find a \textbf{strong positive correlation} with Q1 monosemanticity (Spearman $r_s = 0.71$, $p < 0.001$) but only a \textbf{weak correlation} with overall monosemanticity ($r_s = 0.36$, $p = 0.012$). Despite the strong Q1 correlation, top-10 overlap is near-zero (1--2/10).

\paragraph{Implications.} The three-metric composite is a reasonable proxy for Q1 monosemanticity, useful when practitioners care primarily about a feature's strongest activations. For full activation-range monosemanticity or extreme optimization, direct evaluation remains necessary.

\subsection{C1: Non-Redundancy}
\label{appendix:c1_redundancy}

\begin{table}[t]
\centering
\begin{tabular}{lrrrr}
\toprule
Architecture & Q1 & Q2 & Q3 & Q4 \\
\midrule
Standard ReLU & 3.81 & 2.58 & 2.59 & 2.81 \\
TopK & 1.79 & 0.29 & 0.12 & 0.15 \\
Gated & 1.62 & 0.88 & 0.69 & 0.61 \\
JumpReLU & 1.10 & 0.17 & 0.18 & 0.44 \\
BatchTopK & 0.94 & 0.10 & 0.07 & 0.05 \\
PAnneal & 0.38 & 0.33 & 0.23 & 0.09 \\
Matryoshka & 0.19 & 0.04 & 0.00 & 0.01 \\
\bottomrule
\end{tabular}

\caption{Percentage of features with at least one redundant pair by architecture and quantile (decoder cosine similarity $\geq 0.7$).}
\label{tab:features_with_redundancy}
\end{table}

\begin{table}[t]
\centering
\begin{tabular}{lrrr}
\toprule
Architecture & Q1 (\%) & ANY (\%) & Pairs Within \\
\midrule
BatchTopK & 40.33 & 65.67 & 2.6 \\
TopK & 35.44 & 68.78 & 5.0 \\
JumpReLU & 28.61 & 81.94 & 2.5 \\
Standard ReLU & 26.11 & 93.33 & 4.7 \\
Matryoshka & 25.00 & 33.61 & 2.1 \\
PAnneal & 21.94 & 63.61 & 3.1 \\
Gated & 20.83 & 53.61 & 1.6 \\
\bottomrule
\end{tabular}

\caption{Top 10 features involvement in redundancy by architecture. Q1 (\%): percentage of top 10 features involved in Q1 redundant pairs. ANY (\%): percentage involved in any redundant pair. Pairs Within: average redundant pairs formed within top 10.}
\label{tab:top10_redundancy}
\end{table}

\paragraph{Complete versus within-quartile redundancy graphs.}
With full-dictionary dominance quartiles and the symmetric active-only filter of Section~\ref{subsec:redundancy}, median Q1 endpoint enrichment is $1.64\times$ on the complete graph (95\% SAE-bootstrap CI 1.51--1.74; 71.5\% of SAEs above $1\times$), compared with $1.87\times$ when only within-quartile edges are retained (1.74--1.99; 72.8\%). Both use the same active-feature-share normalization and symmetric filtering. The historical pair-share analyses below instead filter only within-Q4 pairs.

\paragraph{Within-quartile redundant-pair shares.}
Restricting to redundant pairs whose two features fall in the same dominance quartile, Q1 hosts a share of
redundant pairs above the 25\% uniform baseline in every architecture: the
per-SAE-averaged Q1 share ranges 30--61\% unfiltered. After excluding
within-Q4 pairs that involve a dead feature (many of these dead features share
decoder directions that remain close to their random initialization,
producing apparent rather than semantic redundancy), the per-SAE-averaged Q1 share rises
to 52--86\% (Table~\ref{tab:redundancy_combined}, left). The concentration is
statistically significant at the per-SAE level (Wilcoxon signed-rank test:
overall $p = 2.2 \times 10^{-57}$; 79.7\% of SAEs have Q1 hosting the most
redundant pairs; per-architecture results in
Table~\ref{tab:redundancy_combined}, right). This statistic counts within-quartile pairs only and filters dead-feature
pairs in Q4 only; the endpoint-enrichment analysis in Section~\ref{subsec:redundancy} uses the complete graph and a symmetric filter.

\begin{table}[h]
\centering
\caption{Redundancy results across SAE architectures (per-SAE averaged, Q4 dead-feature pairs filtered). \textbf{Left:} per-architecture share of redundant pairs by dominance quantile, at decoder cosine similarity $\geq 0.7$. \textbf{Right:} per-SAE Wilcoxon signed-rank test of Q1 vs.\ Q2--Q4 redundant-pair counts; positive $d$ indicates Q1 has more redundant pairs.}
\label{tab:redundancy_combined}
\begin{minipage}[t]{0.49\linewidth}
  \centering\footnotesize
  \begin{tabular}{lrrrr}
\toprule
Architecture & Q1 (\%) & Q2 (\%) & Q3 (\%) & Q4 (\%) \\
\midrule
Standard ReLU & 63.74 & 8.33 & 10.50 & 17.43 \\
PAnneal & 86.35 & 7.71 & 4.23 & 1.72 \\
Gated & 59.02 & 16.37 & 10.90 & 13.71 \\
TopK & 77.22 & 10.29 & 3.11 & 9.38 \\
BatchTopK & 66.42 & 18.56 & 8.21 & 6.80 \\
Matryoshka & 83.65 & 14.18 & 1.82 & 0.35 \\
JumpReLU & 51.58 & 17.77 & 10.47 & 20.18 \\
\bottomrule
\end{tabular}

\end{minipage}\hfill
\begin{minipage}[t]{0.49\linewidth}
  \centering\footnotesize
  \begin{tabular}{lrr}
    \toprule
    Architecture & $p$ & Cohen's $d$ \\
    \midrule
    Standard ReLU & $3.0{\times}10^{-29}$ & $+0.11$ \\
    TopK       & $1.7{\times}10^{-11}$ & $+0.20$ \\
    Gated      & $7.3{\times}10^{-6}$  & $+0.21$ \\
    PAnneal    & $7.9{\times}10^{-7}$  & $+0.69$ \\
    BatchTopK  & $2.0{\times}10^{-5}$  & $+0.62$ \\
    Matryoshka & $3.0{\times}10^{-7}$  & $+0.59$ \\
    JumpReLU   & $1.9{\times}10^{-3}$  & $+0.22$ \\
    \midrule
    \textbf{Overall} & $\mathbf{2.2{\times}10^{-57}}$ & --- \\
    \bottomrule
  \end{tabular}
\end{minipage}
\end{table}

\paragraph{Dead-feature pair inflation and filter rates.}
The within-quartile shares above are reported after filtering
out pairs that involve a dead feature (firing count $= 0$ in the redundancy
evaluation sample). \textbf{The filter only removes within-Q4 pairs.} We filter
for two reasons. First, a feature that never fires contributes nothing to
model computation, so a high decoder-cosine match to such a feature does not
represent semantic redundancy. Second, dead features tend to retain decoder
directions close to their random initialization, and pairs of features with
frozen-init decoder directions can spuriously match each other above the
cosine threshold---an artifact of training dynamics, not of true semantic
duplication. The unfiltered quantile share for some architectures is therefore
dominated by these dead-feature pair counts in Q4, which masks the Q1 effect. Per-architecture pair filter rates and the unfiltered
quantile shares are reported in
Table~\ref{tab:redundancy_by_architecture_unfiltered}: filter rates range from
$<1\%$ (Gated) to over $50\%$ (JumpReLU). The unfiltered shares for BatchTopK,
Matryoshka, and JumpReLU place a substantial fraction of pairs in Q4---once
dead-feature pairs are removed, the Q1 concentration becomes apparent in every
architecture (compare with Table~\ref{tab:redundancy_combined}, left).
Note in particular Standard ReLU (unfiltered Q1 $\approx$ Q4 at 38\%) and
JumpReLU (unfiltered Q4 $>$ Q1): these two architectures have the highest
dead-pair filter rates (38.85\% and 51.73\%, respectively), so their unfiltered
Q4 column is heavily inflated by dead-feature pairs; after filtering, Q1
dominates for both.
\begin{table}[t]
\centering
\footnotesize
\caption{\textbf{Unfiltered} version of Table~\ref{tab:redundancy_combined} (left): per-architecture share of redundant pairs by dominance quantile, per-SAE averaged, with pairs involving a dead feature \emph{not} excluded. Last column is the per-SAE-averaged percentage of pairs that would be removed by dead-feature filtering. Without filtering, dead-feature pairs inflate Q4 and mask the Q1 concentration for BatchTopK / Matryoshka / JumpReLU; after filtering (Table~\ref{tab:redundancy_combined}), all 7 architectures are Q1-dominant.}
\label{tab:redundancy_by_architecture_unfiltered}
\begin{tabular}{lrrrrr}
\toprule
Architecture & Q1 (\%) & Q2 (\%) & Q3 (\%) & Q4 (\%) & Pair filter rate (\%) \\
\midrule
Standard ReLU & 38.48 & 6.92 & 16.04 & 38.56 & 38.85 \\
PAnneal & 53.98 & 15.88 & 17.91 & 12.23 & 35.89 \\
Gated & 58.99 & 16.35 & 10.90 & 13.77 & 0.08 \\
TopK & 61.43 & 11.40 & 8.01 & 19.16 & 21.54 \\
BatchTopK & 41.62 & 18.71 & 11.43 & 28.25 & 36.94 \\
Matryoshka & 49.31 & 26.29 & 7.62 & 16.78 & 31.11 \\
JumpReLU & 30.42 & 15.74 & 19.45 & 34.38 & 51.73 \\
\bottomrule
\end{tabular}

\end{table}

\begin{figure}[t]
  \centering
  \includegraphics[width=0.85\linewidth]{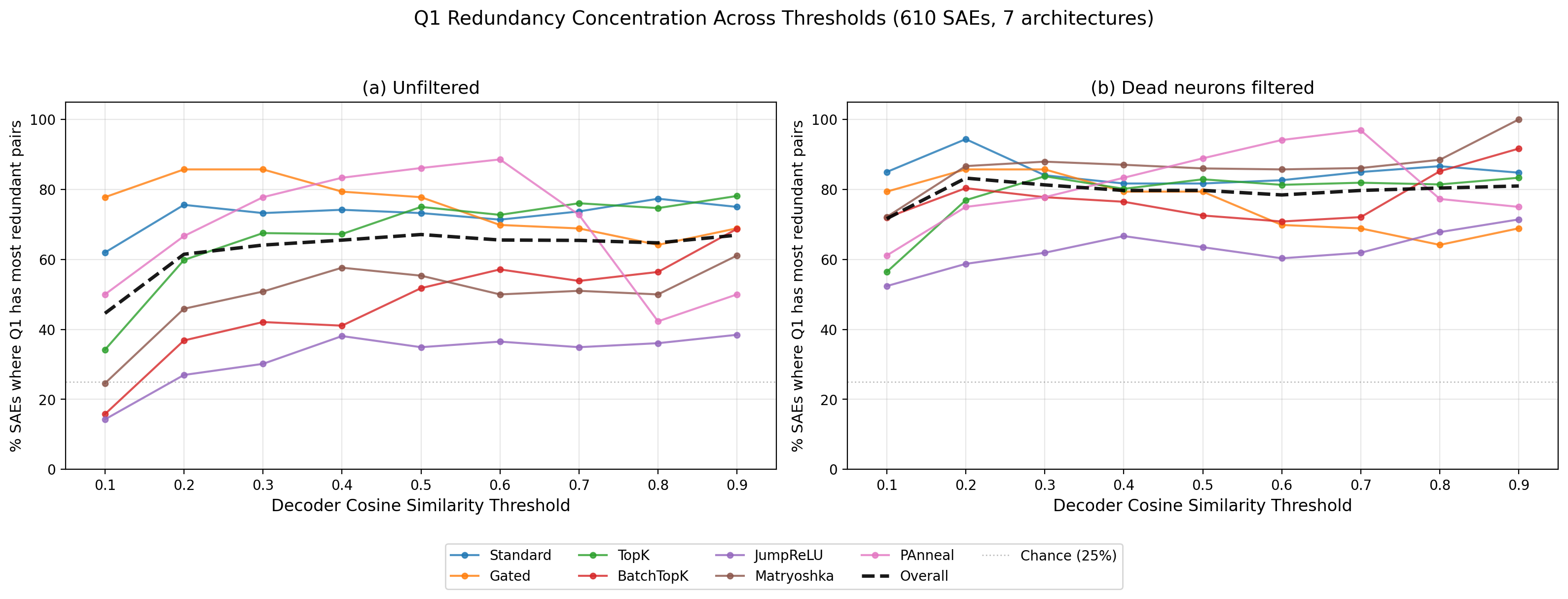}
  \caption{Q1 redundancy concentration across decoder cosine thresholds $\theta \in [0.1, 0.9]$, with and without dead-feature pair filtering. Filtered: 72--83\% of SAEs Q1-dominant at every threshold. Unfiltered: 45--67\%.}
  \label{fig:redundancy_threshold_sweep}
\end{figure}

\paragraph{Threshold robustness.}
The Q1 redundancy concentration claim depends on a choice of decoder cosine
threshold ($\theta = 0.7$, following \citet{karvonen2025saebench}).
To verify the claim does not hinge on this specific cutoff, we sweep $\theta$
across $\{0.1, 0.2, \ldots, 0.9\}$ and recompute the per-SAE Q1 dominance
indicator (whether Q1 hosts the most redundant pairs for that SAE).
Figure~\ref{fig:redundancy_threshold_sweep} shows that, after dead-feature
filtering, 72--83\% of SAEs are Q1-dominant at every threshold; without
filtering 45--67\%. The claim is therefore not threshold-sensitive.

\paragraph{Matched-sparsity control.}
A potential concern is that the Q1 effect is confounded by sparsity:
architectures with lower median $L_0$ have stronger dominance imbalances, so
the Q1 finding could reflect sparsity differences rather than dominance per se.
To rule this out, we bin all evaluated SAEs by effective $L_0$ into ranges
$\{[0,100), [100,200), [200,300), [300,500), [500,1000), [1000,5000)\}$ and
recompute the Q1 monosemanticity deficit and Q1 redundancy concentration
within each bin. Both effects persist at all sparsity levels: the per-bin
fraction of SAEs showing significant Q1 deficit ranges 50--68\%, and the
per-bin median Q1 redundancy fraction ranges 0.93--1.00.

\begin{table}[t]
\centering
\begin{tabular}{lrrr}
\toprule
Architecture & Q1 (\%) & ANY (\%) & Pairs Within \\
\midrule
BatchTopK & 13.87 & 18.50 & 10.7 \\
TopK & 14.52 & 20.93 & 17.7 \\
JumpReLU & 14.53 & 29.22 & 19.3 \\
Standard ReLU & 20.53 & 71.36 & 296.3 \\
Matryoshka & 2.72 & 3.58 & 2.3 \\
PAnneal & 5.33 & 16.89 & 10.6 \\
Gated & 10.86 & 25.56 & 9.9 \\
\bottomrule
\end{tabular}

\caption{Top 100 features involvement in redundancy by architecture. Q1 (\%): percentage of top 100 features involved in Q1 redundant pairs. ANY (\%): percentage involved in any redundant pair. Pairs Within: average redundant pairs formed within top 100.}
\label{tab:top100_redundancy}
\end{table}

\begin{table}[t]
\centering
\begin{tabular}{lrr}
\toprule
Quantile & Mean & Median \\
\midrule
Q1 & 0.0561 & 0.0239 \\
Q2 & 0.1090 & 0.0313 \\
Q3 & 0.1249 & 0.0385 \\
Q4 & 0.1347 & 0.0527 \\
\bottomrule
\end{tabular}
\caption{Encoder similarity (all pairs) by quantile across all SAEs. Q1 has the lowest encoder similarity, indicating more encoder space partitioning.}
\label{tab:encoder_similarity_by_quantile}
\end{table}

\begin{figure}[t]
\centering
\includegraphics[width=0.85\linewidth]{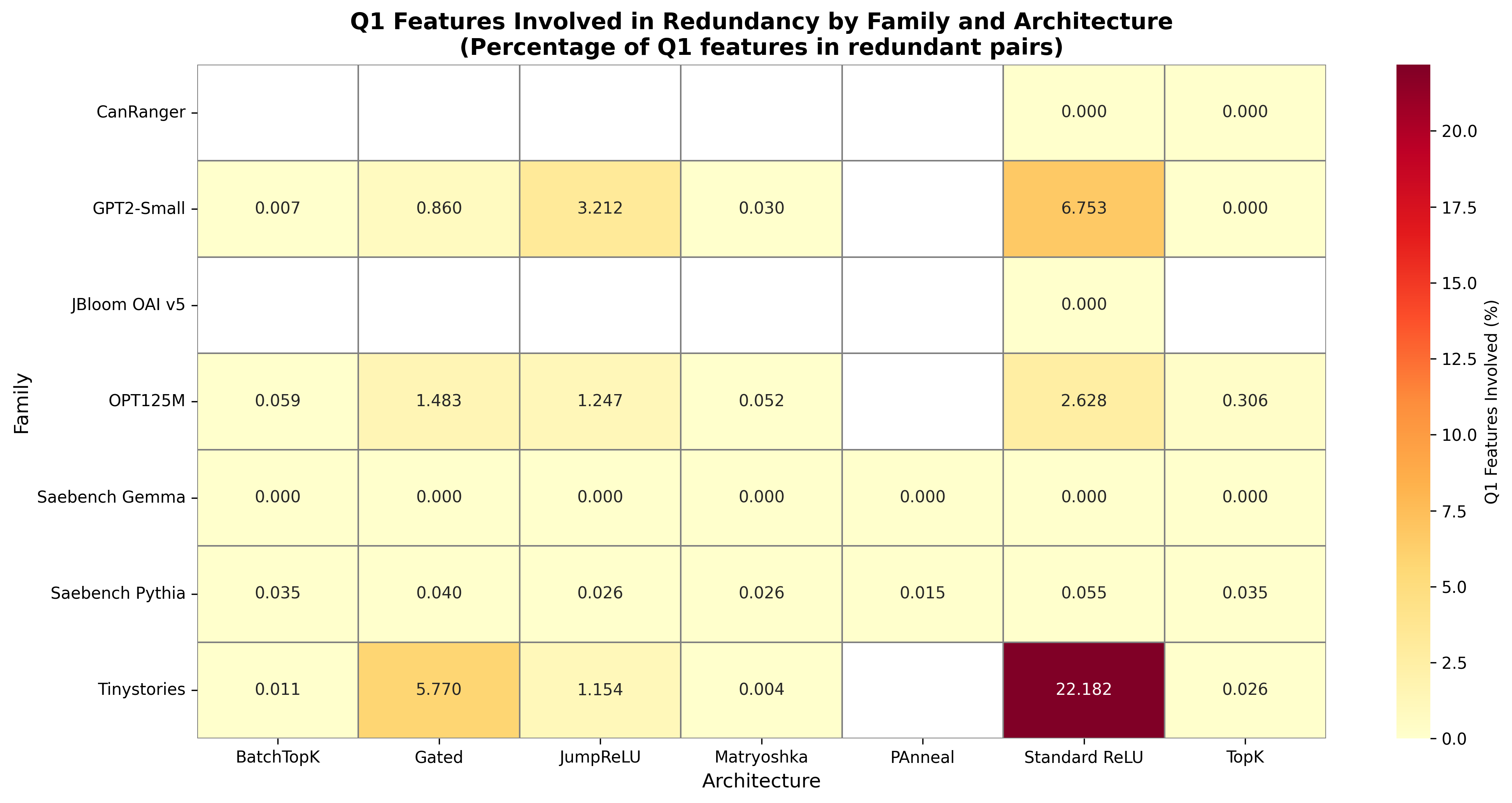}
\caption{Q1 features involved in redundancy (percentage) by family and architecture. Standard ReLU shows the highest Q1 redundancy, particularly in TinyStories (up to 28.6\%) and GPT2-Small (up to 10.3\%), while JumpReLU shows minimal redundancy across all families. Smaller models show more severe redundancy than larger models.}
\label{fig:family_architecture_q1_redundancy}
\end{figure}

\paragraph{Family-Architecture Redundancy Patterns.}
The concentration of redundancy in Q1 features varies substantially across family-architecture combinations (Figure~\ref{fig:family_architecture_q1_redundancy}). \textbf{TinyStories} with Standard ReLU shows the highest Q1 redundancy, with up to 28.6\% of Q1 features involved in redundant pairs at 65k width, while \textbf{JumpReLU} architectures show minimal Q1 redundancy across families. \textbf{GPT2-Small} with Standard ReLU also shows high Q1 redundancy (up to 10.3\% at 65k width). This architecture-dependent variation suggests that Q1 redundancy is not universal but depends on both the base model and SAE architecture, with smaller models (TinyStories, GPT2-Small) showing more severe redundancy than larger models (SAEBench Pythia, SAEBench Gemma).

\begin{figure}[t]
\centering
\includegraphics[width=0.85\linewidth]{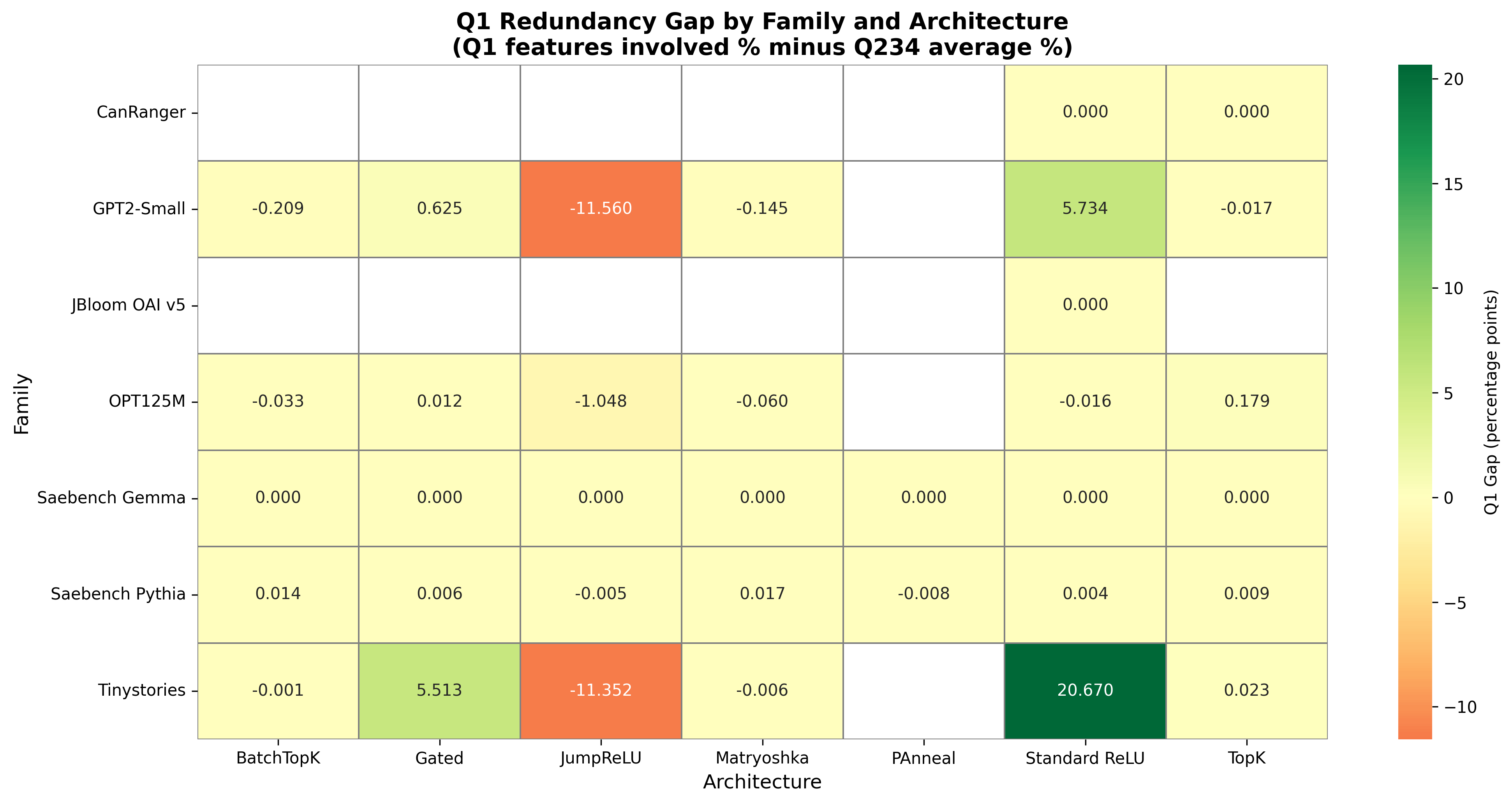}
\caption{Q1 redundancy gap (Q1 features involved \% minus Q2--Q4 average \%) by family and architecture. Positive values indicate Q1 has more redundancy than Q2--Q4. Standard ReLU shows the largest positive gaps, indicating stronger Q1 redundancy concentration.}
\label{fig:family_architecture_q1_redundancy_gap}
\end{figure}

\paragraph{Q1 Redundancy Gap Analysis.}
The Q1 redundancy gap---measuring how much more redundant Q1 features are compared to Q2--Q4---reveals clear architecture-dependent patterns (Figure~\ref{fig:family_architecture_q1_redundancy_gap}). \textbf{Standard ReLU} architectures show the largest positive gaps (up to +27.3 percentage points in TinyStories), indicating that Q1 features are substantially more redundant than Q2--Q4. In contrast, \textbf{JumpReLU} shows negative gaps (down to -18.6 percentage points), suggesting that Q1 features are actually less redundant than Q2--Q4 in these architectures. The gap is most pronounced in smaller models (TinyStories, GPT2-Small) compared to larger models.

\begin{figure}[t]
\centering
\includegraphics[width=0.85\linewidth]{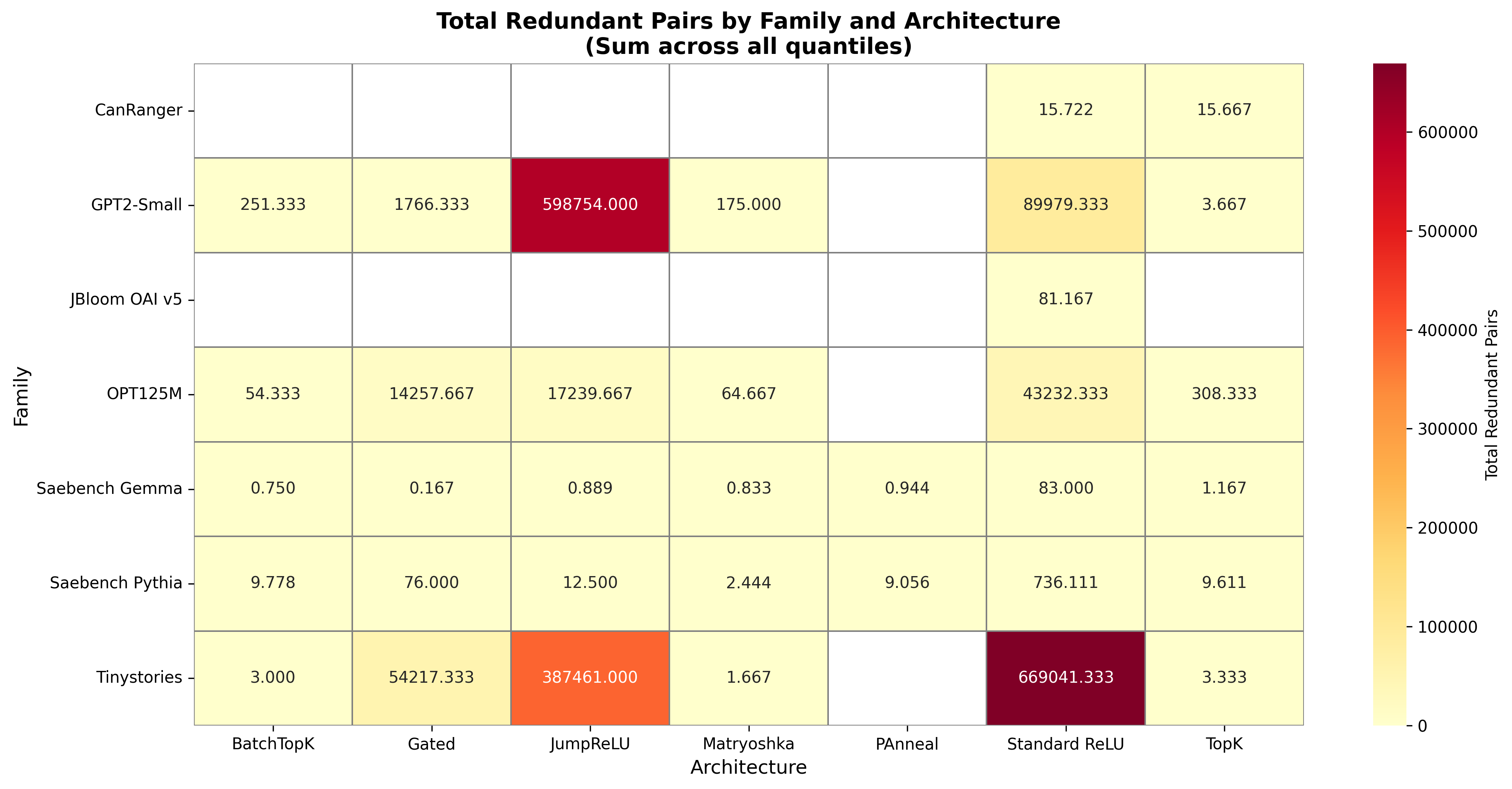}
\caption{Total redundant pairs (sum across all quantiles) by family and architecture. Standard ReLU shows the highest total redundancy, particularly in TinyStories (over 1.3 million pairs at 65k) and GPT2-Small. JumpReLU shows high total redundancy but with different quantile distribution patterns.}
\label{fig:family_architecture_total_redundant_pairs}
\end{figure}

\paragraph{Total Redundancy by Architecture.}
Figure~\ref{fig:family_architecture_total_redundant_pairs} shows the total number of redundant pairs across all quantiles by family and architecture. \textbf{TinyStories} with Standard ReLU at 65k width shows the highest total redundancy (over 1.3 million redundant pairs), followed by \textbf{GPT2-Small} with JumpReLU at 16k (over 1 million pairs, though with low Q1 concentration). \textbf{JumpReLU} shows high total redundancy in some configurations but with minimal Q1 concentration, suggesting redundancy is distributed across quantiles rather than concentrated in Q1. Standard ReLU architectures thus show the most redundancy overall, particularly in smaller models, while JumpReLU shows different redundancy patterns.

\begin{figure}[t]
\centering
\includegraphics[width=0.85\linewidth]{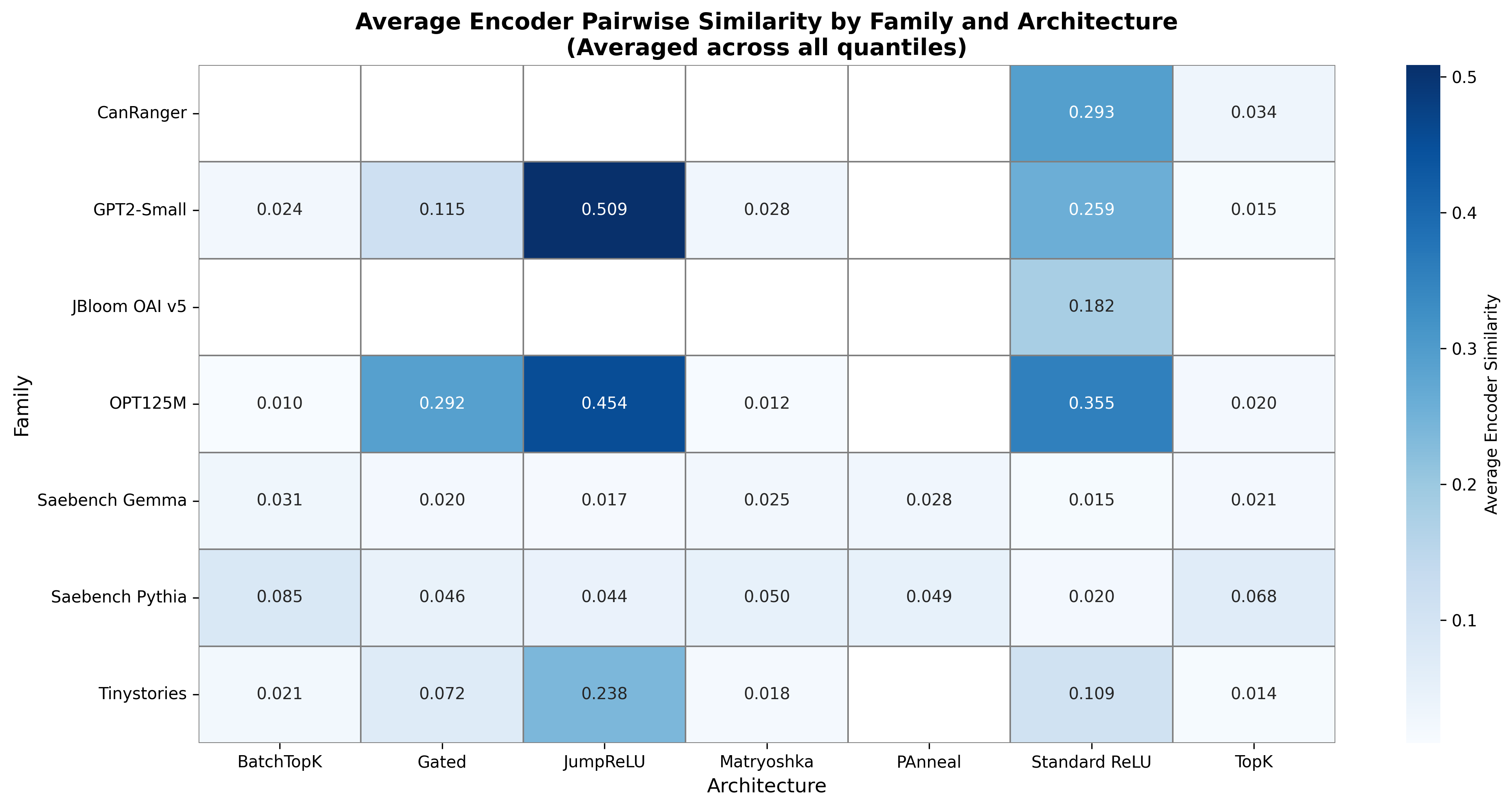}
\caption{Average encoder pairwise similarity (across all quantiles) by family and architecture. Lower values indicate more encoder space partitioning.}
\label{fig:family_architecture_encoder_similarity}
\end{figure}

\begin{figure}[t]
\centering
\includegraphics[width=0.85\linewidth]{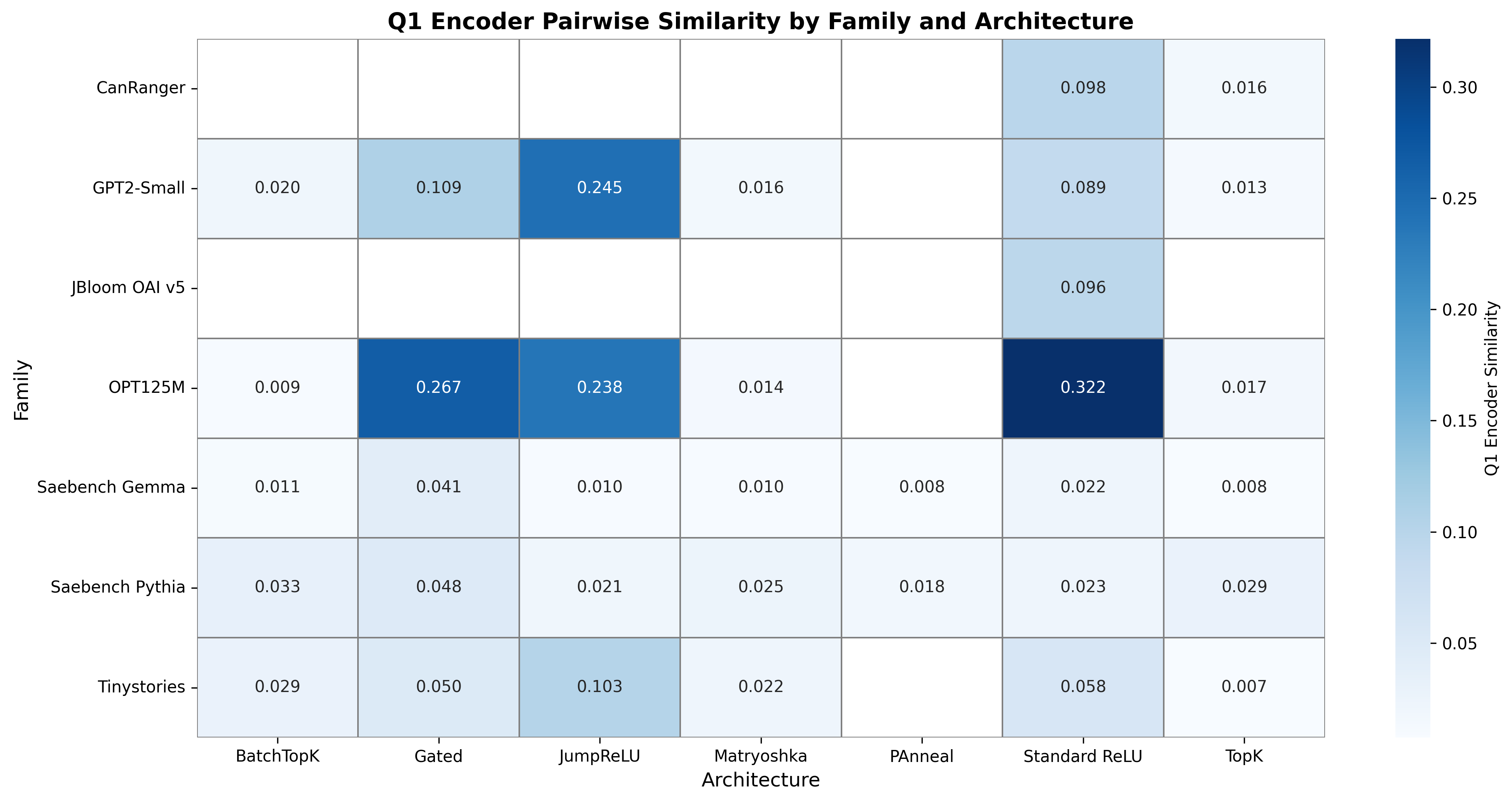}
\caption{Q1 encoder pairwise similarity by family and architecture. Q1 consistently shows lower encoder similarity than Q2--Q4, indicating greater encoder space partitioning for dominant features.}
\label{fig:family_architecture_q1_encoder_similarity}
\end{figure}

\paragraph{Encoder Similarity Patterns.}
The encoder similarity heatmaps (Figures~\ref{fig:family_architecture_encoder_similarity} and~\ref{fig:family_architecture_q1_encoder_similarity}) reveal that Q1 features consistently show lower encoder similarity than Q2--Q4, confirming the encoder space partitioning pattern. However, the magnitude of this difference varies by architecture: Standard ReLU shows the largest Q1 vs.\ Q2--Q4 encoder similarity gap, while JumpReLU shows more uniform encoder similarity across quantiles.

\paragraph{Top Redundant SAE Configurations.}
Tables~\ref{tab:top10_total_redundant_pairs},~\ref{tab:top10_q1_redundant_pairs}, and~\ref{tab:top10_q1_features_pct} identify the top 10 SAE configurations across three redundancy metrics: total redundant pairs, Q1 redundant pairs, and Q1 features involved percentage. These tables reveal several important patterns.

\textbf{Top 10 Total Redundant Pairs:} \textbf{TinyStories} with Standard ReLU at 65k width shows the highest total redundancy (1,340,724 pairs), with Q1 contributing 1,328,227 pairs (99.1\%). \textbf{GPT2-Small} with JumpReLU also appears in the top 10, but with only 0.5\% of redundancy in Q1, indicating a different redundancy pattern. Standard ReLU dominates the most redundant configurations, particularly in smaller models.

\textbf{Top 10 Q1 Redundant Pairs:} \textbf{TinyStories} Standard ReLU configurations dominate, with the top configuration (65k width) having 1,328,227 Q1 redundant pairs, representing 28.6\% of Q1 features involved. \textbf{GPT2-Small} Standard ReLU at 65k shows 188,940 Q1 redundant pairs (10.3\% of Q1 features).

\textbf{Top 10 Q1 Features Involved Percentage:} \textbf{TinyStories} Standard ReLU shows the highest percentages (28.6\% at 65k, 21.2\% at 4k, 16.7\% at 16k), followed by \textbf{GPT2-Small} Standard ReLU (up to 10.3\%). \textbf{Gated} architecture also appears in the top 10 for TinyStories (8.98\% at 65k). This metric reveals that smaller models (TinyStories, GPT2-Small) show much higher Q1 redundancy percentages than larger models (SAEBench Pythia, SAEBench Gemma), suggesting that model size is a key factor in Q1 redundancy.

\begin{table}[t]
\centering
\begin{tabular}{lrrrr}
\toprule
Family & Arch & Width & Total Redundant Pairs & Q1 Redundant Pairs \\
\midrule
Tinystories & Standard ReLU & 65k & 1340724 & 1328227 \\
GPT2-Small & JumpReLU & 16k & 1020088 & 5573 \\
Tinystories & JumpReLU & 16k & 665199 & 7438 \\
Tinystories & Standard ReLU & 16k & 561217 & 545472 \\
GPT2-Small & JumpReLU & 4k & 430005 & 18665 \\
GPT2-Small & JumpReLU & 65k & 346169 & 5714 \\
Tinystories & JumpReLU & 4k & 287253 & 1998 \\
Tinystories & JumpReLU & 65k & 209931 & 11 \\
GPT2-Small & Standard ReLU & 65k & 195754 & 188940 \\
Tinystories & Gated & 65k & 126588 & 125887 \\
\bottomrule
\end{tabular}

\caption{Top 10 SAE configurations by total redundant pairs (sum across all quantiles). TinyStories Standard ReLU at 65k shows the highest total redundancy (1.3 million pairs), with 99.1\% concentrated in Q1. GPT2-Small JumpReLU also appears but with different quantile distribution.}
\label{tab:top10_total_redundant_pairs}
\end{table}

\begin{table}[t]
\centering
\begin{tabular}{lrrrr}
\toprule
Family & Arch & Width & Q1 Redundant Pairs & Q1 Features Involved (\%) \\
\midrule
Tinystories & Standard ReLU & 65k & 1328227 & 28.63 \\
Tinystories & Standard ReLU & 16k & 545472 & 16.72 \\
GPT2-Small & Standard ReLU & 65k & 188940 & 10.28 \\
Tinystories & Gated & 65k & 125887 & 8.98 \\
Tinystories & Standard ReLU & 4k & 101376 & 21.19 \\
GPT2-Small & Standard ReLU & 16k & 54739 & 5.03 \\
Tinystories & Gated & 16k & 30837 & 4.10 \\
OPT125M & Standard ReLU & 65k & 27057 & 4.43 \\
GPT2-Small & JumpReLU & 4k & 18665 & 6.32 \\
OPT125M & Gated & 65k & 9938 & 2.60 \\
\bottomrule
\end{tabular}

\caption{Top 10 SAE configurations by Q1 redundant pairs. TinyStories Standard ReLU dominates, with up to 1.3 million Q1 redundant pairs (28.6\% of Q1 features) at 65k width. GPT2-Small Standard ReLU also shows high Q1 redundancy (up to 188,940 pairs, 10.3\% of Q1 features).}
\label{tab:top10_q1_redundant_pairs}
\end{table}

\begin{table}[t]
\centering
\begin{tabular}{lrrrr}
\toprule
Family & Arch & Width & Q1 Features Involved (\%) & Q1 Redundant Pairs \\
\midrule
Tinystories & Standard ReLU & 65k & 28.63 & 1328227 \\
Tinystories & Standard ReLU & 4k & 21.19 & 101376 \\
Tinystories & Standard ReLU & 16k & 16.72 & 545472 \\
GPT2-Small & Standard ReLU & 65k & 10.28 & 188940 \\
Tinystories & Gated & 65k & 8.98 & 125887 \\
GPT2-Small & JumpReLU & 4k & 6.32 & 18665 \\
GPT2-Small & Standard ReLU & 16k & 5.03 & 54739 \\
GPT2-Small & Standard ReLU & 4k & 4.95 & 9231 \\
OPT125M & Standard ReLU & 65k & 4.43 & 27057 \\
Tinystories & Gated & 4k & 4.23 & 4815 \\
\bottomrule
\end{tabular}

\caption{Top 10 SAE configurations by Q1 features involved percentage. TinyStories Standard ReLU shows the highest percentages (28.6\% at 65k, 21.2\% at 4k, 16.7\% at 16k), followed by GPT2-Small Standard ReLU (up to 10.3\%). Smaller models show dramatically higher Q1 redundancy than larger models.}
\label{tab:top10_q1_features_pct}
\end{table}

\paragraph{Q1 Redundancy Rate Analysis.}
Table~\ref{tab:top10_q1_redundancy_rate} shows the top 10 configurations by Q1 redundancy rate (redundant pairs per Q1 feature). This metric normalizes for the number of Q1 features, revealing which architectures produce the most redundant pairs per feature. \textbf{TinyStories} Standard ReLU configurations dominate, with rates up to 147.6 redundant pairs per Q1 feature at 65k width, far exceeding rates in larger models. \textbf{GPT2-Small} Standard ReLU also shows high rates (up to 21.0 per feature at 65k). Redundancy is thus particularly concentrated in dominant features for Standard ReLU architectures, with smaller models showing dramatically higher rates.

\begin{table}[t]
\centering
\begin{tabular}{lrrrr}
\toprule
Family & Arch & Width & Q1 Redundancy Rate & Q1 Redundant Pairs \\
\midrule
Tinystories & Standard ReLU & 65k & 147.581 & 1328227 \\
Tinystories & Standard ReLU & 16k & 60.608 & 545472 \\
Tinystories & Standard ReLU & 4k & 33.000 & 101376 \\
GPT2-Small & Standard ReLU & 65k & 20.993 & 188940 \\
Tinystories & Gated & 65k & 13.987 & 125887 \\
GPT2-Small & Standard ReLU & 16k & 6.082 & 54739 \\
GPT2-Small & JumpReLU & 4k & 6.076 & 18665 \\
Tinystories & Gated & 16k & 3.426 & 30837 \\
OPT125M & Standard ReLU & 65k & 3.006 & 27057 \\
GPT2-Small & Standard ReLU & 4k & 3.005 & 9231 \\
\bottomrule
\end{tabular}

\caption{Top 10 SAE configurations by Q1 redundancy rate (redundant pairs per Q1 feature). TinyStories Standard ReLU shows the highest rates (up to 147.6 pairs per feature at 65k), far exceeding larger models. GPT2-Small Standard ReLU also shows high rates (up to 21.0 per feature).}
\label{tab:top10_q1_redundancy_rate}
\end{table}

\paragraph{Q1 Redundancy Gap.}
Table~\ref{tab:top10_q1_redundancy_gap} shows the top 10 configurations by Q1 redundancy gap (Q1 redundant pairs minus Q2--Q4 average). This metric identifies configurations where Q1 redundancy is most concentrated relative to Q2--Q4. \textbf{TinyStories} Standard ReLU configurations show the largest gaps, with the top configuration (65k width) showing a gap of 1,324,000 pairs, indicating that Q1 has substantially more redundancy than Q2--Q4. \textbf{GPT2-Small} Standard ReLU also shows large gaps (up to 187,000 pairs). The gap is most pronounced in smaller models.

\begin{table}[t]
\centering
\begin{tabular}{lrrrrr}
\toprule
Family & Arch & Width & Q1 Gap (pairs) & Q1 Redundant Pairs & Total Redundant Pairs \\
\midrule
Tinystories & Standard ReLU & 65k & 1324061.3 & 1328227 & 1340724 \\
Tinystories & Standard ReLU & 16k & 540223.7 & 545472 & 561217 \\
GPT2-Small & Standard ReLU & 65k & 186668.7 & 188940 & 195754 \\
Tinystories & Gated & 65k & 125653.3 & 125887 & 126588 \\
Tinystories & Standard ReLU & 4k & 100107.0 & 101376 & 105183 \\
GPT2-Small & Standard ReLU & 16k & 51776.0 & 54739 & 63628 \\
Tinystories & Gated & 16k & 30716.7 & 30837 & 31198 \\
OPT125M & Standard ReLU & 65k & 22578.7 & 27057 & 40492 \\
GPT2-Small & Standard ReLU & 4k & 8789.3 & 9231 & 10556 \\
OPT125M & Gated & 65k & 7410.3 & 9938 & 17521 \\
\bottomrule
\end{tabular}

\caption{Top 10 SAE configurations by Q1 redundancy gap (Q1 redundant pairs minus Q2--Q4 average). TinyStories Standard ReLU shows the largest gaps (up to 1.3 million pairs at 65k), indicating extreme Q1 redundancy concentration. GPT2-Small Standard ReLU also shows large gaps (up to 186,669 pairs).}
\label{tab:top10_q1_redundancy_gap}
\end{table}

\paragraph{Architecture-Specific Redundancy Patterns.}
The comprehensive redundancy analysis across over 600 SAEs from 7 families reveals clear architecture-dependent and model-size-dependent patterns. \textbf{Standard ReLU} shows the highest overall redundancy and the strongest Q1 concentration. This is most severe in smaller models: \textbf{TinyStories} Standard ReLU shows up to 28.6\% of Q1 features involved in redundancy, while \textbf{GPT2-Small} shows up to 10.3\%. In contrast, larger models (\textbf{SAEBench Pythia}, \textbf{SAEBench Gemma}) show much lower Q1 redundancy percentages despite using the same architecture. \textbf{JumpReLU} shows high total redundancy in some configurations but with negative Q1 gaps, indicating redundancy is distributed across quantiles rather than concentrated in Q1. \textbf{Gated} shows intermediate patterns, with moderate redundancy that is less concentrated in Q1 than Standard ReLU. These patterns suggest that both architectural choices and model size affect Q1 redundancy, with smaller models showing more severe redundancy in Standard ReLU architectures.

\section{SAE Feature Activation Behavior and Dominance Imbalance}
\label{sec:sae_feature_firing_behavior}

Section~\ref{sec:local-to-global} uses mean dominance $\bar{\rho}_i$ to stratify features by their relative contribution to individual inputs. Here we define the complementary measures of firing frequency $\nu_i$ and total activation magnitude $M_i$, explain how they differ from mean dominance, and document these imbalance patterns across over 600 SAEs spanning multiple model families, architectures, sparsity levels, and dictionary sizes.

\subsection{Firing Frequency Imbalance ($\nu_i$)}

Firing frequency is $\nu_i = \mathbb{E}_x\bigl[\mathbf{1}[|f_i(x)| \ge \epsilon]\bigr]$, the proportion of inputs where feature $i$ activates for a fixed threshold $\epsilon$.

Even under high sparsity constraints, feature firing frequencies exhibit dramatic imbalance. Some features activate on nearly all inputs, while many rarely fire.

\begin{figure}[t]
    \centering
    \includegraphics[width=0.8\linewidth]{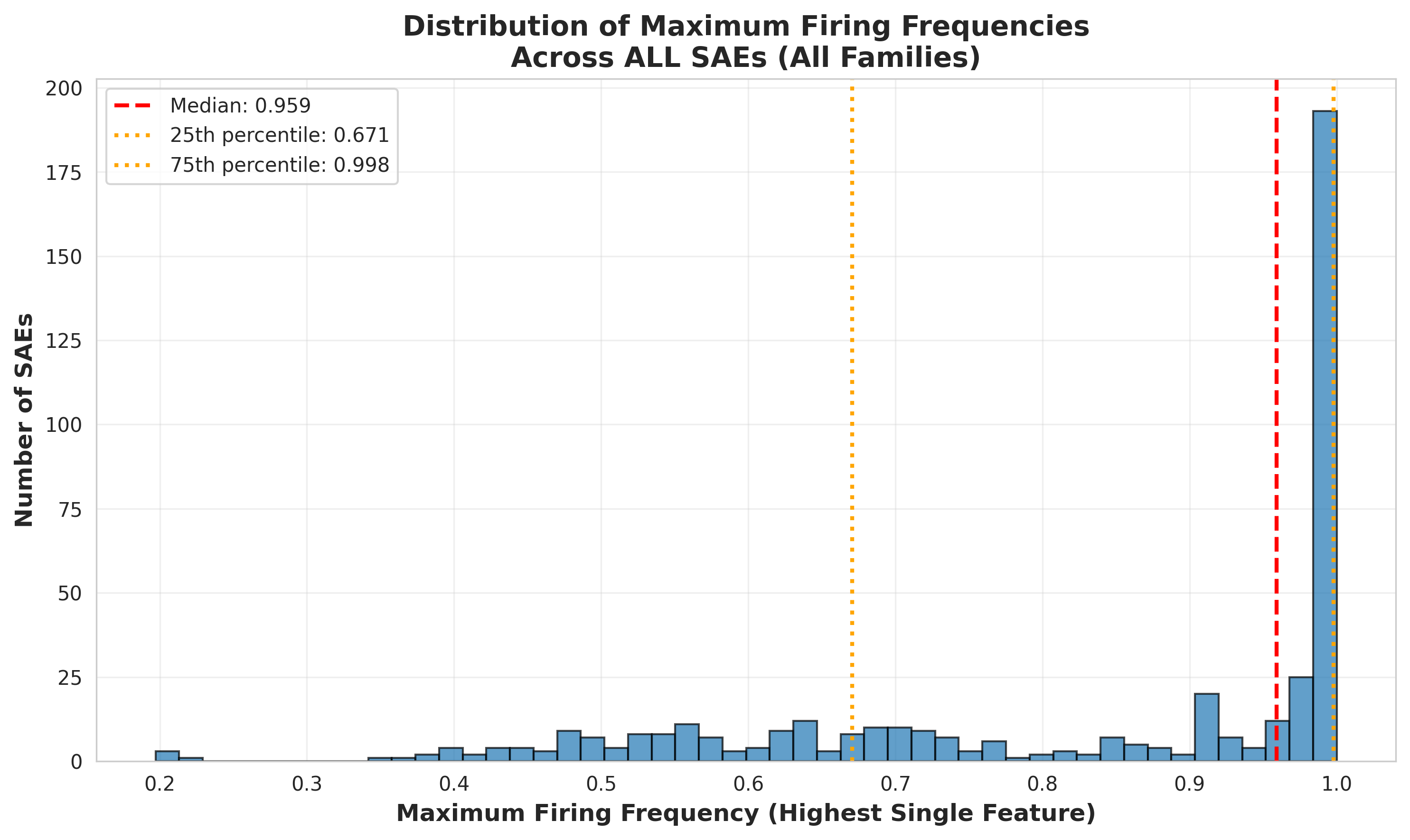}
    \caption{Distribution of maximum firing frequencies across all SAEs (all families). The median maximum firing frequency across 600+ SAEs is shown, with 75th percentile indicating that many SAEs have features that activate on a substantial fraction of inputs despite nominal sparsity constraints.}
    \label{fig:firing_freq_aggregation_all}
\end{figure}

Figure~\ref{fig:firing_freq_aggregation_all} shows the distribution of maximum firing frequencies across all evaluated SAEs. This aggregation reveals that imbalance is pervasive: across all architectures and model families, a substantial fraction of SAEs exhibit features that activate on a large proportion of inputs.

\begin{figure}[t]
    \centering
    \includegraphics[width=0.85\linewidth]{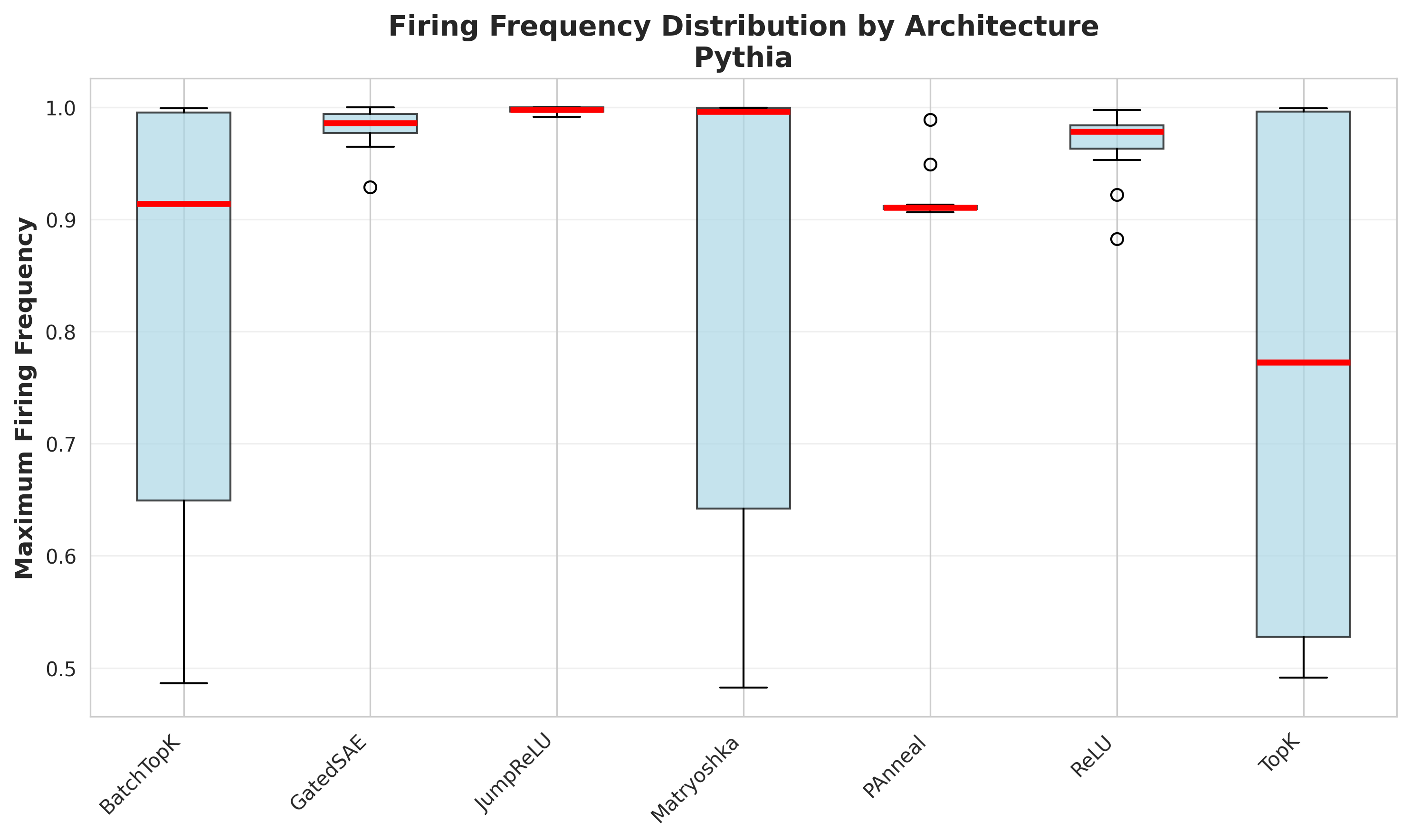}
    \includegraphics[width=0.85\linewidth]{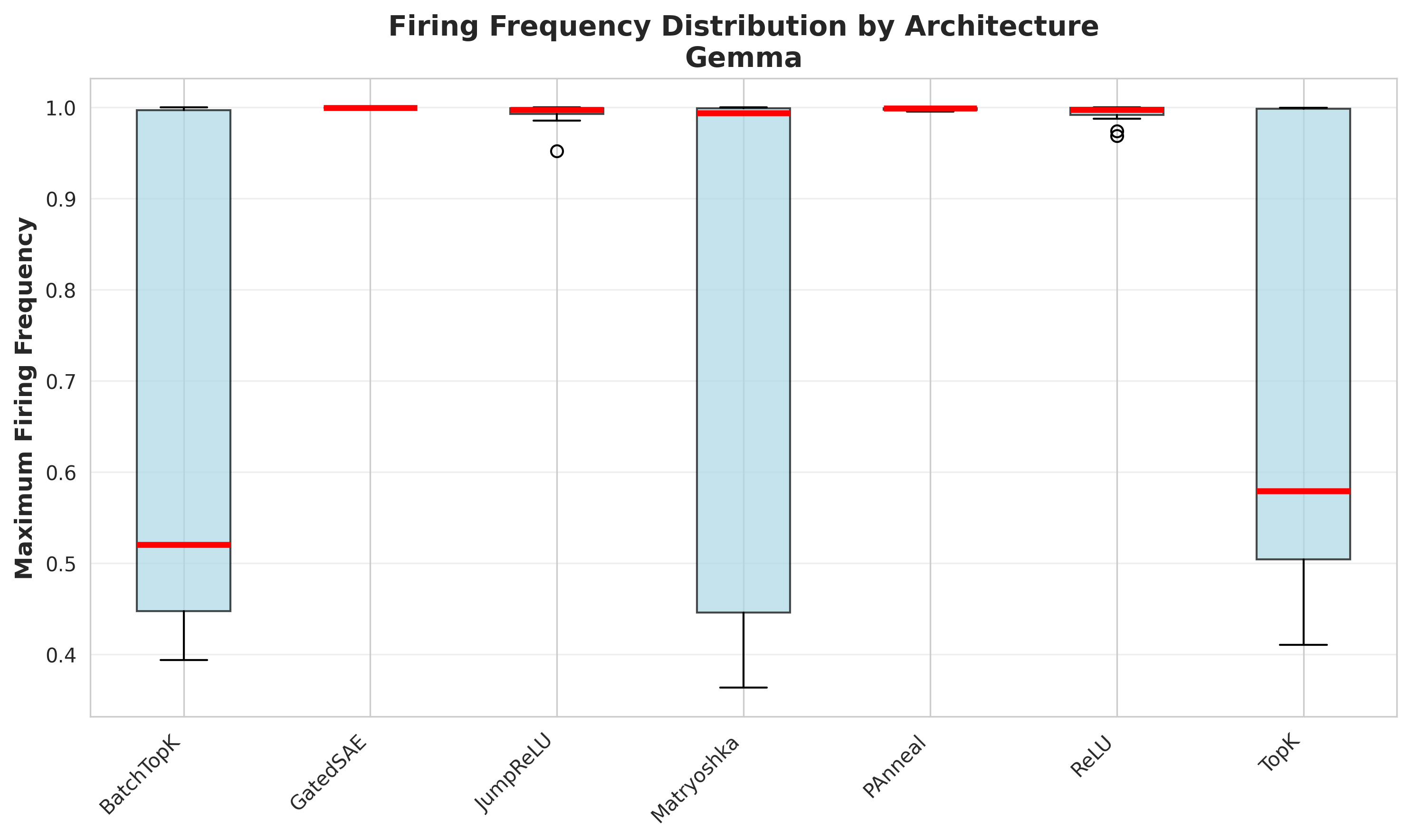}
    \caption{Firing frequency distributions by architecture for SAEBench Pythia (top) and Gemma (bottom) SAEs, shown as examples of patterns observed across all families. Violin plots show the distribution of maximum firing frequencies across all SAE configurations within each architecture, with the red line indicating the median. ReLU, PAnneal, Gated, and JumpReLU architectures consistently exhibit higher median maximum firing frequencies with lower variability compared to TopK and BatchTopK across all model families, which show lower medians but wider distributions.}
    \label{fig:firing_freq_by_architecture}
\end{figure}

\begin{figure}[t]
    \centering
    \includegraphics[width=0.8\linewidth]{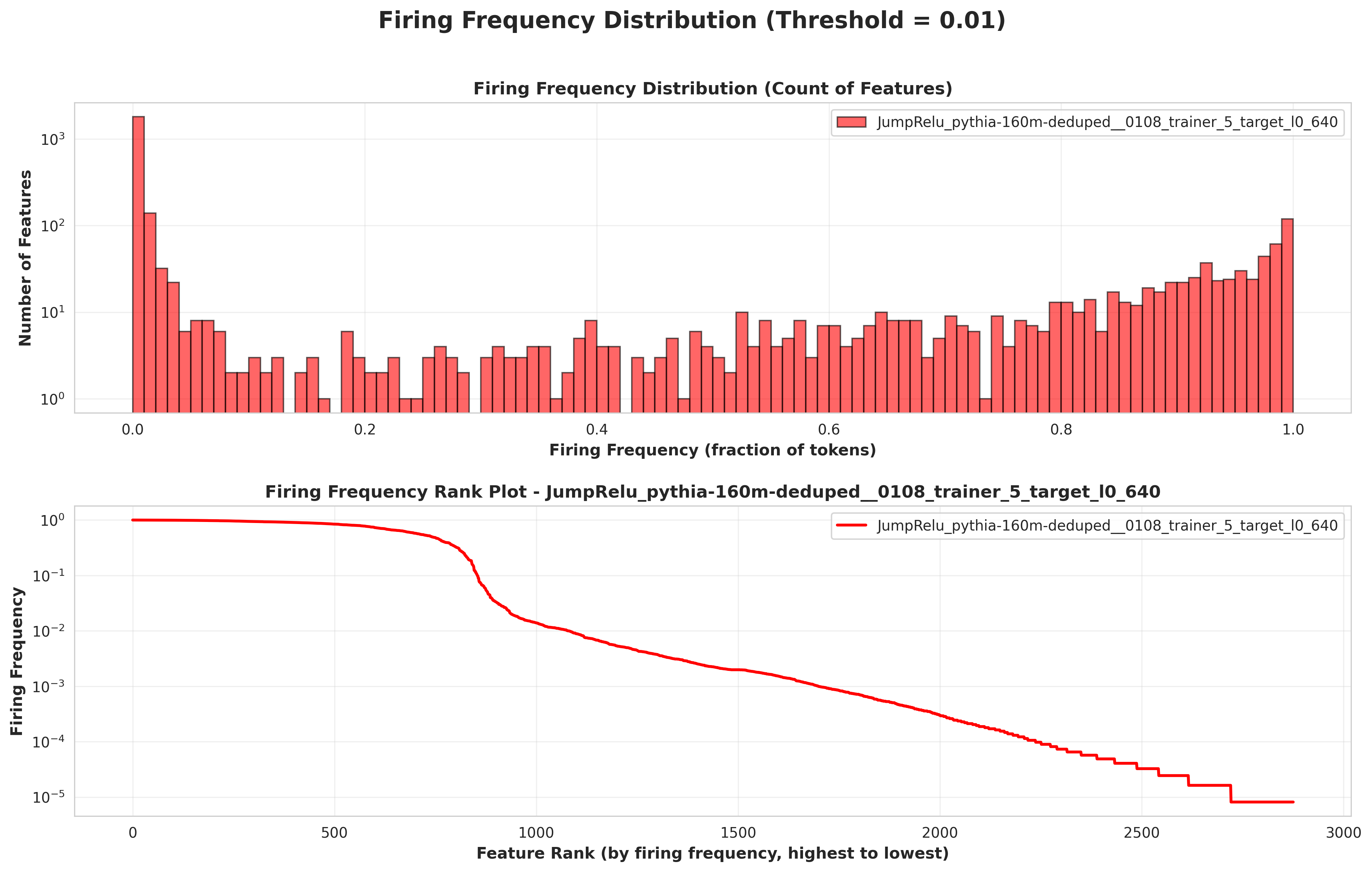}
    \caption{Extreme case: Firing frequency distribution for a JumpRelu Pythia SAE where 10.18\% of features fire more than 90\% of the time (threshold $\epsilon = 0.01$). This represents the most extreme case observed across all evaluated SAEs, demonstrating that high-frequency features can dominate a substantial fraction of the feature dictionary in pathological cases.}
    \label{fig:extreme_firing_frequency}
\end{figure}

\begin{figure}[t]
    \centering
    \includegraphics[width=1\linewidth]{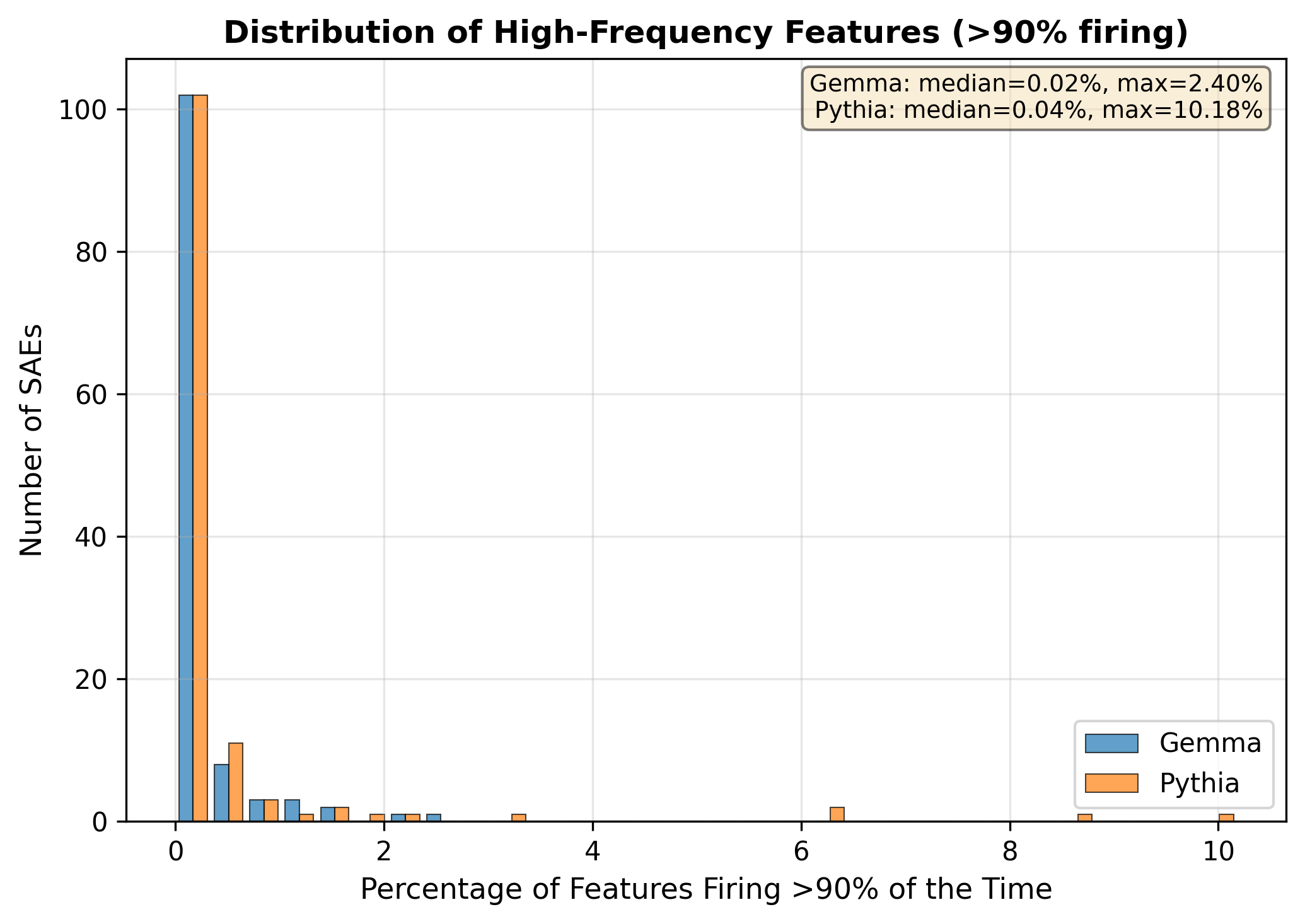}
    \caption{Distribution of the percentage of features that fire more than 90\% of the time across all SAEs (shown here for SAEBench Pythia and Gemma families as examples). While median values are low (0.02--0.04\%) across all families, maximum values reveal extreme cases where up to 10.18\% of features violate sparsity constraints by firing on nearly all inputs.}
    \label{fig:high_firing_freq_distribution}
\end{figure}

To quantify the prevalence of high-frequency features across all SAEs, Figure~\ref{fig:high_firing_freq_distribution} shows the distribution of the percentage of features that fire more than 90\% of the time. While the median percentage is low (typically 0.02--0.04\% for features firing more than 90\% of the time) across all SAE families, the maximum values reveal extreme cases: up to 10.18\% of features fire more than 90\% of the time in some SAEs, demonstrating that high-frequency features can dominate a substantial fraction of the feature dictionary in pathological cases. This pattern is observed across all model families, with the most extreme cases occurring in ReLU, PAnneal, and JumpReLU architectures. Figure~\ref{fig:extreme_firing_frequency} illustrates the most extreme case observed: a JumpReLU Pythia SAE where 10.18\% of features fire more than 90\% of the time, representing a pathological example where high-frequency features dominate a substantial fraction of the feature dictionary.

Architecture-specific analysis (Figure~\ref{fig:firing_freq_by_architecture}, Table~\ref{tab:max_firing_freq_by_architecture}) reveals differences across architectures. ReLU, PAnneal, Gated, and JumpReLU architectures exhibit high median maximum firing frequencies (0.990--0.999) across all SAE families, while TopK shows the lowest median (0.597). BatchTopK shows an intermediate median (0.749). As an illustrative example, SAEBench Pythia SAEs show ReLU (median 0.98), PAnneal (median 0.91), and Gated (median 0.99) with high medians and low variability, while TopK (median 0.78) and BatchTopK (median 0.92) show wider distributions. JumpReLU consistently shows maximum firing frequencies (median 1.0) with minimal variability across families. These architectural patterns are observed across GPT-2, OPT, TinyStories, CanRager, and JBloom, suggesting that architectural choices may influence both the level and distribution of maximum firing frequencies.

\subsection{Magnitude Distribution Imbalance ($M_i$)}

Total activation magnitude is $M_i = \mathbb{E}_x\bigl[|f_i(x)|\bigr]$, the expected absolute activation of feature $i$ across inputs.

The concentration of total activation magnitude across features is equally extreme. While feature dictionaries may contain thousands of features, only a small fraction contribute substantially to model computation.

\begin{figure}[t]
    \centering
    \includegraphics[width=1\linewidth]{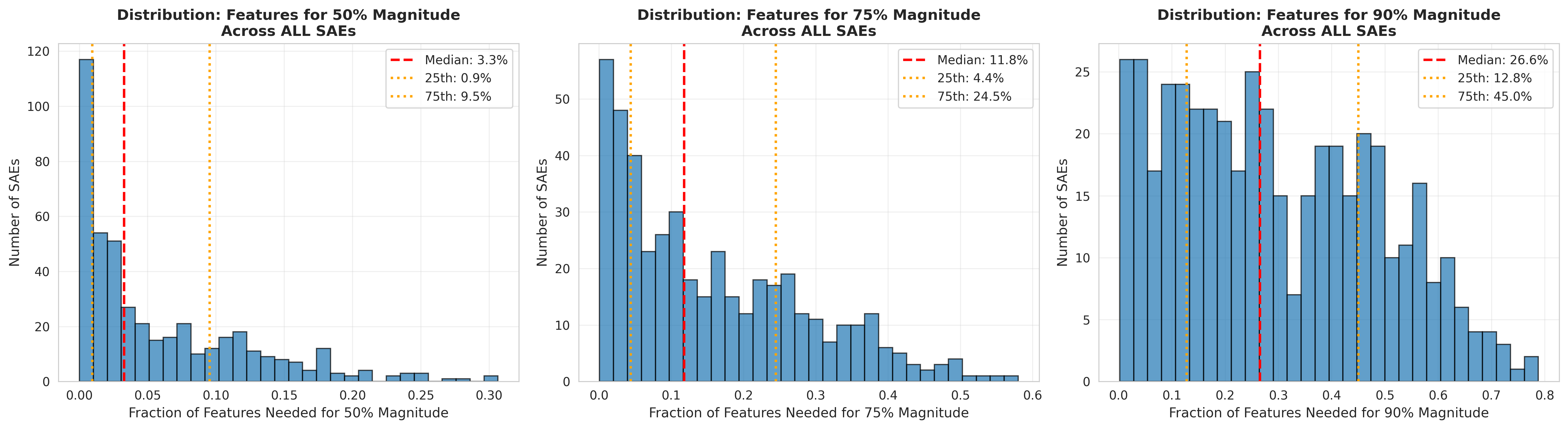}
    \caption{Distribution of feature fraction needed to account for 50\%, 75\%, and 90\% of total activation magnitude across all SAEs. The median SAE requires 3.3\% of features to account for 50\% of total magnitude, 11.8\% to account for 75\%, and 26.6\% to account for 90\%, demonstrating extreme concentration where the vast majority of features contribute negligibly.}
    \label{fig:magnitude_aggregation_all}
\end{figure}

Figure~\ref{fig:magnitude_aggregation_all} aggregates cumulative magnitude distributions across all SAEs. The median SAE requires 3.3\% of features to account for 50\% of total activation magnitude, 11.8\% to account for 75\%, and 26.6\% to account for 90\%. This concentration is consistent across architectures, though the specific percentages vary.

\begin{figure}[t]
    \centering
    \includegraphics[width=0.85\linewidth]{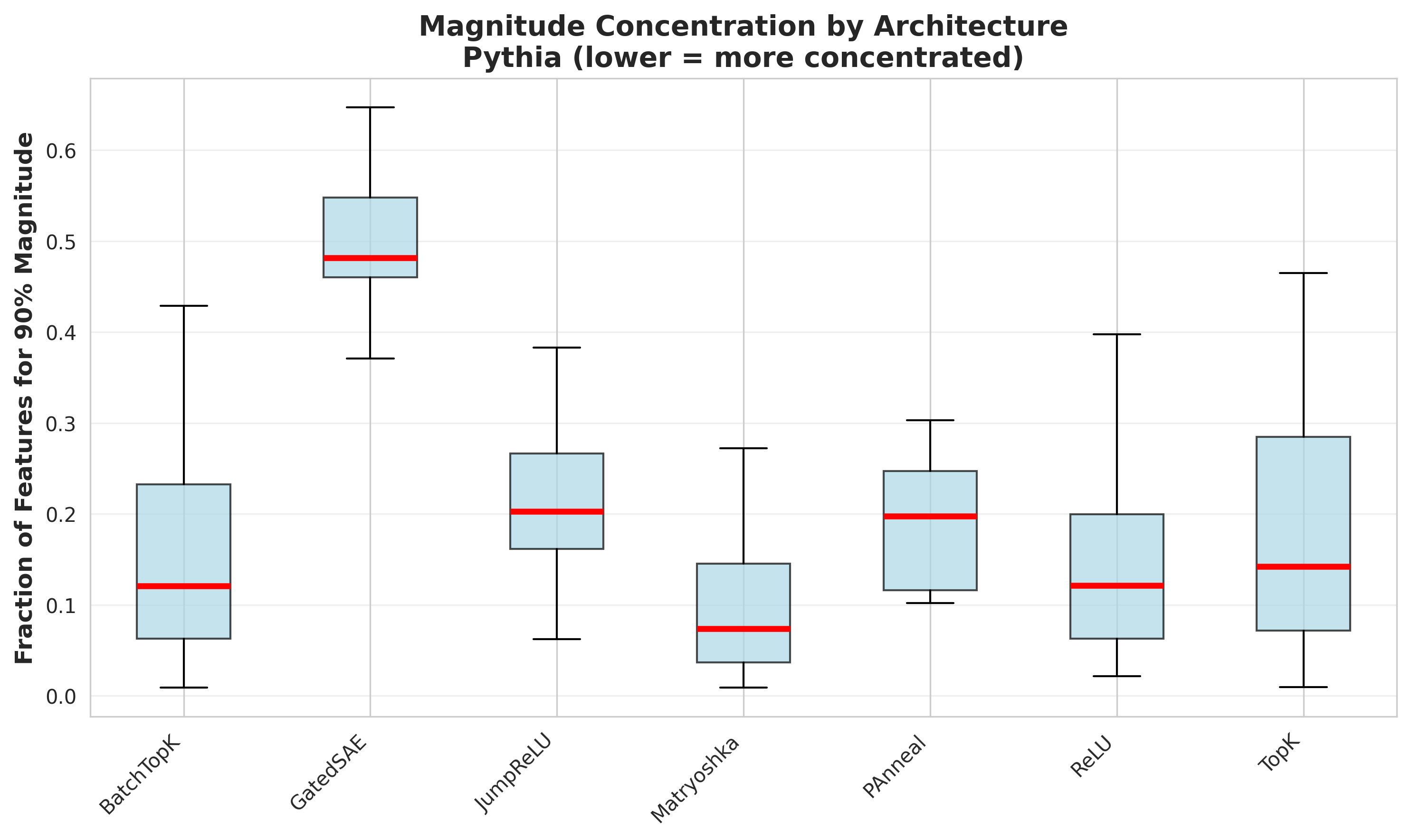}
    \includegraphics[width=0.85\linewidth]{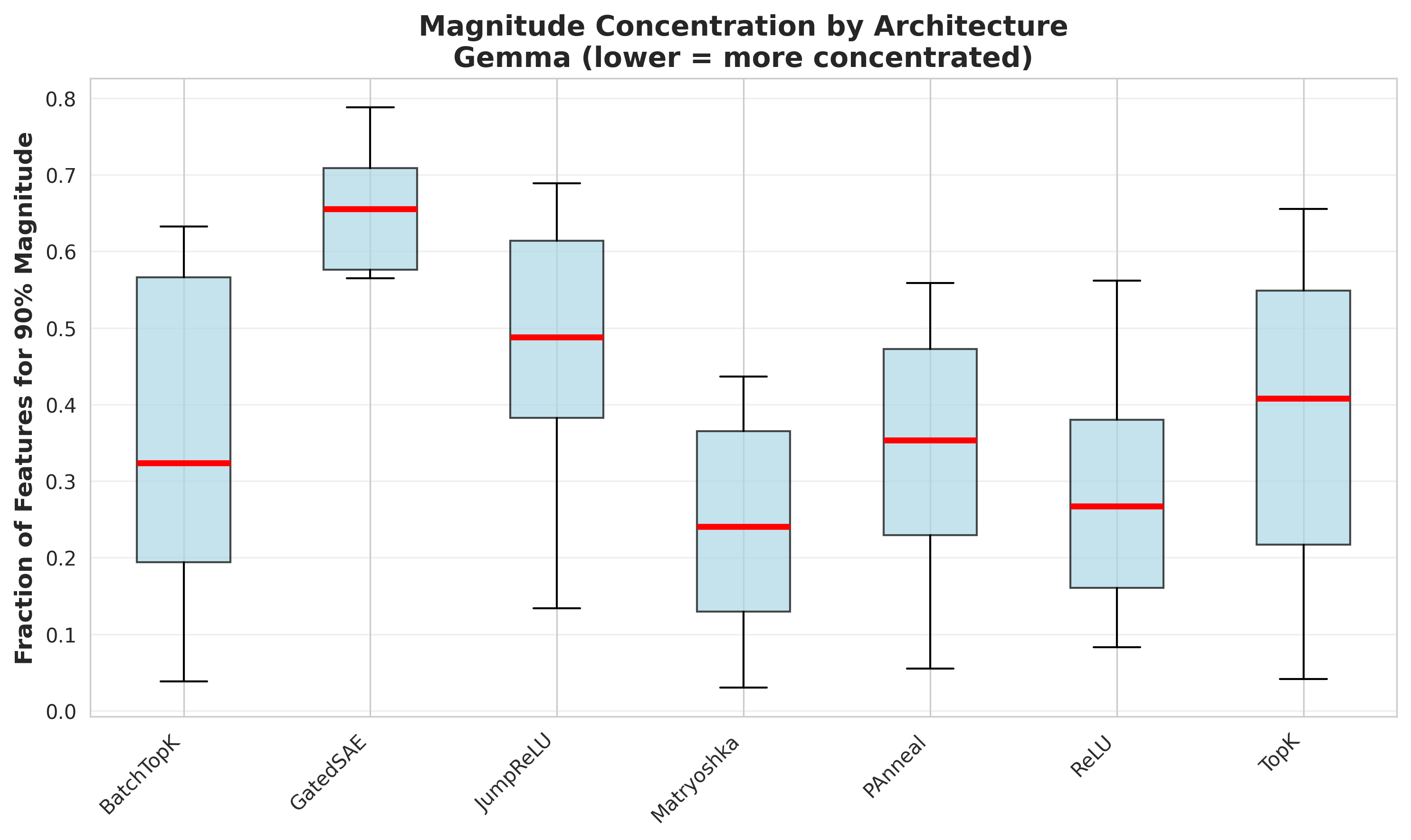}
    \caption{Magnitude concentration by architecture for SAEBench Pythia (top) and Gemma (bottom) SAEs, shown as examples of patterns observed across all families. Violin plots show the distribution of the fraction of features needed to account for 90\% of total magnitude across all SAE configurations within each architecture, with the red line indicating the median. Lower values indicate higher concentration (fewer features account for most magnitude).}
    \label{fig:magnitude_by_architecture}
\end{figure}

Architecture comparison (Figure~\ref{fig:magnitude_by_architecture}, Table~\ref{tab:magnitude_90pct_by_architecture}) reveals differences in magnitude concentration patterns across architectures. Matryoshka exhibits the highest concentration (median fraction 0.139) across all families, while Gated shows the lowest concentration (median fraction 0.575). BatchTopK (0.189) and ReLU (0.177) show relatively high concentration, while TopK (0.228), PAnneal (0.235), and JumpReLU (0.297) show moderate concentration. As illustrated by SAEBench Pythia and Gemma examples, these architectural patterns hold consistently: Matryoshka shows the highest concentration, Gated the lowest, with BatchTopK and ReLU showing intermediate-high concentration. These patterns are observed across GPT-2, OPT, TinyStories, CanRager, and JBloom, suggesting that architectural choices may influence how activation magnitude is distributed across features.

\begin{figure}[t]
    \centering
    \includegraphics[width=0.8\linewidth]{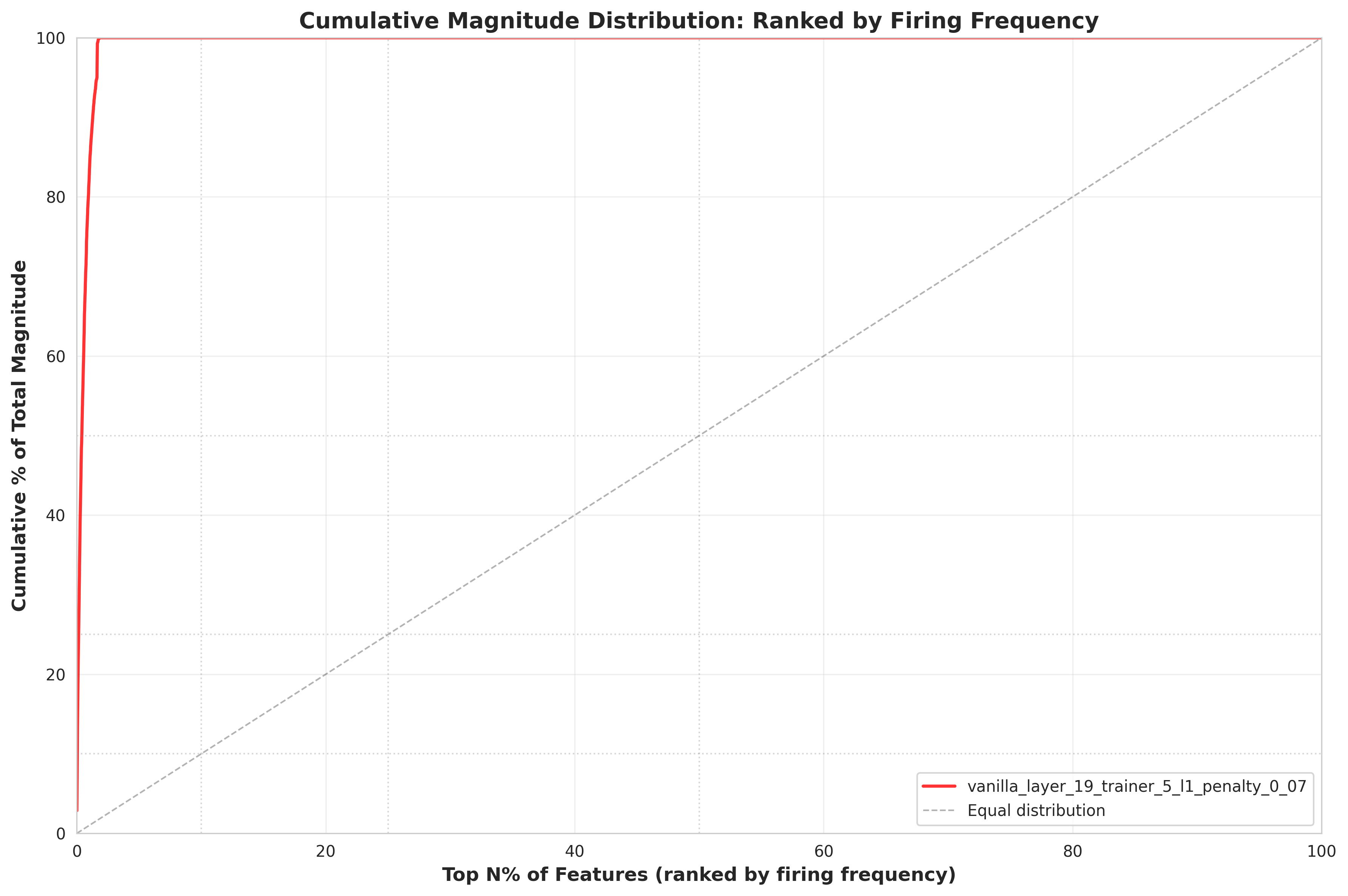}
    \caption{Extreme case: ReLU SAE from the CanRager family (layer 19, trainer 5, l1 penalty 0.07) cumulative magnitude distribution. Only $\sim$2.5\% of features account for nearly 100\% of cumulative activation magnitude---an extreme example where the vast majority of the feature dictionary contributes negligibly to actual computation.}
    \label{fig:extreme_cumulative_canager}
\end{figure}

Extreme cases demonstrate even more pronounced concentration. Figure~\ref{fig:extreme_cumulative_canager} shows a ReLU SAE from the CanRager SAE family (layer 19, trainer 5, l1 penalty 0.07) where only 2.5\% of features account for nearly 100\% of cumulative magnitude, representing an extreme but not isolated example of magnitude concentration.

\subsection{Per-Input Dominance ($\bar{\rho}_i$)}

Mean dominance is $\bar{\rho}_i = \mathbb{E}_x[\rho_i(x)]$, where $\rho_i(x) = |f_i(x)| / \sum_{j=1}^m |f_j(x)|$ is feature $i$'s fraction of total activation on input $x$. Firing frequency and activation magnitude both expose uneven feature usage, but capture different aspects: frequency measures how often a feature fires, while activation mass measures its aggregate magnitude. Neither alone provides an input-normalized share. Mean dominance instead normalizes each feature's activation by the total activation on the same input before averaging across inputs. We use it for stratification because a non-atomic feature occupying a large share of a representation can distort a large portion of the interpretation of that input.

\begin{figure}[t]
    \centering
    \includegraphics[width=0.55\linewidth]{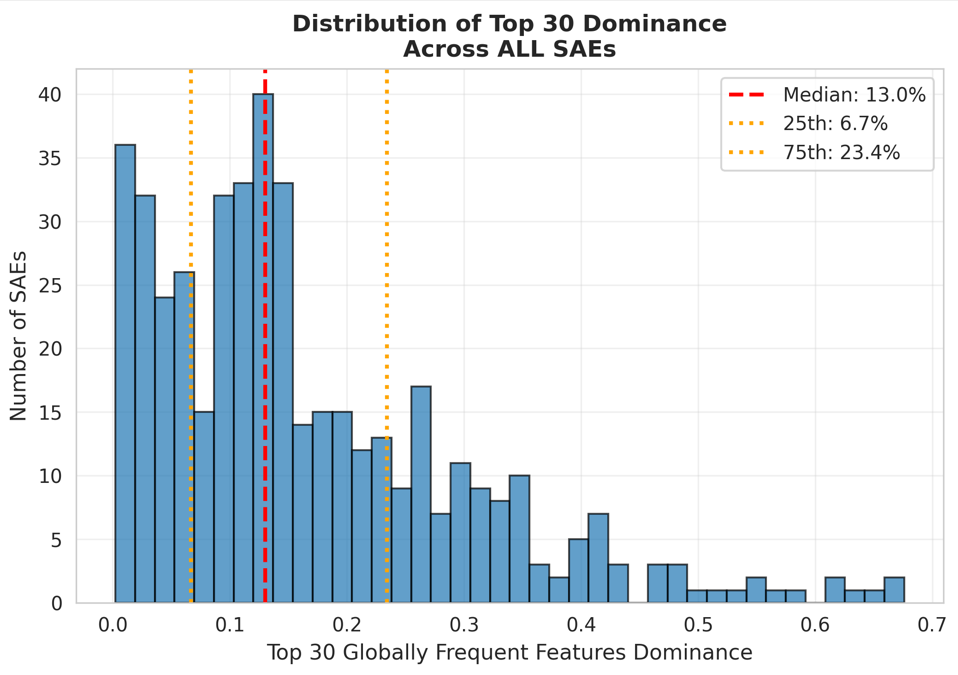}
    \caption{Distribution of top-30 feature dominance across all evaluated SAEs.}
    \label{fig:dominance_top30}
\end{figure}

Beyond aggregate firing and magnitude statistics, per-input dominance reveals structural patterns where the same small set of features persistently dominates individual input reconstructions. The top-10 distribution is shown in the main text (Figure~\ref{fig:dominance_top10}); the top-30 distribution is shown here in Figure~\ref{fig:dominance_top30}. The top 10 globally frequent features contribute a median of 7.8\% of per-input activation (25th percentile: 3.8\%, 75th percentile: 12.4\%), while the top 30 features contribute a median of 13.0\% (25th percentile: 6.7\%, 75th percentile: 23.4\%). Both distributions are right-skewed, with the majority of SAEs showing relatively low dominance values, but a long tail reveals many SAEs with extreme dominance---where a small number of features dominate reconstructions across most or all inputs, with some SAEs showing top 10 dominance exceeding 30\% and top 30 dominance exceeding 60\%.

\begin{figure}[t]
    \centering
    \includegraphics[width=0.85\linewidth]{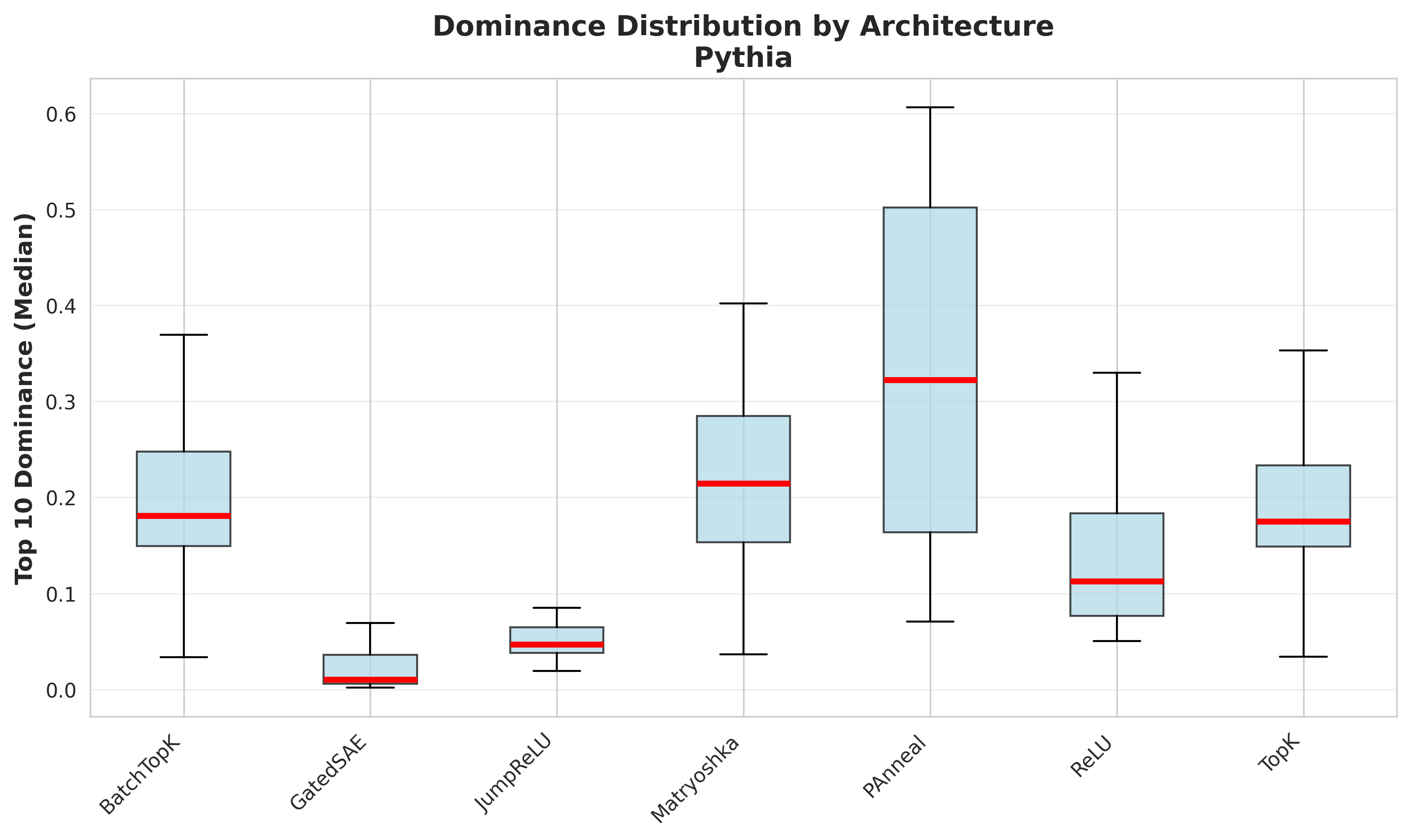}
    \includegraphics[width=0.85\linewidth]{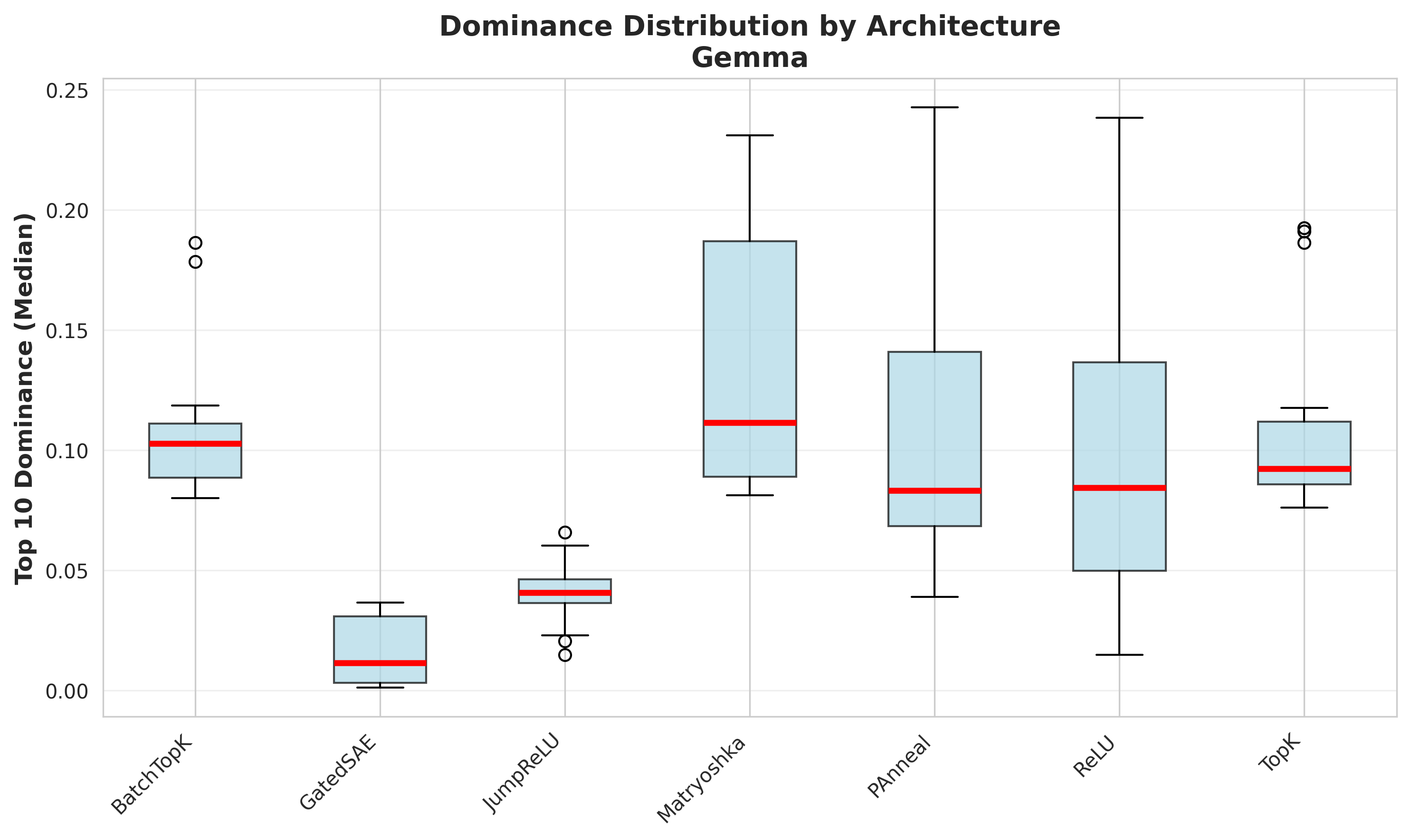}
    \caption{Dominance distributions by architecture for SAEBench Pythia (top) and Gemma (bottom) SAEs, shown as examples of patterns observed across all families. Violin plots show the distribution of top 10 dominance across all SAE configurations within each architecture, with the red line indicating the median. Matryoshka, BatchTopK, and PAnneal architectures exhibit higher median dominance compared to TopK and ReLU across all families.}
    \label{fig:dominance_by_architecture}
\end{figure}

Architecture comparisons (Figure~\ref{fig:dominance_by_architecture}, Table~\ref{tab:top10_dominance_by_architecture}) reveal architectural differences across all evaluated model families. Matryoshka shows the highest median top 10 dominance (0.177), followed by BatchTopK (0.151) and PAnneal (0.155). ReLU (0.102) and TopK (0.120) show moderate dominance, while Gated (0.011), JumpReLU (0.045), and JBloom (0.046) show the lowest dominance. As illustrated by SAEBench Pythia and Gemma examples, these patterns are consistent: Matryoshka, BatchTopK, and PAnneal show higher dominance, while TopK, ReLU, Gated, and JumpReLU show lower dominance. These architectural patterns are observed across GPT-2, OPT, TinyStories, CanRager, and JBloom, suggesting that architectural choices may influence dominance patterns.

\begin{figure}[t]
    \centering
    \includegraphics[width=0.8\linewidth]{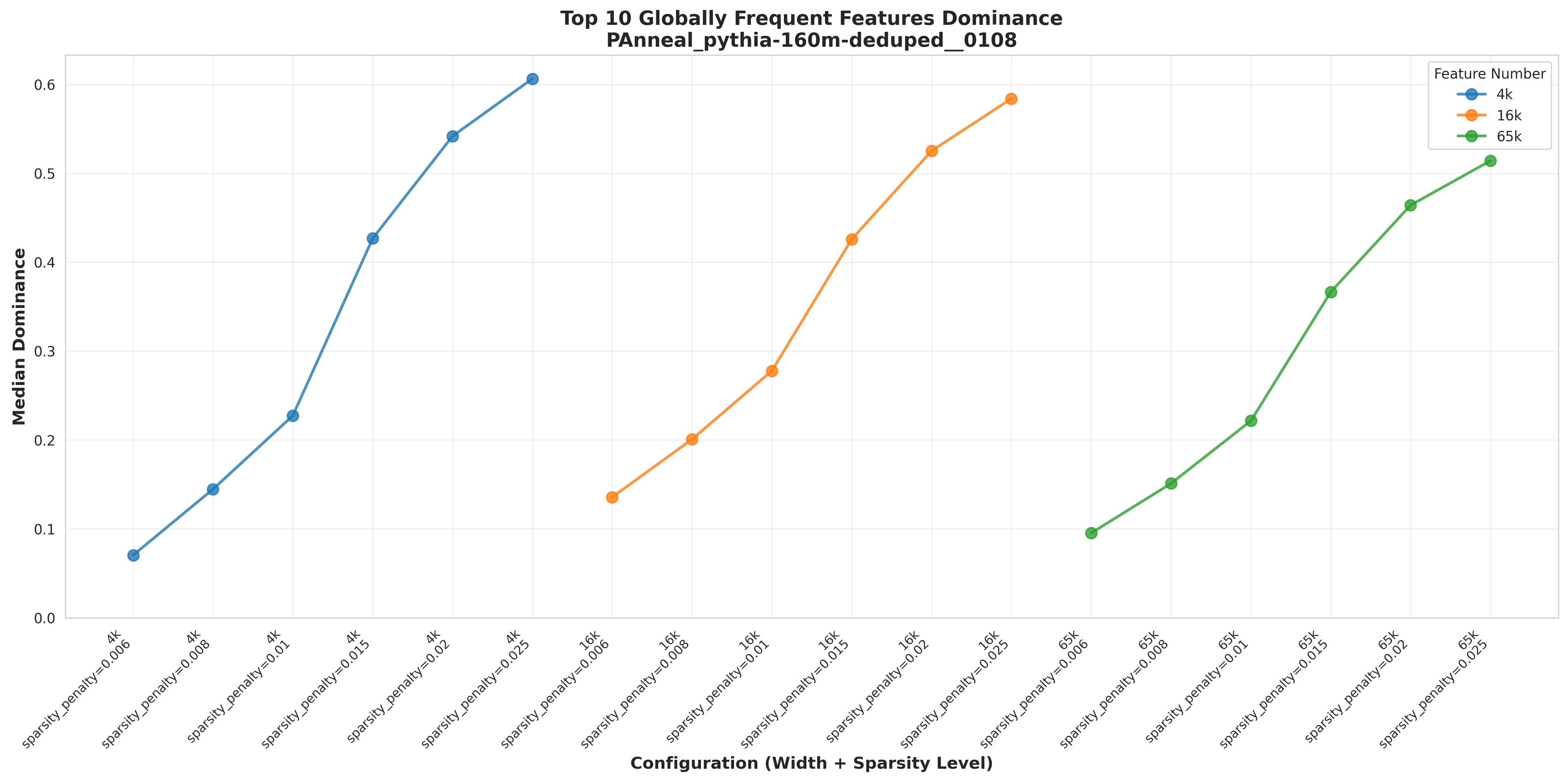}
    \caption{Extreme case: Mean dominance by the most frequent feature. For some SAEs, the single most frequently firing feature contributes up to 60\% of total activation magnitude on average across all inputs, demonstrating persistent structural dominance where individual features encode multiple concepts.}
    \label{fig:extreme_per_input_dominance}
\end{figure}

Figure~\ref{fig:extreme_per_input_dominance} illustrates an extreme case where the single most frequent feature contributes up to 60\% of per-input activation on average. Such extreme dominance indicates that a small number of features persistently encode diverse input patterns, directly violating the monosemanticity objective.

\subsection{Dominance-Sparsity Relationship}

An intuitive expectation is that higher sparsity constraints (fewer features active per input) should promote more distributed feature usage, with each feature making a smaller relative contribution. However, we observe the opposite pattern in many architectures.

\begin{figure}[t]
    \centering
    \includegraphics[width=0.9\linewidth]{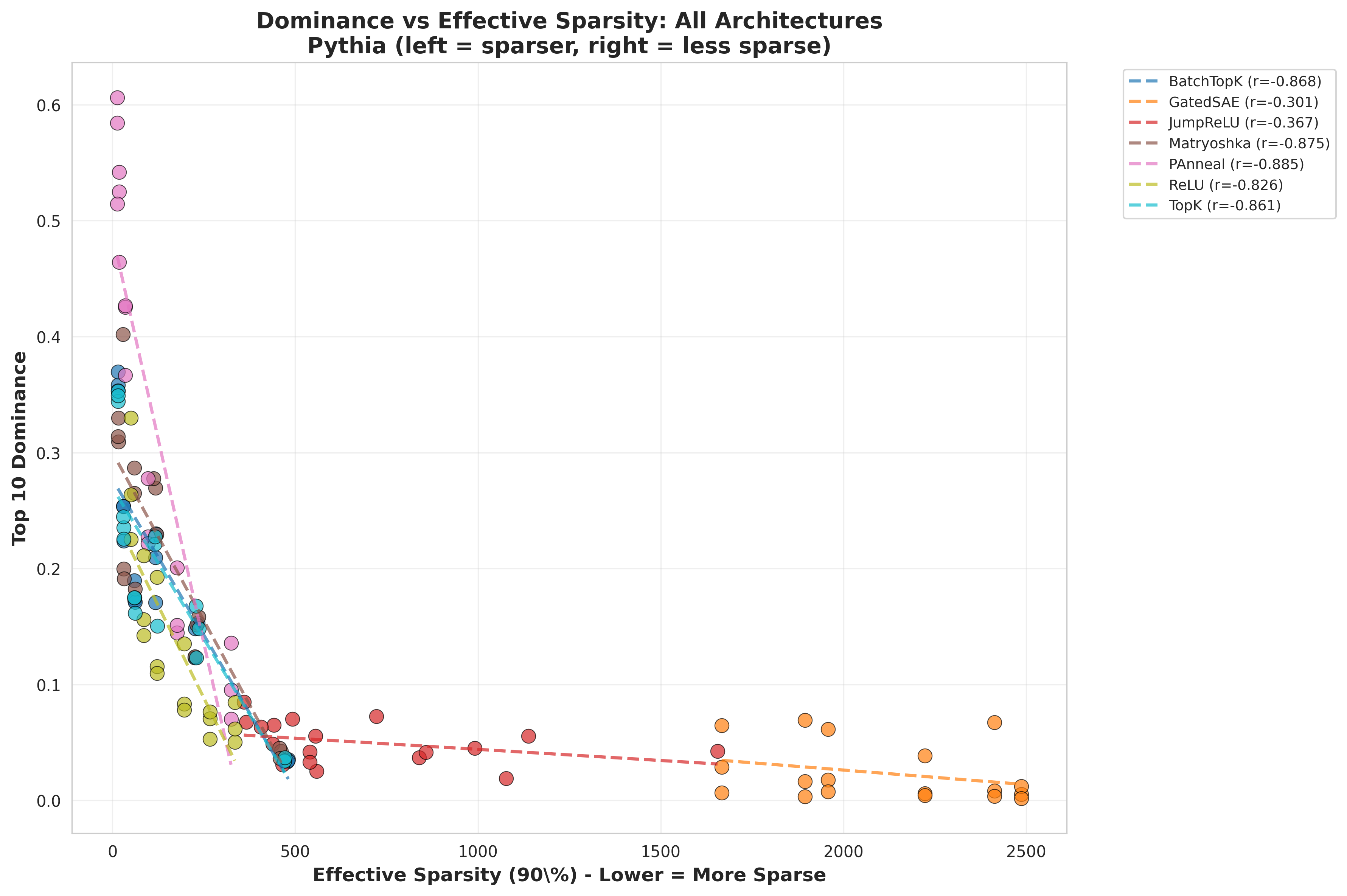}
    \includegraphics[width=0.9\linewidth]{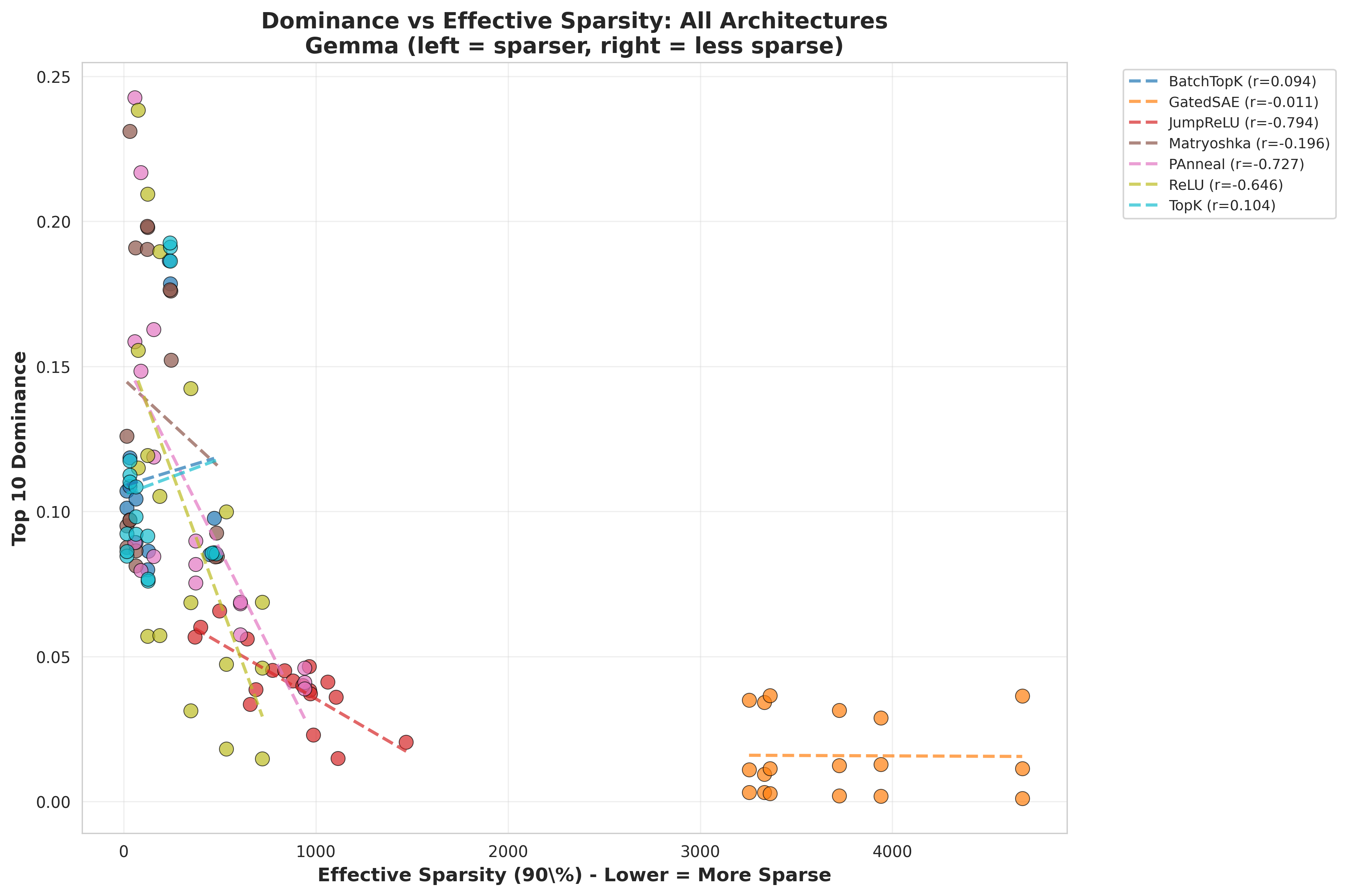}
    \caption{Dominance vs. effective sparsity (90\%) for all architectures in SAEBench Pythia (top) and Gemma (bottom), shown as examples of patterns observed across all families. The x-axis shows effective active features (90\%), where lower values indicate higher sparsity. All architectures are shown together, revealing that ReLU and PAnneal exhibit strong positive correlations (higher sparsity $\to$ higher dominance) across all families, while TopK and BatchTopK show weaker or mixed patterns. This counterintuitive relationship indicates that sparsity constraints alone are insufficient to promote distributed feature usage.}
    \label{fig:dominance_sparsity_relationship}
\end{figure}

Across all evaluated SAE families, top 10 dominance shows relationships with effective sparsity that vary by architecture. Using effective active features (90\%) as a unified sparsity measure allows direct comparison across architectures with different sparsity mechanisms. ReLU and PAnneal architectures consistently exhibit strong positive correlations (higher sparsity $\to$ higher dominance) across all model families, while TopK and BatchTopK show weaker or mixed relationships. As illustrated by SAEBench Pythia and Gemma examples (Figure~\ref{fig:dominance_sparsity_relationship}), ReLU and PAnneal show strong positive correlations, while TopK and BatchTopK show weaker patterns. These architectural patterns are consistent across GPT-2, OPT, TinyStories, CanRager, and JBloom.

To rigorously test this relationship while controlling for confounding factors, we compute correlations between effective sparsity and dominance separately for each architecture and dictionary size (width) across all model families. Tables~\ref{tab:sparsity_dominance_correlation_pythia} and~\ref{tab:sparsity_dominance_correlation_gemma} present these correlations for SAEBench Pythia and Gemma as examples, isolating the effect of sparsity on dominance while holding architecture and dictionary size constant.

The tables reveal several key patterns that hold across all evaluated families. First, ReLU and PAnneal architectures consistently show strong positive correlations ($r > 0.85$) across all model families and dictionary sizes, confirming the counterintuitive pattern: as sparsity increases (fewer effective active features), dominance increases. Second, the strength of correlations varies by model family, with some families (e.g., Pythia, GPT-2, OPT, TinyStories) showing stronger and more consistent positive correlations across architectures compared to others. Third, Gated and JumpReLU architectures show weaker or mixed correlations, varying by model family. Fourth, TopK and BatchTopK architectures show weak correlations (sometimes negative, sometimes positive), suggesting these architectures may distribute computation more evenly as sparsity increases in some families, though the effect is generally weak. Fifth, the patterns are consistent across different dictionary sizes (4k, 16k, 65k) within each family, indicating that width is not a confounding factor. Correlations with $|r| \geq 0.5$ are shown in bold.

\begin{table}[t]
\centering
\small
\begin{tabular}{lcc|r}
\toprule
\textbf{Arch} & \textbf{Width} & \textbf{N} & \textbf{r} \\
\midrule
BatchTopK & 16k & 6 & \textbf{0.877} \\
BatchTopK & 4k & 6 & \textbf{0.863} \\
BatchTopK & 65k & 6 & \textbf{0.867} \\
Gated & 16k & 6 & \textbf{0.934} \\
Gated & 4k & 6 & \textbf{0.629} \\
Gated & 65k & 6 & \textbf{0.694} \\
JumpReLU & 16k & 6 & -0.367 \\
JumpReLU & 4k & 6 & \textbf{0.853} \\
JumpReLU & 65k & 6 & 0.129 \\
Matryoshka & 16k & 6 & \textbf{0.855} \\
Matryoshka & 4k & 6 & \textbf{0.897} \\
Matryoshka & 65k & 6 & \textbf{0.930} \\
PAnneal & 16k & 6 & \textbf{0.901} \\
PAnneal & 4k & 6 & \textbf{0.896} \\
PAnneal & 65k & 6 & \textbf{0.896} \\
Standard ReLU & 16k & 6 & \textbf{0.876} \\
Standard ReLU & 4k & 6 & \textbf{0.913} \\
Standard ReLU & 65k & 6 & \textbf{0.856} \\
TopK & 16k & 6 & \textbf{0.882} \\
TopK & 4k & 6 & \textbf{0.828} \\
TopK & 65k & 6 & \textbf{0.881} \\
\bottomrule
\end{tabular}
\caption{Correlation between sparsity level and top 10 dominance for Saebench Pythia SAEs, controlling for width and architecture. For each architecture and dictionary size (width), we compute Pearson correlation between sparsity level (higher values = more sparse, fewer active features) and dominance across different sparsity parameter settings. A \textbf{positive} correlation indicates that as sparsity increases, dominance increases---more sparse SAEs concentrate computation in fewer dominant features. Correlations with $|r| \geq 0.5$ are shown in \textbf{bold}.}
\label{tab:sparsity_dominance_correlation_pythia}
\end{table}

\begin{table}[t]
\centering
\small
\begin{tabular}{lcc|r}
\toprule
\textbf{Arch} & \textbf{Width} & \textbf{N} & \textbf{r} \\
\midrule
BatchTopK & 4k & 6 & -0.174 \\
BatchTopK & 65k & 6 & -0.021 \\
Gated & 16k & 6 & -0.375 \\
Gated & 4k & 6 & 0.038 \\
Gated & 65k & 6 & \textbf{0.954} \\
JumpReLU & 16k & 6 & -0.462 \\
JumpReLU & 4k & 6 & \textbf{0.910} \\
JumpReLU & 65k & 6 & \textbf{0.727} \\
Matryoshka & 16k & 6 & -0.074 \\
Matryoshka & 4k & 6 & \textbf{0.707} \\
Matryoshka & 65k & 6 & -0.036 \\
PAnneal & 16k & 6 & \textbf{0.947} \\
PAnneal & 4k & 6 & \textbf{0.910} \\
PAnneal & 65k & 6 & \textbf{0.953} \\
Standard ReLU & 16k & 6 & \textbf{0.920} \\
Standard ReLU & 4k & 6 & \textbf{0.988} \\
Standard ReLU & 65k & 6 & \textbf{0.839} \\
TopK & 16k & 6 & -0.129 \\
TopK & 4k & 6 & -0.103 \\
TopK & 65k & 6 & -0.081 \\
\bottomrule
\end{tabular}
\caption{Correlation between sparsity level and top 10 dominance for Saebench Gemma SAEs, controlling for width and architecture. For each architecture and dictionary size (width), we compute Pearson correlation between sparsity level (higher values = more sparse, fewer active features) and dominance across different sparsity parameter settings. A \textbf{positive} correlation indicates that as sparsity increases, dominance increases---more sparse SAEs concentrate computation in fewer dominant features. Correlations with $|r| \geq 0.5$ are shown in \textbf{bold}.}
\label{tab:sparsity_dominance_correlation_gemma}
\end{table}

To assess the generality of this phenomenon across all SAE families, Tables~\ref{tab:sparsity_dominance_correlation_gpt2}--\ref{tab:sparsity_dominance_correlation_jbloom} present correlations for GPT-2, OPT-125M, TinyStories, CanRager, and JBloom families separately. Across all families, ReLU and PAnneal architectures consistently show strong positive correlations, confirming the general pattern. GPT-2, OPT-125M, and TinyStories show consistently positive correlations ($r \geq 0.7$) across all architectures, though statistical significance is limited by smaller sample sizes ($n = 3$). CanRager and JBloom show weaker correlations overall, consistent with their different training configurations, but still exhibit the same architectural patterns (ReLU and PAnneal showing stronger positive correlations than TopK-based methods). These results demonstrate that the sparsity-dominance relationship is a general phenomenon across diverse SAE families, with architectural differences being the primary determinant. Correlations with $|r| \geq 0.5$ are shown in bold.

\begin{table}[t]
\centering
\small
\begin{tabular}{lcc|r}
\toprule
\textbf{Arch} & \textbf{Width} & \textbf{N} & \textbf{r} \\
\midrule
BatchTopK & 16k & 3 & \textbf{0.860} \\
BatchTopK & 4k & 3 & \textbf{0.853} \\
BatchTopK & 65k & 3 & \textbf{0.834} \\
Gated & 16k & 3 & \textbf{0.831} \\
Gated & 4k & 3 & \textbf{0.878} \\
Gated & 65k & 3 & \textbf{0.792} \\
JumpReLU & 16k & 3 & \textbf{0.796} \\
JumpReLU & 4k & 3 & \textbf{0.830} \\
JumpReLU & 65k & 3 & \textbf{0.885} \\
Matryoshka & 16k & 3 & \textbf{0.855} \\
Matryoshka & 4k & 3 & \textbf{0.796} \\
Matryoshka & 65k & 3 & \textbf{0.902} \\
Standard ReLU & 16k & 3 & \textbf{0.819} \\
Standard ReLU & 4k & 3 & \textbf{0.853} \\
Standard ReLU & 65k & 3 & \textbf{0.821} \\
TopK & 16k & 3 & \textbf{0.826} \\
TopK & 4k & 3 & \textbf{-0.888} \\
TopK & 65k & 3 & \textbf{0.952} \\
\bottomrule
\end{tabular}
\caption{Correlation between sparsity level and top 10 dominance for GPT2-Small SAEs, controlling for width and architecture. For each architecture and dictionary size (width), we compute Pearson correlation between sparsity level (higher values = more sparse, fewer active features) and dominance across different sparsity parameter settings. A \textbf{positive} correlation indicates that as sparsity increases, dominance increases---more sparse SAEs concentrate computation in fewer dominant features. Correlations with $|r| \geq 0.5$ are shown in \textbf{bold}.}
\label{tab:sparsity_dominance_correlation_gpt2}
\end{table}

\begin{table}[t]
\centering
\small
\begin{tabular}{lcc|r}
\toprule
\textbf{Arch} & \textbf{Width} & \textbf{N} & \textbf{r} \\
\midrule
Gated & 16k & 3 & \textbf{0.958} \\
Gated & 4k & 3 & \textbf{0.956} \\
Gated & 65k & 3 & \textbf{0.965} \\
JumpReLU & 16k & 3 & \textbf{0.982} \\
JumpReLU & 4k & 3 & \textbf{0.922} \\
JumpReLU & 65k & 3 & \textbf{0.996} \\
Standard ReLU & 16k & 3 & \textbf{0.973} \\
Standard ReLU & 4k & 3 & \textbf{0.965} \\
Standard ReLU & 65k & 3 & \textbf{0.984} \\
TopK & 16k & 3 & \textbf{0.933} \\
TopK & 4k & 3 & \textbf{-0.559} \\
TopK & 65k & 3 & \textbf{0.996} \\
\bottomrule
\end{tabular}
\caption{Correlation between sparsity level and top 10 dominance for OPT125M SAEs, controlling for width and architecture. For each architecture and dictionary size (width), we compute Pearson correlation between sparsity level (higher values = more sparse, fewer active features) and dominance across different sparsity parameter settings. A \textbf{positive} correlation indicates that as sparsity increases, dominance increases---more sparse SAEs concentrate computation in fewer dominant features. Correlations with $|r| \geq 0.5$ are shown in \textbf{bold}.}
\label{tab:sparsity_dominance_correlation_opt125m}
\end{table}

\begin{table}[t]
\centering
\small
\begin{tabular}{lcc|r}
\toprule
\textbf{Arch} & \textbf{Width} & \textbf{N} & \textbf{r} \\
\midrule
BatchTopK & 16k & 3 & \textbf{0.858} \\
BatchTopK & 4k & 3 & \textbf{0.844} \\
BatchTopK & 65k & 3 & \textbf{0.876} \\
Gated & 16k & 3 & \textbf{0.714} \\
Gated & 4k & 3 & \textbf{0.802} \\
Gated & 65k & 3 & \textbf{0.714} \\
JumpReLU & 16k & 3 & \textbf{0.939} \\
JumpReLU & 4k & 3 & \textbf{0.940} \\
JumpReLU & 65k & 3 & \textbf{0.952} \\
Matryoshka & 16k & 3 & \textbf{0.871} \\
Matryoshka & 4k & 3 & \textbf{0.784} \\
Standard ReLU & 16k & 3 & \textbf{0.859} \\
Standard ReLU & 4k & 3 & \textbf{0.912} \\
Standard ReLU & 65k & 3 & \textbf{0.855} \\
TopK & 16k & 3 & \textbf{0.888} \\
TopK & 4k & 3 & \textbf{0.782} \\
TopK & 65k & 3 & \textbf{0.908} \\
\bottomrule
\end{tabular}
\caption{Correlation between sparsity level and top 10 dominance for Tinystories SAEs, controlling for width and architecture. For each architecture and dictionary size (width), we compute Pearson correlation between sparsity level (higher values = more sparse, fewer active features) and dominance across different sparsity parameter settings. A \textbf{positive} correlation indicates that as sparsity increases, dominance increases---more sparse SAEs concentrate computation in fewer dominant features. Correlations with $|r| \geq 0.5$ are shown in \textbf{bold}.}
\label{tab:sparsity_dominance_correlation_tinystories}
\end{table}

\begin{table}[t]
\centering
\small
\begin{tabular}{lcc|r}
\toprule
\textbf{Arch} & \textbf{Width} & \textbf{N} & \textbf{r} \\
\midrule
TopK & 16k & 18 & -0.140 \\
TopK & 4k & 18 & -0.043 \\
TopK & 65k & 18 & -0.132 \\
\bottomrule
\end{tabular}
\caption{Correlation between sparsity level and top 10 dominance for Canrager SAEs, controlling for width and architecture. For each architecture and dictionary size (width), we compute Pearson correlation between sparsity level (higher values = more sparse, fewer active features) and dominance across different sparsity parameter settings. A \textbf{positive} correlation indicates that as sparsity increases, dominance increases---more sparse SAEs concentrate computation in fewer dominant features. Correlations with $|r| \geq 0.5$ are shown in \textbf{bold}.}
\label{tab:sparsity_dominance_correlation_canranger}
\end{table}

\begin{table}[t]
\centering
\small
\begin{tabular}{lcc|r}
\toprule
\textbf{Arch} & \textbf{Width} & \textbf{N} & \textbf{r} \\
\midrule
Standard ReLU & 128k & 48 & \textbf{0.674} \\
Standard ReLU & 32k & 48 & \textbf{0.732} \\
\bottomrule
\end{tabular}
\caption{Correlation between sparsity level and top 10 dominance for Jbloom SAEs, controlling for width and architecture. For each architecture and dictionary size (width), we compute Pearson correlation between sparsity level (higher values = more sparse, fewer active features) and dominance across different sparsity parameter settings. A \textbf{positive} correlation indicates that as sparsity increases, dominance increases---more sparse SAEs concentrate computation in fewer dominant features. Correlations with $|r| \geq 0.5$ are shown in \textbf{bold}.}
\label{tab:sparsity_dominance_correlation_jbloom}
\end{table}

This counterintuitive sparsity-dominance relationship (where it occurs) suggests that globally frequent features become even more dominant as fewer features are allowed to activate per input, creating a ``winner-take-most'' dynamic where the same small set of features encodes diverse patterns. This directly undermines the interpretability objective: higher sparsity should distribute computation, but instead concentrates it further.

\subsection{Feature Number Effects}

Expanding the feature dictionary size (from 4k to 16k to 65k features) might be expected to reduce dominance, as additional capacity becomes available to distribute computational load. However, this relationship is weak and inconsistent across architectures. TopK and BatchTopK architectures in particular show little change---or even slight increases---in dominance as dictionary size grows, suggesting that additional capacity is not effectively utilized to distribute activation. ReLU-based architectures generally show the expected pattern (dominance decreases with larger dictionaries), though the effect is modest. Overall, feature number shows weaker correlation with dominance compared to sparsity level, indicating that capacity alone is insufficient to address imbalance.

\begin{table}[t]
\centering
\caption{Median maximum firing frequency by architecture across all SAE families.}
\label{tab:max_firing_freq_by_architecture}
\begin{tabular}{lc}
\toprule
Architecture & Median Max Firing Freq. \\
\midrule
ReLU        & 0.990 \\
TopK        & 0.597 \\
BatchTopK   & 0.749 \\
PAnneal     & 0.994 \\
JumpReLU    & 0.998 \\
Gated    & 0.999 \\
Matryoshka  & 0.996 \\
JBloom      & 0.702 \\
\bottomrule
\end{tabular}
\end{table}

\begin{table}[t]
\centering
\caption{Median fraction of features needed to account for 90\% of total magnitude by architecture across all SAE families. Lower values indicate higher concentration.}
\label{tab:magnitude_90pct_by_architecture}
\begin{tabular}{lc}
\toprule
Architecture & Median Fraction (90\%) \\
\midrule
ReLU       & 0.177 \\
TopK       & 0.228 \\
BatchTopK  & 0.189 \\
PAnneal    & 0.235 \\
JumpReLU   & 0.297 \\
Gated   & 0.575 \\
Matryoshka & 0.139 \\
JBloom     & 0.288 \\
\bottomrule
\end{tabular}
\end{table}

\begin{table}[t]
\centering
\caption{Median top 10 globally frequent features dominance by architecture across all SAE families.}
\label{tab:top10_dominance_by_architecture}
\begin{tabular}{lc}
\toprule
Architecture & Median Top 10 Dominance \\
\midrule
ReLU       & 0.102 \\
TopK       & 0.120 \\
BatchTopK  & 0.151 \\
PAnneal    & 0.155 \\
JumpReLU   & 0.045 \\
Gated   & 0.011 \\
Matryoshka & 0.177 \\
JBloom     & 0.046 \\
\bottomrule
\end{tabular}
\end{table}

\section{Dominant Features: Explanation Analysis}
\label{appendix:dominant_redundant_explanations}

This appendix provides a detailed analysis of the \emph{text explanations} of features, examining linguistic patterns, semantic content, and relationships between explanations for dominant features. Unlike quantitative metrics, explanation text reveals what language models identify as the semantic content of features, offering insights into interpretability patterns.

\subsection{Q1 Dominant Features: Explanation Patterns by DAME Score}

We analyze explanations for Q1 (most dominant) features, comparing those with high versus low DAME scores to understand what makes some dominant features more interpretable than others.

\subsubsection{Explanation Characteristics}

Table~\ref{tab:explanation_metrics} compares explanation metrics across categories using Q2--Q3 (medium dominance) as a baseline. While word counts are similar (22.1--23.9 words), there are important differences: Q1 low DAME features have the longest explanations (23.9 words, +1.3 vs baseline) despite the lowest DAME scores (0.425, -0.257 vs baseline). In contrast, Q4 high DAME features have the shortest explanations (22.1 words, -0.5 vs baseline) and highest DAME scores (0.862, +0.180 vs baseline).

\begin{table}[h]
\centering
\caption{Explanation Metrics by Category (Q2-Q3 as Baseline)}
\label{tab:explanation_metrics}
\begin{tabular}{lcccc}
\toprule
Category & Word Count & Specificity & Unique Ratio & DAME Score \\
\midrule
Q2-Q3 (Baseline) & 22.6 & 0.738 & 0.988 & 0.682 \\
Q1 High DAME & 22.3 & 0.737 & 0.992 & 0.828 \\
Q1 Low DAME & 23.9 & 0.747 & 0.969 & 0.425 \\
Q4 High DAME & 22.1 & 0.747 & 0.993 & 0.862 \\
Q4 Low DAME & 23.3 & 0.717 & 0.986 & 0.558 \\
\bottomrule
\end{tabular}
\end{table}

However, qualitative analysis reveals important differences in \emph{content} and \emph{semantic focus}. High DAME Q1 features tend to have explanations that describe clear, generalizable patterns:

\begin{itemize}
\item \textbf{High DAME Q1:} ``This feature in a sparse autoencoder detects patterns of commas separating elements in lists or sequences within text.'' (DAME: 1.0)
\item \textbf{High DAME Q1:} ``This feature detects patterns related to conjunctions, prepositions, and modal verbs, indicating transitions or connections between parts of a sentence.'' (DAME: 0.9)
\item \textbf{High DAME Q1:} ``This feature detects patterns of temporal sequencing and conjunctions in text, specifically focusing on tokens indicating the start, continuation, or completion of actions or events.'' (DAME: 1.0)
\end{itemize}

In contrast, low DAME Q1 features often focus on domain-specific or structural patterns that may be less semantically coherent:

\begin{itemize}
\item \textbf{Low DAME Q1:} ``This feature detects patterns related to XML tags and their attributes, specifically the presence of certain tags and attributes such as `\texttt{<datatype>}', `\texttt{IS}', `\texttt{minOccurs}', `\texttt{maxOccurs}'.'' (DAME: 0.0)
\item \textbf{Low DAME Q1:} ``This feature detects patterns related to empty or missing data tags in XML text, specifically the presence of `\texttt{</field>}' and `\texttt{minOccurs}' or `\texttt{maxOccurs}' with value `\texttt{0}'.'' (DAME: 0.2)
\end{itemize}

\subsubsection{Word Frequency Patterns}

\begin{figure}[h]
\centering
\includegraphics[width=0.9\textwidth]{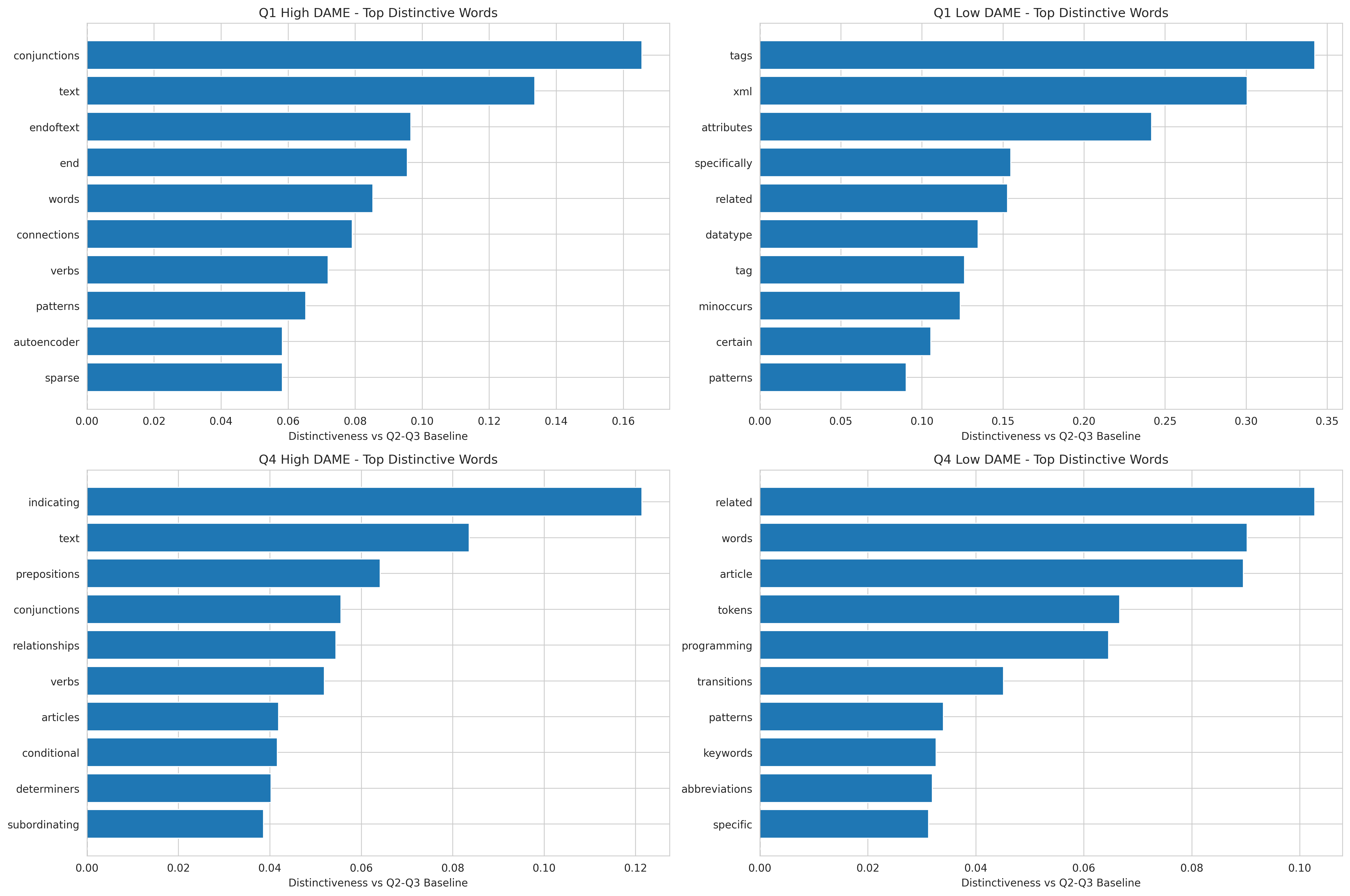}
\caption{Top distinctive words in explanations by category. Q1 high DAME explanations emphasize general linguistic structures (conjunctions, articles), while Q1 low DAME explanations focus on technical patterns (symbols, XML-related terms).}
\label{fig:explanation_word_frequencies}
\end{figure}

Figure~\ref{fig:explanation_word_frequencies} shows distinctive word usage patterns relative to the Q2--Q3 baseline. Q1 high DAME explanations more frequently mention \emph{conjunctions} (+16.6\% vs baseline), \emph{text} (+13.4\%), and \emph{endoftext} (+9.7\%), suggesting they focus on general linguistic structures and text boundaries. Q1 low DAME explanations more frequently mention \emph{XML tags} (+34.2\% vs baseline), \emph{attributes} (+24.2\%), and domain-specific terms like \emph{datatype} and \emph{minOccurs}, indicating a focus on technical markup patterns rather than semantic content. Q4 high DAME explanations emphasize grammatical elements (\emph{indicating}, \emph{prepositions}, \emph{conjunctions}), while Q4 low DAME explanations use more generic terms (\emph{related}, \emph{words}, \emph{tokens}).

\subsubsection{Explanation Complexity Patterns}

\begin{figure}[h]
\centering
\includegraphics[width=0.9\textwidth]{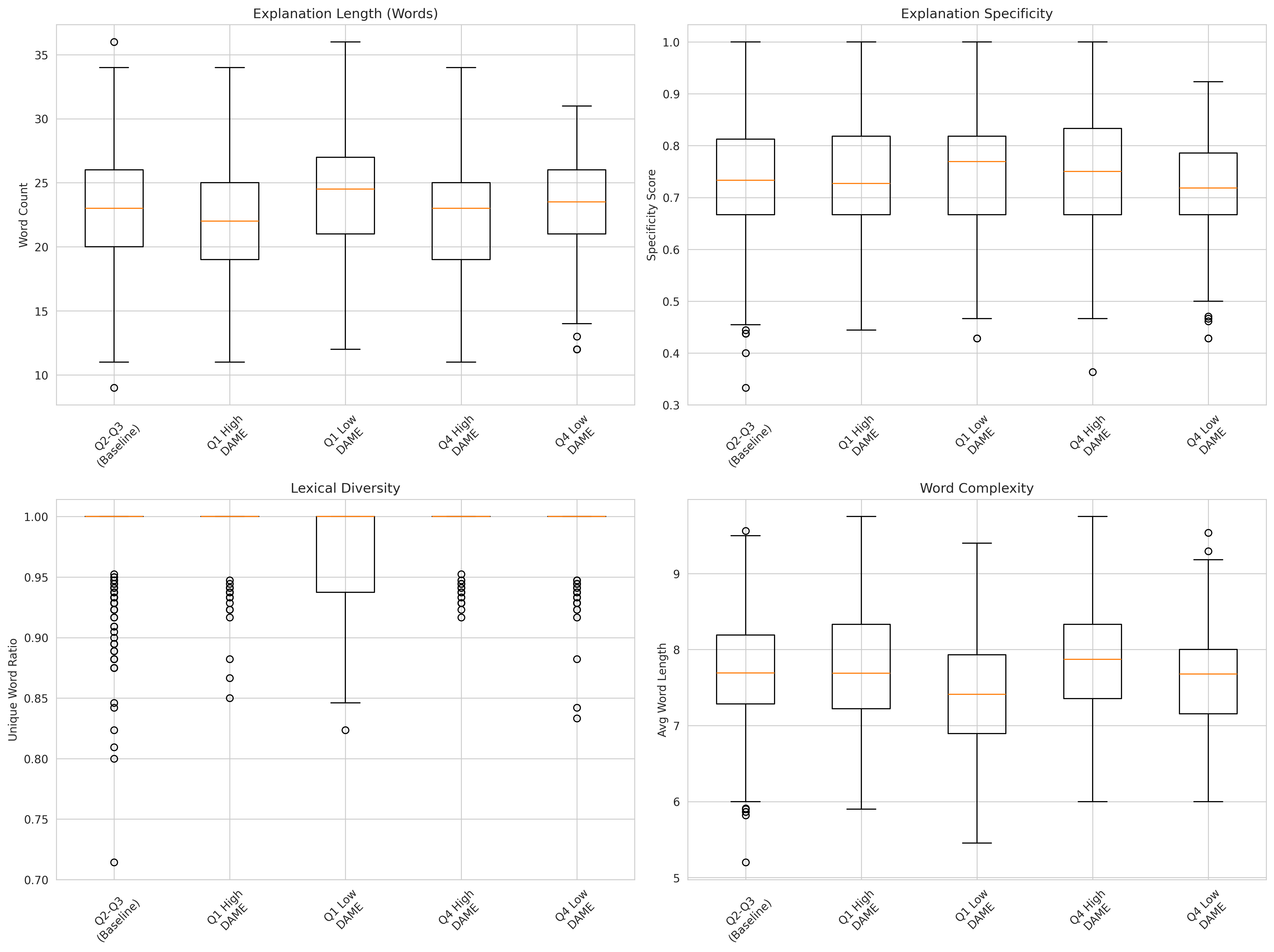}
\caption{Explanation complexity metrics by category. Metrics are similar across categories, but distributions reveal subtle differences in consistency.}
\label{fig:explanation_complexity_metrics}
\end{figure}

Figure~\ref{fig:explanation_complexity_metrics} visualizes explanation complexity metrics across categories. While mean values are similar, the distributions reveal subtle differences: Q1 high DAME features show slightly more consistent specificity scores, while Q1 low DAME features have more variable word counts, suggesting less coherent explanation patterns.

\subsection{Domain-Specific Patterns}

\begin{figure}[h]
\centering
\includegraphics[width=0.9\textwidth]{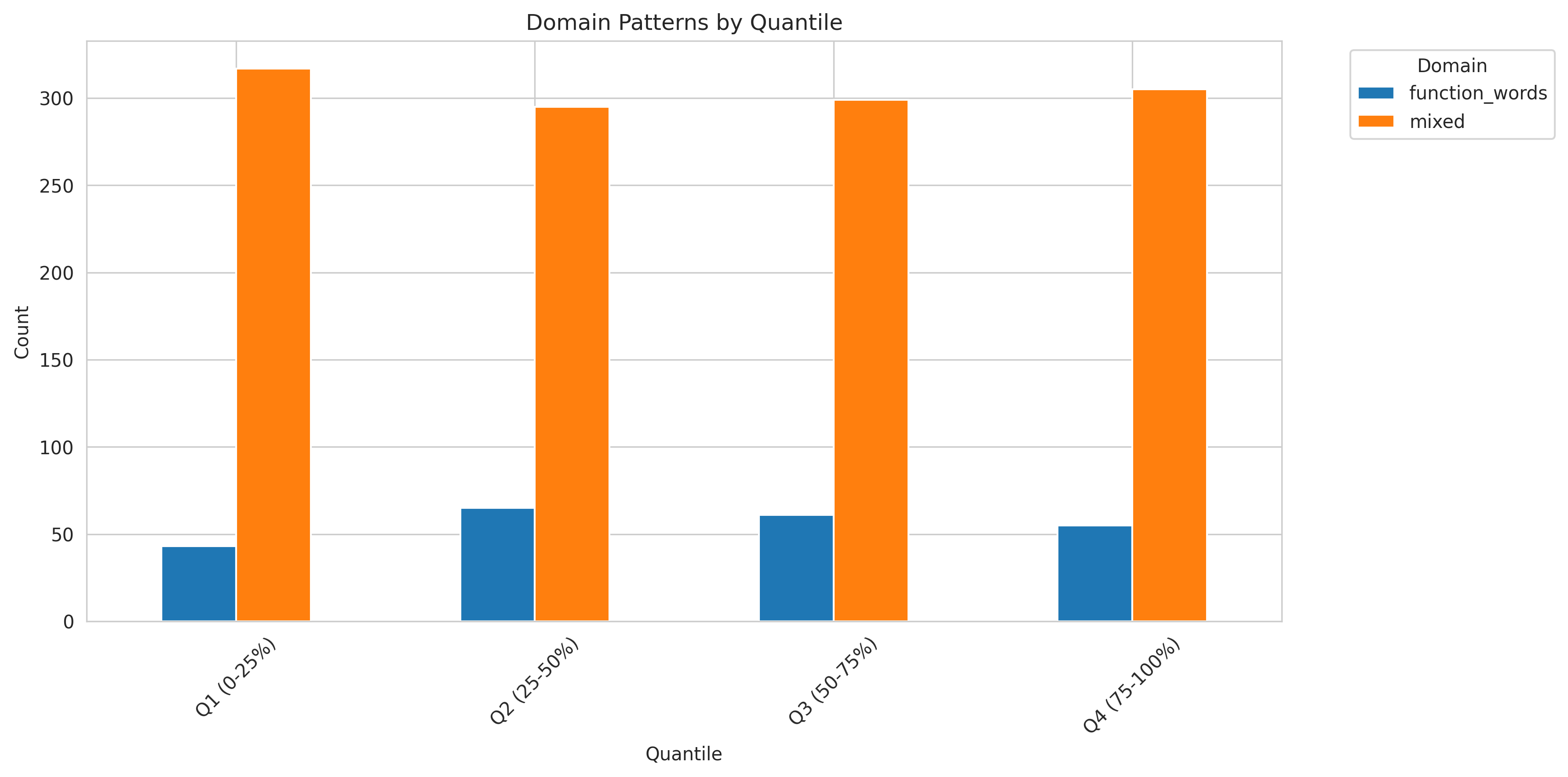}
\caption{Domain patterns by quantile. Q1 features show high proportion of mixed domains, combining multiple pattern types, which may contribute to interpretability challenges.}
\label{fig:domain_patterns_by_quantile}
\end{figure}

Figure~\ref{fig:domain_patterns_by_quantile} shows domain classification patterns across quantiles. Explanations are classified into domains based on keyword matching: \emph{XML} (tags, attributes, datatype), \emph{punctuation} (commas, marks, quotes), \emph{structure} (sentence, sequence, beginning, end), \emph{grammar} (conjunction, preposition, verb, noun), and \emph{function words} (common words like ``the'', ``and'', ``or'', ``of'', ``to'', ``in''). Explanations matching keywords from multiple domains are classified as \emph{mixed}. Q1 features show a high proportion of ``mixed'' domain classifications, combining XML, punctuation, structure, grammar, and function word patterns. This mixed nature may contribute to lower average DAME scores: features that combine multiple domain patterns may be harder to interpret coherently.

\subsection{Discussion}

Our explanation analysis reveals several important patterns:

\begin{enumerate}
\item \textbf{Explanation length/complexity does not predict interpretability:} Q1 high DAME, Q1 low DAME, and Q4 features have similar explanation metrics, suggesting that interpretability differences stem from \emph{content} rather than \emph{form}.

\item \textbf{High DAME Q1 features focus on general linguistic patterns:} Explanations emphasize conjunctions, articles, and general structural elements (commas, transitions), suggesting these features capture coherent, generalizable semantic patterns.

\item \textbf{Low DAME Q1 features focus on domain-specific patterns:} Explanations emphasize XML tags, technical markup, and domain-specific structures, suggesting these features capture less generalizable, context-dependent patterns.

\end{enumerate}

\section{SAE Geometry: Mean-Direction Clustering and Encoder–Decoder Mismatch}
\label{appendix:decoder-sim}

This appendix provides detailed geometric analysis supporting our focus on decoder cosine similarity rather than encoder similarity for evaluating non-redundancy. The key insight is that encoder weights function as gating mechanisms that must orient orthogonally to typical activations to satisfy sparsity constraints, while decoder weights define the semantic content written into the residual stream.

We present evidence from three complementary geometric analyses: (1) activation space clustering behavior, (2) encoder-decoder-input angular relationships, and (3) learned weight distributions relative to the activation mean direction. Together, these analyses demonstrate that encoder similarity is fundamentally unsuitable for measuring feature redundancy, while decoder cosine similarity captures the semantic relationships that matter for interpretability.

\subsection{Activation Space Clustering and Its Implications}

Transformer activations in high-dimensional spaces exhibit strong clustering behavior around a dominant mean direction. This geometric property has important implications for how encoder weights must be oriented to achieve sparsity.

\begin{figure}[h]
    \centering
    \includegraphics[width=0.9\linewidth]{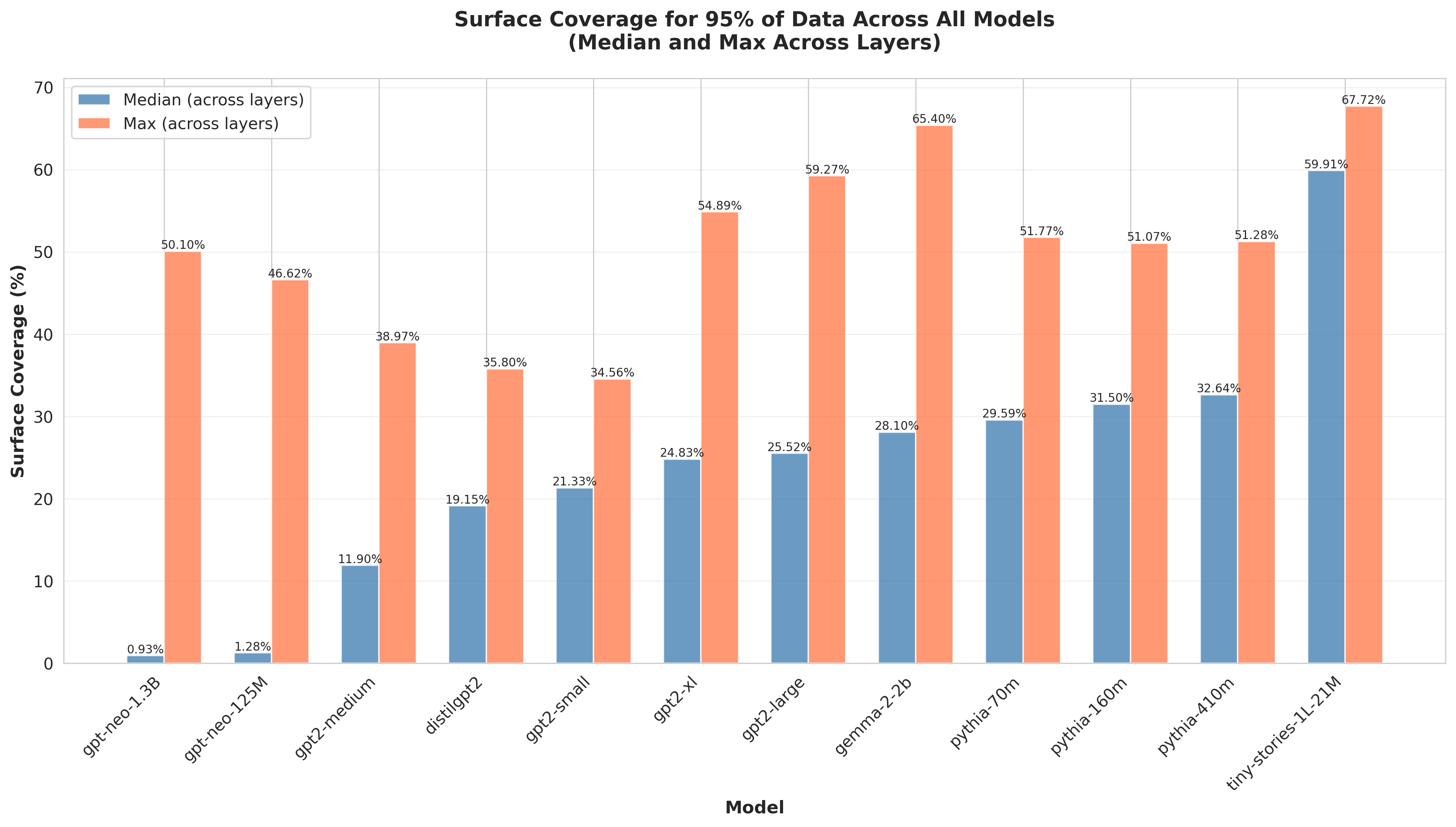}
    \caption{
    Surface coverage for 95\% of data across all evaluated models. The plot shows median and maximum surface coverage across layers for each model. For comparison, a uniform distribution on a sphere would require 95\% surface coverage to contain 95\% of the data. Actual activation data requires much smaller surface coverage (typically 1--33\%), demonstrating that activations cluster near a dominant mean direction rather than being uniformly distributed. This clustering behavior forces encoder weights to orient nearly orthogonally to the mean direction to achieve sparsity.
    }
    \label{fig:surface_coverage_95pct}
\end{figure}

The clustering phenomenon is consistent across all layers and models we evaluated. Figure~\ref{fig:surface_coverage_95pct} shows that even the maximum surface coverage across layers remains well below what would be expected from a uniform distribution. This geometric constraint is fundamental: if activations clustered uniformly on the sphere, encoder weights could point in any direction and still achieve sparsity. Instead, the strong clustering near a mean direction forces encoders to carve out narrow firing regions by orienting nearly orthogonally to this mean.

\begin{figure}[h]
    \centering
    \includegraphics[width=0.9\linewidth]{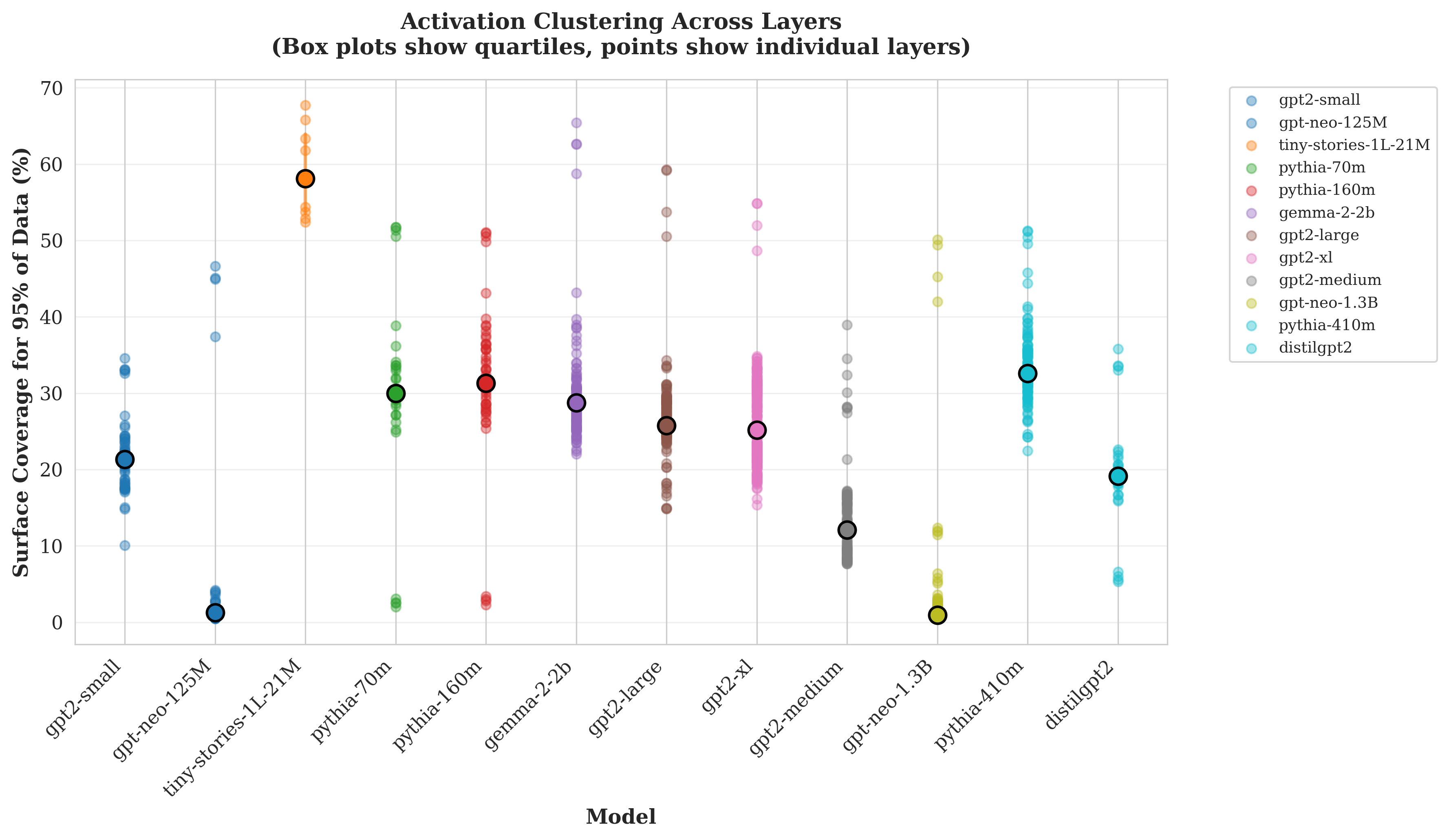}
    \caption{
    Distribution of surface coverage values across layers for each model. Each point represents one layer's surface coverage for 95\% of data. Box plots show quartiles, with median values marked. The consistent low surface coverage across all layers and models demonstrates that activation clustering is a universal property of transformer representations, not specific to particular layers or architectures. This geometric structure is what necessitates the orthogonal orientation of encoder weights.
    }
    \label{fig:analysis1_layer_evolution}
\end{figure}

Figure~\ref{fig:analysis1_layer_evolution} shows the distribution of surface coverage values across layers, revealing that the clustering behavior is consistent throughout the network depth. The median surface coverage remains low (typically 5--15\%) across all layers, with some variation but no systematic trend toward more uniform distributions in deeper layers. This geometric structure is universal across the models we evaluated, establishing that encoder weights must consistently orient orthogonally to the activation mean direction to satisfy sparsity constraints.

\subsection{Encoder-Decoder-Input Angular Relationships}

The clustering of activations near a mean direction creates systematic angular relationships between encoder weights, decoder weights, and input activations. Understanding these relationships is crucial for understanding why encoder similarity fails as a redundancy metric.

\begin{figure}[h]
    \centering
    \includegraphics[width=0.9\linewidth]{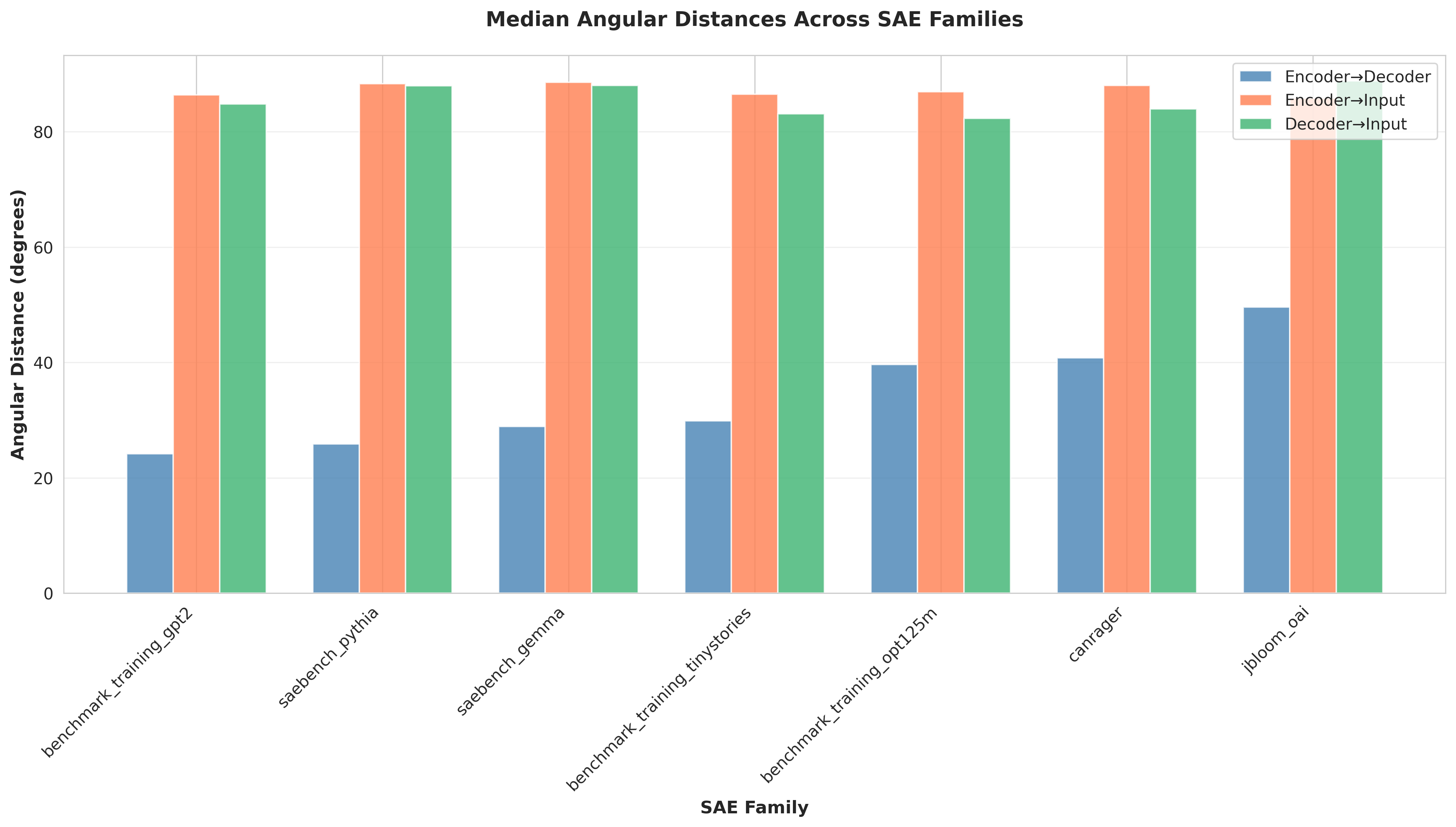}
    \caption{
    Median angular distances between encoder–decoder, encoder–input, and decoder–input vectors across all SAE families. Encoder–decoder distances range from 24--50$^\circ$ (median), showing systematic mismatch between encoder and decoder directions. Both encoder–input and decoder–input distances are large (82--88$^\circ$), demonstrating that both encoder and decoder vectors orient nearly orthogonally to typical input activations. The encoder–input distances are typically slightly larger than decoder–input distances, consistent with encoders needing to be more orthogonal to satisfy sparsity constraints.
    }
    \label{fig:angular_distances_aggregation}
\end{figure}

Figure~\ref{fig:angular_distances_aggregation} shows the three key angular relationships. The encoder–decoder mismatch (24--50$^\circ$) demonstrates that encoders and decoders serve fundamentally different roles: encoders act as gating mechanisms, while decoders specify semantic content. Both encoder–input and decoder–input distances are large (82--88$^\circ$), showing that both weight types must diverge from typical activation directions, but for different reasons: encoders must be orthogonal to achieve sparsity, while decoders encode semantic directions distinct from the mean.

\begin{figure}[h]
    \centering
    \includegraphics[width=0.9\linewidth]{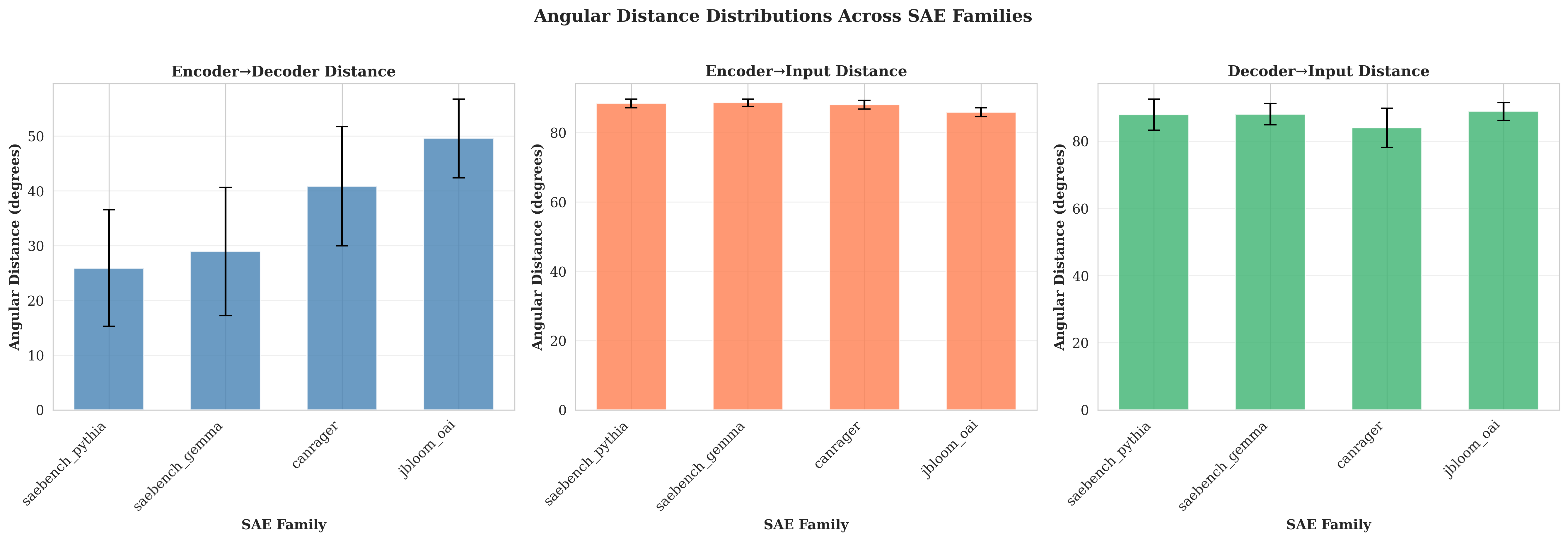}
    \caption{
    Distribution of angular distances across SAE families, showing encoder–decoder, encoder–input, and decoder–input distances separately. Error bars show standard deviation across SAEs within each family. The consistent pattern across families demonstrates that these geometric relationships are universal properties of SAE training, not artifacts of specific architectures or training procedures. Encoder–input distances are consistently larger than decoder–input distances, reflecting the stronger orthogonality requirement for encoders.
    }
    \label{fig:analysis2_family_comparison}
\end{figure}

Figure~\ref{fig:analysis2_family_comparison} breaks down these relationships by SAE family, revealing consistent patterns across different architectures and training procedures. The encoder–decoder mismatch varies by family (ranging from 25--50$^\circ$), but encoder–input and decoder–input distances remain consistently large across all families. This universality suggests these geometric constraints are fundamental to the SAE optimization problem, not implementation-specific artifacts.

\begin{figure}[h]
    \centering
    \includegraphics[width=0.7\linewidth]{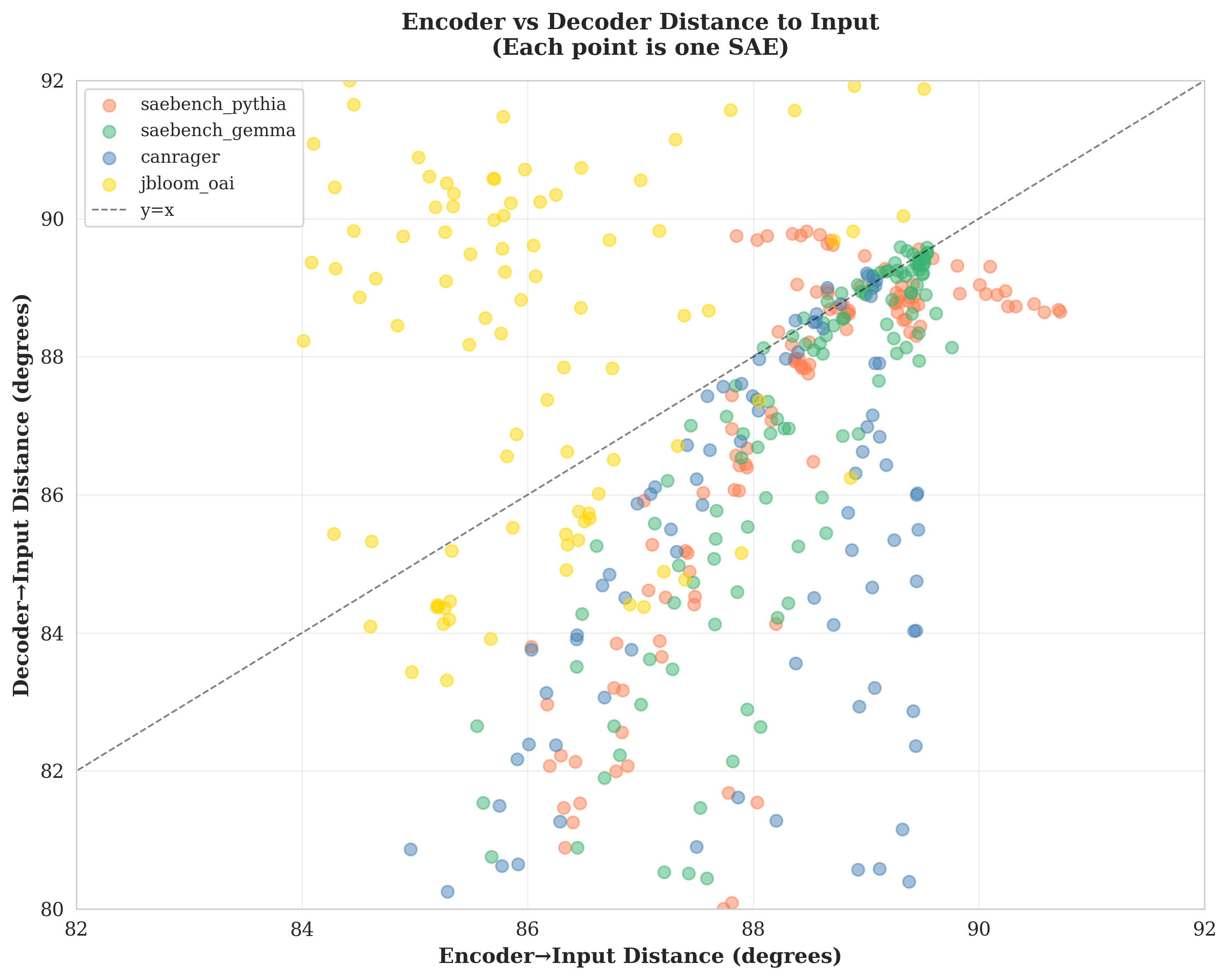}
    \caption{
    Scatter plot comparing encoder–input vs decoder–input angular distances for individual SAEs. Each point represents one SAE. Points above the diagonal (dashed line) indicate SAEs where encoder–input distance exceeds decoder–input distance. The majority of points lie above the diagonal, demonstrating that encoders are systematically more orthogonal to input activations than decoders. This pattern holds across all SAE families, with canrager showing the largest differences (5--6$^\circ$) and jbloom\_oai showing smaller but still consistent differences.
    }
    \label{fig:analysis2_enc_vs_dec_to_input}
\end{figure}

Figure~\ref{fig:analysis2_enc_vs_dec_to_input} directly compares encoder–input and decoder–input distances for individual SAEs. The majority of points lie above the diagonal, confirming that encoders are systematically more orthogonal to input activations than decoders. This pattern is strongest in canrager SAEs (median difference 5--6$^\circ$) but present across all families. This systematic difference reflects the different roles: encoders must be more orthogonal to satisfy sparsity constraints, while decoders can be slightly less orthogonal because they encode semantic content rather than gating decisions.

The key insight from this analysis is that encoder–input and decoder–input mismatches are both large and natural, but they serve different purposes. The encoder–input mismatch is a \emph{constraint} imposed by sparsity requirements, while the decoder–input mismatch reflects the semantic diversity of features. This distinction explains why encoder similarity cannot capture redundancy: two features with similar encoders may still encode different concepts if their decoders differ, while two features with different encoders but similar decoders are genuinely redundant.

\subsection{Learned Weight Distributions Relative to Activation Mean}

The clustering of activations near a mean direction causes both encoder and decoder weights to diverge substantially from this mean direction, but for different reasons.

\begin{figure}[h]
    \centering
    \includegraphics[width=0.9\linewidth]{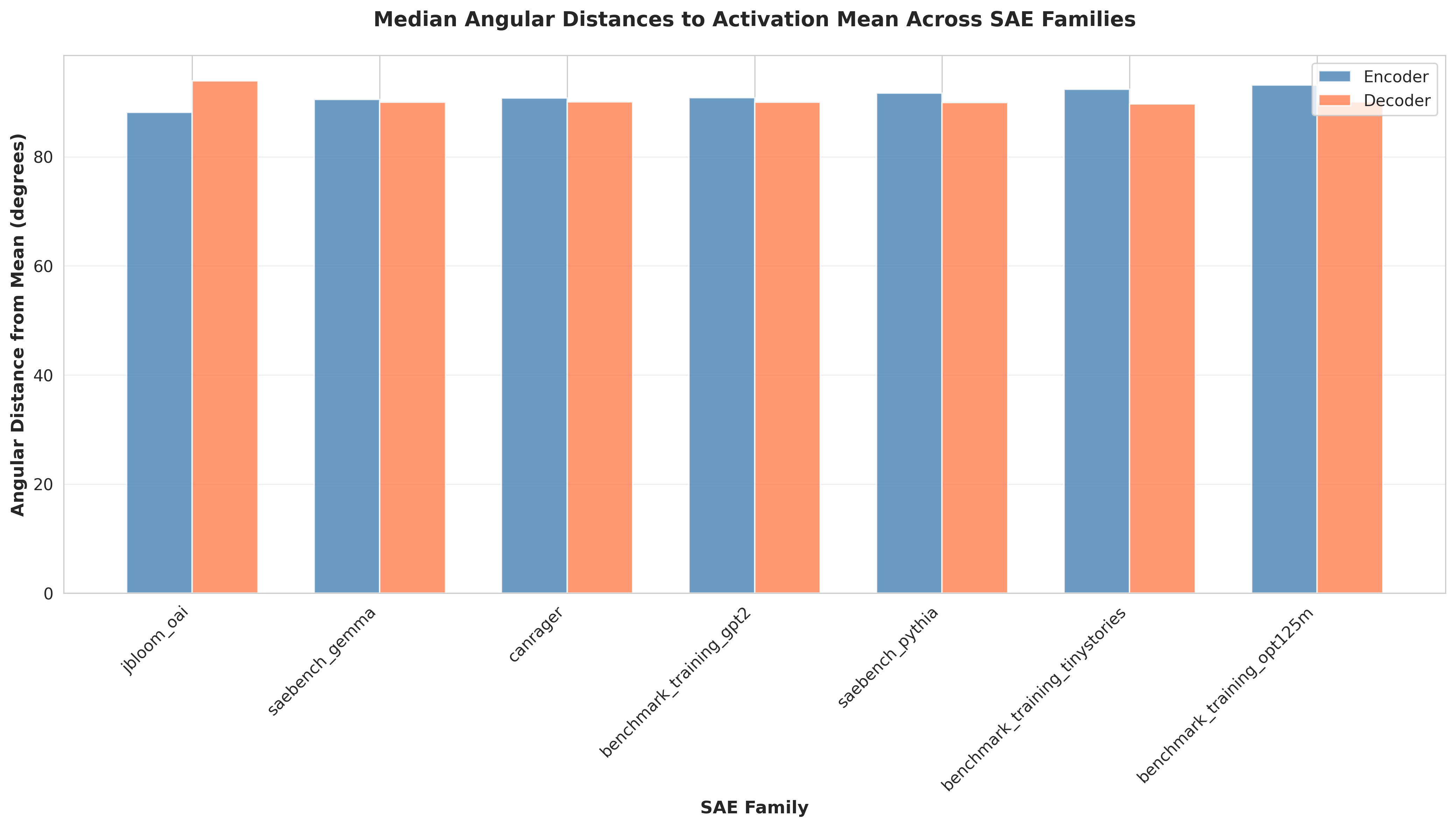}
    \caption{
    Median angular distances of encoder and decoder weights from the activation mean direction across all SAE families. Encoder weights are nearly orthogonal to the mean direction (median angles of 88--93$^\circ$), demonstrating that they must orient orthogonally to satisfy sparsity constraints. Decoder weights also diverge substantially (median angles of 90--94$^\circ$), but this divergence serves a different purpose: encoding distinct semantic directions that are not aligned with the mean activation direction. This geometric separation between encoder/decoder weights and the activation mean is natural and necessary for SAEs to function effectively.
    }
    \label{fig:mean_direction_divergence}
\end{figure}

Figure~\ref{fig:mean_direction_divergence} shows that both encoder and decoder weights diverge substantially from the activation mean direction, with median angles of 88--93$^\circ$ for encoders and 90--94$^\circ$ for decoders. This near-orthogonality is necessary: if weights were aligned with the mean direction, features would activate on nearly all inputs, violating sparsity constraints. However, the reasons for divergence differ: encoders must be orthogonal to achieve sparsity, while decoders encode semantic directions distinct from the mean.

\begin{figure}[h]
    \centering
    \includegraphics[width=0.9\linewidth]{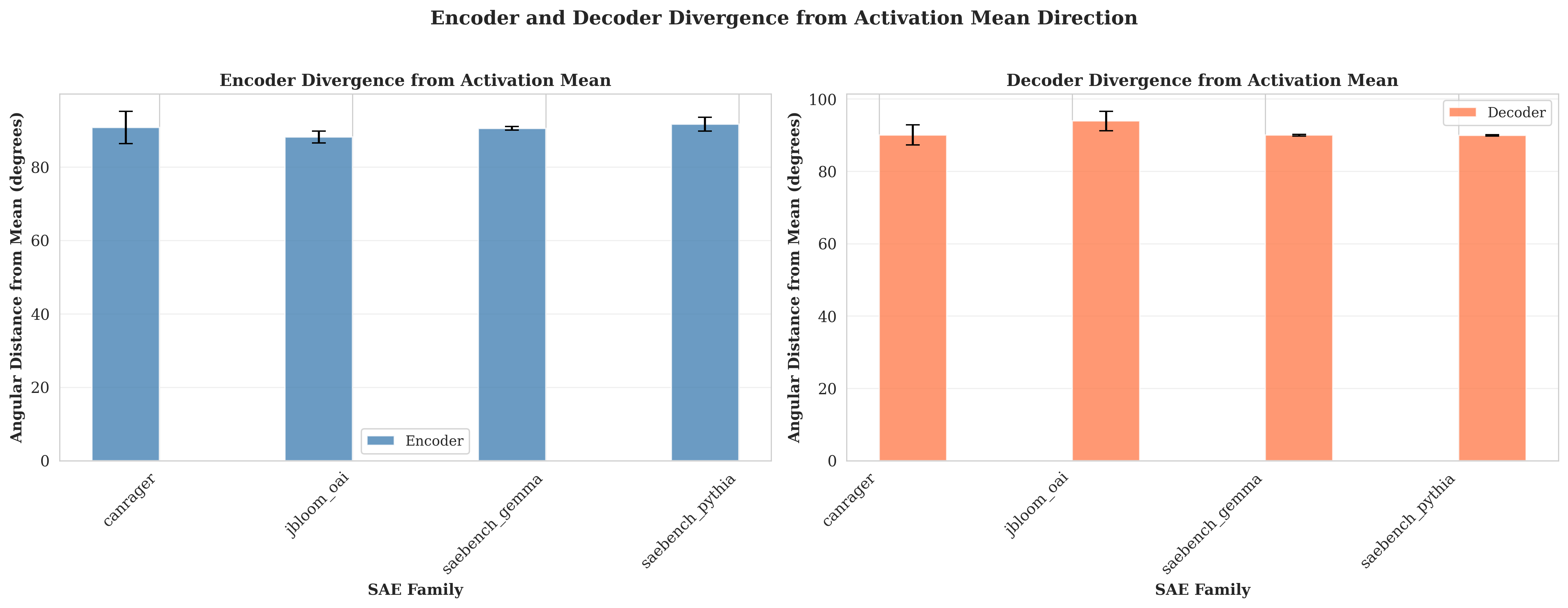}
    \caption{
    Distribution of encoder and decoder divergence from activation mean across SAE families. Error bars show standard deviation across SAEs within each family. Both encoder and decoder weights show consistent near-orthogonality to the mean direction across all families, with encoders typically slightly more orthogonal than decoders. This pattern demonstrates that the geometric constraints identified in the activation clustering and angular relationship analyses manifest consistently in the learned weight distributions.
    }
    \label{fig:analysis5_family_distribution}
\end{figure}

Figure~\ref{fig:analysis5_family_distribution} shows the distribution of divergence values across SAE families. Both encoder and decoder weights consistently diverge from the mean direction, with median angles typically in the 88--94$^\circ$ range. The slight difference between encoder and decoder divergence (encoders typically 1--2$^\circ$ more orthogonal) reflects the stronger orthogonality requirement for encoders to satisfy sparsity constraints. This pattern is consistent across all SAE families, demonstrating that these geometric properties are universal outcomes of SAE training.

\begin{figure}[h]
    \centering
    \includegraphics[width=0.7\linewidth]{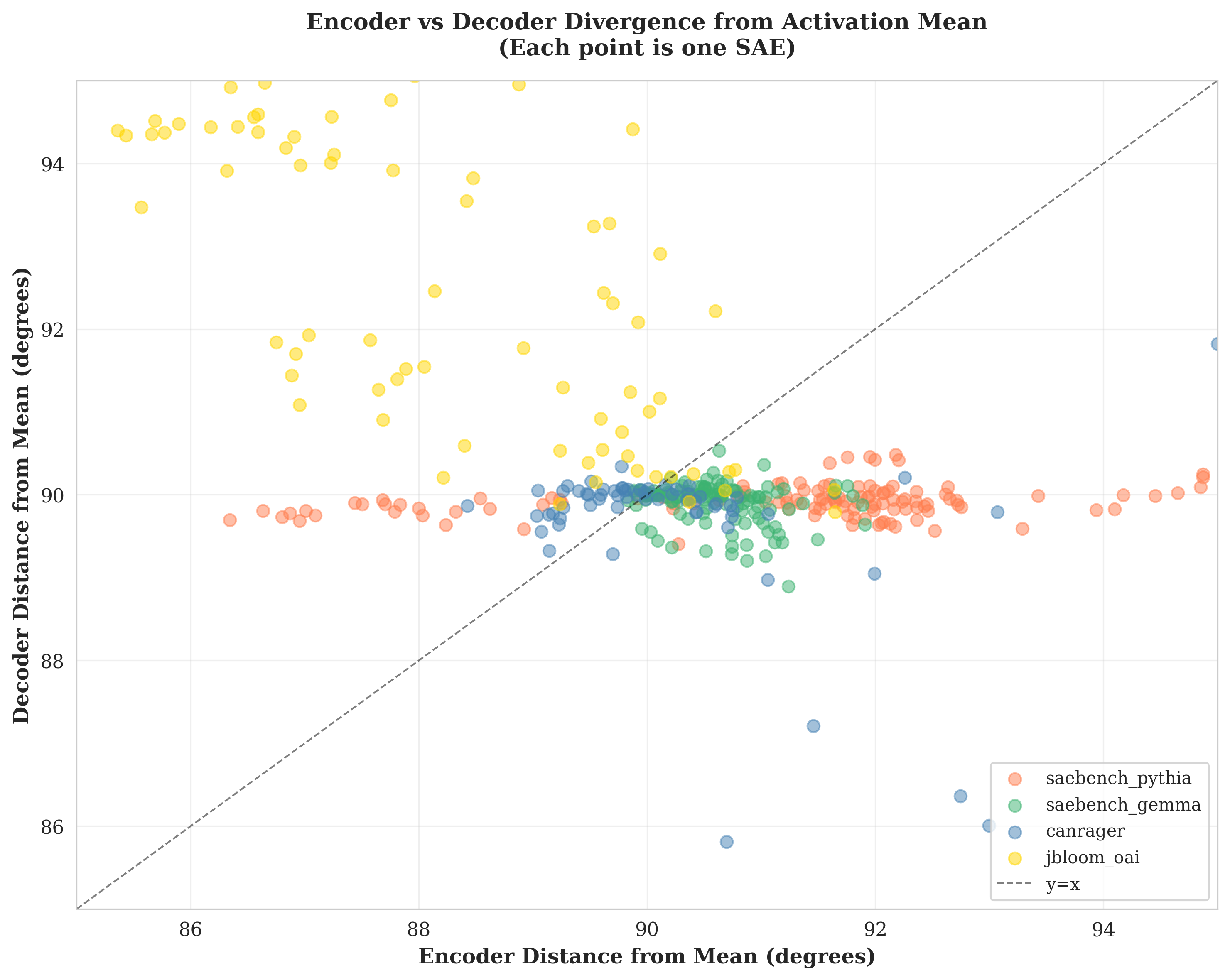}
    \caption{
    Scatter plot comparing encoder vs decoder divergence from activation mean for individual SAEs. Each point represents one SAE. Points above the diagonal (dashed line) indicate SAEs where encoder divergence exceeds decoder divergence. The majority of points cluster near the diagonal but slightly above it, showing that encoders are consistently slightly more orthogonal to the mean than decoders. This systematic difference reflects the different roles: encoders must satisfy sparsity constraints (requiring stronger orthogonality), while decoders encode semantic content (allowing slightly less orthogonality).
    }
    \label{fig:analysis5_enc_vs_dec_divergence}
\end{figure}

Figure~\ref{fig:analysis5_enc_vs_dec_divergence} directly compares encoder and decoder divergence for individual SAEs. The points cluster near the diagonal but with a slight upward bias, confirming that encoders are consistently slightly more orthogonal to the mean than decoders. This systematic difference is small (typically 1--2$^\circ$) but consistent across SAE families, reflecting the different optimization pressures: encoders must satisfy sparsity constraints, while decoders must encode diverse semantic directions.

\subsection{Synthesis: Why Decoder Cosine Similarity, Not Encoder Similarity}

The three analyses presented above provide complementary evidence for why decoder cosine similarity, not encoder similarity, is the appropriate metric for evaluating feature redundancy:

\textbf{From the activation clustering analysis:} Activation clustering forces encoder weights to orient orthogonally to the mean direction. This geometric constraint means that encoder directions are primarily determined by sparsity requirements, not semantic content. Two features with similar encoders may have similar firing patterns, but this similarity reflects shared sparsity constraints rather than shared semantic content.

\textbf{From the angular relationship analysis:} Encoder–decoder mismatch (24--50$^\circ$) demonstrates that encoders and decoders serve fundamentally different roles. Encoder–input and decoder–input distances are both large (82--88$^\circ$), but for different reasons: encoders must be orthogonal to satisfy sparsity, while decoders encode semantic directions. The systematic difference (encoders typically 1--6$^\circ$ more orthogonal than decoders) reflects the stronger orthogonality requirement for encoders.

\textbf{From the weight distribution analysis:} Both encoder and decoder weights diverge substantially from the activation mean (88--94$^\circ$), but encoders are consistently slightly more orthogonal than decoders. This pattern reflects the different optimization pressures: encoders must satisfy sparsity constraints, while decoders must encode diverse semantic directions.

Together, these analyses establish that:
\begin{enumerate}
    \item Encoder directions are primarily constrained by sparsity requirements, not semantic content.
    \item Encoder–decoder mismatch is natural and necessary, reflecting their different roles.
    \item Encoder similarity cannot distinguish between features that share sparsity constraints versus features that share semantic content.
    \item Decoder cosine similarity directly measures semantic redundancy, which is what matters for interpretability.
\end{enumerate}

This geometric understanding supports the formal argument in Section~\ref{sec:criteria}: encoder non-redundancy does not imply decoder non-redundancy (two features can have different encoders but similar decoders, leading to semantic redundancy), while decoder non-redundancy implies encoder non-redundancy (two features with different decoders cannot have identical encoders in an optimal SAE). Therefore, decoder cosine similarity is the appropriate metric for evaluating feature redundancy.

\clearpage
\section{Additional Validation and Robustness Analyses}
\label{appendix:additional-validation}

\subsection{Controlled Comparison of Per-Feature Scoring Rules}
\label{appendix:scorer-comparison}

We compare per-feature scoring rules in a controlled synthetic setting in which the generator, rather than any evaluator, fixes which explanation is correct. For each of 80 independently generated synthetic features (40 lexical-strength and 40 contextual-strength features), the generator defines a three-level activation rule, $H > M > Z$. For every feature we construct a matched $E_{\text{exact}}$ explanation that states this relationship and an $E_{\text{reverse}}$ explanation that preserves the trigger, active support, explanation frame, and comparable length, but reverses only the high-versus-middle relationship ($M > H > Z$). Generator truth therefore fixes $E_{\text{exact}} > E_{\text{reverse}}$ before any scorer is queried.

This isolates the case in which two scenes have the same binary support but differ in activation strength: both explanations identify the high and middle scenes as active and the zero scene as inactive, but they disagree about which active scene has greater activation. Every scorer receives the same feature, explanation, and held-out evidence. Each feature is evaluated on 18 permutation-balanced held-out $H/M/Z$ triplets, so the scene types do not occupy fixed display positions. Methods are compared only through a common within-feature endpoint: whether the method assigns a higher native held-out score to $E_{\text{exact}}$ than to $E_{\text{reverse}}$; raw scores are not compared across output spaces. The experiment comprises 14,400 requests, all successful, and the inferential unit is the synthetic feature ($N = 80$), not an individual request.

\begin{figure}[htbp]
\centering
\includegraphics[width=\linewidth]{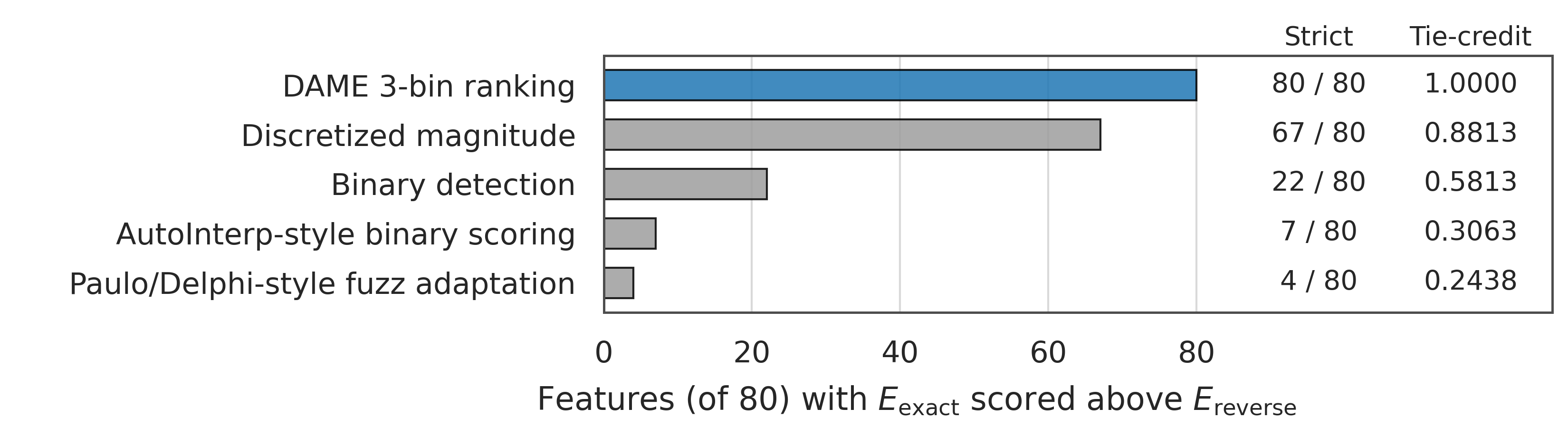}
\caption{Controlled comparison of per-feature scoring rules on 80 synthetic features with generator-fixed ground truth. Bars and the ``Strict'' column give the number of features whose correct explanation is scored strictly above the reversed one; the tie-credit rate counts a tie as one half.}
\label{fig:scorer_comparison}
\end{figure}

Figure~\ref{fig:scorer_comparison} shows that DAME ranks the correct explanation above the reversed one for all 80 features, compared with 67 for discretized magnitude and at most 22 for the binary and fuzz-style methods. Under a family-stratified feature bootstrap, DAME's tie-credit advantage over discretized magnitude is $+0.11875$ (95\% interval $[+0.0625, +0.18125]$); its advantages over binary detection, AutoInterp-style binary scoring, and the fuzz adaptation are $+0.41875$, $+0.69375$, and $+0.75625$, respectively, with all intervals excluding zero. The fuzz comparator is our synthetic marked-cue adaptation inspired by \citet{paulo2024automatically}. The comparison shows the benefit of preserving high-versus-middle activation information rather than collapsing all nonzero activations into one binary class.

\subsection{DAME and Binary AutoInterp on the Same Features}
\label{appendix:dame-vs-autointerp}

Prior AutoInterp-style binary assays evaluate features individually and summarize them feature-uniformly; without conditioning on dominance, such an aggregate does not ask whether its errors are concentrated in the small subset of features that carries most of the representation. We test this on two SAEs from the main benchmark with a DAME Q1 deficit, one Gemma and one Pythia, by applying a prior-style binary marked-token AutoInterp assay to the same features. Rather than comparing the two assays' raw scores, we ask whether each reproduces the Q1-versus-Q2--Q4 gap. DAME shows Q1 deficits of $-11.83$ pp and $-9.50$ pp, whereas the binary assay shows $+0.13$ pp and $+0.35$ pp. On these two pre-selected SAEs, the dominance-conditioned gap exposed by DAME is not reproduced by the feature-uniform binary assay.

\subsection{Explainer and Judge Sensitivity}
\label{appendix:cross-judge}

We cross two explainers with two judges on six SAEs from the main benchmark: the original Mistral-7B explanations or explanations regenerated with GPT-5.6 Sol, each judged by Mistral-7B or by GPT-5.6 Sol. Table~\ref{tab:cross_judge} reports the mean difference in perfect ranking accuracy between Q1 and Q2--Q4 over the six SAEs and the number of SAEs with a Q1 deficit. The Q1 deficit appears in every cell.

\begin{table}[h]
\centering
\caption{Explainer-by-judge check on six SAEs.}
\label{tab:cross_judge}
\small
\begin{tabular}{llcc}
\toprule
Explainer & Judge & Mean Q1 $-$ Q2--Q4 (pp) & SAEs with a Q1 deficit\\
\midrule
Original Mistral-7B & Mistral-7B & $-11.26$ & 6/6\\
GPT-5.6 Sol & Mistral-7B & $-3.25$ & 6/6\\
Original Mistral-7B & GPT-5.6 Sol & $-4.04$ & 5/6\\
GPT-5.6 Sol & GPT-5.6 Sol & $-18.84$ & 6/6\\
\bottomrule
\end{tabular}
\end{table}

\subsection{Explanation Length Budget}
\label{appendix:length-budget}

To test whether the 15-word explanation instruction drives the Q1 deficit, we reran a fixed panel of 28 SAEs (2,240 features) under matched explanation instructions of 15 words / 40 tokens, 30 words / 80 tokens, and 60 words / 160 tokens, where the first quantity is the word budget stated in the prompt and the second is the generation-token cap (Table~\ref{tab:length_budget}). Realized explanation lengths increased from 23.02 to 29.27 to 35.45 words, while the Q1-minus-Q2--Q4 contrast in token-based perfect ranking accuracy remained negative at $-4.53$, $-4.05$, and $-4.78$ pp, with Q1 lower in 21, 20, and 23 of the 28 SAEs. The respective 95\% hierarchical-bootstrap intervals for these token-accuracy contrasts are $[-7.31,-1.80]$, $[-7.01,-1.14]$, and $[-7.90,-1.79]$ pp, resampling SAEs and then features within each quartile. Relative to the 15-word instruction, the Q1-specific changes were $+0.48$ pp at 30 words (95\% CI $[-1.44, +2.38]$) and $-0.25$ pp at 60 words ($[-2.53, +1.94]$), giving no evidence of Q1-specific recovery with additional budget. Overall strict ranking accuracy rose from 57.32\% to 58.94\% to 60.18\%, a modest general benefit rather than a Q1-only gain.

\begin{table}[h]
\centering
\caption{Explanation length budget on a fixed panel of 28 SAEs (2,240 features).}
\label{tab:length_budget}
\footnotesize
\begin{tabular}{lccccc}
\toprule
Explanation & Q1 $-$ Q2--Q4 & SAEs with & Q1 mean & Q2--Q4 mean & Q1 $-$ Q2--Q4 length\\
instruction & accuracy (pp) & lower Q1 & words & words & (paired $p$)\\
\midrule
15 words / 40 tokens & $-4.53$ & 21/28 & 23.63 & 22.82 & $+0.81$ (.00049)\\
30 words / 80 tokens & $-4.05$ & 20/28 & 30.07 & 29.00 & $+1.07$ (.00064)\\
60 words / 160 tokens & $-4.78$ & 23/28 & 35.80 & 35.33 & $+0.47$ (.443)\\
\bottomrule
\end{tabular}
\end{table}

\subsection{AND/OR Alignment in the Synthetic Benchmark}
\label{appendix:and-or}

The correlation-gap statistic of Appendix~\ref{appendix:synthetic-validation} does not distinguish conjunctive (AND) from disjunctive (OR) features. We therefore ran a held-out analysis over all 105 synthetic SAEs. The evaluation set is split in half; for each feature, the first half identifies its most-correlated candidate in each of three families (a single ground-truth variable, an AND target $\min(c_a, c_b)$, and an OR target $\max(c_a, c_b)$), and Table~\ref{tab:and_or} reports the selected target's correlation on the independent second half. Features are more OR-aligned than single-variable- or AND-aligned, and the Q1-minus-Q2--Q4 separation is largest for OR ($+.253$, versus $+.174$ for single variables and $+.075$ for AND), so the dominance result does not reflect preferential conjunctive composition.

The highest single-variable correlation measures best-match alignment rather than monosemanticity: it ignores a feature's other associations, and because correlation is symmetric, a feature confined to a coherent subset of a concept can correlate weakly with that concept despite never firing outside it. The synthetic benchmark contains no meaningful redundancy or sharding population: across the 105 SAEs, only 81 of 18,574,976 alive within-SAE feature pairs (0.00044\%) have decoder cosine similarity $\ge 0.70$. We therefore use it only for monosemanticity calibration.

\begin{table}[h]
\centering
\caption{Held-out correlation with the selected single-variable, AND, and OR targets on 105 synthetic SAEs.}
\label{tab:and_or}
\small
\begin{tabular}{lccc}
\toprule
Group & Highest single-variable correlation & AND & OR\\
\midrule
Q1 & .447 & .143 & .616\\
Q2--Q4 & .273 & .067 & .363\\
Q1 $-$ Q2--Q4 & $+.174$ & $+.075$ & $+.253$\\
\bottomrule
\end{tabular}
\end{table}

\subsection{Held-Out Support Containment}
\label{appendix:support-containment}

We also evaluate held-out support containment on the synthetic benchmark: when a feature fires (its encoded activation is positive), is the same ground-truth variable consistently present? For a ground-truth variable $c$, let $r_c$ denote raw containment, the number of examples on which the feature fires and $c$ is active divided by the number on which the feature fires, and let $p_c$ denote the fraction of examples in which $c$ is active. The prevalence-corrected score is $(r_c - p_c)/(1 - p_c)$. For each feature we select the variable with the highest corrected score on 25,000 discovery examples and report its score on 25,000 disjoint held-out examples. Under this score, each firing without the selected variable lowers the score; a feature that fires only on a coherent subset of a concept, and never outside it, scores one; and a feature that fires independently of $c$ scores zero in expectation.

We evaluate 210 SAEs: the 105 of Appendix~\ref{appendix:synthetic-validation} at widths 256, 512, and 1024, and 105 matched SAEs trained at widths 128, 192, and 768, with 35 SAEs at each width. Table~\ref{tab:support_containment} reports the mean held-out score over all alive features. Across the 210 paired SAE-level contrasts, Q1 has lower containment than Q2--Q4 (one-sided exact sign test, $p = 1.6 \times 10^{-12}$).

\begin{table}[h]
\centering
\caption{Prevalence-corrected held-out support containment by SAE width (35 SAEs per width).}
\label{tab:support_containment}
\small
\begin{tabular}{lccc}
\toprule
SAE width & Q1 & Q2--Q4 & Q1 $-$ Q2--Q4\\
\midrule
128 & .094 & .104 & $-.010$\\
192 & .088 & .105 & $-.018$\\
256 & .087 & .106 & $-.019$\\
512 & .084 & .108 & $-.023$\\
768 & .082 & .106 & $-.024$\\
1024 & .085 & .104 & $-.019$\\
\bottomrule
\end{tabular}
\end{table}

\subsection{Frequency-Matched Control}
\label{appendix:frequency-matched}

To separate dominance from firing frequency, within each SAE we match every Q1 feature, without replacement, to the nearest Q2--Q4 feature in firing-frequency rank, where the rank is each feature's percentile among alive features in the same SAE. A rank caliper of $c$ requires an absolute percentile-rank difference of at most $c$ between matched features (for example, 0.05 permits a difference of at most five percentile points); a Q1 feature is omitted only when no unused Q2--Q4 feature falls within the caliper, and an SAE enters a matched analysis only if it retains at least 10 pairs. Table~\ref{tab:frequency_matched} reports strict (0.05), primary (0.10), and loose (0.20) calipers. Thus, across the full range of matching thresholds, firing frequency alone does not explain the observed token-DAME gap. This analysis matches firing frequency, not feature type, and therefore does not rule out differences in syntactic, positional, or semantic content across dominance quantiles.

\begin{table}[h]
\centering
\caption{Frequency-matched comparison of Q1 features with Q2--Q4 control features.}
\label{tab:frequency_matched}
\footnotesize
\begin{tabular}{lccc}
\toprule
Comparison & Matched Q1--control & Firing-frequency standardized & Q1 $-$ control accuracy\\
 & pairs retained & mean difference & (pp, 95\% CI)\\
\midrule
Unmatched Q1 versus Q2--Q4 & --- & 0.715 & $-1.59$ $(-2.30, -0.89)$\\
Matched, 0.05 rank caliper & 55.2\% & 0.034 & $-2.45$ $(-3.67, -1.24)$\\
Matched, 0.10 rank caliper & 69.6\% & 0.080 & $-2.07$ $(-3.07, -1.07)$\\
Matched, 0.20 rank caliper & 86.2\% & 0.165 & $-2.00$ $(-2.85, -1.16)$\\
\bottomrule
\end{tabular}
\end{table}

\subsection{Sharding-Compatible Pairs at Low $L_0$}
\label{appendix:low-l0-sharding}

We search for sharding-compatible pairs in a held-out panel covering all SAE families, using three conditions: aligned decoder directions (cosine $\ge 0.70$), separated encoder directions (cosine $\le 0.40$ and decoder-minus-encoder cosine $\ge 0.45$), and complementary activation support (the two features share at most 10\% of their union of firing positions and at most 20\% of either feature's own firing positions). Table~\ref{tab:low_l0_sharding} reports, by held-out mean $L_0$, the percentage of decoder-redundant pairs that meet these conditions and the percentage of SAEs with at least one such pair; higher values indicate greater prevalence, not severity. Sharding-compatible pairs occur below $L_0 = 50$, so high $L_0$ is not necessary for this pattern.

\begin{table}[h]
\centering
\caption{Sharding-compatible pairs by held-out mean $L_0$.}
\label{tab:low_l0_sharding}
\small
\begin{tabular}{lcc}
\toprule
Held-out mean $L_0$ & Sharding-compatible pair rate & SAEs with at least one such pair\\
\midrule
$<50$ & 2.6\% & 56.3\%\\
50--200 & 3.8\% & 42.9\%\\
$\ge 200$ & 3.3\% & 30.4\%\\
\bottomrule
\end{tabular}
\end{table}

\subsection{Functional Sharding Case: Intervention Protocol}
\label{appendix:functional-sharding-protocol}

\paragraph{Model, SAE, and natural support.}
The case in Section~\ref{subsec:redundancy} uses features 41902 and 46571 of a 65,536-feature TopK SAE ($k=80$, checkpoint \texttt{trainer\_2}) on the layer-8 residual stream of \texttt{EleutherAI/pythia-160m-deduped}, revision \texttt{step143000}. Model inference uses float16, and SAE encoding uses float32. Their decoder and encoder cosine similarities are 0.7405 and 0.2066, respectively. The natural-support measurement uses 192 held-out WikiText-103 validation texts, truncated to 128 tokens. Firing is determined by the SAE's exact positive TopK output, rather than encoder preactivations; support overlap is measured at shared token positions. The intervention confirmation below uses separate control-calibration texts and a fixed, hand-authored prompt family.

\paragraph{Matched controls.}
We calibrate support on WikiText-103 test: strip texts, retain those with at least 160 characters, shuffle with seed 20260726, and take the first 96. We tokenize in batches of eight with right padding and truncation to 128 tokens, excluding padding positions from support counts. For each candidate $i$, we select a different, active feature $c$ with $|\cos(d_i,d_c)|\leq0.10$, minimizing
$[\log(1+s_c)-\log(1+s_i)]^2+[\log\|d_c\|_2-\log\|d_i\|_2]^2$,
where $s$ is the exact TopK support count. Candidate features are excluded from the control pool, and the second control must differ from the first. The selected controls are 12364 and 1537; each matches its candidate's support count and approximately unit decoder norm, with candidate--control cosine similarities 0.0252 and 0.0834.

\paragraph{Prompts, gate check, and intervention.}
Table~\ref{tab:functional_sharding_prompts} gives the complete prompt family, fixed after exploratory discovery of the target. The target is the single next token \texttt{aim} with a leading space (token ID 4388). At the last non-padding position, we retain a prompt only if both candidate features have zero natural TopK activity; the two excluded prompts are marked in the table. For each eligible prompt, we independently add each candidate or control decoder direction to the layer-8 residual stream after the block, then run the actual downstream model. Each addition has the same positive $L_2$ magnitude,
$0.05\,\operatorname{median}_p\|h_{8,p}\|_2=1.2307$,
where the median is over eligible prompts' unmodified final-position residuals. Thus, the edit for feature $i$ is $h_{8,p}\leftarrow h_{8,p}+1.2307\,d_i/\|d_i\|_2$; all other positions are unchanged.

\begin{table}[h]
\centering
\caption{Complete fixed target-prompt family. $\dagger$ marks prompts excluded because feature 46571 was naturally active at the intervention position.}
\label{tab:functional_sharding_prompts}
\small
\begin{tabular}{ll}
\toprule
The committee's primary & His ultimate\\
The organization has one clear & The campaign's central\\
The proposal's central & The department's stated\\
The team's shared$^\dagger$ & The paper's broader\\
The research program has a single & The initiative's guiding\\
Her stated & Our main\\
The policy's underlying$^\dagger$ & The reform's practical\\
The strategy's main & The new plan has one\\
The report clarifies the & The committee articulated its\\
Every project needs a clear & The program's original\\
\bottomrule
\end{tabular}
\end{table}

\paragraph{Readout and scope.}
For each prompt and edit, the outcome is the target's next-token logit minus its unmodified-baseline logit. The effect-vector cosine in the main text compares the two candidates' target-logit changes across eligible prompts, not their full-vocabulary logit vectors. The matched-control contrast first averages the two candidate effects and subtracts the average of the two control effects within each prompt, then averages across prompts; its 95\% interval is the percentile bootstrap interval over prompts, keeping all four edits paired. The stored prompt-level effects reproduce the reported $+0.472$ contrast and $[+0.347,+0.597]$ interval. These interventions isolate decoder writes and bypass native gates, so they test the shared downstream write of the two features. Coherent noun-phrase continuations remain compatible with the observed article-versus-possessive contexts.

\subsection{Extension to Larger Models}
\label{appendix:larger-models}

We extend the token-based DAME evaluation to public BatchTopK SAEs for Qwen3-8B, Qwen3-14B, and Qwen3-32B, and to Gemma Scope residual-stream SAEs for Gemma-2-9B. As in Section~\ref{subsec:dame}, Q1 is the highest mean-dominance quartile, and an SAE has a Q1 deficit when its mean token perfect-ranking accuracy in Q1 is lower than the pooled mean in Q2--Q4. Table~\ref{tab:larger_model_dame} reports the percentage of evaluated SAEs with this deficit separately for each model and SAE family. These percentages describe the direction of the per-SAE contrast, not the percentage with a statistically significant deficit.

\begin{table}[h]
\centering
\caption{Q1 token perfect-ranking deficits in public larger-model SAE panels. Percentages are computed within each model's evaluated SAE panel.}
\label{tab:larger_model_dame}
\small
\begin{tabular}{llc}
\toprule
Base model & Public SAE family & SAEs with a Q1 deficit\\
\midrule
Qwen3-8B & BatchTopK & 41.7\%\\
Qwen3-14B & BatchTopK & 50.0\%\\
Qwen3-32B & BatchTopK & 41.7\%\\
Gemma-2-9B & Gemma Scope residual & 75.0\%\\
\bottomrule
\end{tabular}
\end{table}

For Qwen3, features were sampled uniformly with a fixed seed within each dominance quartile; frozen reserves were promoted only to satisfy question-completeness requirements, without using scoring outcomes. The deficit occurs in a subset of evaluated SAEs at every tested scale, extending its observed scope beyond the primary benchmark's 70M--2.7B range. The Qwen3 sizes come from a single model family; broader prevalence across models, layers, and SAE families remains open (Appendix~\ref{appendix:limitations}).

\subsection{Activation-Support Geometry and Blinded Context Audit}
\label{appendix:activation-support-audit}

\paragraph{Support geometry.}
We analyze native positive-activation supports for the frozen feature-pair panel used in the decoder--explanation comparison (Appendix~\ref{appendix:decoder-to-embedding}), covering five SAE families. The primary analysis retains checkpoints with more than ten sampled pairs and pairs whose two features each fire at least 20 times in the pooled evaluation corpus; it does not require support in each half of that corpus. Decoder-aligned pairs have cosine similarity $\geq0.7$. For supports $S_i,S_j$, we measure Jaccard overlap $|S_i\cap S_j|/|S_i\cup S_j|$ and directional containment $|S_i\cap S_j|/|S_i|$ and $|S_i\cap S_j|/|S_j|$. We classify pairs as duplicate-like when both containments are at least 0.8; nested-like when the larger containment is at least 0.8 and the smaller at most 0.6; partition-like when Jaccard is at most 0.1 and both containments are at most 0.2; and partial overlap otherwise. These categories describe support shape only.

Table~\ref{tab:support_geometry_audit} first computes the fraction of eligible pairs in each category within each SAE, then averages across contributing SAEs with equal weight. The 95\% percentile intervals resample SAEs within family--architecture strata (10,000 replicates, seed 20260724), not individual pairs. In a separate matched comparison, high-cosine pairs are matched one-to-one without replacement to existing low-cosine pairs ($<0.3$) in the same SAE, minimizing distance between sorted log-transformed endpoint support sizes. Its high-cosine column therefore uses the matched subset, not the full eligible pool.

\begin{table}[h]
\centering
\caption{Support geometry: mean within-SAE pair percentages. Intervals describe the full eligible high-cosine pool; the last two columns use matched pairs.}
\label{tab:support_geometry_audit}
\small
\begin{tabular}{lccc}
\toprule
Support pattern & \shortstack{Full high-cosine\\Mean [95\% CI]} & \shortstack{Matched\\high-cosine} & \shortstack{Low-cosine\\controls} \\
\midrule
Partition-like & 21.5\% [17.6, 25.6] & 21.6\% & 80.0\% \\
Nested-like & 15.3\% [11.5, 19.3] & 16.1\% & 1.2\% \\
Duplicate-like & 29.9\% [25.1, 34.9] & 27.4\% & 2.1\% \\
Partial overlap & 33.3\% [28.8, 37.8] & 34.9\% & 16.7\% \\
\bottomrule
\end{tabular}

\end{table}

Partitioning also occurs, and is more frequent, among the low-cosine controls. Sharding therefore requires decoder alignment and context-level evidence in addition to separated supports; likewise, absorption additionally requires that a general-concept latent fails to activate while another feature takes over its role.

\paragraph{Blinded semantic audit.}
We audit all partition-screened, decoder-aligned targets from the primary analysis, together with low-cosine controls selected from the existing matches using frozen family--architecture quotas. Fresh WikiText-103 validation contexts are collected separately from the original selection corpus. Each feature's card contains its strongest activation contexts from distinct documents, with up to eight contexts per feature. GPT-5.6 Sol judges the public cards twice, with feature orientation reversed between passes and without access to feature IDs, model identity, selection geometry, or target/control role. A stable sibling-subclass reading requires both passes to judge the features assessable and to identify distinct subclasses under a named common parent; this is agreement in the relation category, not a requirement that the two parent descriptions be verbatim identical.

Table~\ref{tab:semantic_support_audit} reports descriptive percentages of cards in each stated panel. Cards with insufficient evidence remain in each panel's denominator. The full target pool and the smaller control-matched target subset are shown separately. These are pair-level percentages within the audited panels, unlike the SAE-averaged quantities in Table~\ref{tab:support_geometry_audit}.

\begin{table}[h]
\centering
\caption{Two-pass context judgments. ``Fresh support'' additionally requires the partition criterion to replicate in the new context corpus. The insufficient column means that at least one pass could not assess both features.}
\label{tab:semantic_support_audit}
\small
\begin{tabular}{lccc}
\toprule
Audit panel & \shortstack{Stable sibling\\reading} & \shortstack{Also fresh\\support} & \shortstack{Insufficient\\evidence} \\
\midrule
All screened targets & 15.8\% & 13.3\% & 51.4\% \\
Control-matched targets & 13.3\% & 13.3\% & 60.0\% \\
Low-cosine controls & 3.3\% & 3.3\% & 56.7\% \\
\bottomrule
\end{tabular}

\end{table}

\paragraph{Illustrative context distinctions.}
Two cases with fresh-support replication illustrate the sibling readings. In Pythia-160M, a BatchTopK-4k SAE (\texttt{trainer\_0}) has features 2081 and 3844 with decoder cosine 0.759 and fresh Jaccard 0: the first activates on \emph{they}, the second on \emph{he}/\emph{she}. Both passes describe a grammatical-number distinction among third-person pronouns. Table~\ref{tab:pronoun_activation_contexts} shows selected real contexts from this pair's saved audit card.

\begin{table}[htbp]
\centering
\caption{Selected activation contexts for the Pythia-160M BatchTopK-4k pair (decoder cosine 0.759; fresh-support Jaccard 0). Bold marks the activating token. Contexts are independently selected for each feature; ellipses mark cropped text.}
\label{tab:pronoun_activation_contexts}
\small
\begin{tabularx}{\linewidth}{@{}>{\raggedright\arraybackslash}X >{\raggedright\arraybackslash}X@{}}
\toprule
\textbf{Feature 2081: \emph{they}} & \textbf{Feature 3844: \emph{he}/\emph{she}} \\
\midrule
\ldots\ the NBA, \textbf{they} like them young, \ldots &
\ldots\ declared that \textbf{he} will be a better leader for the sake of his friends \ldots \\
\addlinespace[0.6em]
\ldots\ after having suffered heavy casualties, \textbf{they} returned to Tagaung. &
Radcliffe has supported various charities. \textbf{He} designed \ldots \\
\addlinespace[0.6em]
\ldots\ have stated numerous times that \textbf{they} do not condone \ldots &
\ldots\ Madonna starts fighting with the men. \textbf{She} starts bending backwards \ldots \\
\bottomrule
\end{tabularx}

\end{table}

In Gemma-2-2B, a PAnneal-4k SAE (\texttt{trainer\_5}) has features 1352 and 2535 with decoder cosine 0.794 and fresh Jaccard 0.088. Recorded contexts include ``It \textbf{is} set in common time'' and ``He \textbf{was} promoted to flight lieutenant''; both passes identify forms of the verb \emph{be}. The latter contrast also permits a copula-versus-passive-auxiliary reading. Both are coherent grammatical distinctions. They provide context-level support beyond geometry, complementing the decoder-write evidence of the intervention case in Appendix~\ref{appendix:functional-sharding-protocol}.

\clearpage

\section{Notation and Terminology}
\label{appendix:notation}

\begin{longtable}{@{}p{0.25\linewidth}p{0.71\linewidth}@{}}
\toprule
Symbol / term & Meaning \\
\midrule
\endhead
$x$ & An input token position. \\
$\phi(x) \in \mathbb{R}^{d_{\text{in}}}$ & Model activation at the hooked layer; $d_{\text{in}}$ is the activation dimension. \\
$m$ & SAE width: the number of features in the dictionary. \\
$f_i(x)$, $d_i$, $w_i$ & Activation, decoder direction, and encoder row of feature $i$. \\
$\mathcal{C}$, $U$ & Target concept set fixed for an analysis at a stated explanatory granularity; set of interpretable units to which attribution is assigned. \\
Atomic explanatory unit & A basic coordinate to which attribution is assigned directly, without further decomposition within the chosen explanatory interface. \\
AIU basis & Atomic interpretable unit basis: a coordinate system in which attribution and intervention are semantically meaningful. \\
C1--C5 & Non-redundancy, completeness, monosemanticity, and no dead units (semantic-correspondence requirements); per-input sparsity (input-conditioned accessibility requirement). \\
Firing frequency $\nu_i$ & Proportion of inputs on which the activation of feature $i$ exceeds the threshold $\epsilon$. \\
Per-input dominance $\rho_i(x)$ & Fraction of total feature activation on input $x$ contributed by feature $i$. \\
Mean dominance $\bar{\rho}_i$ & $\rho_i(x)$ averaged over inputs. \\
Q1--Q4 & Dominance quantiles of an SAE's features; Q1 contains the most dominant features. \\
$L_0$ & Number of features active on an input, averaged over inputs. \\
SAE family; architecture & A collection of SAEs from one source for one base model (e.g., SAEBench Pythia, CanRager); the SAE type (Standard ReLU, TopK, BatchTopK, Matryoshka, JumpReLU, Gated, PAnneal). \\
DAME & Dominance-Aware Monosemanticity Evaluation: dominance-stratified feature sampling with a three-bin ranking validation. \\
Explainer; judge & The LLM that writes a feature's explanation; the LLM that ranks held-out examples given the explanation. \\
Perfect ranking accuracy & Fraction of validation questions in which the judge recovers the exact activation order of the three examples. \\
Q1 deficit & An SAE has a Q1 deficit when its mean token-based perfect ranking accuracy is lower for Q1 than for Q2--Q4. \\
Redundant pair & Two features whose decoder cosine similarity is at least $\theta$ ($\theta = 0.7$ unless stated). \\
Feature sharding & A pairwise representational pattern: similar decoder directions expressed through complementary firing contexts. \\
Feature absorption & A functional claim \citep{chanin2024absorption}: a latent for a general concept fails on hierarchical child cases while other latents take over its role. \\
Dead feature & A feature whose firing frequency is below the dead-feature threshold on the evaluation data. \\
\bottomrule
\end{longtable}

\clearpage
\section*{NeurIPS Paper Checklist}

\begin{enumerate}

\item {\bf Claims}
    \item[] Question: Do the main claims made in the abstract and introduction accurately reflect the paper's contributions and scope?
    \item[] Answer: \answerYes{}
    \item[] Justification: The abstract and introduction claim three contributions---the atomic interpretable unit framework, the DAME methodology, and the empirical finding that dominant features are systematically less monosemantic and more redundant---each of which is delivered in the corresponding section of the paper.

\item {\bf Limitations}
    \item[] Question: Does the paper discuss the limitations of the work performed by the authors?
    \item[] Answer: \answerYes{}
    \item[] Justification: Yes; we provide a dedicated Limitations appendix (Appendix~\ref{appendix:limitations}) covering four items: model scale (a small- to medium-model primary benchmark with limited larger-model checks), the conditional rather than absent nature of ground-truth concept sets in real LLMs, the proxy nature of LLM-based monosemanticity scoring (a limitation shared by all auto-interpretability methods, including ours), and hyperparameter-transfer caveats across LLMs.

\item {\bf Theory assumptions and proofs}
    \item[] Question: For each theoretical result, does the paper provide the full set of assumptions and a complete (and correct) proof?
    \item[] Answer: \answerNA{}
    \item[] Justification: The paper does not include formal theorems or lemmas requiring proofs. The atomic interpretable unit framework and the encoder--decoder asymmetry argument are presented as definitions and conceptual derivations in the main body, not as numbered theoretical results.

    \item {\bf Experimental result reproducibility}
    \item[] Question: Does the paper fully disclose all the information needed to reproduce the main experimental results of the paper to the extent that it affects the main claims and/or conclusions of the paper (regardless of whether the code and data are provided or not)?
    \item[] Answer: \answerYes{}
    \item[] Justification: Yes; Appendix~\ref{appendix:experiment_details} details the experimental setup, all SAE families and architectures evaluated, and training hyperparameters for our newly trained SAEs, while Appendix~\ref{appendix:other_metrics} specifies the per-criterion proxy-metric definitions. We also release the evaluation code and result data as supplementary material.

\item {\bf Open access to data and code}
    \item[] Question: Does the paper provide open access to the data and code, with sufficient instructions to faithfully reproduce the main experimental results, as described in supplemental material?
    \item[] Answer: \answerYes{}
    \item[] Justification: {\def\UrlBreaks{\do\/\do-}\def\UrlBigBreaks{\do\:}Yes; we release a public code repository (\url{https://github.com/ritengzhang/SAE-AIU}) containing the evaluation framework, training pipeline, and aggregation scripts, alongside per-SAE result data and a README documenting installation and reproduction commands, and the checkpoints of our newly trained SAEs (\url{https://huggingface.co/ritengzhang/SAE-AIU}).}

\item {\bf Experimental setting/details}
    \item[] Question: Does the paper specify all the training and test details (e.g., data splits, hyperparameters, how they were chosen, type of optimizer) necessary to understand the results?
    \item[] Answer: \answerYes{}
    \item[] Justification: Yes; Appendix~\ref{appendix:experiment_details} lists the training schedule, optimizer settings, activation storage and feature-management hyperparameters, evaluation batches, and per-architecture variants for the SAEs we trained, and credits the SAEBench / jbloom / CanRager release configurations for the externally loaded families.

\item {\bf Experiment statistical significance}
    \item[] Question: Does the paper report error bars suitably and correctly defined or other appropriate information about the statistical significance of the experiments?
    \item[] Answer: \answerYes{}
    \item[] Justification: Yes; the main DAME results (Section~\ref{subsec:dame}) are accompanied by per-SAE Wilcoxon signed-rank tests (computed via \texttt{scipy.stats.wilcoxon}, one-sided, with each SAE as the unit of variability) with reported $p$-values and Cohen's $d$ effect sizes; the main redundancy results (Section~\ref{subsec:redundancy}) report endpoint enrichment with 95\% SAE-bootstrap confidence intervals; the SAEBench correlation analysis (Appendix~\ref{appendix:saebench-correlations}) reports Spearman $\rho$ (\texttt{scipy.stats.spearmanr}) with significance markers, and the decoder--explanation similarity validation (Appendix~\ref{appendix:decoder-to-embedding}) reports both pooled-pair and per-SAE Spearman $\rho$.

\item {\bf Experiments compute resources}
    \item[] Question: For each experiment, does the paper provide sufficient information on the computer resources (type of compute workers, memory, time of execution) needed to reproduce the experiments?
    \item[] Answer: \answerYes{}
    \item[] Justification: Yes; Appendix~\ref{appendix:compute_resources} reports the GPU type and memory (4$\times$ NVIDIA Quadro RTX 6000, 24~GB each), per-SAE training-time range, and approximate total compute budget for both training and evaluation.

\item {\bf Code of ethics}
    \item[] Question: Does the research conducted in the paper conform, in every respect, with the NeurIPS Code of Ethics \url{https://neurips.cc/public/EthicsGuidelines}?
    \item[] Answer: \answerYes{}
    \item[] Justification: Yes; the work is an evaluation methodology for sparse autoencoders, conducted using publicly released models and datasets, and we have reviewed the NeurIPS Code of Ethics. The work does not introduce any new dataset, deployed system, or human-subject component.

\item {\bf Broader impacts}
    \item[] Question: Does the paper discuss both potential positive societal impacts and negative societal impacts of the work performed?
    \item[] Answer: \answerYes{}
    \item[] Justification: Positive impact: our framework and DAME methodology raise the rigor of SAE evaluation, which directly supports interpretability-based safety oversight (circuit tracing, feature attribution, intervention) used in alignment research. Negative impact is minimal: we release an evaluation framework rather than a deployable model or dataset, do not improve any generative or surveillance capability, and use only previously released models and datasets.

\item {\bf Safeguards}
    \item[] Question: Does the paper describe safeguards that have been put in place for responsible release of data or models that have a high risk for misuse (e.g., pre-trained language models, image generators, or scraped datasets)?
    \item[] Answer: \answerNA{}
    \item[] Justification: We do not release any high-risk assets. The paper releases evaluation code, per-SAE result files, and the SAE checkpoints we trained; we do not release new model checkpoints with high misuse potential, and these SAE checkpoints are interpretability artifacts attached to small- to medium-scale models with no demonstrated dual-use risk.

\item {\bf Licenses for existing assets}
    \item[] Question: Are the creators or original owners of assets (e.g., code, data, models), used in the paper, properly credited and are the license and terms of use explicitly mentioned and properly respected?
    \item[] Answer: \answerYes{}
    \item[] Justification: Yes; Appendix~\ref{appendix:licenses} lists each external asset (SAE families, models, datasets, libraries, and the LLM explainer and judge models) with its license, including those for which the public release does not declare an explicit license, and all usage complies with the corresponding terms. We do not redistribute any third-party model weights, SAE checkpoints, or datasets beyond pointers to their original public locations.

\item {\bf New assets}
    \item[] Question: Are new assets introduced in the paper well documented and is the documentation provided alongside the assets?
    \item[] Answer: \answerYes{}
    \item[] Justification: Yes; the public code release (training pipeline, evaluation framework, aggregation scripts, and sample result data) is documented via a top-level README and per-folder READMEs covering installation, the data layout, and reproduction commands, and the released SAE checkpoints are documented by a model card.

\item {\bf Crowdsourcing and research with human subjects}
    \item[] Question: For crowdsourcing experiments and research with human subjects, does the paper include the full text of instructions given to participants and screenshots, if applicable, as well as details about compensation (if any)?
    \item[] Answer: \answerNA{}
    \item[] Justification: The paper does not involve crowdsourcing or research with human subjects.

\item {\bf Institutional review board (IRB) approvals or equivalent for research with human subjects}
    \item[] Question: Does the paper describe potential risks incurred by study participants, whether such risks were disclosed to the subjects, and whether Institutional Review Board (IRB) approvals (or an equivalent approval/review based on the requirements of your country or institution) were obtained?
    \item[] Answer: \answerNA{}
    \item[] Justification: The paper does not involve human subjects research; no IRB approval is applicable.

\item {\bf Declaration of LLM usage}
    \item[] Question: Does the paper describe the usage of LLMs if it is an important, original, or non-standard component of the core methods in this research? Note that if the LLM is used only for writing, editing, or formatting purposes and does \emph{not} impact the core methodology, scientific rigor, or originality of the research, declaration is not required.
    \item[] Answer: \answerYes{}
    \item[] Justification: Yes; DAME (Section~\ref{mono_dom}) uses \texttt{mistralai/Mistral-7B-Instruct-v0.2}, run locally, to judge feature explanations, following \citet{bills2023language} and \citet{paulo2024automatically}. Appendices~\ref{appendix:cross-judge} and~\ref{appendix:activation-support-audit} describe the additional GPT-5.6 Sol explainer/judge comparison and blinded context judgments.

\end{enumerate}

\end{document}